\documentclass[10pt,twocolumn,letterpaper]{article}

\usepackage[pagenumbers]{cvpr} %

\usepackage{caption}
\usepackage{subcaption}
\usepackage{multirow}

\usepackage[x11names]{xcolor}
\definecolor{darkred}{rgb}{0.7,0.1,0.1}
\definecolor{darkgreen}{rgb}{0.1,0.5,0.1}
\definecolor{cyan}{rgb}{0.7,0.0,0.7}
\definecolor{otherblue}{rgb}{0.1,0.4,0.8}
\definecolor{maroon}{rgb}{0.76,.13,.28}
\definecolor{burntorange}{rgb}{0.81,.33,0}
\definecolor{othergreen}{rgb}{0.29,0.49,0.07}
\definecolor{orange}{rgb}{1.0,0.65,0.0}
\definecolor{darkorange}{rgb}{1.0, 0.45, 0.0}

\usepackage[pagebackref,breaklinks,colorlinks]{hyperref}

\title{SceNeRFlow: Time-Consistent Reconstruction of General Dynamic Scenes}

\author{Edith Tretschk$^1$~~~Vladislav Golyanik$^1$~~~Michael Zollh\"ofer$^2$\\
~~~Alja\v{z} Bo\v{z}i\v{c}$^2$~~~~~~Christoph Lassner$^2$~~~~Christian Theobalt$^1$
\\ \\
$^1$Max Planck Institute for Informatics, Saarland Informatics Campus~~~~$^2$Meta Reality Labs Research
}

\begin{document}
\maketitle
\begin{abstract}
    
Existing methods for the 4D reconstruction of general, non-rigidly deforming objects focus on novel-view synthesis and neglect correspondences. 
However, time consistency enables advanced downstream tasks like 3D editing, motion analysis, or virtual-asset creation. 
We propose \emph{SceNeRFlow} to reconstruct a general, non-rigid scene in a time-consistent manner. 
Our dynamic-NeRF method takes multi-view RGB videos and background images from static cameras with known camera parameters as input. 
It then reconstructs the deformations of an estimated canonical model of the geometry and appearance in an online fashion. 
Since this canonical model is time-invariant, we obtain correspondences even for long-term, long-range motions. 
We employ neural scene representations to parametrize the components of our method. 
Like prior dynamic-NeRF methods, we use a backwards deformation model. 
We find non-trivial adaptations of this model necessary to handle larger motions: We decompose the deformations into a strongly regularized coarse component and a weakly regularized fine component, where the coarse component also extends the deformation field into the space surrounding the object, which enables tracking over time. 
We show experimentally that, unlike prior work that only handles small motion, our method enables the reconstruction of studio-scale motions.
\end{abstract}

\vspace{-1em}

\section{Introduction}\label{sec:introduction}

\begin{figure}
    \centering
    
    \includegraphics[width=\columnwidth]{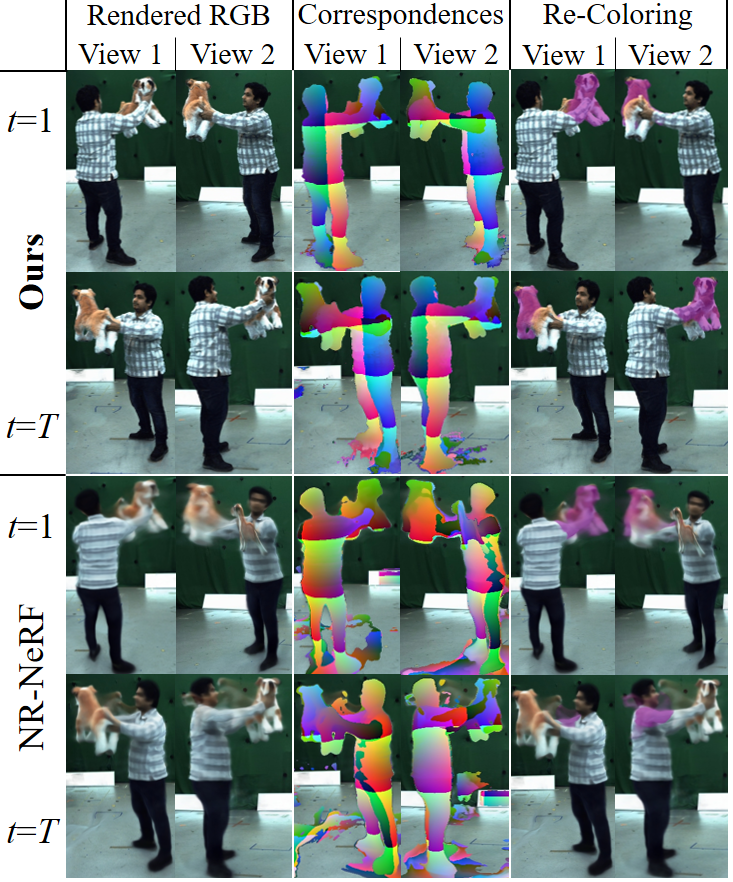}
    
    \caption{\textbf{SceNeRFlow.}
    Our NeRF-based method reconstructs a general non-rigid scene from multi-view videos \emph{with time consistency}. 
    Here, a person with a plush dog rotates $180^{\circ}$ from time $t{=}1$ until $t{=}T$. 
    Novel view $1$ at $t{=}1$ is consistent with novel view $2$ at $t{=}T$  (placed opposite of novel view $1$) for our method, but not for NR-NeRF~\cite{tretschk2021non}. %
    This enables, \emph{e.g.}, time- and view-consistent re-coloring. %
    We color the correspondences according to 3D positions in static canonical space, which differs between both methods. 
    }
    \label{fig:teaser}
    \vspace{-2em}
\end{figure}

One criterion for the proper reconstruction of a deforming scene is time consistency, \emph{i.e.} correspondences across time. 
Reconstructing general dynamic objects from RGB input in a time-consistent manner is a little %
explored~\cite{cagniart2010probabilistic,mustafa2016temporally}, but challenging and highly relevant research direction. 
Establishing correspondences is equivalent to factorizing the reconstruction into time-varying deformations and a time-invariant geometry model. 
High-level tasks that go beyond novel-view synthesis benefit immensely from such a deeper analysis of the scene, namely a reconstruction with long-range, long-term dense 3D correspondences. 
For example, having access to such a virtual dynamic 3D object is a crucial step towards sophisticated 3D editing or estimating a motion model for virtual-asset creation~\cite{Chen_VEO}. 
Almost all existing work on reconstructing general dynamic objects either only obtains time consistency for small motion or relaxes it to very small time windows. 
In this work, we explore the setting of time consistency for large motions.

Prior work~\cite{park2021nerfies,li2021neural,tretschk2021non,pumarola2021d,xian2021space,du2021neural,park2021hypernerf,gao2021dynamic} for non-rigid 3D reconstruction based on NeRF~\cite{mildenhall2020nerf} focuses on novel-view synthesis and does not aim for long-range correspondences. 
Almost all dynamic-NeRF %
papers (except for \cite{pumarola2021d,tretschk2021non,liu2022devrf}, which only handle small motions) weaken the long-term consistency of the reconstruction by design by having a time-varying geometry and/or appearance model. 
Thus, only rather simple tasks like replaying the input scene under novel views are straightforward. 
Furthermore, as we will show in our experiments, this time dependency improves the novel-view rendering quality but loosens long-term correspondences. 
In contrast, we explore the other extreme, where \emph{only} deformations are time-variant. %

Classical, non-NeRF-based work on 4D reconstruction only handles small motion~\cite{Russell2012,Kumar2018,salzmann2007deformable,Bartoli2015,kair2022sft} or is not time-consistent~\cite{newcombe2015dynamicfusion,dou2016fusion4d,Yoon_2020_CVPR,bansal20204d}.
Similarly, scene flow~\cite{zhai2021optical} focuses on the deformations and not a full reconstruction, and exhibits correspondence drift \cite{hung2013consistent}, similar to optical flow.

In this paper, we propose \emph{SceNeRFlow} (Scene Flow + NeRF) to tackle time-consistent reconstruction of a general, non-rigidly deforming scene; see Fig.~\ref{fig:teaser}. 
Our method is NeRF-based, trained per scene, and by design estimates each timestamp's deformation of a time-invariant canonical model, thereby counteracting correspondence drift. 
As is common for dynamic NeRFs~\cite{park2021nerfies,pumarola2021d,tretschk2021non}, we employ a backward deformation model. 
Standard reconstruction techniques like online optimization, coarse-and-fine deformation decomposition, and rigidity regularization turn out to be insufficient for large motion. 
While category-specific priors like an articulated human skeleton would help by normalizing out large motion, this work tackles the general setting, where such prior knowledge is not available. 
Instead, we crucially find that carefully ``extending the deformation field'' is the minimally necessary change to make backwards deformation modeling work for large non-rigid motion.

Since ours is the first NeRF-based work with time consistency for large motion, we want to focus on the core problem of long-term correspondences. 
We therefore use multi-view input to avoid the need for correctly modeling the deformations of occluded geometry that arises in the monocular setting. 
Importantly, our scenes exhibit significantly larger motion than any prior time-consistent general reconstruction method could handle. 
Our experiments show how SceNeRFlow enables the time-consistent reconstruction even of studio-scale motion without any category-specific priors, \emph{e.g.} without a human skeleton.

In summary, our \textbf{contributions} are (1) an end-to-end differentiable, time-consistent 4D reconstruction method for general dynamic scenes from multi-view RGB input from static cameras, which is built around (2) a general approach to make backward deformation models work for larger motion via ``extending the deformation field'' %
(Sec.~\ref{subsec:tracking}).

\section{Related Work}\label{sec:related_work}

\noindent\textbf{3D Correspondence Estimation.} 
For temporal data, Vedula \textit{et al.}'s seminal work~\cite{vedula1999three} introduced scene flow as the 3D equivalent of optical flow, which led to many approaches tackling the problem~\cite{huguet2007variational,basha2013multi,quiroga2014dense,vogel20153d,yoon20183d,teed2021raft}, as a recent survey~\cite{zhai2021optical} summarizes. 
Lately, learning-based methods~\cite{liu2019flownet3d,gu2019hplflownet,Mittal_2020_CVPR,li2021neuralsceneflow} focus on 3D point clouds as input. 
More generally, non-rigid 3D point-cloud registration~\cite{deng2022survey} estimates correspondences of provided 3D point clouds. 
We go beyond just estimating 3D correspondences to create a full reconstruction (including appearance) in 3D space from only 2D input data, in an end-to-end differentiable pipeline.

\noindent\textbf{4D Reconstruction of General Non-Rigid Scenes.}
Multiple lines of research target 4D reconstruction without relying on NeRF. 
As a recent survey~\cite{TretschkNonRigidSurvey} discusses, monocular RGB input has traditionally been tackled by Non-Rigid Structure-from-Motion~\cite{Russell2012,Garg2013,Kumar2018,Parashar_2020_CVPR,Grasshof2022} and Shape-from-Template methods~\cite{salzmann2007deformable,Bartoli2015,Ngo2015,casillas2021isowarp,kair2022sft}, although learning-based approaches~\cite{cmrKanazawa18,vmr2020,wu2021dove,Kokkinos_2021_CVPR,Yoon_2020_CVPR,yang2022banmo} have grown in popularity over the last years. 
Due to the restrictive input setting, they either only handle small motions or lack time consistency. 
Slightly larger motions are possible with a single RGB-D camera~\cite{zollhofer2018state}, although these methods~\cite{zollhoefer2014deformable,newcombe2015dynamicfusion,guo2015robust,innmann2016volume,guo2017realtime,slavcheva2017killingfusion,Slavcheva_2018_CVPR,bozic2020neural,bozic2020deepdeform,lin2022occlusionfusion,Cai2022NDR} all update~\cite{curless1996volumetric} both previously unseen \emph{and previously seen} parts of the canonical model over time during their online optimization, thereby losing time consistency (except for the template-based~\cite{zollhoefer2014deformable}). 
Multi-view input has seen comparatively less work over the years~\cite{zhang2003spacetime,goldluecke2004space,larsen2007temporally,tung2009complete}. 
Fusion4D~\cite{dou2016fusion4d,orts2016holoportation} operates in real time but also modifies the canonical model over time. 
Mustafa \textit{et al.}~\cite{mustafa2016temporally} track a mesh through a temporal sequence according to optical flow and refine its topology for each timestamp, but do not reconstruct the appearance. 
Unlike such multi-view mesh-tracking approaches~\cite{cagniart2010probabilistic,zhao2022human}, our method is end-to-end differentiable and can easily integrate recent advances in neural scene representations~\cite{tewari2021advances}. 
Bansal \textit{et al.}~\cite{bansal20204d} obtain impressive but time-inconsistent novel-view results.

\noindent\textbf{Dynamic NeRFs.}
2D neural rendering~\cite{sotaNeuralRendering} and 3D neural scene representations~\cite{tewari2021advances} based on NeRF~\cite{mildenhall2020nerf} have in recent years seen great success. 
Apart from early work~\cite{Lombardi2019neuralvol}, current methods~\cite{park2021nerfies,li2021neural,tretschk2021non,pumarola2021d,xian2021space,du2021neural,park2021hypernerf,gao2021dynamic,Menapace2022PlayableEnvironments} for dynamic scene reconstruction with neural representations build on NeRF. 
Some approaches focus on surface~\cite{johnson2022ub4d}, fast~\cite{fang2022tineuvox,Guo_2022_NDVG_ACCV}, generalizable~\cite{wang2022fourier}, or depth-supported~\cite{attal2021torf} reconstruction. 
Our method is most related to this field of work but differs in its goal: we do not primarily aim for novel-view synthesis but rather for a \emph{time-consistent} reconstruction. 
A few prior methods~\cite{pumarola2021d,tretschk2021non,liu2022devrf} do obtain time consistency but either only in synthetic~\cite{pumarola2021d} or small-motion~\cite{tretschk2021non,liu2022devrf} settings. 
Furthermore, most works use monocular input~\cite{gao2022dynamic} and only a few~\cite{wang2022fourier,li2022dynerf,uii_eccv22_pref,liu2022devrf} explore multi-view input, where the latter forego long-term time consistency or only handle small motion~\cite{liu2022devrf}.

\section{Method}\label{sec:method} 

\begin{figure*} 
    \centering 
    \includegraphics[width=\textwidth]{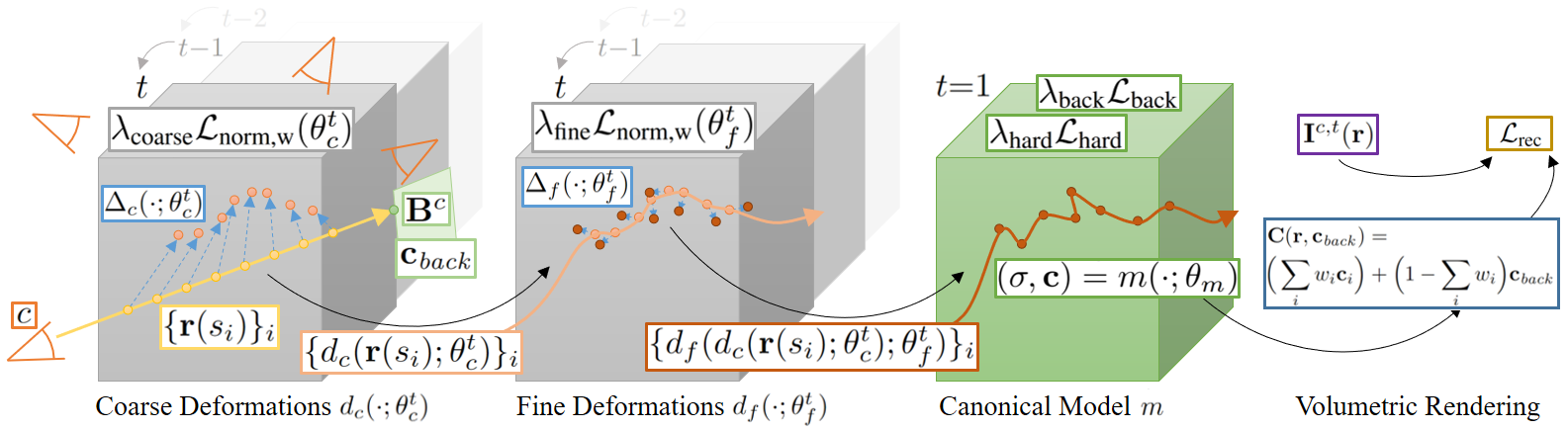} 
    \caption{\textbf{SceNeRFlow Overview.} %
    \emph{Input:} We take as input a {\color{Purple4}multi-view RGB video $\{\mathbf{I}^{c,t}\}_{c,t}$} of a general dynamic scene from $C$ cameras with {\color{DarkOliveGreen3}background images $\{\mathbf{B}^c\}_c$} and known camera parameters. 
    \emph{Output:} From this, we build a NeRF-based~\cite{mildenhall2020nerf}, time-consistent 4D reconstruction. 
    \emph{Model:} 
    To render a {\color{Gold2}ray $\mathbf{r}$} from {\color{Chocolate1} camera $c$}, we first sample {\color{Gold2}discrete points $\{\mathbf{r}(s_i)\}_i$} along the straight ray (with {\color{DarkOliveGreen3}background color $\mathbf{c}_{\text{back}}$} at the end), then coarsely bend the ray with the {\color{Sienna2}coarse deformations $d_c$}, and then finely bend the resulting ray with the {\color{Sienna4}fine deformations $d_f$}. 
    We then query the {\color{Chartreuse4}\emph{canonical model} $m$} at the resulting positions. 
    $m$ represents geometry (opacity $\sigma$) and appearance (color $\mathbf{c}$) volumetrically for any 3D point, in a time-invariant manner. 
    Finally, we use NeRF-style volumetric rendering to obtain the {\color{SteelBlue4}ray's color $\mathbf{C}$}. 
    We parametrize all three components with HashMLPs~\cite{mueller2022instant}. 
    \emph{Training:} We construct $m$ from time $t{=}1$ using a {\color{Goldenrod3} reconstruction loss} ({\color{Goldenrod3}$\mathcal{L}_{\text{rec}}$} w.r.t.\ the {\color{Purple4} ground-truth color $\mathbf{I}^{c,t}(\mathbf{r})$}) and geometric regularizers ($\mathcal{L}_{\text{back}}$ and $\mathcal{L}_{\text{hard}}$), and keep $m$'s parameters $\theta_m$ fixed for future timestamps, which leads to time consistency. 
    For $t{>}1$, we split the deformations into a strongly regularized coarse component $\Delta_c$ and a weakly regularized fine component $\Delta_f$. 
    Our online optimization performs frame-wise tracking to handle large motion and we thus initialize the deformation parameters $\theta^t_c$ and $\theta^t_f$ at time $t$ with $\theta^{t{-}1}_c$ and $\theta^{t{-}1}_f$. 
    To handle a peculiarity of backward deformation models in this tracking setting, we regularize $\Delta_c$ to ``extend the deformation field'' ($\mathcal{L}_{\text{norm,w}}$). 
    }
    \label{fig:pipeline}
\end{figure*}

SceNeRFlow takes as input multi-view RGB images of size $h{\times}w$ over $T$ consecutive timestamps from $C$ static cameras with known extrinsics and intrinsics, \emph{i.e.} $\{\mathbf{I}^{c,t}\in[0,1]^{h\times w \times 3}\}^{C,T}_{c=1,t=1}$, and associated background images $\{\mathbf{B}^c\}_c$. 
In a time-consistent manner, it then reconstructs the general dynamic scene as a time-invariant, NeRF-style canonical model of the geometry and appearance, with time-dependent deformations. 
The optimization proceeds in an online manner: we build a canonical model from the first timestamp (Sec.~\ref{subsec:canonical}) and then track it frame-by-frame through the temporal input sequence (Sec.~\ref{subsec:tracking}). 
Fig.~\ref{fig:pipeline} provides an overview of the pipeline. 

\subsection{Constructing the Canonical Model}\label{subsec:canonical}

\noindent\textbf{Canonical Model.} 
The canonical model $m$ encodes the geometry and appearance in a time-independent manner. 
Specifically, we use a \emph{HashMLP}~\cite{mueller2022instant} to represent opacity $\sigma$ and RGB color $\mathbf{c}\in[0,1]^3$ for any 3D point $\mathbf{x}$: $(\sigma,\mathbf{c}) = m(\mathbf{x})$.  %
A HashMLP combines a \emph{hash grid} $v$ with a subsequent shallow MLP $M$: $m(\mathbf{x}) = M(v(\mathbf{x}))$, which is significantly faster than NeRF's~\cite{mildenhall2020nerf} pure MLP. 
A hash grid consists of about a dozen voxel grids of increasing resolution, each containing learnable features. 
To evaluate, each grid is queried via trilinear interpolation and the resulting features from all grids are concatenated. 
Crucially, each voxel grid is implemented via hashed indexing into an array of feature vectors rather than as a dense voxel grid.

\noindent\textbf{Rendering.} 
Since we take 2D images as input, we need to render $m$ into 2D. 
To this end, we follow the quadrature discretization of volumetric rendering from NeRF~\cite{mildenhall2020nerf} for a ray $\mathbf{r}(s) = \mathbf{o} + s\mathbf{d}$ with origin $\mathbf{o}\in\mathbb{R}^3$ and direction $\mathbf{d}\in\mathbb{R}^3$: 

{\footnotesize 
\begin{equation}\label{eq:rendering}
    \mathbf{C}(\mathbf{r}) = \sum_{i=1}^S w_i\mathbf{c}_i, \text{ with } w_i =  \exp \Bigg(-\sum_{j=1}^{i-1}\sigma_j \delta_j\Bigg)(1-\exp(-\sigma_i \delta_i)),
\end{equation}
}
\hspace{-6pt} 
where $i$ indexes $S$ discrete samples $\{\mathbf{r}(s_i)\}_i$ along the ray; ($\sigma_i$,$\mathbf{c}_i)=m(\mathbf{r}(s_i))$ are the opacity and color of the $i$-th sample, respectively; and $\delta_i = s_{i+1} - s_i$. 
Like NeRF~\cite{mildenhall2020nerf}, we use stratified sampling of $S$ evenly sized intervals between the near plane and far plane to obtain $\{s_i\}_i$.  

When a background color $\mathbf{c}_{\mathit{back}}$ is provided (\textit{e.g.} from $\mathbf{B}^c$), we composite it at the end of the ray: 
\begin{equation}\label{eq:renderback}
\mathbf{C}(\mathbf{r},\mathbf{c}_{\mathit{back}}) = \mathbf{C}(\mathbf{r}) + \Big(1-\sum_i w_i\Big)\mathbf{c}_{\mathit{back}}.
\end{equation}

\noindent\textbf{Losses.} 
To obtain the canonical model, we iteratively optimize it on images $\{\mathbf{I}^{c,1}\}_c$ from $t{=}1$ for $20k$ iterations. 
Each iteration uses a batch of $R$ rays with an $\ell_1$ reconstruction loss w.r.t.\ the ground-truth color $\mathbf{I}^{c,1}(\mathbf{r}_r)$: 
\begin{equation}
    \mathcal{L}_{\text{rec}} = \frac{1}{R} \sum_{r=1}^R \lVert \mathbf{C}(\mathbf{r}_r,\mathbf{c}_{\mathit{back},r}) - \mathbf{I}^{c,1}(\mathbf{r}_r) \rVert_1.
\end{equation}

However, using only $\mathcal{L}_{\text{rec}}$ leads to a canonical model with floating geometry artifacts. 
We remove them 
by steering the model to commit to foreground \emph{or} background by encouraging $\sum_i w_i$ to be $1$ or $0$ via the beta distribution \cite{Lombardi2019neuralvol}: 
\begin{equation}
\mathcal{L}_{\text{back}} = \frac{1}{R} \sum_r \log\Big(\sum_i w_{r,i}\Big) + \log\Big(1-\sum_i w_{r,i}\Big),
\end{equation}
where $w_{r,i}$ is $w_i$ of ray $r$. 
We also discourage transparent geometry by encouraging each $w_i$ to be $0$ or $1$ via a mixture of two Laplacian distributions \cite{rebain2022lolnerf}: 
\begin{equation}
    \mathcal{L}_{\text{hard}} = -\frac{1}{RS} \sum_r \sum_i \log\Big(e^{-w_{r,i}} + e^{-(1-w_{r,i})}\Big).
\end{equation}
We note that these losses do not use foreground masks; the canonical model needs to learn on its own whether it can be transparent and use the background image $\mathbf{B}$ or not. 

The total loss for the canonical model is thus: 
\begin{equation}
    \mathcal{L}_{\text{canon}} = \mathcal{L}_{\text{rec}} + \lambda_{\text{back}} \mathcal{L}_{\text{back}} + \lambda_{\text{hard}} \mathcal{L}_{\text{hard}},
\end{equation}
where $\lambda_{\text{back}},\lambda_{\text{hard}}\in\mathbb{R}$ are loss weights. 
Since we want the canonical model to be time-consistent, we keep its parameters fixed after constructing it from the first timestamp. 

\subsection{Optimizing per Timestamp}\label{subsec:tracking}

Given the canonical model from $t{=}1$, we next reconstruct its deformations for the remaining timestamps $t{>}1$. 

\noindent\textbf{Space Warping.} 
We model the deformations at time~$t$ like prior dynamic-NeRF work: we use backwards space warping $d_c(\mathbf{x};\theta_c^t)=\mathbf{x} + \Delta_c(\mathbf{x};\theta_c^t) = \mathbf{x}'$, where $\mathbf{x}$ is a 3D point in deformed world space, $\Delta_c$ outputs a coarse offset (fine offsets $\Delta_f$ will be introduced later), $\mathbf{x}'$ is in undeformed canonical space, and $\theta_c^t$ are the time-dependent parameters of $d_c$/$\Delta_c$. 
We use a HashMLP to parametrize $\Delta_c$, \emph{i.e.} there is one HashMLP per timestamp for coarse deformations. 
To query the canonical model from world space at $t{>}1$, we first undo the deformation: $(\sigma(\mathbf{x},t), \mathbf{c}(\mathbf{x},t))=m(d_c(\mathbf{x};\theta_c^t))$. 
When rendering, this leads to view-consistent ray bending: $\{\mathbf{r}(s_i)\}_i$ in world space becomes $\{d_c(\mathbf{r}(s_i);\theta_c^t)\}_i$ in canonical space. 
We can then apply Eq.~\ref{eq:rendering} or Eq.~\ref{eq:renderback} to the bent ray to render, which in turn enables us to use $\mathcal{L}_{\text{rec}}$ to optimize for the deformation parameters at time $t$ using $\{\mathbf{I}^{c,t}\}_c$. 

\noindent\textbf{Frame-Wise Tracking.} 
However, na\"ively reconstructing the entire scene by optimizing all timestamps at once does not converge to a recognizable canonical model %
when the scene contains large motion. 
Furthermore, keeping all $\{\mathbf{I}^{c,t}\}_{c,t}$ in memory at once is expensive in practice. 
We thus propose online, timestamp-by-timestamp tracking. 
Since $t{=}1$ has, by construction, zero offsets, we set the last layer of $\Delta_c(\cdot;\theta_c^1)$ to zeros~\cite{tretschk2021non} and do not optimize the deformations at $t{=}1$. 
After reconstructing time $t$, we fix $\theta_c^t$ and proceed with $t{+}1$, where we initialize $\theta_c^{t{+}1}$ with $\theta_c^t$. 

\begin{figure}
    \centering
    \includegraphics[width=\columnwidth]{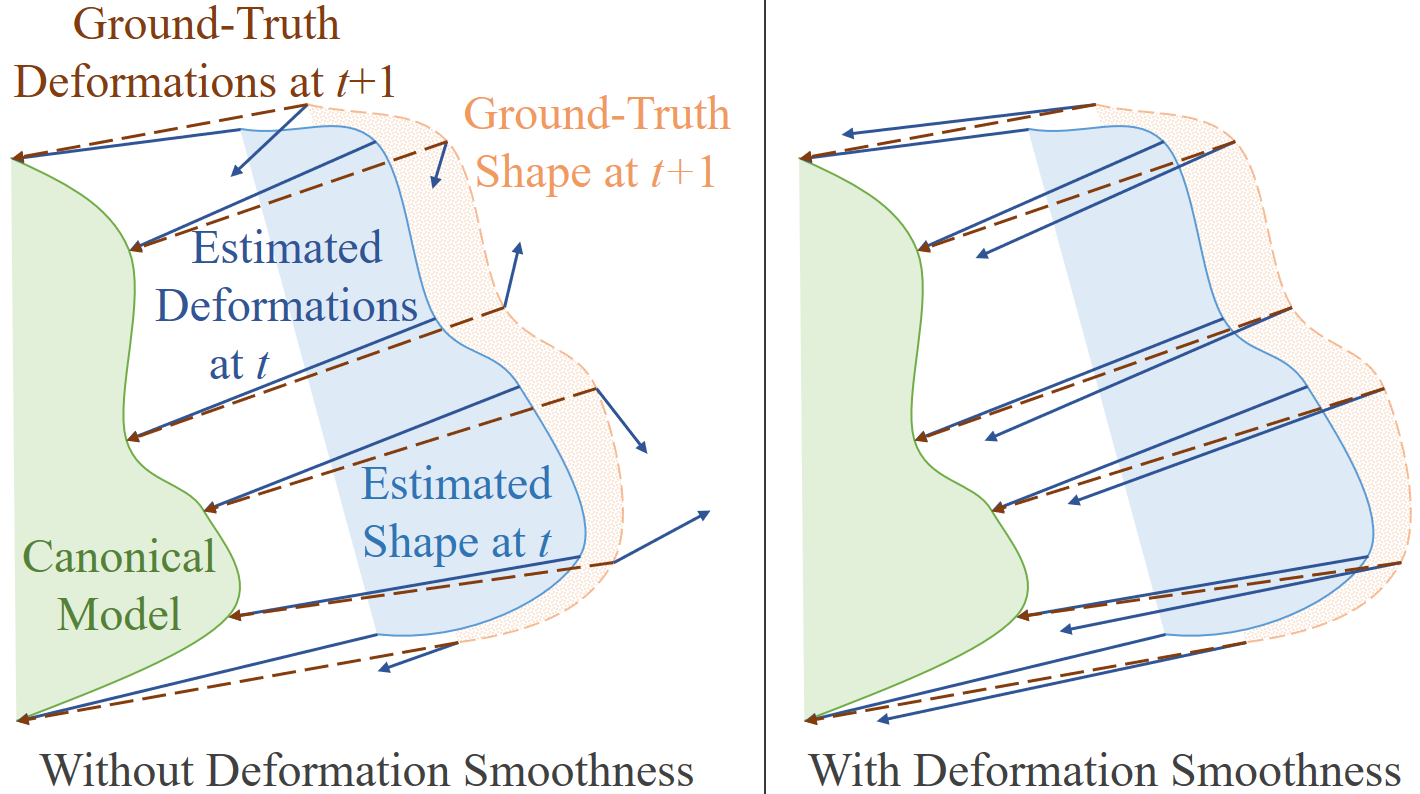}
    \caption{
    \textbf{Extending the Deformation Field for Tracking.} 
    Without smoothness, the estimated backwards deformations at time $t$ give a bad initialization for $t{+}1$. 
    Smoothness initializes the deformations closer to the ground truth. 
    }
    \label{fig:extending}
\end{figure}

\noindent\textbf{Extending the Deformation Field.} 
Unfortunately, na\"ively employing a HashMLP for $d_c$ fails to reconstruct the scene for any deformed state that significantly differs from the canonical model. 
This is because, at time $t{+}1$, the dynamic object resides in a slightly different place in world space, part of which was empty space at time $t$ and is hence not well initialized by $d_c(\cdot;\theta_c^t)$. 
This becomes especially prevalent when the object has undergone large motion. 
We propose to mitigate this failure of the backward model by extending the deformations into the area surrounding the dynamic object at time $t$, as Fig.~\ref{fig:extending} illustrates. 

To this end, we employ two means: 
(1)~We find that reducing the resolution of the coarsest grid of the deformation hash grid to $32^3$ stabilizes tracking. 
We hypothesize that when the object enters a previously empty voxel in a fine grid, the uninitialized latent codes of this voxel are too quickly too influential in the trilinear interpolation of the latent codes. 
These uninitialized latent codes thus do not obtain the right values before being relevant but rather negatively impact the deformations, leading to artifacts and a lack of large-scale smoothness. 
(2)~However, this structural change gives an unpredictable, badly-controlled smoothness, similar to an MLP. 
We encourage well-behavedness via a smoothness loss, which uses the inherent smoothness of the coarse grid to propagate its influence. %

\noindent\textbf{Smoothness Loss.} 
To stabilize the tracking, we propose to impose a smoothness loss on the deformations. 
Because our deformation model is continuous and fully differentiable, we do not need to discretize the loss. 
Instead, we influence the local behavior directly via the (spatial) Jacobian $\mathbf{J}\in\mathbb{R}^{3\times 3}$ of the deformations: $\mathbf{J} = \mathbf{J}_{\mathbf{x}} = \frac{\partial d_c(\mathbf{x};\theta_t)}{\partial \mathbf{x}}$. 
As is common for general reconstruction methods~\cite{TretschkNonRigidSurvey}, we assume the object to deform locally in an as-rigid-as-possible manner~\cite{sorkine2007rigid}. 
Specifically, we take inspiration from Nerfies's elastic loss on $\mathbf{J}$ \cite{park2021nerfies}. 
However, their loss is computationally expensive because it needs to compute all rows of $\mathbf{J}_{\mathbf{r}_r(s_i)}$ for each sample $\mathbf{r}_r(s_i)$ (which takes three backward passes during the forward pass of the loss computation) and then performs an SVD of $\mathbf{J}_{\mathbf{r}_r(s_i)}$. 
Denoting the identity matrix by $\mathbf{I}$, we can avoid the SVD and allow for trivial computation of gradients by relaxing the constraint from the rotation group $\mathit{SO}(3)=\{\mathbf{A}\in\mathbb{R}^{3 \times 3} | \mathbf{A}^\top\mathbf{A}{=}\mathbf{I}, \det{\mathbf{A}}{=}1\}$ to the orthogonal group $\mathit{O}(3)=\{\mathbf{A}\in\mathbb{R}^{3 \times 3} | \mathbf{A}^\top\mathbf{A}{=}\mathbf{I}\}$, which additionally allows reflections since $\det{\mathbf{A}}{=}\pm1$ for $\mathbf{A}\in \mathit{O}(3)$. 
We can then encourage rigidity via: 
\begin{equation}
    \mathcal{L}_{\text{rigid}} = \frac{1}{9RS} \sum_r \sum_i \sum_{j,k} \big\lvert (\mathbf{J}_{\mathbf{r}_r(s_i)}^\top \mathbf{J}_{\mathbf{r}_r(s_i)} - \mathbf{I})_{j,k} \big\rvert,
\end{equation} 
where $\mathbf{A}_{j,k}\in\mathbb{R}$ is the entry of matrix $\mathbf{A}$ at index $(j,k)$. %

We can compute this prior even faster by reducing the need for three backward passes to just one, lowering overall training time by a factor of $2{-}3$. 
To this end, we first note that norm preservation is an equivalent definition of $\mathit{O}(3)$: 
\begin{equation}
    \mathbf{A} \in \mathit{O}(3) \iff \forall \mathbf{e}\in\mathbb{R}^3: \lVert \mathbf{A}\mathbf{e} \rVert_2 = \lVert \mathbf{e}\rVert_2.
\end{equation} 
We next exploit the fact that automatic differentiation~\cite{speelpenning1980compiling,griewank2008evaluating,paszke2017automatic} computes Jacobian-vector products $\mathbf{J}^\top \mathbf{e}$ in a single backward pass, where $\mathbf{e}$ is an arbitrary vector. 
Since $\mathbf{J}^\top = \mathbf{J}$ for $\mathbf{J}\in\mathit{O}(3)$, we can compute the norm of $\mathbf{J}\mathbf{e}$ as $\lVert \mathbf{J} \mathbf{e} \rVert_2 = \lVert \mathbf{J}^\top \mathbf{e} \rVert_2$. 
The following loss then encourages norm preservation and only requires one backward pass: %
\begin{equation}
    \mathcal{L}_{\text{norm}} = \frac{1}{RS} \sum_r \sum_i \mathbb{E}_{\mathbf{e}}\Big[\big\lvert \lVert \mathbf{J}_{\mathbf{r}_r(s_i)}^\top \mathbf{e} \rVert_2 - 1 \big\rvert\Big],
\end{equation}
where $\mathbf{e}$ is distributed uniformly on the unit sphere and hence $\lVert\mathbf{e}\rVert_2=1$. 
We note that it is sufficient to only consider vectors on the unit sphere due to linearity. 
In practice, we approximate this expectation with one random sample. 

\noindent\textbf{Weighting the Smoothness Loss.} 
To extend the deformation field as discussed earlier, we could na\"ively encourage smoothness uniformly everywhere. 
However, this restricts empty space too much, leading it to push back against object deformations. 
We thus focus on the object by weighting $\mathcal{L}_{\text{norm}}$ by $\hat{\sigma}_{r,i}=\exp(-\sigma_{r,i}\delta)$, where $\sigma_{r,i}$ is $\sigma_{i}$ of the $r$-th ray and we do not backpropagate into the opacity~\cite{tretschk2021non}. %

We now need to regularize the space \emph{around} the object.
However, this space is hard to locate efficiently and we approximate it by max-pooling over $\{\hat{\sigma}_{r,i}\}_i$ along ray $r$, with a window size of $1\%$ of the ray length. %

This still makes the space around the object too stiff. 
We thus weaken the regularization by dividing the resulting weight by $u$ if $\hat{\sigma}_{r,i}$ and the weight after max-pooling differ by at least a factor of $u$. 
We empirically set $u{=}10$. 

Finally, regularizing very small offsets, which are widespread during early timestamps, tends to collapse %
the network. 
We hence only regularize deformations that are not very small. 
We denote this weighted loss as $\mathcal{L}_{\text{norm,w}}(\theta_c^t)$. 
The supplement contains a full mathematical description. 

\noindent\textbf{Fine Deformations.} 
In addition to enabling tracking by extending the deformation field, we also use the smoothness loss to stabilize the surface by strongly regularizing its deformations $\Delta_c$.  
However, this leads to a loss of detail because $\Delta_c$ can now only represent \emph{coarse} deformations. 
To counteract this, we add \emph{fine} deformations $\Delta_f(\cdot;\theta_f^t)$ on top, after normalizing out (\emph{i.e.}, applying) the coarse deformations: $\mathbf{x}''=d_f(\mathbf{x}';\theta_f^t)=\mathbf{x}' + \Delta_f(\mathbf{x}';\theta_f^t)$, where $\mathbf{x}'=d_c(\mathbf{x};\theta_c^t)$ for any 3D point $\mathbf{x}$ in world space and we query $m$ at $\mathbf{x}''$. 
Crucially, we allow for details by using a much weaker weight $\lambda_{\text{fine}}$ for $\mathcal{L}_{\text{norm,w}}(\theta_f^t)$, where the Jacobian is w.r.t.\ $\mathbf{x}'$. 
We parametrize $\Delta_f$ with a HashMLP. 

\noindent\textbf{Frame-Wise Tracking Revisited.} 
We apply tracking to the coarse and fine deformations as follows:
At $t{=}1$, we initialize the last layers of $\Delta_c$ and $\Delta_f$ to zeros and do not optimize for the deformations. 
At any $t{>}1$, we initialize $\theta_c^t$ with the final $\theta_c^{t{-}1}$, and optimize for $\Delta_c(\cdot;\theta_c^t)$ for $5k$ iterations while setting $\Delta_f=\mathbf{0}$. 
We then fix $\theta_c^t$, initialize $\theta_f^t$ with the final $\theta_f^{t{-}1}$, and optimize for $\Delta_f(\cdot;\theta_f^t)$ for $5k$ iterations. 

With $\lambda_{\text{coarse}},\lambda_{\text{fine}}\in\mathbb{R}$, the total loss for any $t{>}1$ is: 
\begin{equation}
    \mathcal{L}_{\text{time}} = \mathcal{L}_{\text{rec}} + \lambda_{\text{coarse}} \mathcal{L}_{\text{norm,w}}(\theta_c^t) + \lambda_{\text{fine}} \mathcal{L}_{\text{norm,w}}(\theta_f^t).
\end{equation}

\subsection{Implementation Details}

\begin{figure}
    \centering
    
    \begin{subfigure}{0.22\columnwidth}
    \includegraphics[trim={400 500 700 100},clip,width=\textwidth]{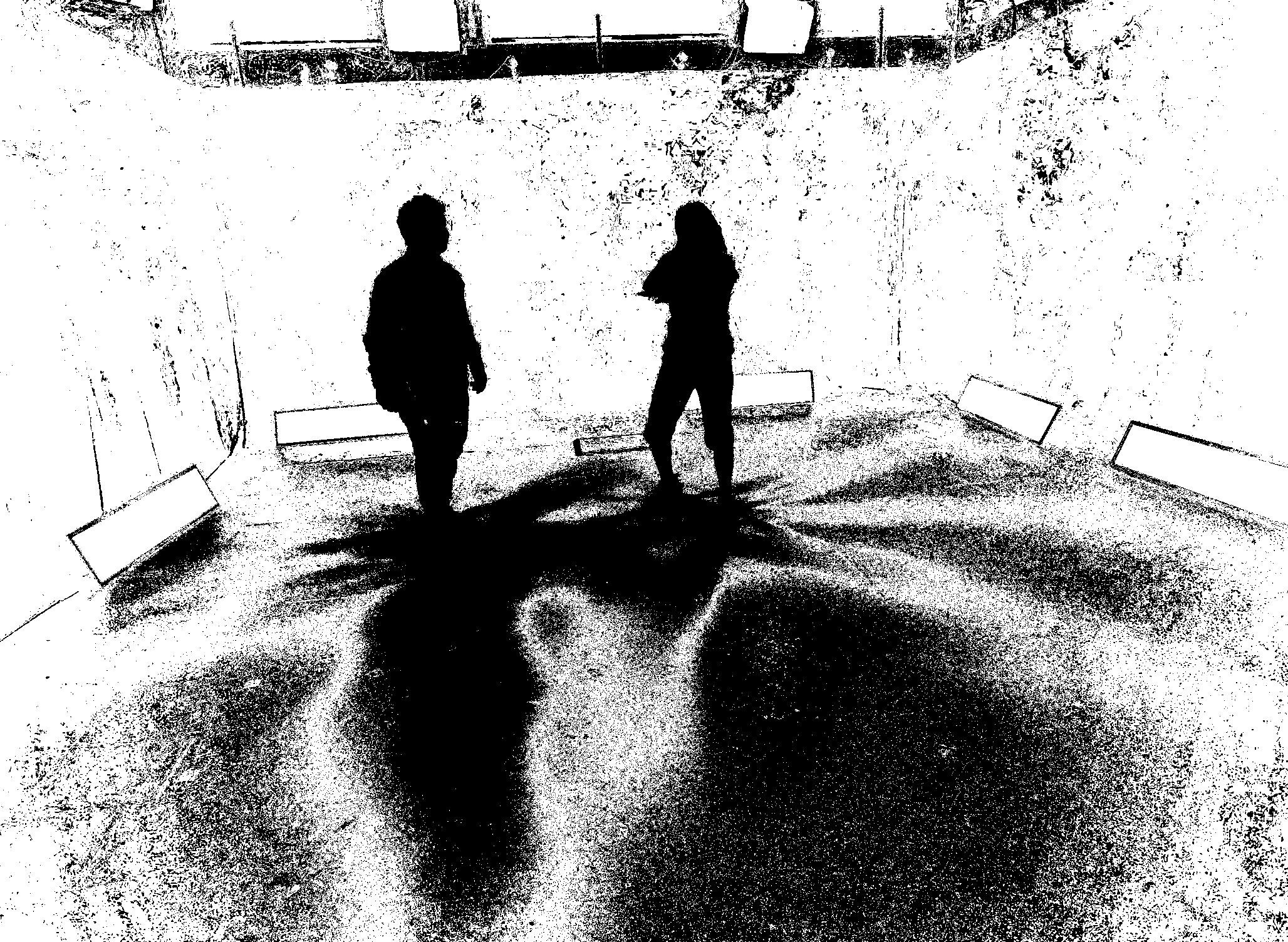}
    \caption{}
    \end{subfigure}
    \begin{subfigure}{0.22\columnwidth}
    \includegraphics[trim={330 100 330 100},clip,width=\textwidth]{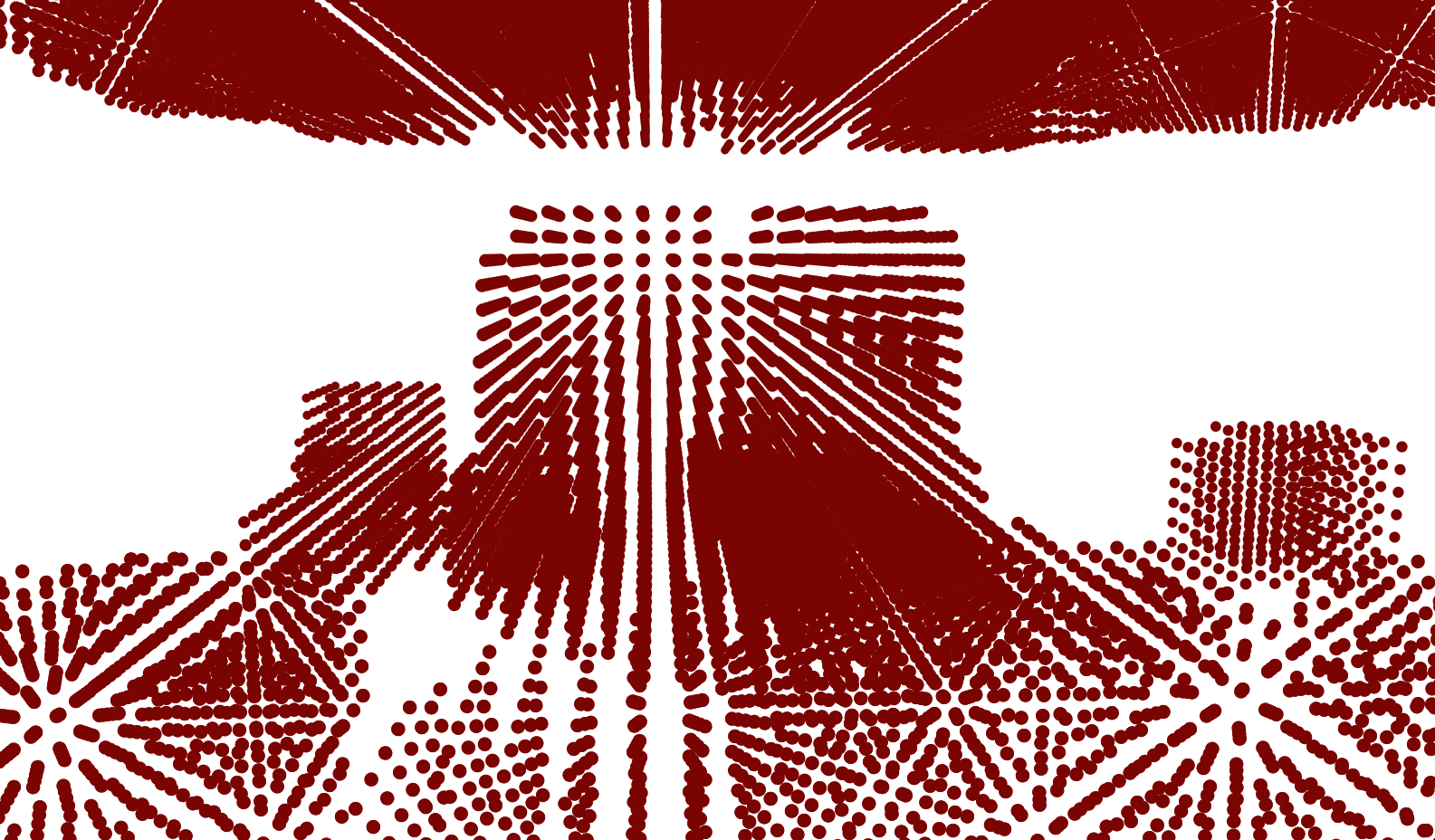}
    \caption{}
    \end{subfigure}
    ~
    \includegraphics[width=0.0065\columnwidth]{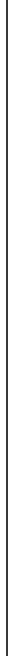}
    ~
    \begin{subfigure}{0.22\columnwidth}
    \includegraphics[trim={0 300 850 270},clip,width=\textwidth]{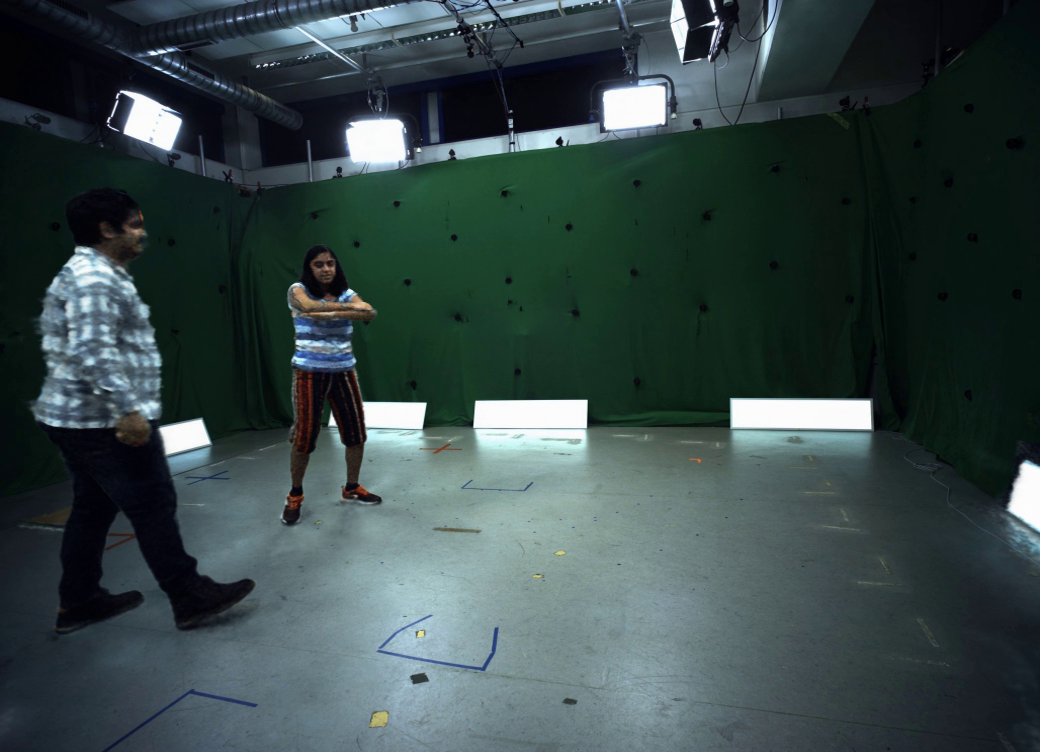}
    \caption{}
    \end{subfigure}
    \begin{subfigure}{0.22\columnwidth}
    \includegraphics[trim={0 300 850 270},clip,width=\textwidth]{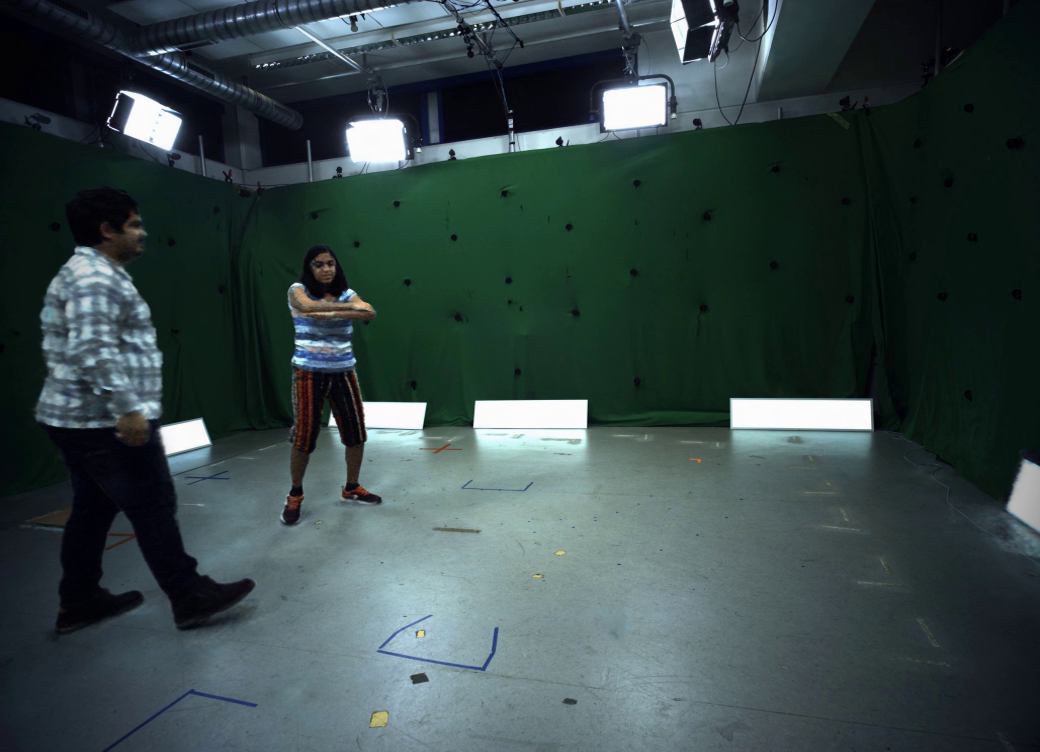}
    \caption{}
    \end{subfigure}
    \caption{\textbf{Implementation.} (a) Foreground $\mathbf{F}^{c,t}$. (b) Pruning grid for (a). (c) W/o and (d) w/ vignetting correction. %
    }
    \label{fig:implementation}
    \vspace{-2em}
\end{figure}

We next describe how we can speed up our method and correct vignetting effects, and provide optimization details. 
Fig.~\ref{fig:implementation} visualizes some of these methods.

\noindent\textbf{Foreground-Focused Batches.} 
To speed up training, we focus the batches on the foreground. 
We first use background subtraction w.r.t.\ $\mathbf{B}^{c}$ to get a rough foreground mask $\mathbf{F}^{c,t}$ for each $\mathbf{I}^{c,t}$. 
We then pick $80\%$ of the rays in the batch from the foreground and the rest from the background. 

\noindent\textbf{Pruning.} 
To speed up rendering, we prune any sample $\mathbf{r}(s_i)$ in world space that does not contain any foreground. 
To determine this for time $t$, we query a binary voxel grid $g_t$ that we overlay over the scene. 
$g_t$ is $1$ for voxels that are potentially in the foreground and $0$ otherwise. 
We fill $g_t$ via space carving~\cite{kutulakos2000theory} with $\{\mathbf{F}^{c,t}\}_c$. 
To be conservative and to let the deformation field extend into the surrounding area, we dilate both the foreground masks and the foreground in the resulting voxel grid. 
To clear out geometry artifacts in empty space, we do not prune when constructing the canonical model, only when optimizing for $t{>}1$. 
Pruning removes ${\sim}80\%$ of samples, making our method four times faster.

\noindent\textbf{Vignetting Correction.} 
Cameras collect less light around the image border, which leads to darkening in the input images. 
We model this vignetting with a radial model \cite{stumpfel2006direct,bal2023image}: 
\begin{equation}
    \mathbf{C}_{vig}(\mathbf{r}) = \mathbf{C}(\mathbf{r})(1 + k_1 p(\mathbf{r}) + k_2 p(\mathbf{r})^2 + k_3 p(\mathbf{r})^3),
\end{equation}
where $p(\mathbf{r})\in\mathbb{R}$ is the squared distance to the camera center in pixel space (we divide the distance by $w$ for resolution invariance), and $k_1,k_2,k_3\in\mathbb{R}$ are correction parameters. 
We initialize these parameters to zeros, optimize for them when constructing the canonical model, and keep them fixed for $t{>}1$. 
We share the same set of parameters across all cameras and timestamps. 
We use $\mathbf{C}_{vig}$ in place of $\mathbf{C}$ in Eq.~\eqref{eq:renderback}. 
The quality improves since the canonical model no longer needs to use artifacts to account for vignetting effects. 

\noindent\textbf{Optimizer.} 
We use AdamW~\cite{loshchilov2018decoupled} with weight decay of $0.01$ to stabilize the training. 
Auto-decoded parameters like hash grids get sparsely non-zero gradients in any given training iteration, which degrades the momentum accumulation. 
We thus use a modified version of AdamW: rather than treating all parameters $\{\theta_i\in\mathbb{R}\}_i$ in a tensor the same, we separately keep track of AdamW's parameters $\theta^\mathit{opt}_i$ (\textit{e.g.} the number of iterations) for each individual parameter $\theta_i$. 
In any given iteration, only $\theta_i$ whose derivative is non-zero have their AdamW parameters $\theta^\mathit{opt}_i$ updated and AdamW applied. 
The supplement describes the learning rates in detail. 

\noindent\textbf{Hyperparameters.} 
All scenes use the same settings: a batch size of $R{=}1024$,  $S{=}3072$ samples per ray, and loss weights $\lambda_{\text{back}}{=}0.001$, $\lambda_{\text{hard}}{=}1$, $\lambda_{\text{coarse}}{=}1000$, and $\lambda_{\text{fine}}{=}30$. 
We use LeakyReLUs~\cite{maas2013rectifier} for the deformations and ReLUs for the canonical model. 
The supplement has more details.

\noindent\textbf{Code.} 
We use PyTorch~\cite{NEURIPS2019_9015} and tiny-cuda-nn~\cite{tiny-cuda-nn} via its Python wrappers for its fast implementation of hash grids. %

\section{Experiments}\label{sec:experiments} 

We evaluate the reconstruction quality and time consistency of our method, perform ablations, show simple scene editing, and discuss limitations and future work. 
The supplemental video contains many additional results. %

\noindent\textbf{Prior Work.} 
We compare with all prior time-consistent general NeRF methods: NR-NeRF~\cite{tretschk2021non}, D-NeRF~\cite{pumarola2021d}, and DeVRF~\cite{liu2022devrf}. 
To evaluate time consistency, we compare against PREF~\cite{uii_eccv22_pref}, a NeRF-based method to estimate correspondences. 

\noindent\textbf{Variants.}
We evaluate reconstruction quality by using novel-view synthesis \emph{as a proxy}. 
Unlike most works, we target \emph{time-consistent} reconstruction and not novel-view synthesis, which leads to a trade-off with novel-view quality. 

Most dynamic-NeRF papers condition the canonical model on time. 
Therefore, we evaluate two variants of our method that do the same and, hence, synthesize better novel views at the cost of correspondences. 
The first variant, \emph{SNF-A}, only makes the appearance time-varying, while the second variant, \emph{SNF-AG}, makes the appearance and canonical geometry time-varying. 
The supplement has further details. 

We emphasize that we do not claim that these variants are competitive with state-of-the-art 4D reconstruction methods that neglect correspondences and focus solely on novel-view synthesis. 
Our design is not tailored to this setting. 

\noindent\textbf{Data.} 
For ease of recording, we evaluate our \emph{general} method on real-world scenes of one or two people. 
Crucially, the recordings also contain a plush dog and loose clothing. 
Furthermore, our method reconstructs scenes with multiple people \emph{without any knowledge that there are multiple entities in the scene nor that they are human}. 
These aspects demonstrate its generality. 
We use a total of $C{=}117$ static cameras in a studio to record six scenes of $4{-}5$sec and one of $12$sec at $25$fps. 
For the test sets, we pick the same two cameras for all scenes. 
We train at an input resolution of $h{=}1504$ and $w{=}2056$.  
In addition, we consider the short multi-view scene from NR-NeRF~\cite{tretschk2021non}, which contains $16$ camera pairs. 

\noindent\textbf{Runtime.} 
On an Nvidia A100 GPU, the canonical model trains at $8$iter/sec, the coarse deformations at $16$iter/sec, and the fine deformations at $14$iter/sec. 
For efficiency, we render all images at half the input resolution, at $10{-}15$sec/image.

\begin{figure}
    \centering
    \footnotesize
    \setlength{\tabcolsep}{1pt}
    \begin{tabular}{cccccc}

    & Ground Truth & Ours & NR-NeRF & SNF-A & SNF-AG\\
    
    \parbox[t]{2mm}{\rotatebox[origin=c]{90}{Rendered RGB~}}
    &
    \raisebox{-0.5\height}{\includegraphics[trim={190 180 520 210},clip,width=0.19\columnwidth]{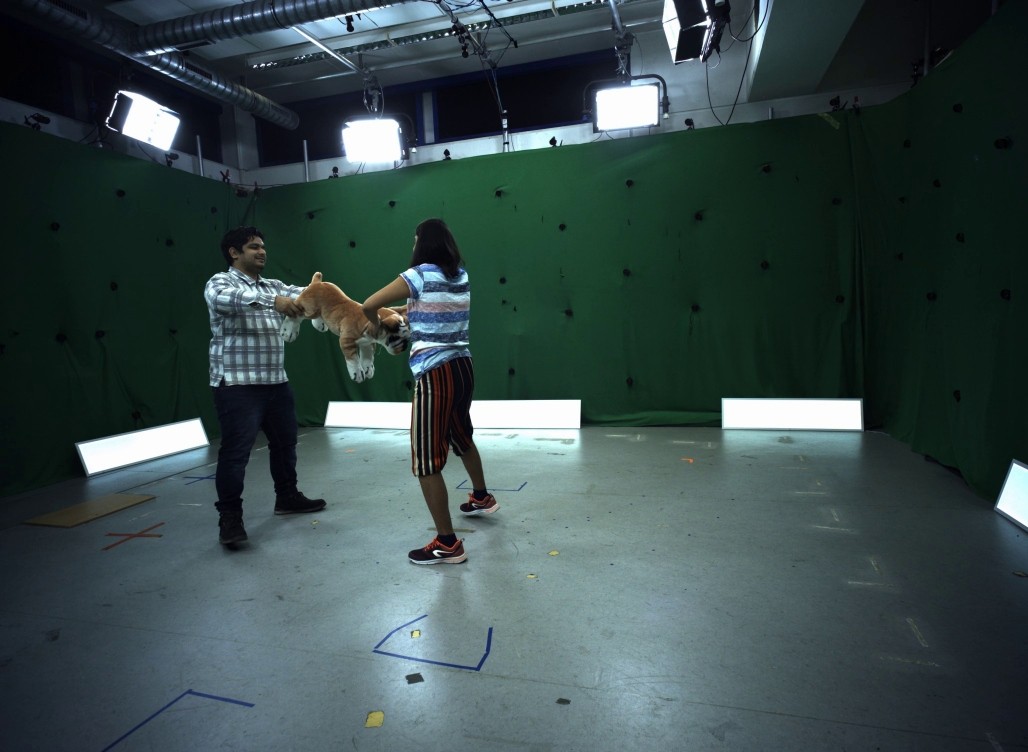}}
    &
    \raisebox{-0.5\height}{\includegraphics[trim={190 180 520 210},clip,width=0.19\columnwidth]{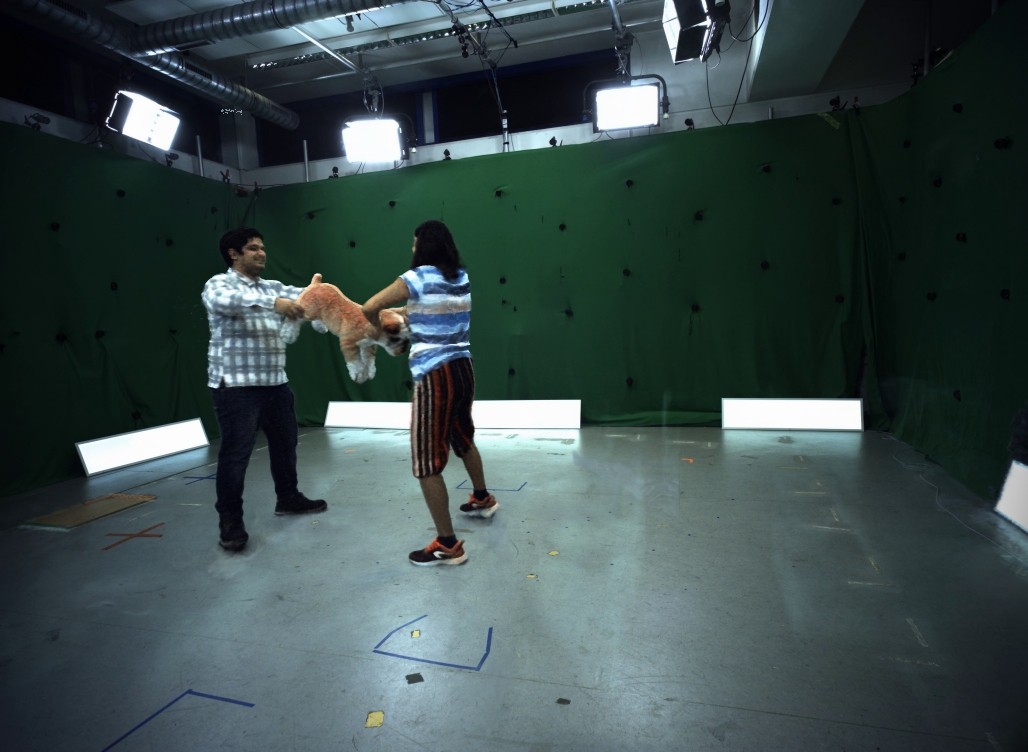}}
    &
    \raisebox{-0.5\height}{\includegraphics[trim={190 180 520 210},clip,width=0.19\columnwidth]{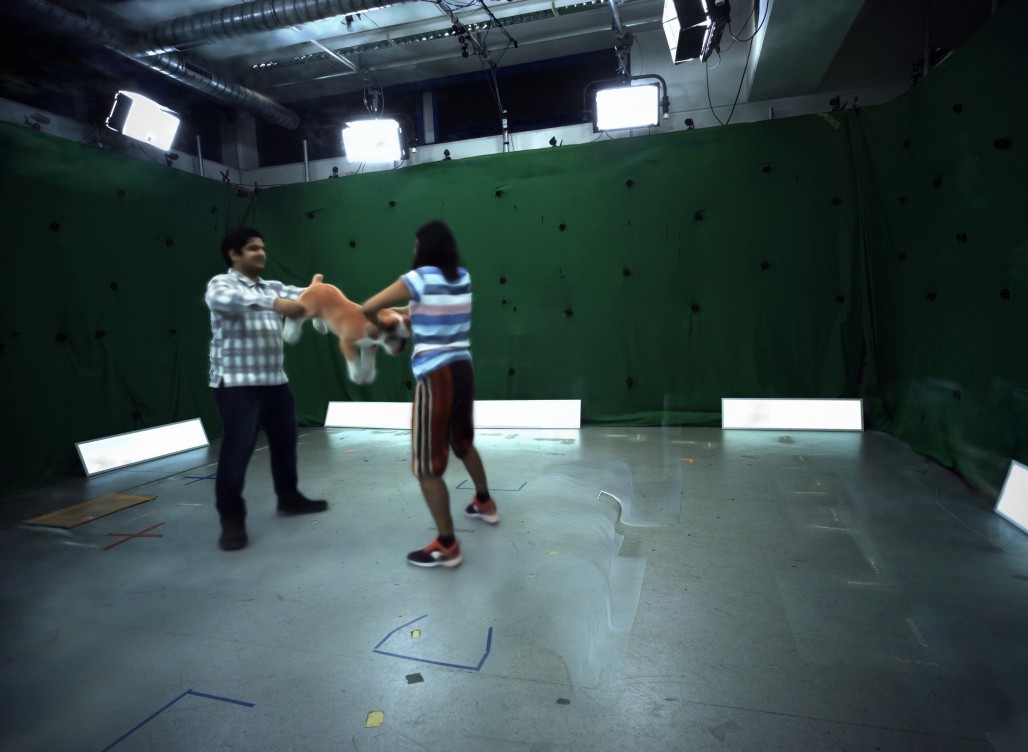}}
    &
    \raisebox{-0.5\height}{\includegraphics[trim={190 180 520 210},clip,width=0.19\columnwidth]{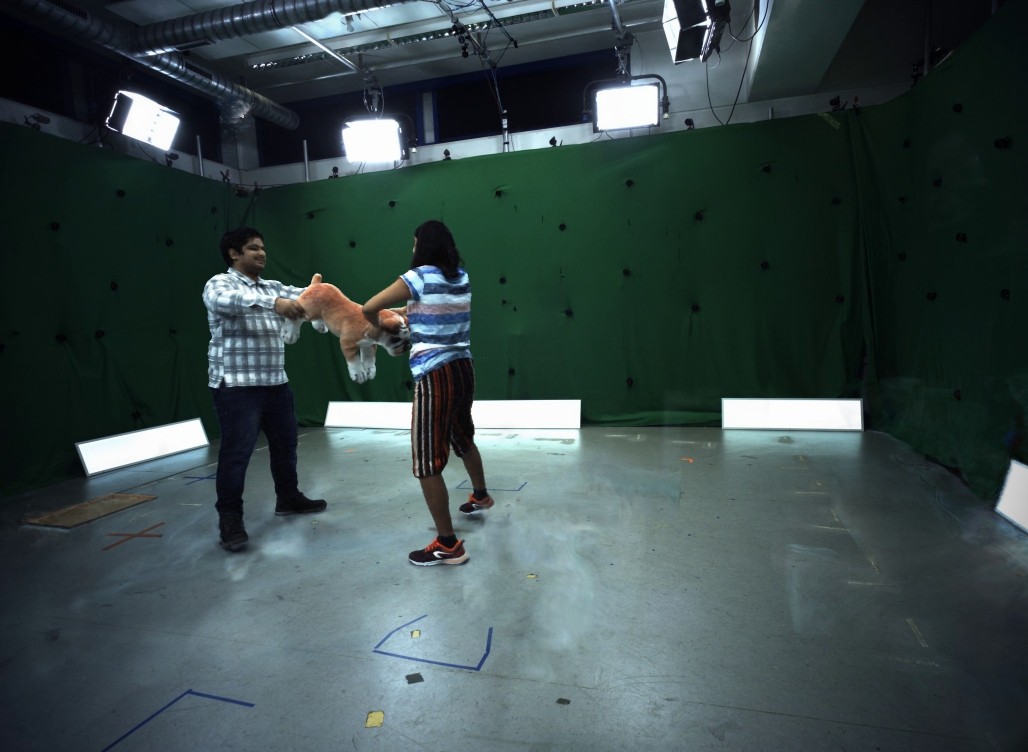}}
    &
    \raisebox{-0.5\height}{\includegraphics[trim={190 180 520 210},clip,width=0.19\columnwidth]{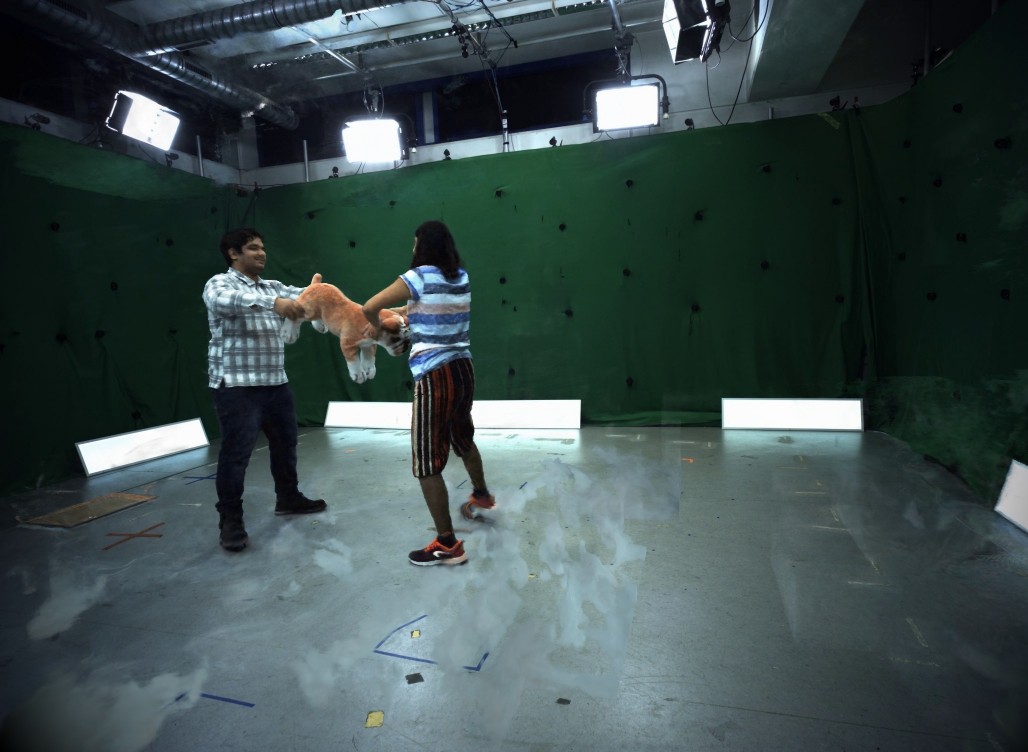}}
    \\

    \parbox[t]{2mm}{\rotatebox[origin=c]{90}{Rendered Depth~}}
    &
    ---
    &
    \raisebox{-0.5\height}{\includegraphics[trim={190 180 520 210},clip,width=0.19\columnwidth]{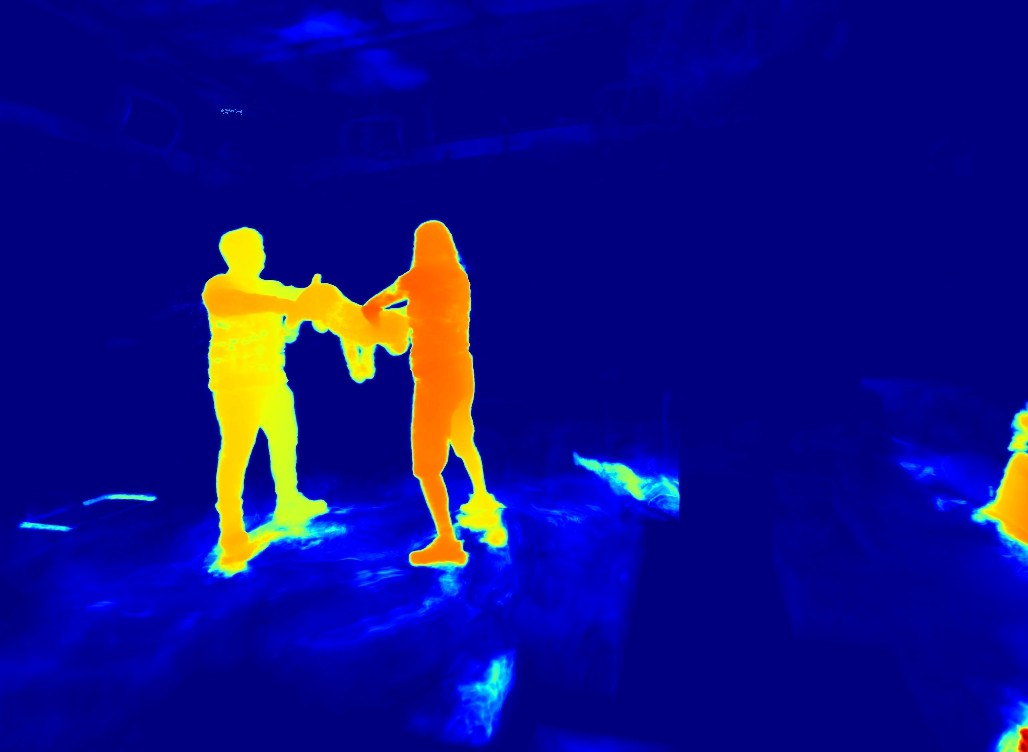}}
    &
    \raisebox{-0.5\height}{\includegraphics[trim={190 180 520 210},clip,width=0.19\columnwidth]{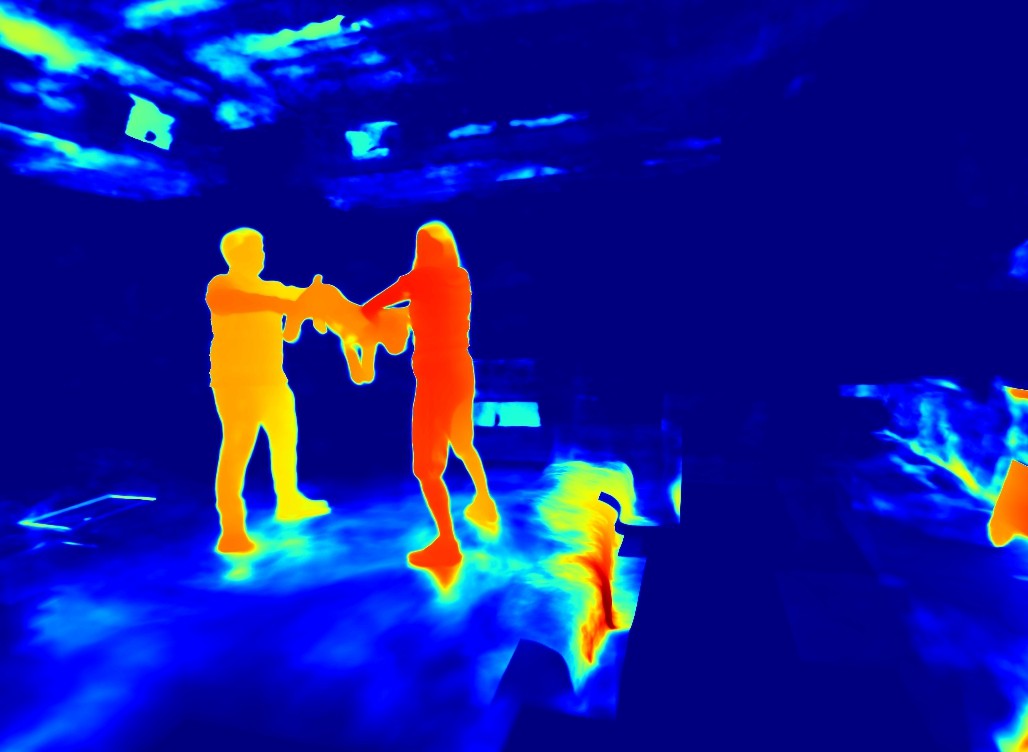}}
    &
    \raisebox{-0.5\height}{\includegraphics[trim={190 180 520 210},clip,width=0.19\columnwidth]{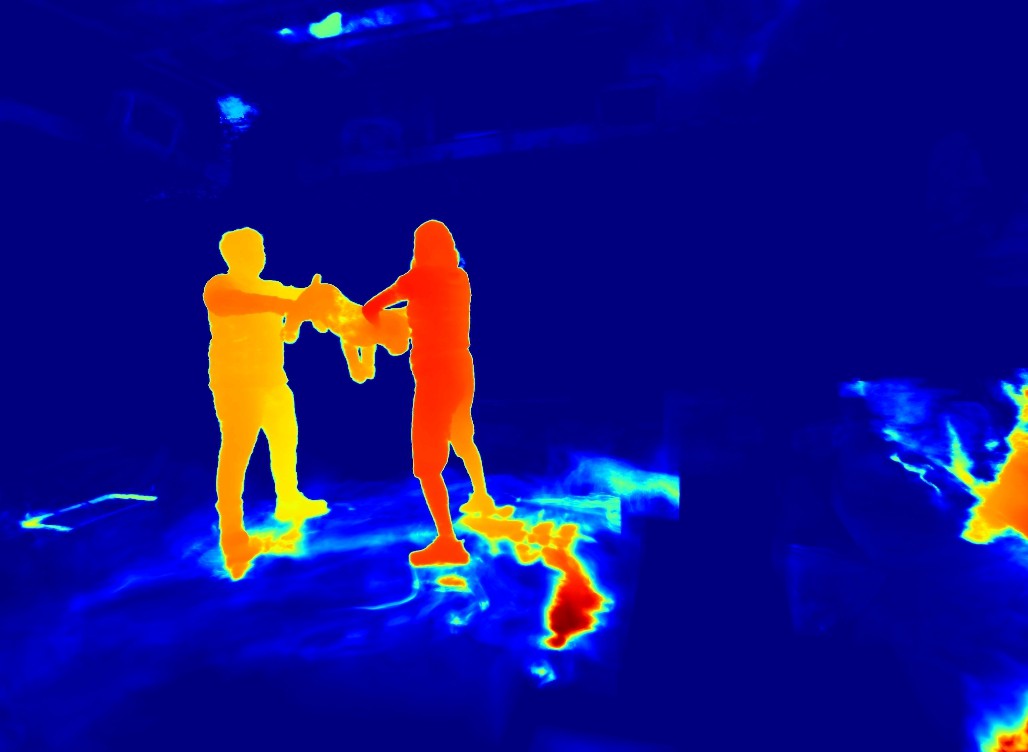}}
    &
    \raisebox{-0.5\height}{\includegraphics[trim={190 180 520 210},clip,width=0.19\columnwidth]{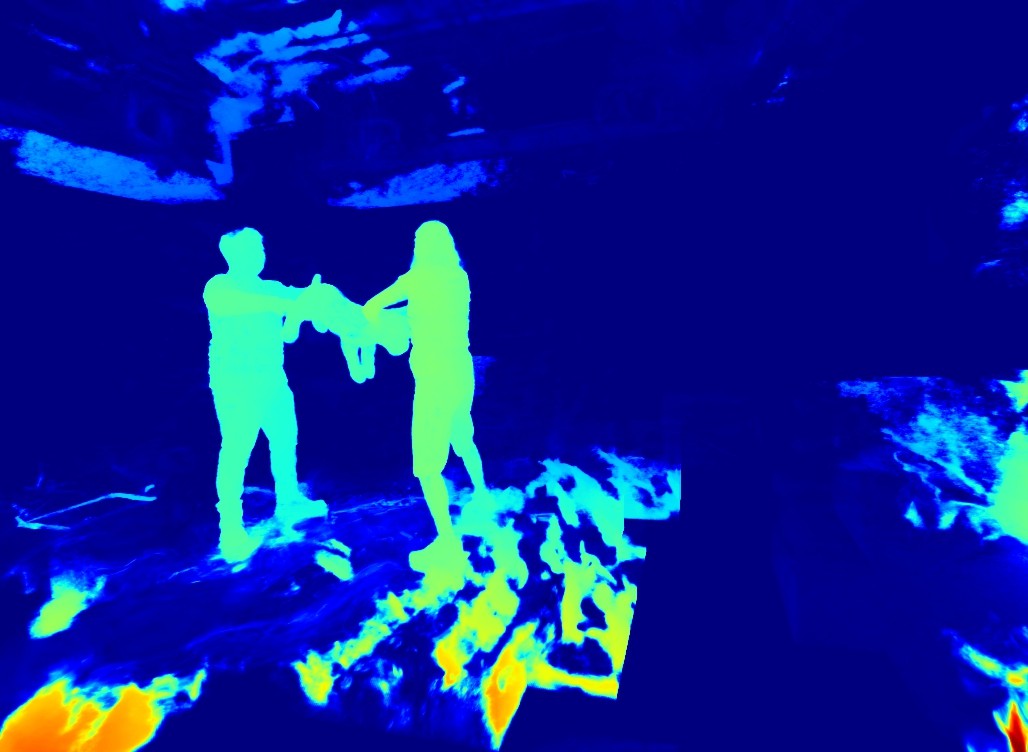}}
    \\
    
    \hline
    \hline

    \parbox[t]{2mm}{\rotatebox[origin=c]{90}{Rendered RGB~~~}}
    &
    \raisebox{-0.5\height}{\includegraphics[trim={430 220 380 230},clip,width=0.19\columnwidth]{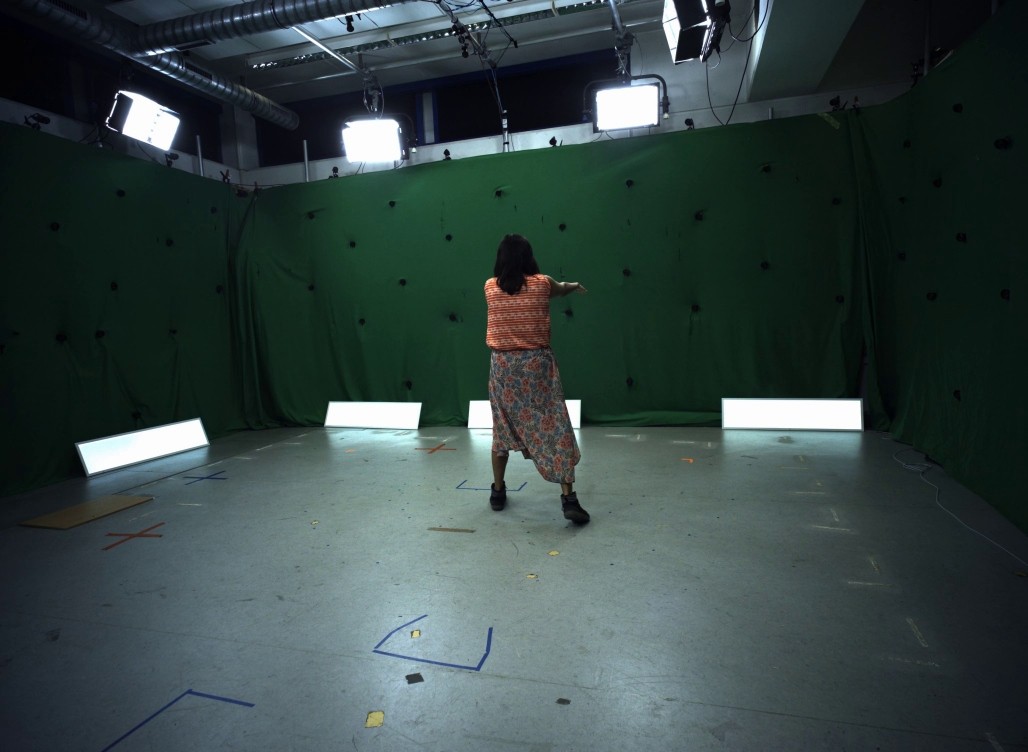}}
    &
    \raisebox{-0.5\height}{\includegraphics[trim={430 220 380 230},clip,width=0.19\columnwidth]{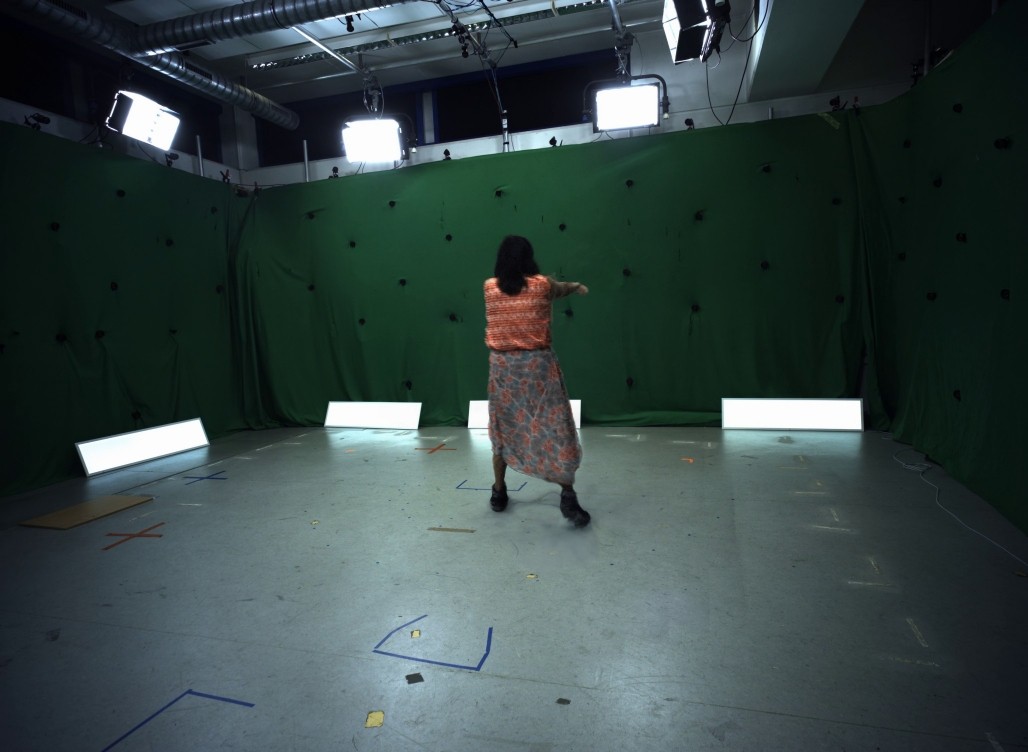}}
    &
    \raisebox{-0.5\height}{\includegraphics[trim={430 220 380 230},clip,width=0.19\columnwidth]{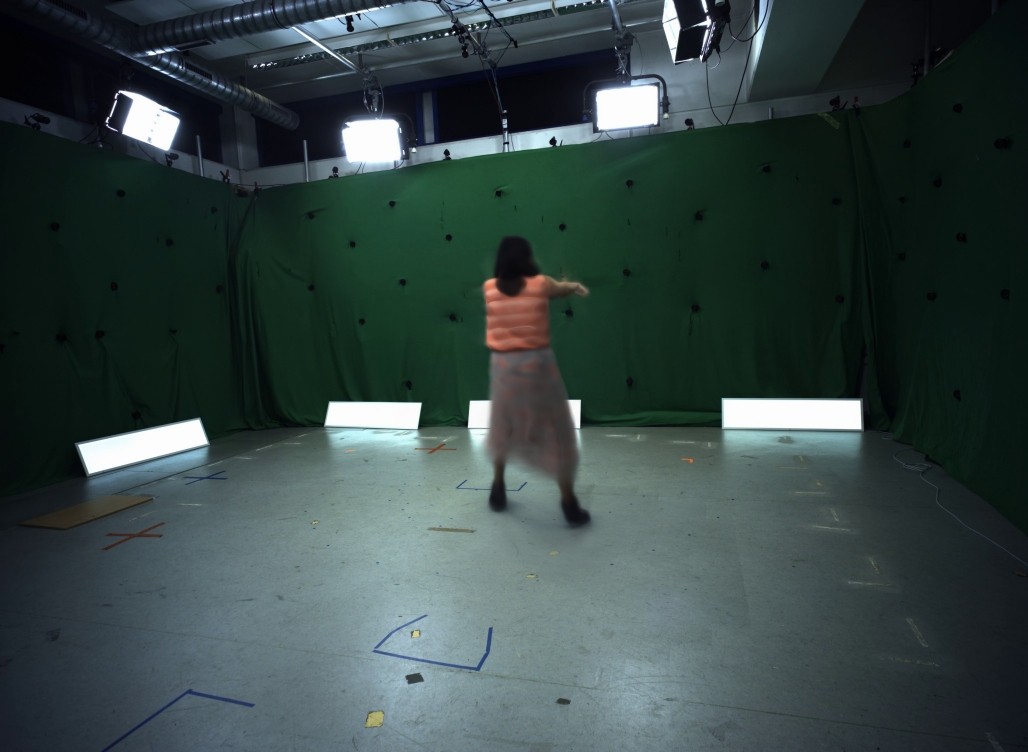}}
    &
    \raisebox{-0.5\height}{\includegraphics[trim={430 220 380 230},clip,width=0.19\columnwidth]{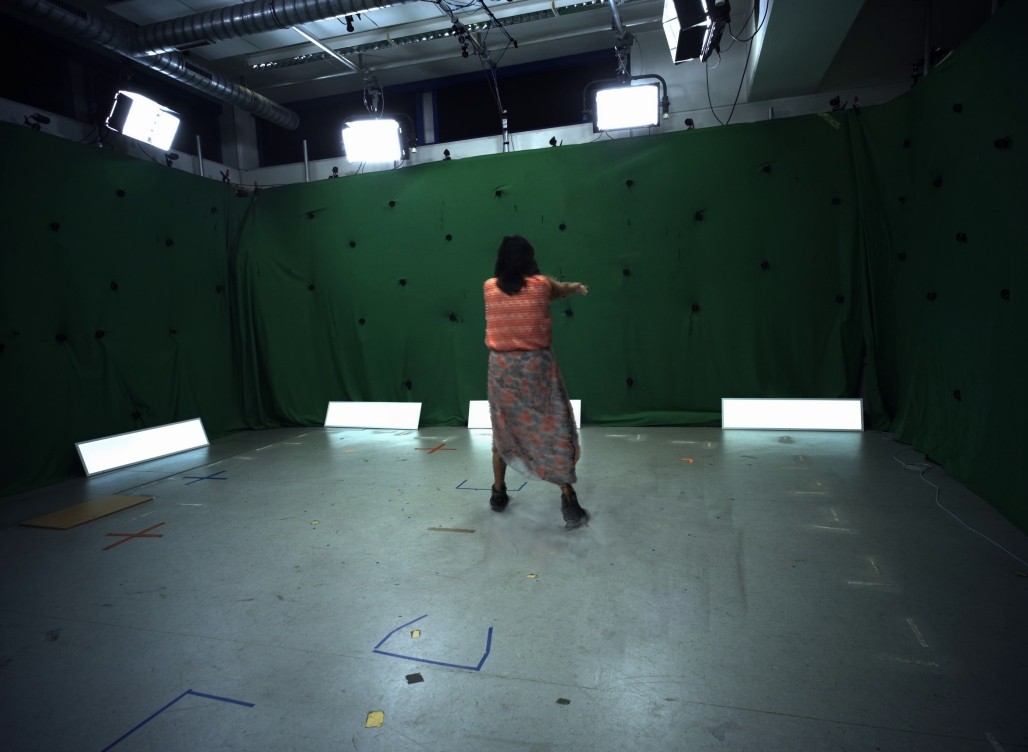}}
    &
    \raisebox{-0.5\height}{\includegraphics[trim={430 220 380 230},clip,width=0.19\columnwidth]{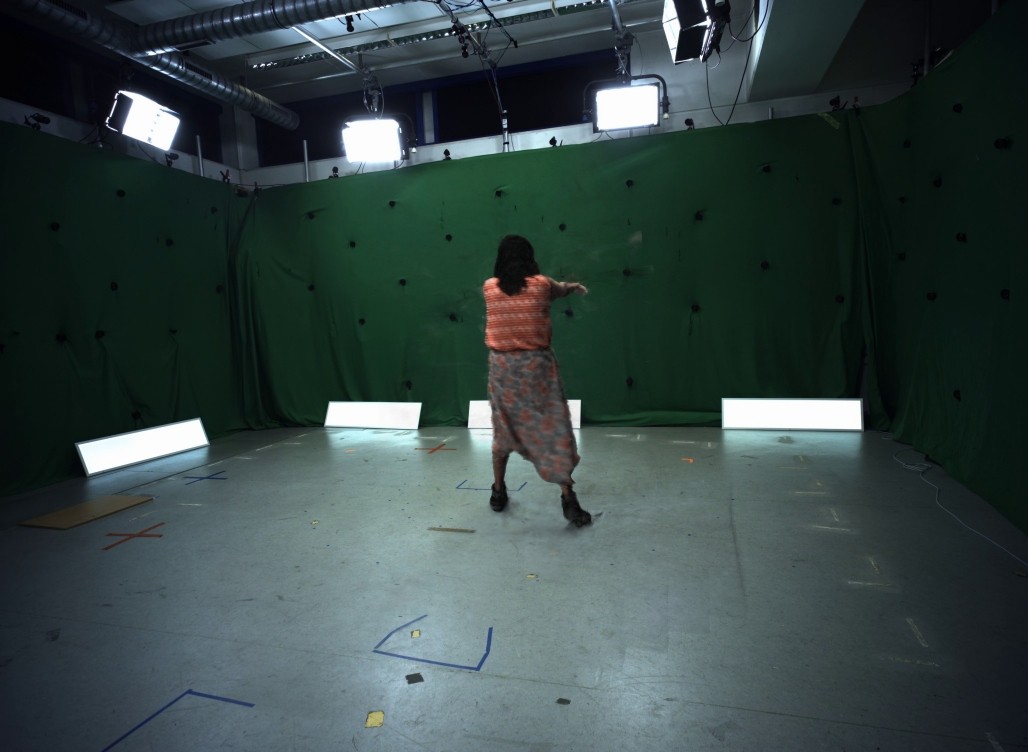}}
    \\

    \parbox[t]{2mm}{\rotatebox[origin=c]{90}{Rendered Depth~~}}
    &
    ---
    &
    \raisebox{-0.5\height}{\includegraphics[trim={430 220 380 230},clip,width=0.19\columnwidth]{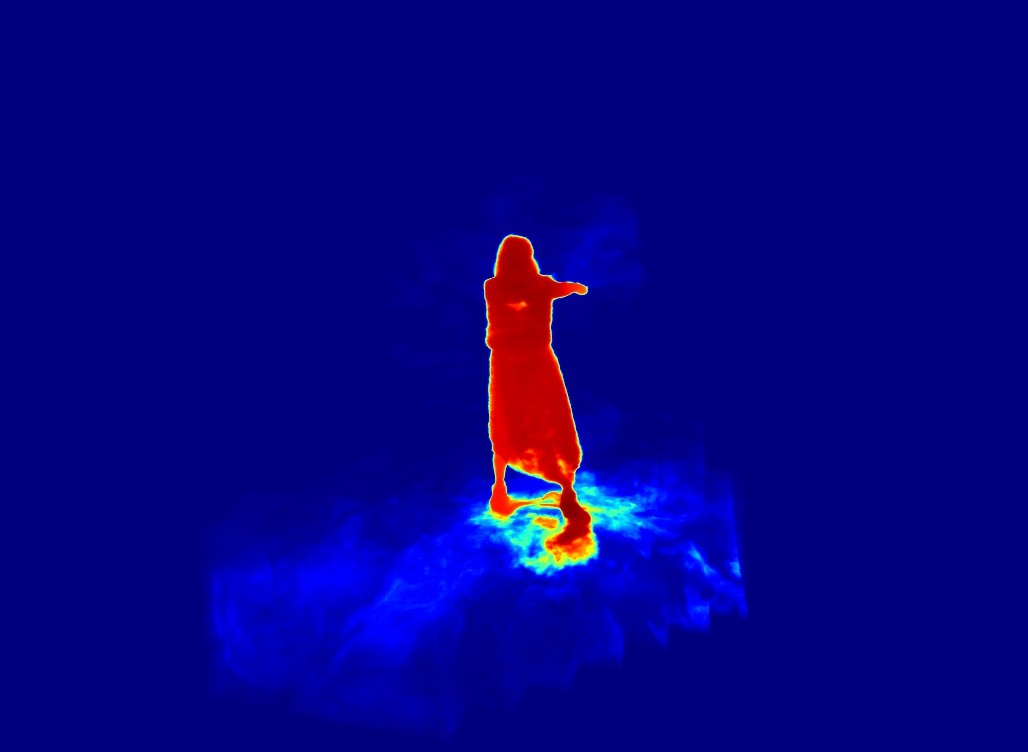}}
    &
    \raisebox{-0.5\height}{\includegraphics[trim={430 220 380 230},clip,width=0.19\columnwidth]{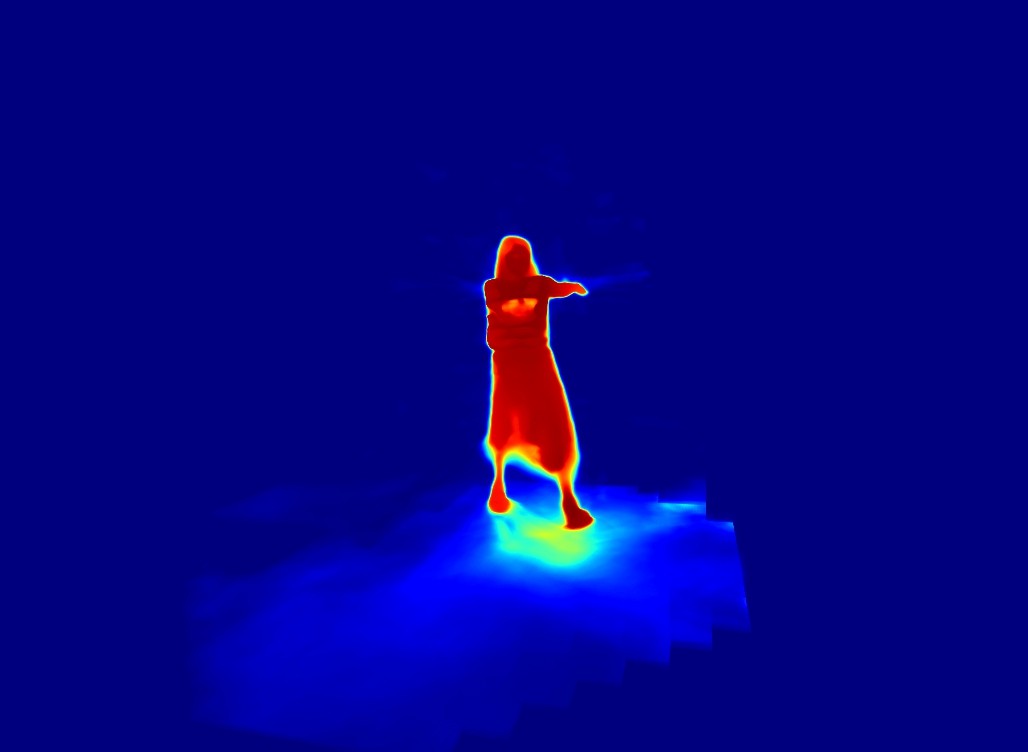}}
    &
    \raisebox{-0.5\height}{\includegraphics[trim={430 220 380 230},clip,width=0.19\columnwidth]{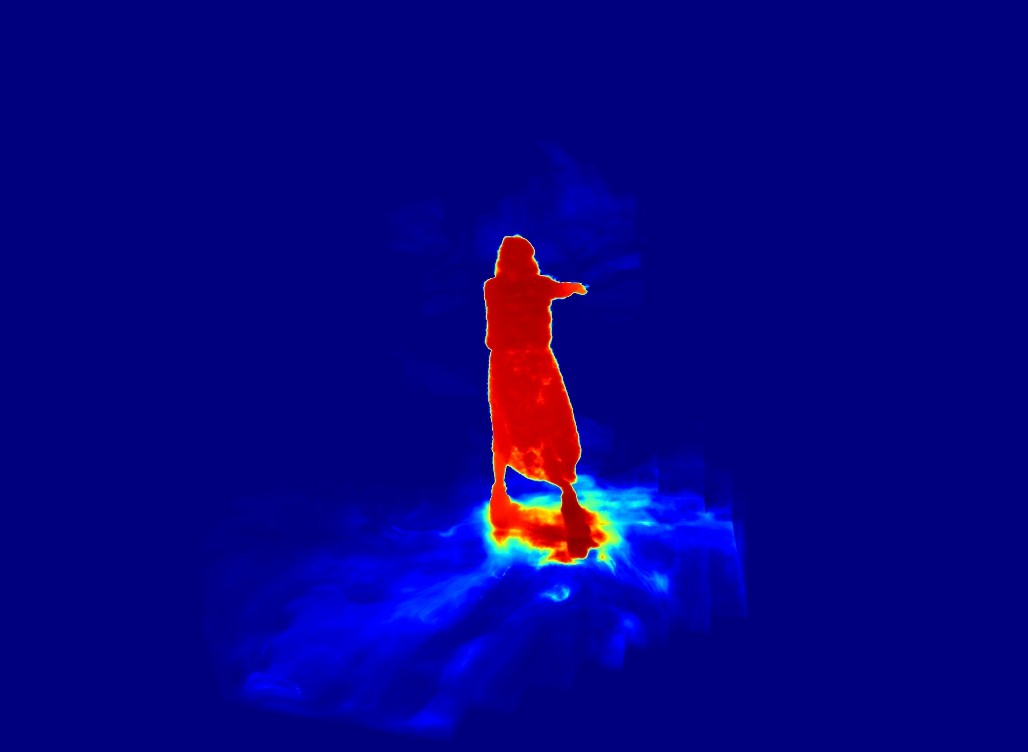}}
    &
    \raisebox{-0.5\height}{\includegraphics[trim={430 220 380 230},clip,width=0.19\columnwidth]{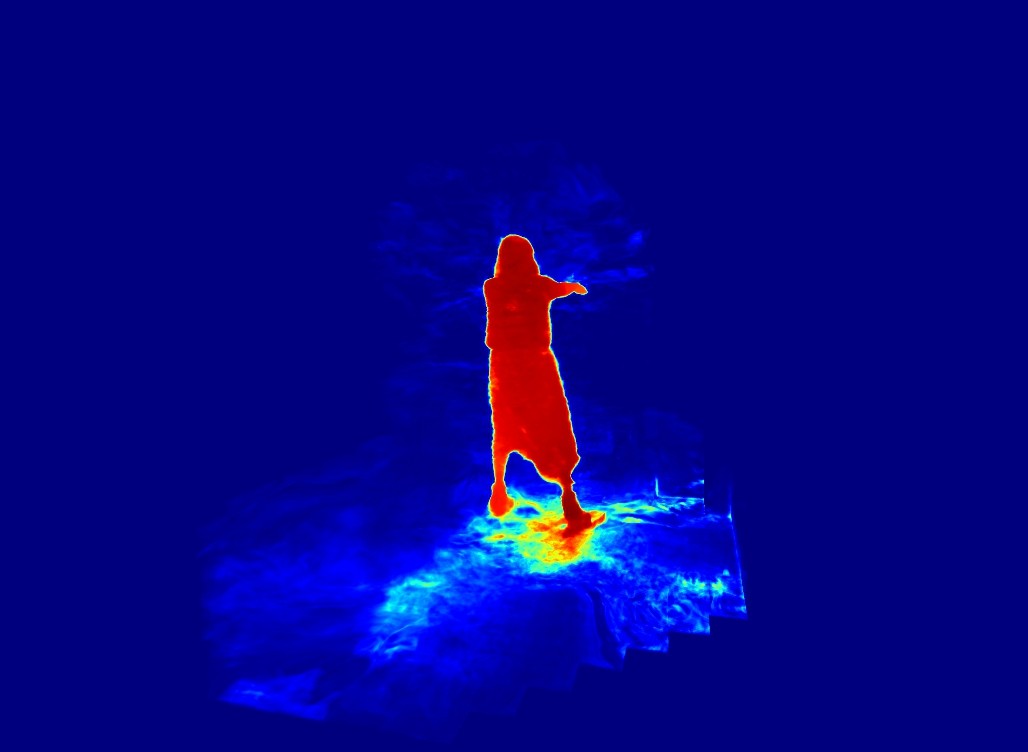}}
    \\
    
    \end{tabular}
    
    \caption{\textbf{Novel-View Synthesis.} (Top) Seq. 1 at $t{=}T$. (Bottom) Seq. 5 at $t{=}38$. The depth color differences between methods are due to normalization. }
    \label{fig:novel_view_and_depth}
    \vspace{-2em}
\end{figure}

\subsection{Qualitative Results}

We first qualitatively evaluate reconstructions at any one point in time, and then examine their consistency over time.

\subsubsection{Volumetric 3D Scene Reconstruction} 
Fig.~\ref{fig:novel_view_and_depth} shows novel views of the reconstruction and its geometry via depth maps. 
While NR-NeRF gives blurry results, our method yields high-quality reconstructions. 
Although blurrier, SNF-A matches the pattern on the dress better than ours. 
SNF-AG reconstructs the non-rigid geometry more accurately as it loosens the correspondences.

\subsubsection{Time Consistency}\label{subsec:tradeoff}

\begin{figure}
    \centering
    \small
    \setlength{\tabcolsep}{1pt}
    \begin{tabular}{ccc|cc}
    & $t{=}1$, view $1$ & $t{=}T$, view $2$ & $t{=}1$, view $2$ & $t{=}T$, view $2$
    \\
    \parbox[t]{2mm}{\rotatebox[origin=c]{90}{Ours}}
    &
    \raisebox{-0.5\height}{\includegraphics[trim={360 180 310 120},clip,width=0.24\columnwidth]{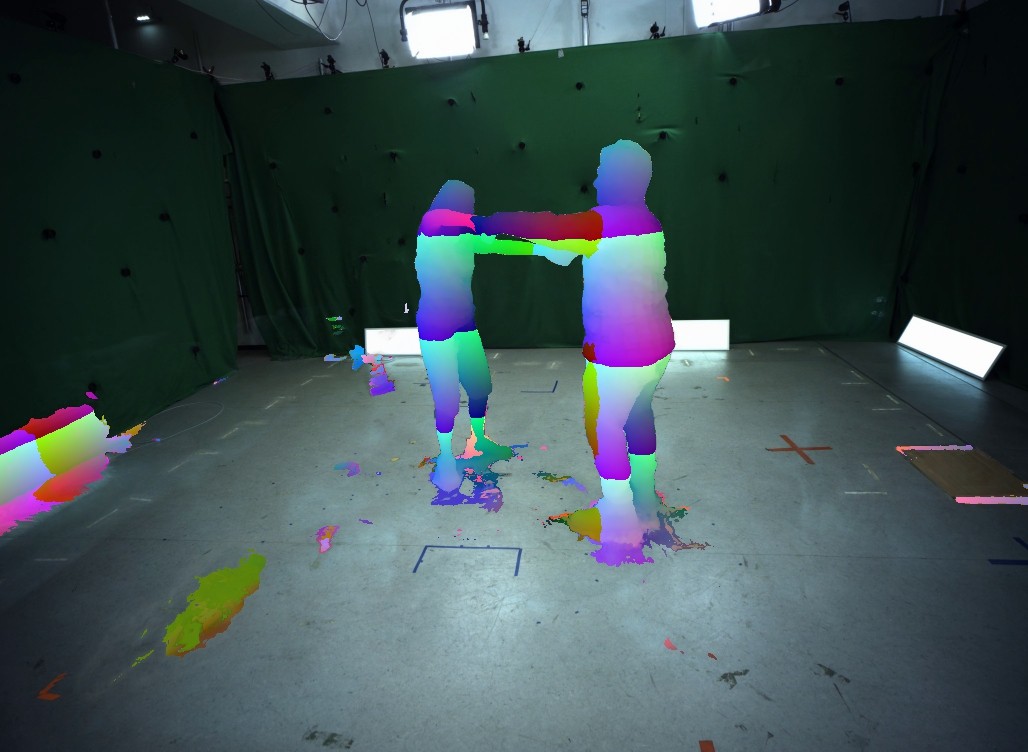}}
    &
    \raisebox{-0.5\height}{\includegraphics[trim={180 150 490 150},clip,width=0.24\columnwidth]{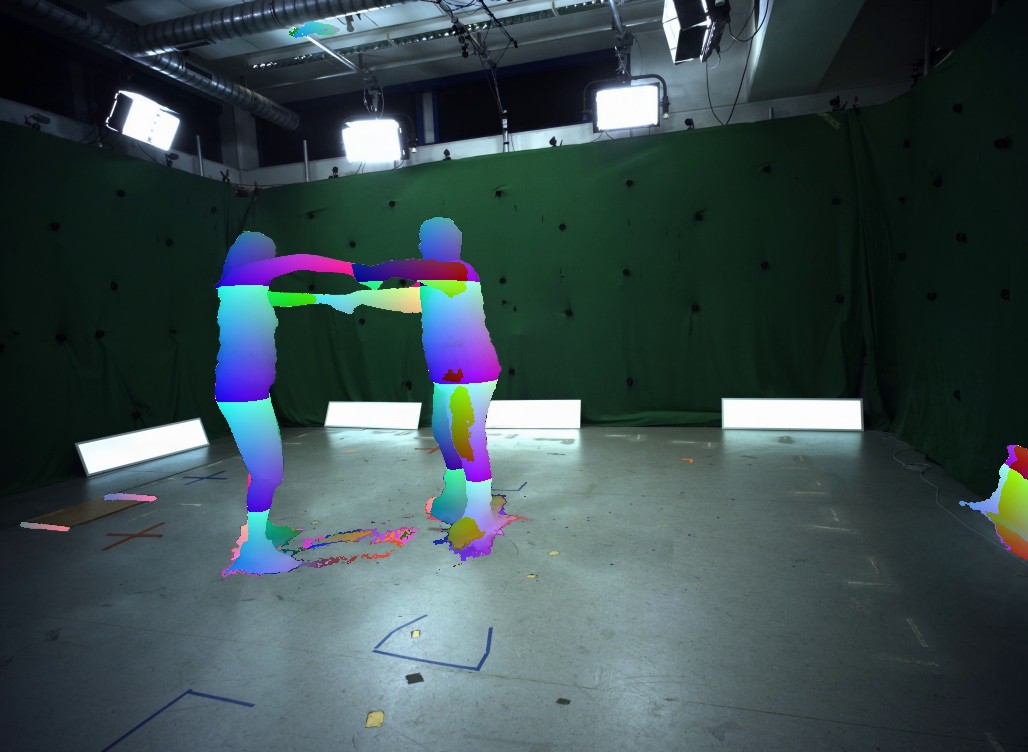}}
    &
    \raisebox{-0.5\height}{\includegraphics[trim={260 170 470 205},clip,width=0.24\columnwidth]{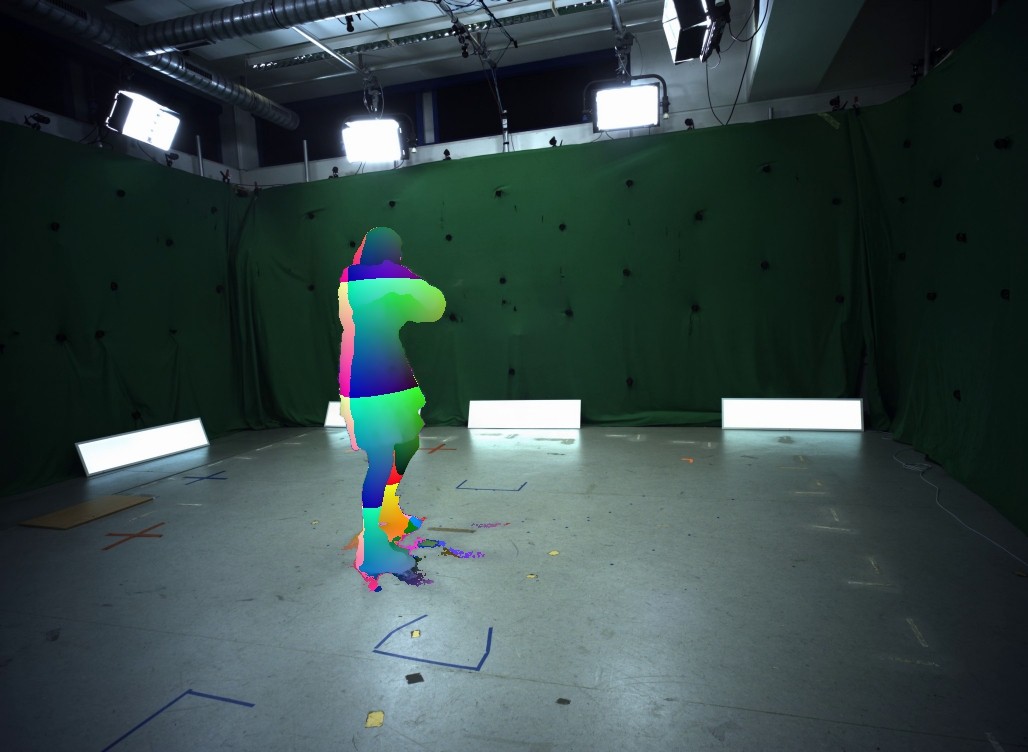}}
    &
    \raisebox{-0.5\height}{\includegraphics[trim={330 205 440 220},clip,width=0.24\columnwidth]{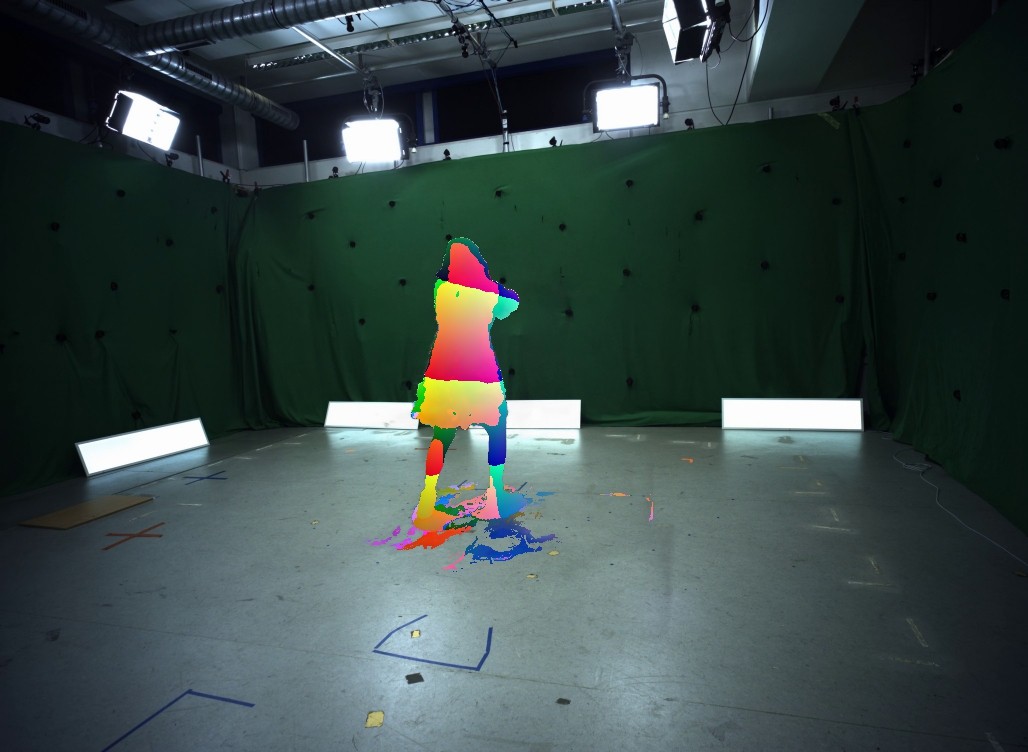}}
    \\
    \parbox[t]{2mm}{\rotatebox[origin=c]{90}{NR-NeRF}}
    &
    \raisebox{-0.5\height}{\includegraphics[trim={360 180 310 120},clip,width=0.24\columnwidth]{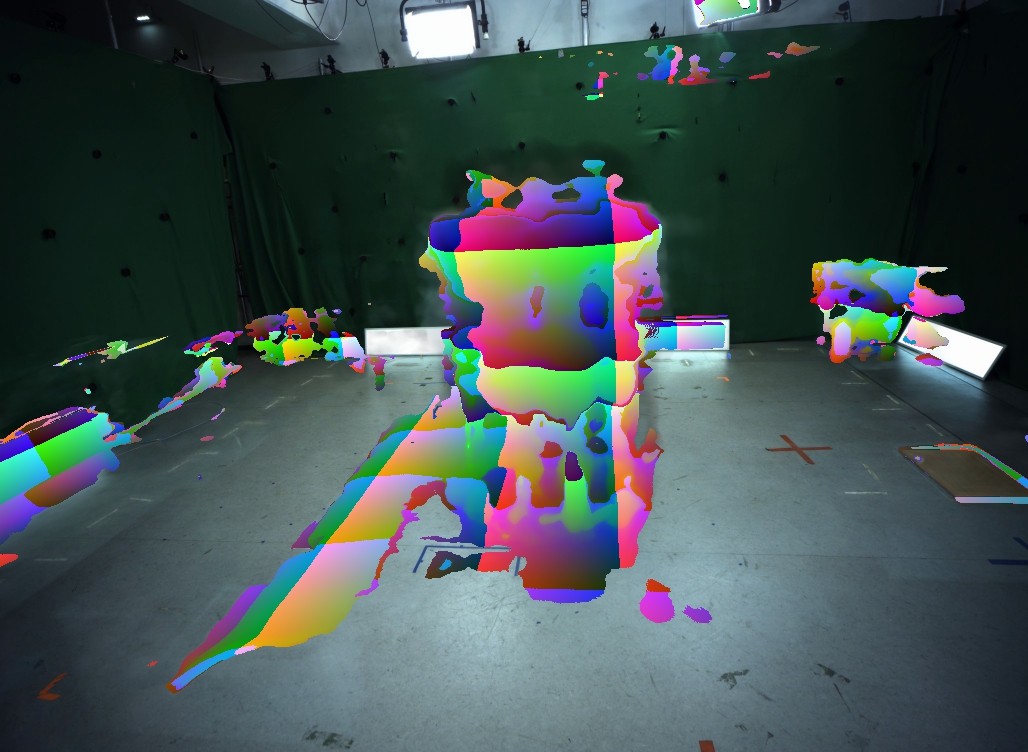}}
    &
    \raisebox{-0.5\height}{\includegraphics[trim={180 150 490 150},clip,width=0.24\columnwidth]{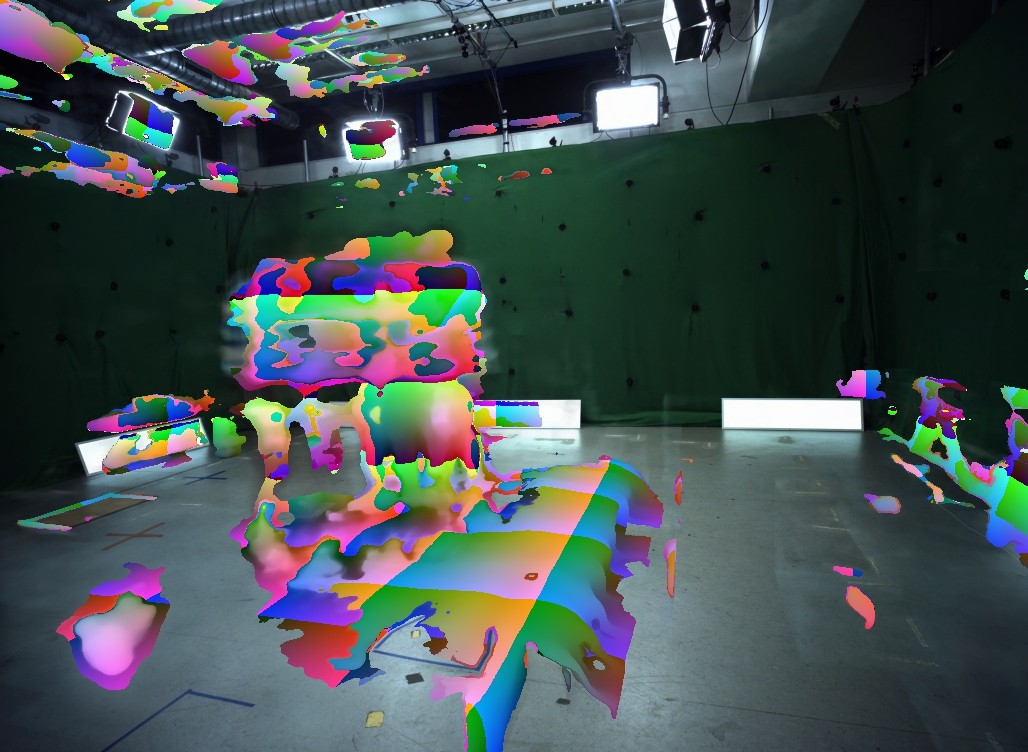}}
    &
    \raisebox{-0.5\height}{\includegraphics[trim={260 170 470 205},clip,width=0.24\columnwidth]{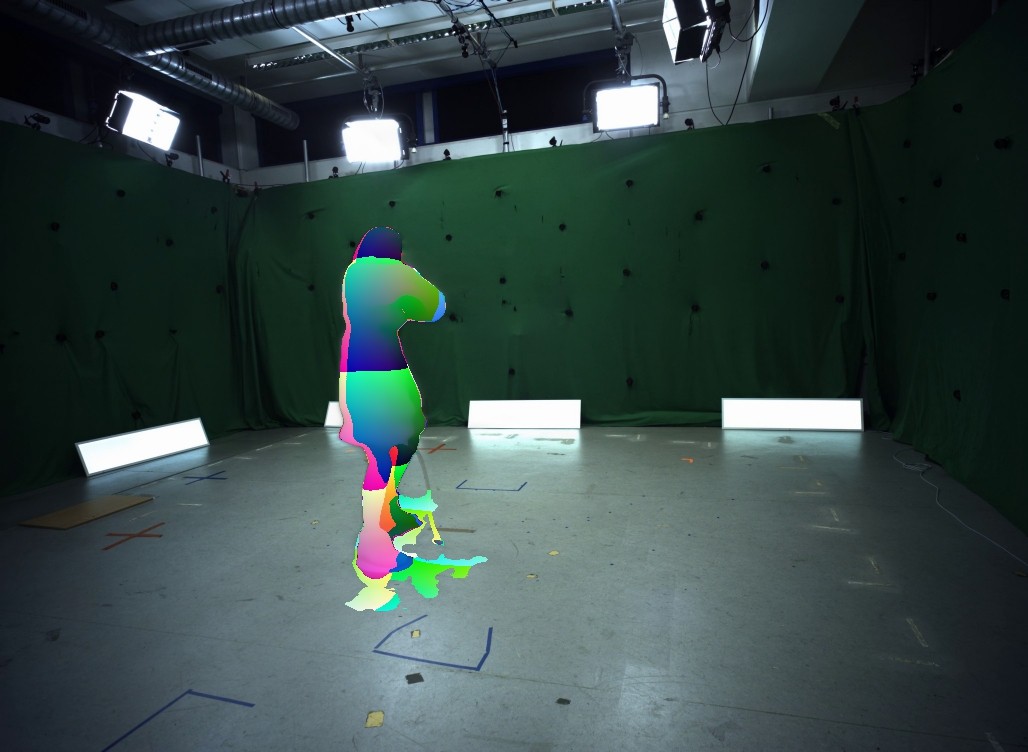}}
    &
    \raisebox{-0.5\height}{\includegraphics[trim={330 205 440 220},clip,width=0.24\columnwidth]{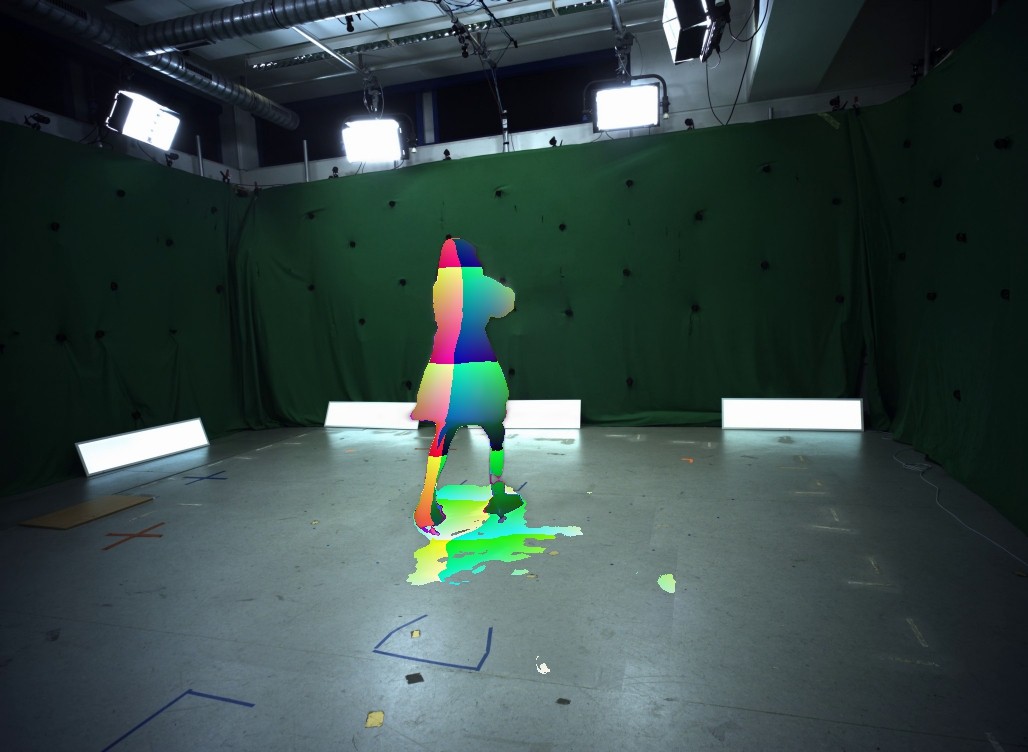}}
    \\
    \end{tabular}
    
    \caption{\textbf{Correspondences.} (Left) Seq.~2, opposite views. NR-NeRF fails. (Right) Seq.~4, same view. Only our correspondences follow the $90^\circ$ left turn; see orange at $t{=}1$ \& $T$. }
    \label{fig:correspondences}
\end{figure}

\noindent\textbf{Correspondence Visualization.} 
Time consistency enables 3D correspondences over time:  
Fig.~\ref{fig:correspondences} shows that our estimated correspondences are temporally stable. 
However, on scenes with large motion, NR-NeRF fails to converge to a recognizable model since this requires simultaneous correspondence estimation in this highly challenging setting.

\noindent\textbf{Comparisons with Prior Work.}
For further evaluation, we follow PREF~\cite{uii_eccv22_pref}, which estimates world-space 3D human joint positions over time: $\{\Tilde{\mathbf{p}}^t_j\in\mathbb{R}^3\}_{t,j}$, where $j$ indexes the $J{=}23$ joints. 
Off-the-shelf commercial systems~\cite{captury} exploit strong human-specific priors to reliably estimate 3D joints, which makes for excellent pseudo-ground-truth long-term 3D correspondences $\{\hat{\mathbf{p}}^t_j\}_{t,j}$. 
(The joints are used only for evaluation; our method does not use human priors.)

To track the joints with our method, we first initialize the estimated positions $\{\Tilde{\mathbf{p}}^1_j\}_{j}$ with the ground-truth joints $\{\hat{\mathbf{p}}^1_j\}_{j}$ at $t{=}1$, which are the positions in the canonical model. 
We then need to invert the backwards deformation field for $t{>}1$. 
To this end, for each joint $j$ at time $t$, we evaluate $d_f(d_c(\cdot;\theta_c^t);\theta_f^t)$ on a voxel grid (with a resolution of $128^3$ and a side length of $40$cm) centered on the estimated position at $t{-}1$, $\Tilde{\mathbf{p}}^{t-1}_j$, and pick the voxel center that lands closest to the ground-truth joint $\hat{\mathbf{p}}^1_j$ in the canonical model as the estimated world-space joint position $\Tilde{\mathbf{p}}^t_j$. 
The supplement contains tracking details for NR-NeRF and PREF. 

\begin{figure}
    \centering
    \setlength{\tabcolsep}{1pt}
    \small
    \begin{tabular}{ccc|ccc}
    \includegraphics[trim={80 30 80 50},clip,width=0.16\columnwidth]{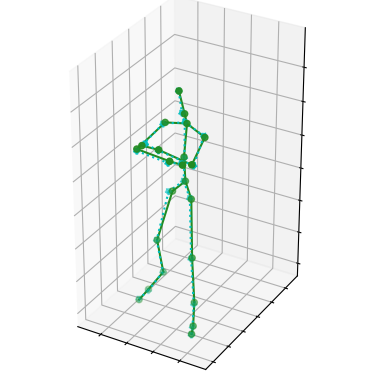}
    &\includegraphics[trim={80 30 80 50},clip,width=0.16\columnwidth]{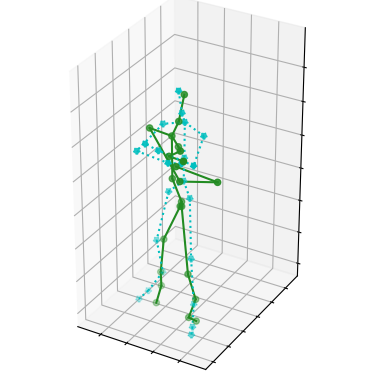}
    &\includegraphics[trim={80 30 80 50},clip,width=0.16\columnwidth]{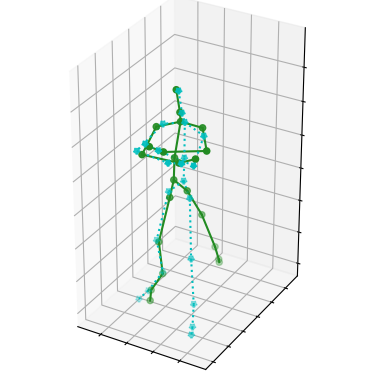}
    &\includegraphics[trim={80 50 80 30},clip,width=0.16\columnwidth]{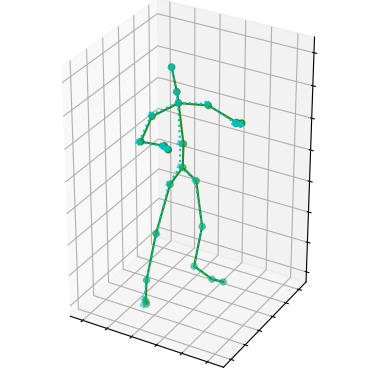}
    &\includegraphics[trim={80 50 80 30},clip,width=0.16\columnwidth]{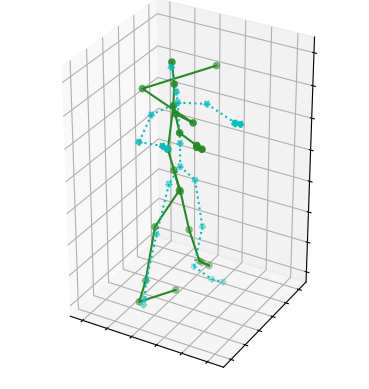}
    &\includegraphics[trim={80 50 80 30},clip,width=0.16\columnwidth]{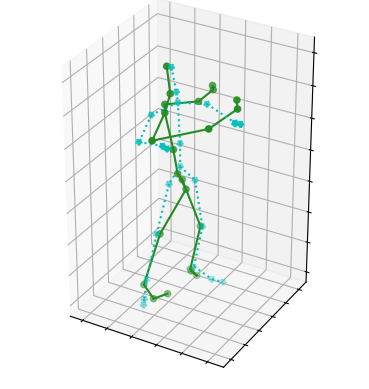}
    \\
    Ours & PREF & NR-NeRF & Ours & PREF & NR-NeRF\\
    \end{tabular}
    
    \caption{\textbf{Time Consistency.} (Left) Seq.~4. (Right) Seq.~6. The solid skeleton is the tracking estimate at $t{=}T$. The dotted skeleton is the pseudo-ground truth at $t{=}T$. }
    \label{fig:joints}
\end{figure}

Fig.~\ref{fig:joints} shows results at $t{=}T$. 
Our method yields stable correspondences, while NR-NeRF does not. 
PREF's correspondences drift strongly over time due to error accumulation from chaining frame-to-frame correspondences. %

\begin{figure}
    \centering
    \footnotesize
    \setlength{\tabcolsep}{1pt}
    \begin{tabular}{ccc|ccc}
    \includegraphics[trim={230 360 740 310},clip,width=0.16\columnwidth]{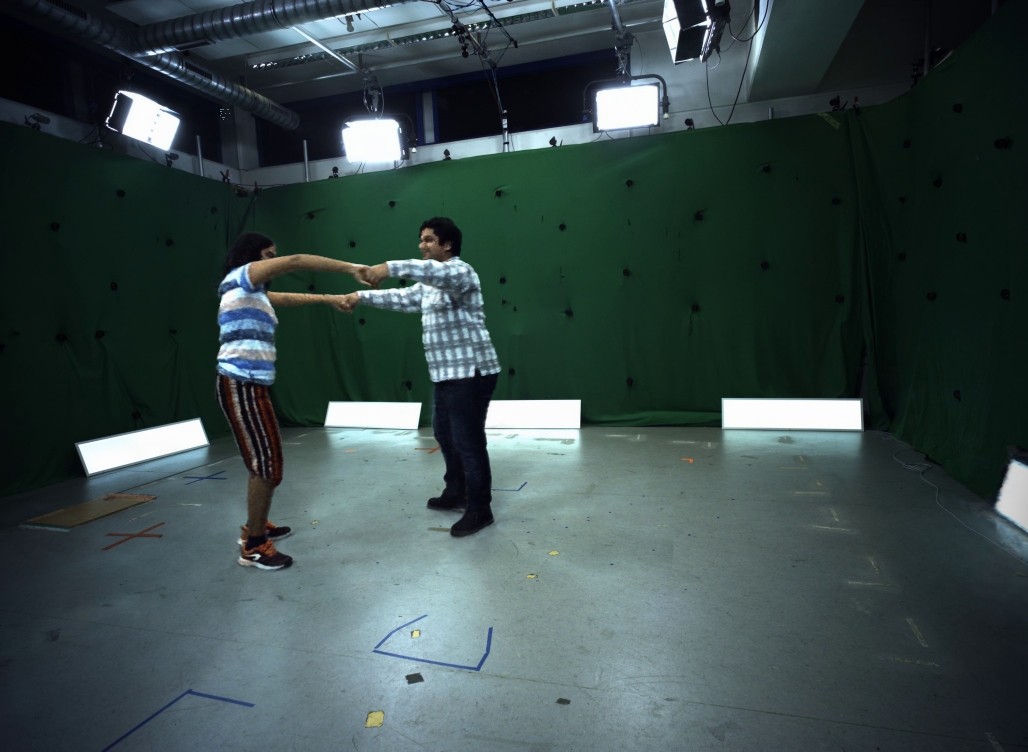}
    &
    \includegraphics[trim={230 360 740 310},clip,width=0.16\columnwidth]{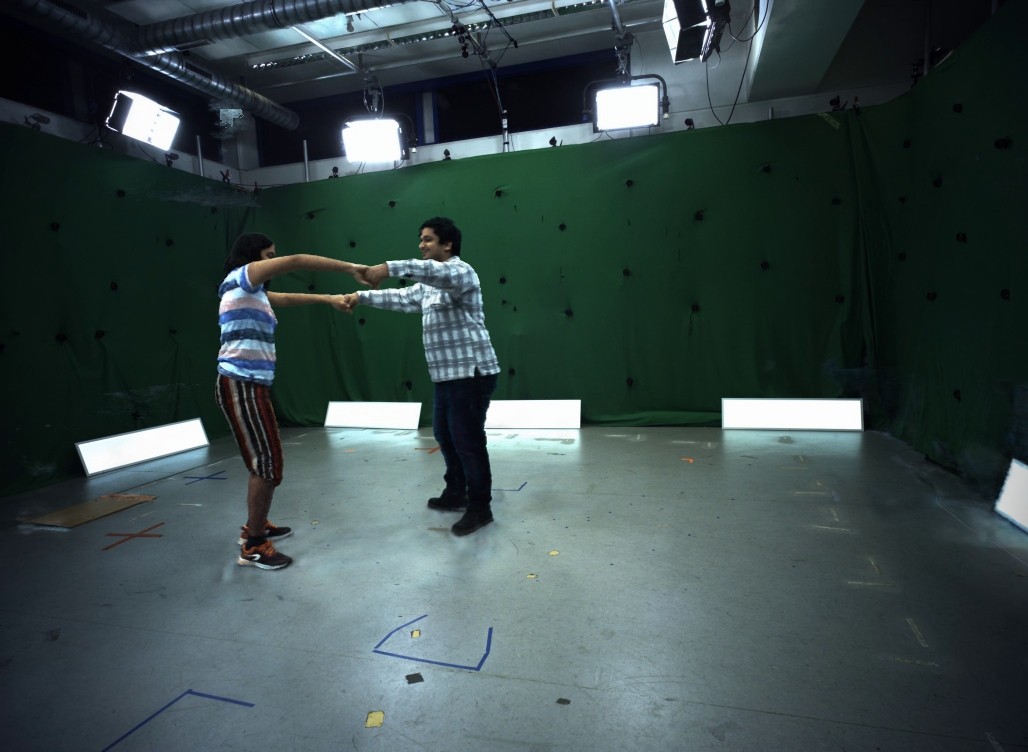}
    &
    \includegraphics[trim={230 360 740 310},clip,width=0.16\columnwidth]{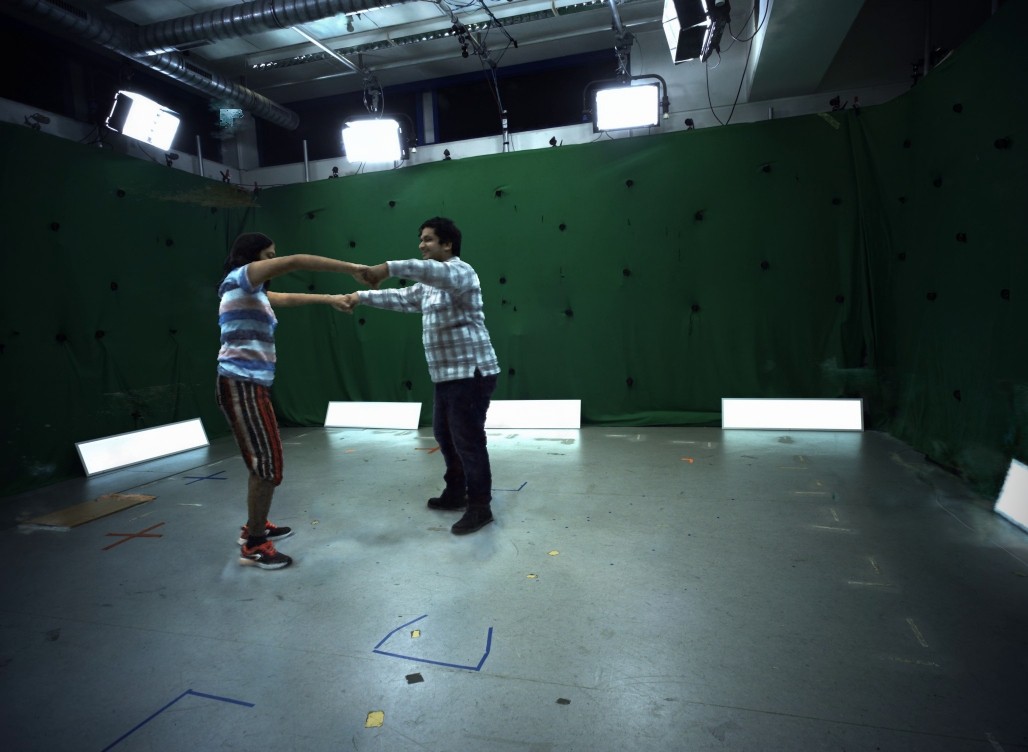}
    &
    \includegraphics[trim={500 360 460 300},clip,width=0.16\columnwidth]{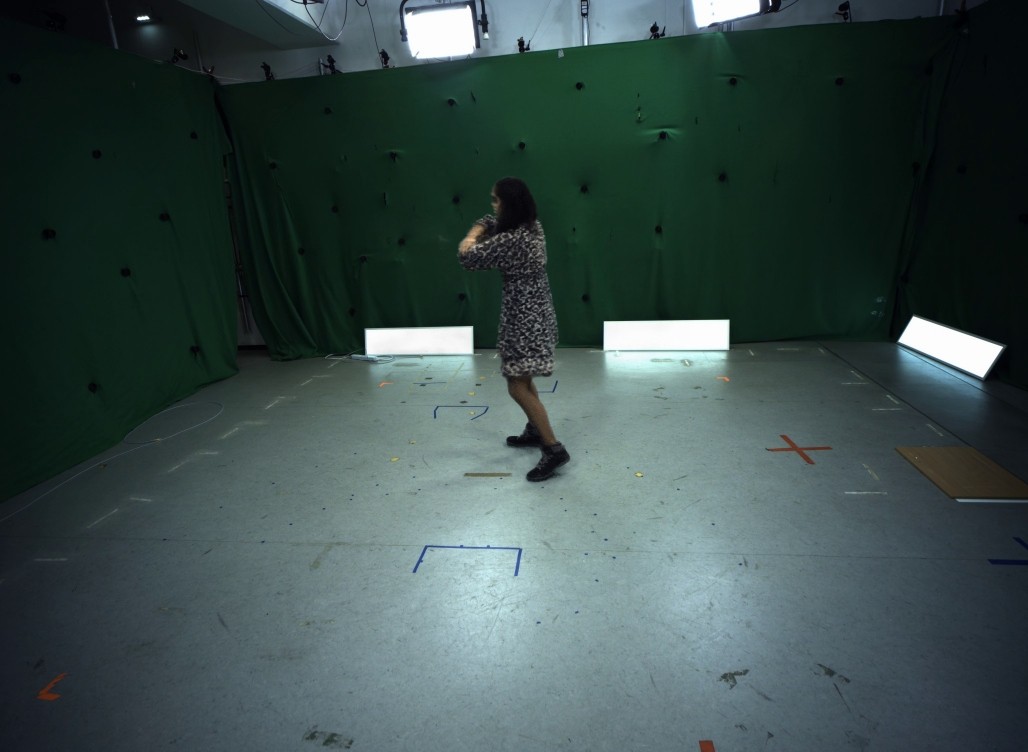}
    &
    \includegraphics[trim={500 360 460 300},clip,width=0.16\columnwidth]{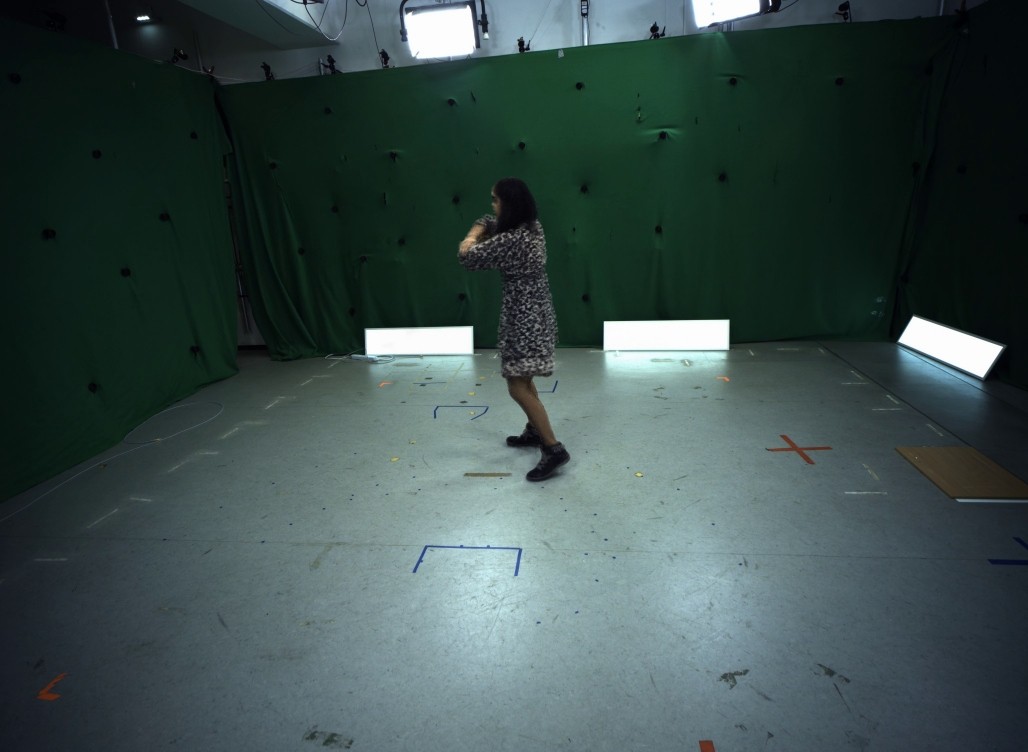}
    &
    \includegraphics[trim={500 360 460 300},clip,width=0.16\columnwidth]{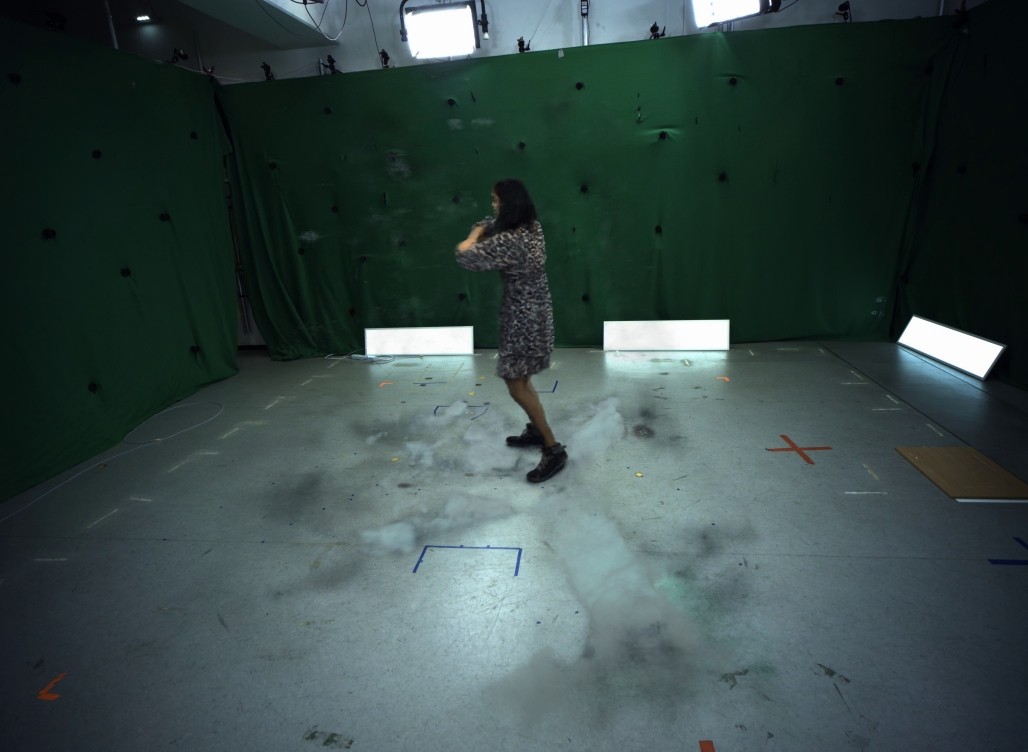}
    \\
    Ours & SNF-A  & SNF-A & Ours & SNF-AG & SNF-AG \\   
    at $t_1$ \& $t_2$ & at $t_1$ & at $t_2$ & at $t_1$ \& $t_2$ & at $t_1$ & at $t_2$\\  
    \end{tabular}
    
    \caption{\textbf{Canonical Model.} Ours uses a static canonical model, while those of SNF-A(G) vary over time. }
    \label{fig:canonical}
\end{figure}

\noindent\textbf{Variants.} 
We contrast the time consistency of our method with its variants. 
To this end, Fig.~\ref{fig:canonical} visualizes the canonical model at two different timestamps. 
Unlike SceNeRFlow, the canonical model of the variants changes over time, \emph{i.e.}, they lack inherent correspondences. 
However, as results in the video show, these additional degrees of freedom improve the variants' novel-view synthesis, showing a trade-off between time consistency and novel-view synthesis.

\subsection{Quantitative Results}

\noindent\textbf{Time Consistency.} 
Like PREF~\cite{uii_eccv22_pref}, we measure time consistency via 3D joint tracking. 
We report the mean per-joint position error (MPJPE)~\cite{sigal2010humaneva} %
over all timestamps $t{>}1$ and the $J$ joints: $\operatorname{MPJPE}=\frac{1}{(T-1)J}\sum_{t=2}^{T} \sum_{j=1}^J \lVert \hat{\mathbf{p}}^t_j - \Tilde{\mathbf{p}}^t_j \rVert_2$. 
Lower is better. 
For scenes with two people, we report the average across both. 
We evaluate all scenes except for the 5-frame Seq.~8 used in NR-NeRF~\cite{tretschk2021non}. 
Tab.~\ref{tab:joints_per_scene} contains the results. 
SceNeRFlow has the lowest error with only marginal drift. 
PREF's frame-wise approach leads to large drift. 
D-NeRF cannot handle large motion.

\begin{table}
    \centering
    \resizebox{\columnwidth}{!}{
        \begin{tabular}{l|ccccccc|c}
        \hline
            Sequence & $1$ & $2$ & $3$ &$4$ &$5$ & $6$ &$7$ & Mean\\
        \hline
        Ours & $\mathbf{0.9}$ & $\mathbf{0.9}$ & $\mathbf{1.8}$& $\mathbf{1.8}$&$\mathbf{2.8}$ &$\mathbf{0.9}$ & $1.4$& $\mathbf{1.5}$\\ 
        PREF & $11.0$&$13.9$ &$47.8$ & $8.9$& $13.4$& $12.1$& $3.9$&$15.9$ \\
        NR-NeRF & $2.2$&$46.4$ & $110.9$& $6.1$&$8.5$ &$14.8$ &$\mathbf{1.2}$ & $27.2$\\
        D-NeRF & $40.7$ & $46.2$ & $110.1$ & $12.2$ & $81.9$ & $35.3$ & $70.0$ & $56.6$
        \end{tabular}
    }
    \caption{\textbf{Time Consistency.} We report per-scene and mean MPJPE in cm. Lower is better.
    }
    \label{tab:joints_per_scene}
    \vspace{-1em}
\end{table}

\begin{table}
    \centering
    \resizebox{\columnwidth}{!}{
        \begin{tabular}{clc|ccccc}
        \hline
           &   &              & Ours & NR-NeRF & SNF-A & SNF-AG & Background \\
        \hline
        \hline
        \parbox[t]{2mm}{\multirow{3}{*}{\rotatebox[origin=c]{90}{\footnotesize Unmasked}}} & 
          PSNR  & $\uparrow$   & $30.32$ & $28.77$ & $\mathbf{30.88}$ & $30.34$ & $21.69$ \\
        & SSIM  & $\uparrow$   & $0.939$ & $0.922$ & $\mathbf{0.940}$ & $0.929$ & $0.920$\\
        & LPIPS & $\downarrow$ & $\mathbf{0.054}$ & $0.099$ & $0.057$ & $0.089$ & $0.101$\\
        \hline
        \parbox[t]{2mm}{\multirow{3}{*}{\rotatebox[origin=c]{90}{Masked}}} 
        & PSNR  & $\uparrow$   & $32.38$ & $30.80$ & $\mathbf{33.31}$ & $33.22$ & --- \\
        & SSIM  & $\uparrow$   & $0.976$ & $0.965$ & $\mathbf{0.978}$ & $0.977$ &---\\
        & LPIPS & $\downarrow$ & $0.018$ & $0.041$ & $\mathbf{0.016}$ & $0.018$ & ---\\
        \end{tabular}
    }
    \caption{\textbf{Novel-View Synthesis.} Mean PSNR, SSIM, and LPIPS across scenes. SNF-A(G) are variants of ours. 
    }
    \label{tab:novel_view}
    \vspace{-0.5em}
\end{table}

\noindent\textbf{Reconstruction.} 
We quantify the reconstruction quality by using novel-view quality as a proxy. 
Like NeRF~\cite{mildenhall2020nerf}, we measure the latter with PSNR and SSIM~\cite{ssim2004} (for both, higher is better), and the neural perceptual metric LPIPS~\cite{johnson2016perceptual} (lower is better). 
We focus on the dynamic scene part by (1) also providing these metrics w.r.t.\ the static background image, and (2) reporting \emph{masked scores} where we mask out the renderings with foreground masks estimated from the ground truth. 
The results are in Tab.~\ref{tab:novel_view}.
Our method outperforms NR-NeRF. %
The variants have artifacts in empty space and only outperform our method on the masked scores. 
The supplement has per-scene results and further comparisons with D-NeRF and DeVRF.

\subsection{Ablations}

\begin{figure}
    \centering

    \small
    \setlength{\tabcolsep}{1pt}
    \begin{tabular}{ccc|ccc}
    
    \includegraphics[trim={300 390 590 200},clip,width=0.16\columnwidth]{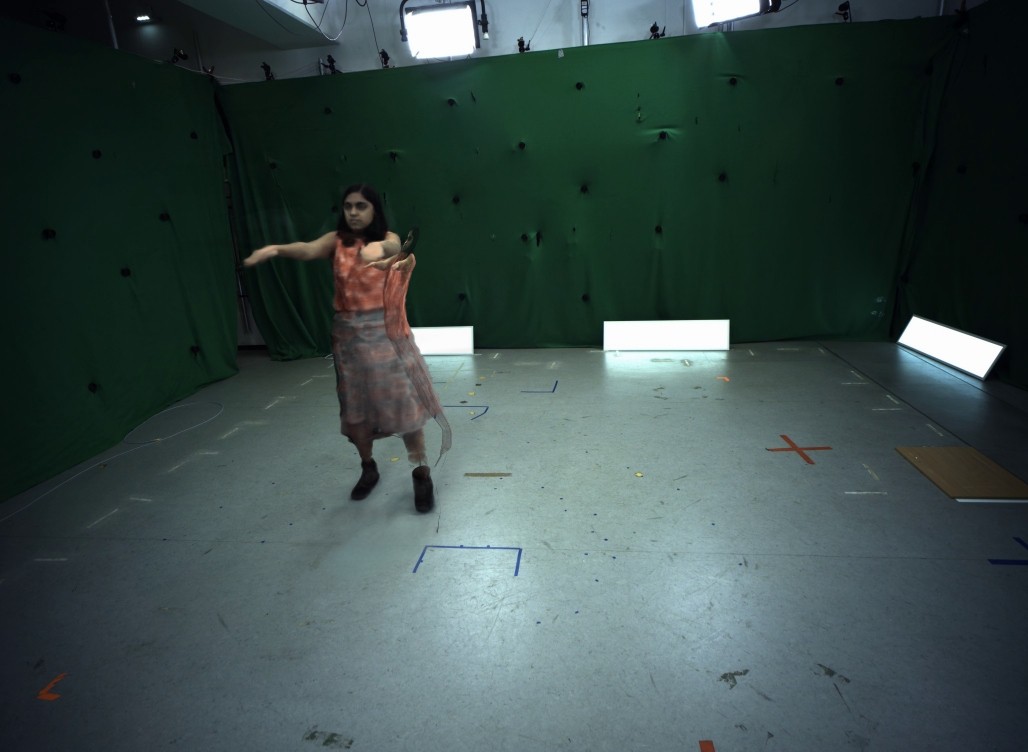}
    &
    \includegraphics[trim={300 390 590 200},clip,width=0.16\columnwidth]{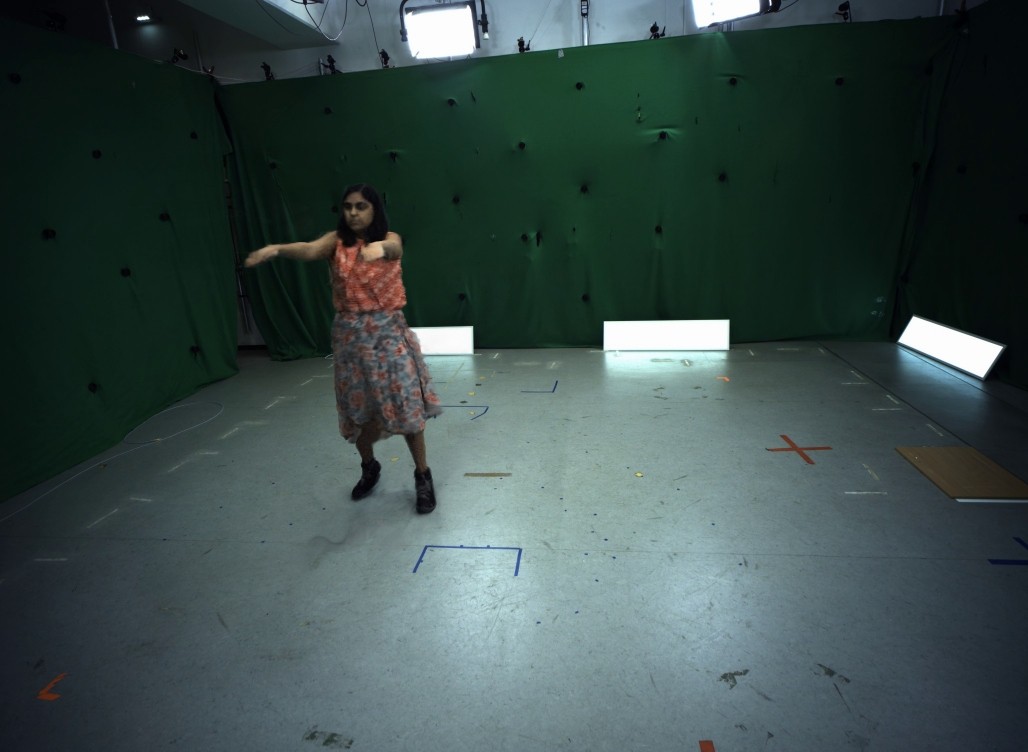}
    &
    \includegraphics[trim={300 390 590 200},clip,width=0.16\columnwidth]{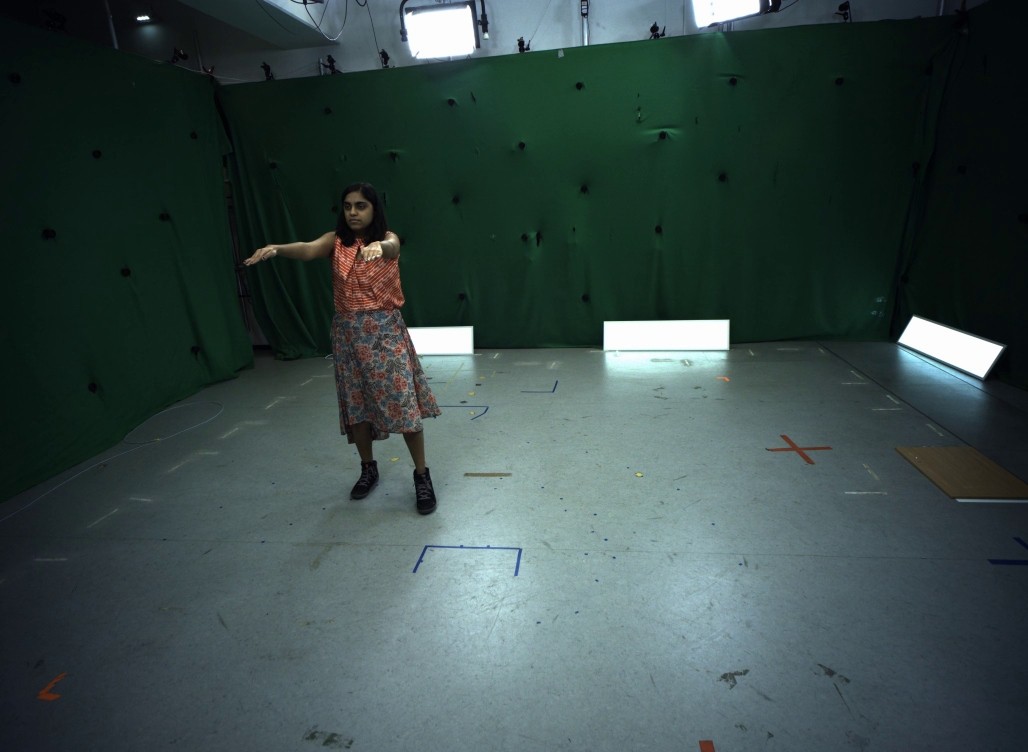} 
    & 
    \includegraphics[trim={400 370 510 230},clip,width=0.16\columnwidth]{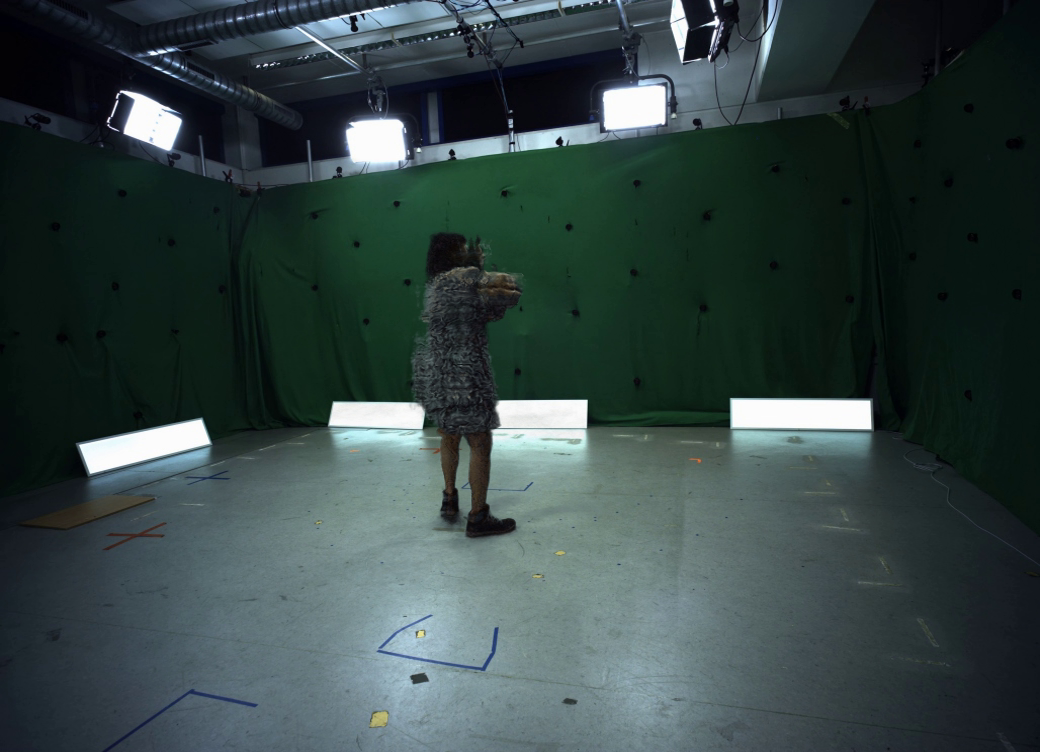}
    &
    \includegraphics[trim={400 370 510 230},clip,width=0.16\columnwidth]{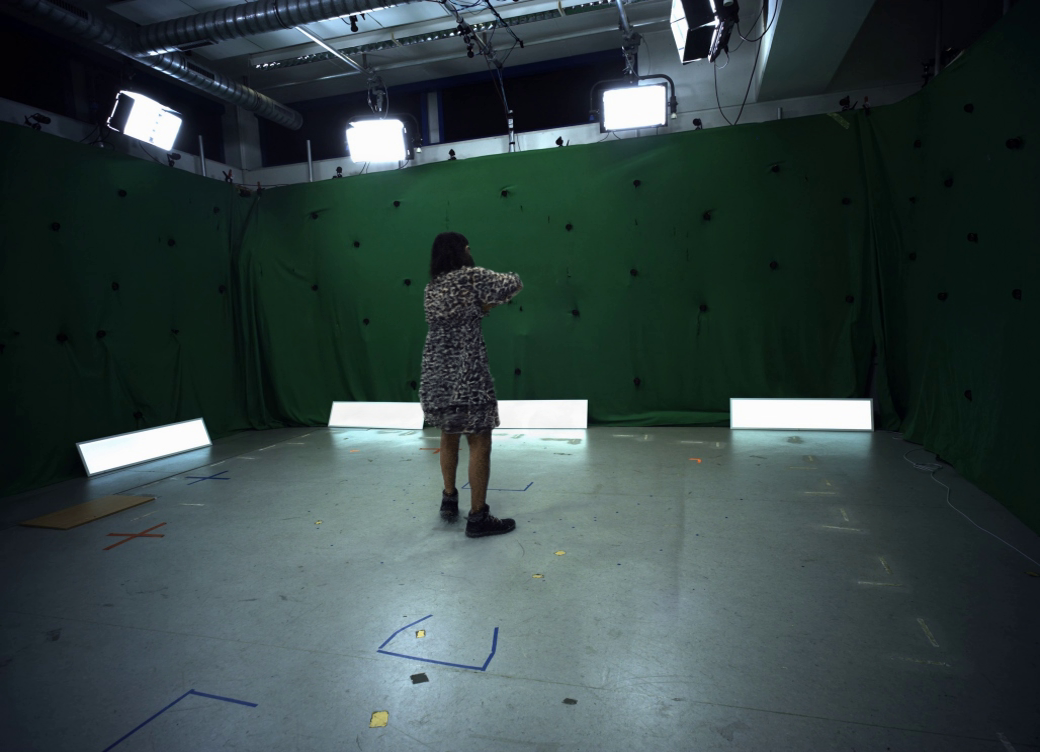}
    &
    \includegraphics[trim={400 370 510 230},clip,width=0.16\columnwidth]{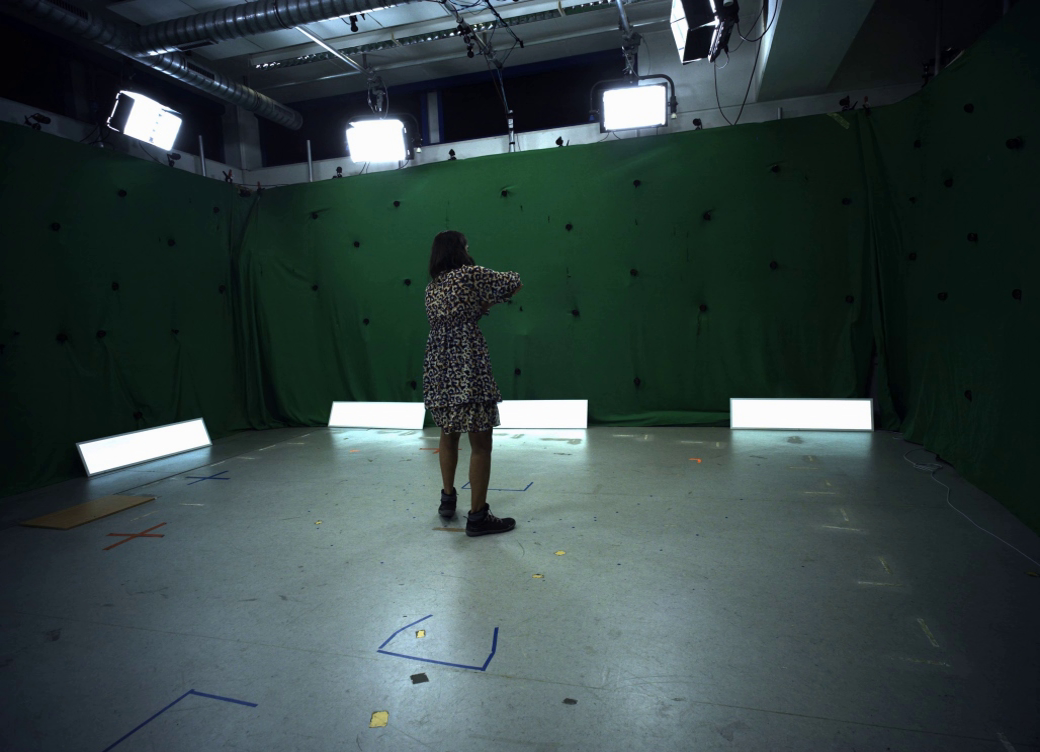}
    \\
    No Online &  Ours &  GT & No Extend & Ours & GT\\
    \hline
    \end{tabular}

    \begin{tabular}{cccc}
    \includegraphics[trim={410 250 480 330},clip,width=0.24\columnwidth]{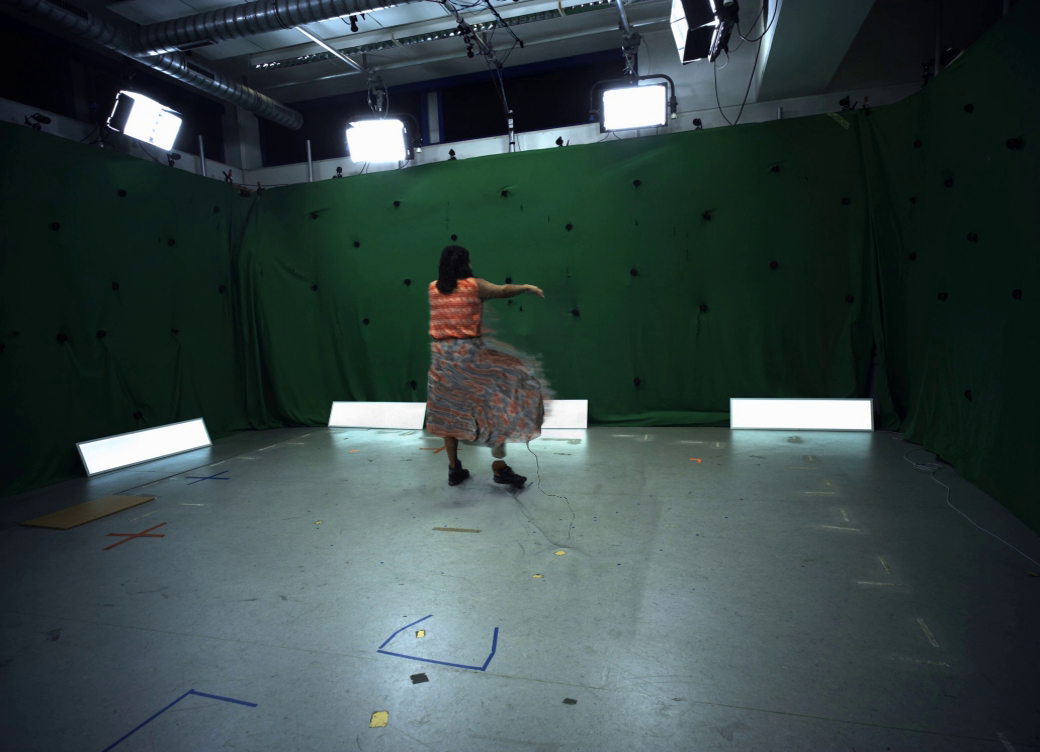}
    &
    \includegraphics[trim={410 250 480 330},clip,width=0.24\columnwidth]{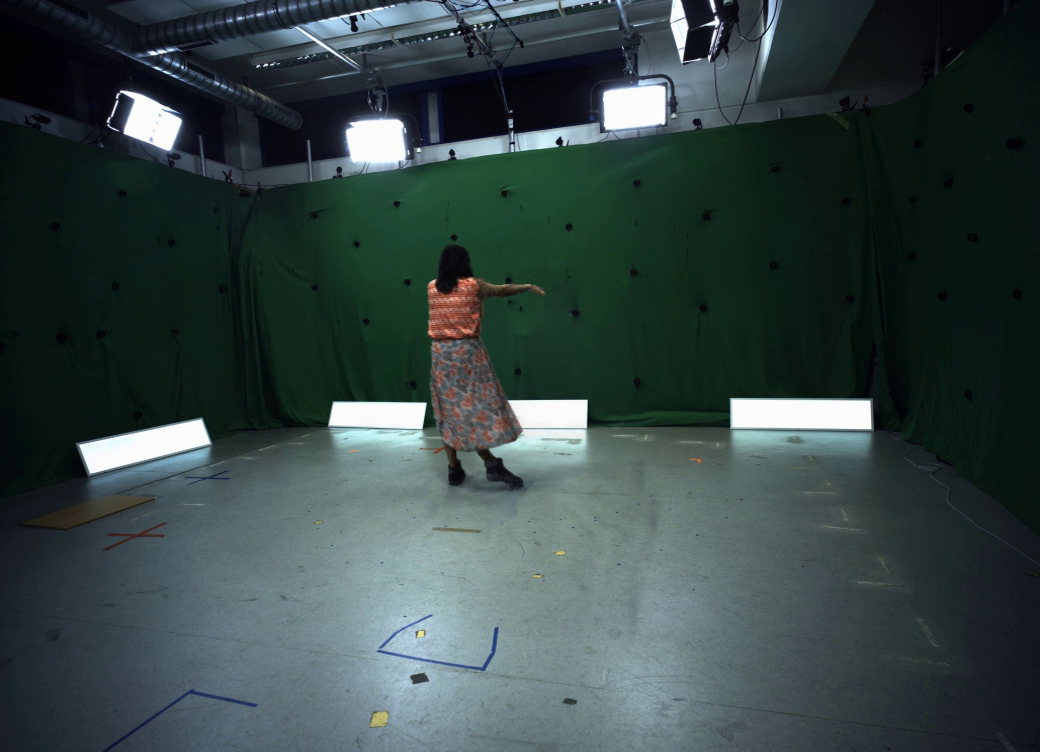}
    &
    \includegraphics[trim={410 250 480 330},clip,width=0.24\columnwidth]{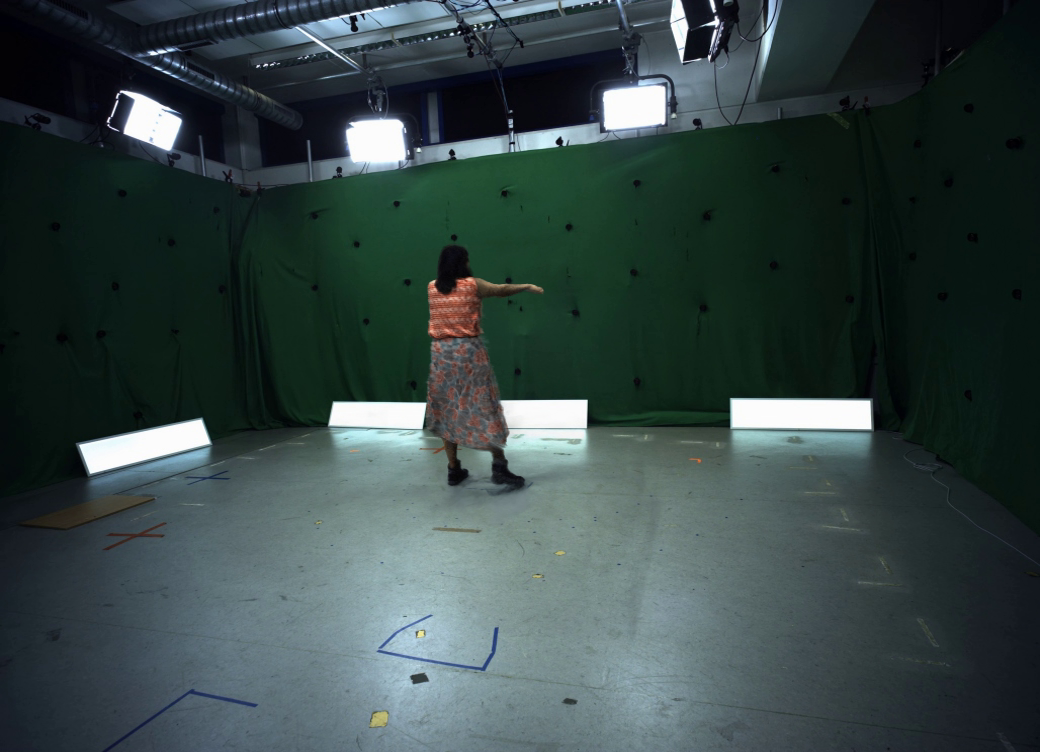}
    &
    \includegraphics[trim={410 250 480 330},clip,width=0.24\columnwidth]{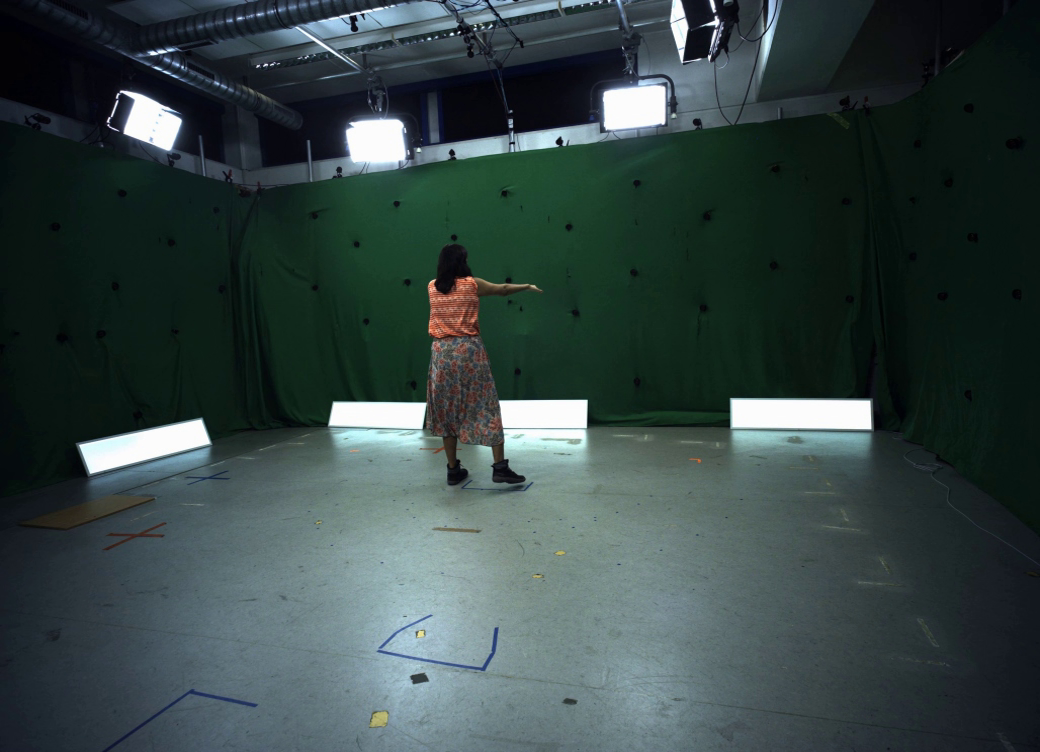}
    \\
    No Coarse & No Fine & Ours & GT\\
    \end{tabular}
    
    \caption{\textbf{Ablations.} (Top left) Online optimization, (top right) extending the deformations, (bottom) coarse and fine deformations.}
    \label{fig:ablations}
    \vspace{-0.5em}
\end{figure}

We ablate the core components of our method. 
We investigate the relevance of (1)~optimizing in an online manner (by optimizing all timestamps at once),  (2)~``extending the deformation field'' (by using a coarse resolution of $512^3$ and $\mathcal{L}_{\text{norm,w}}$ without max-pooling%
), (3)~the coarse deformation model (by setting $\Delta_c{=}\mathbf{0}$ at training time), and (4)~the fine deformation model (by setting $\Delta_f{=}\mathbf{0}$ at test time). 
Fig.~\ref{fig:ablations} shows the results, the supplement confirms these quantitatively: 
(1) Optimizing online simplifies implicit correspondence estimation, avoiding ghosting artifacts; (2) extending the deformation field helps the online optimization by initializing the deformations for the next timestamp well; (3) the coarse deformations stabilize the reconstruction; and (4) the fine deformations add details.

\subsection{Simple Editing}

\begin{figure}
    \centering
    \footnotesize
    \setlength{\tabcolsep}{1pt}
    \begin{tabular}{cc|cc}
    \multicolumn{2}{c|}{View $1$}  &  \multicolumn{2}{c}{View $2$} \\\hline
        $t{=}1$ &  $t{=}T$ &  $t{=}1$ &  $t{=}T$\\
        \includegraphics[trim={20 170 640 170},clip,height=0.3\columnwidth]{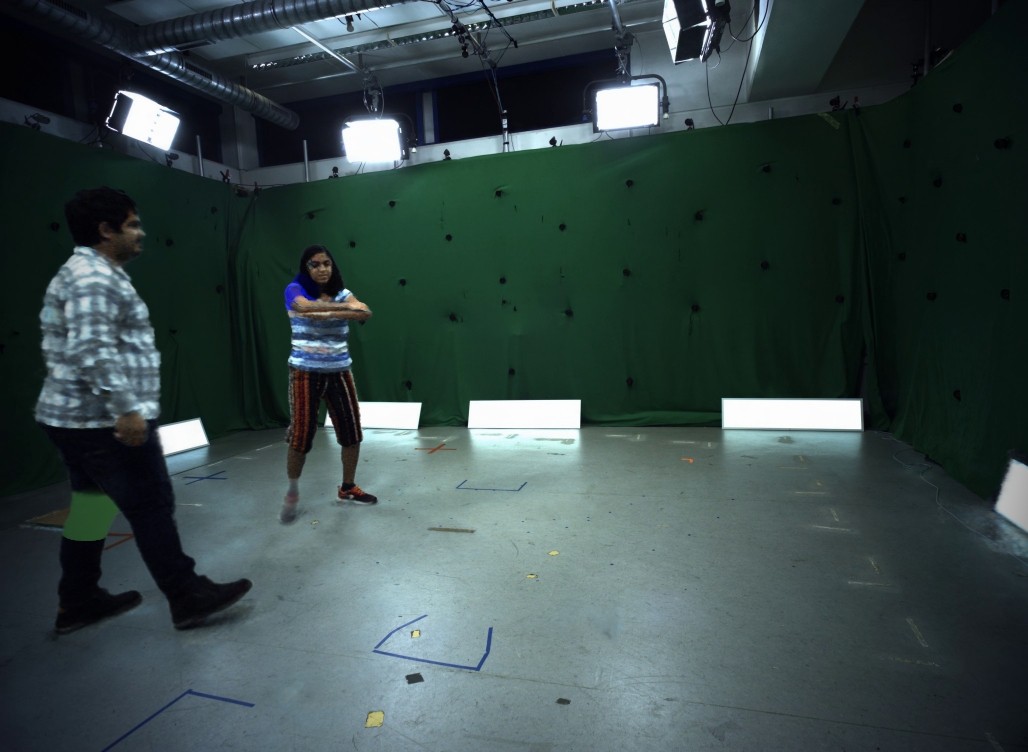}
        &
        \includegraphics[trim={110 170 600 170},clip,height=0.3\columnwidth]{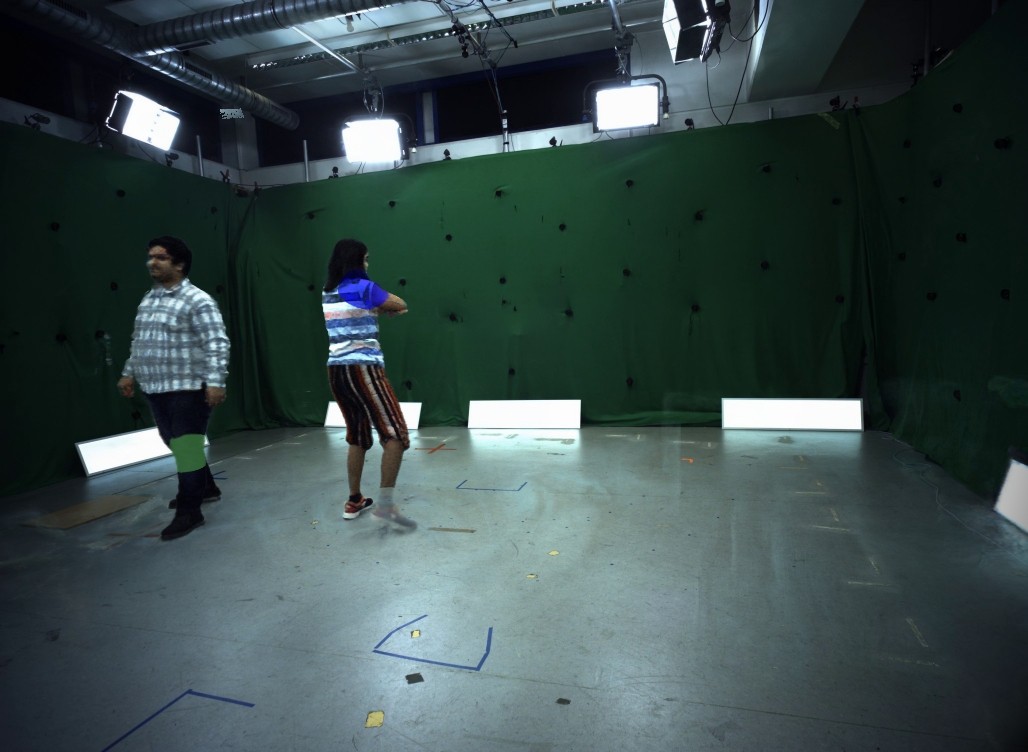}
        &
        \includegraphics[trim={540 220 280 120},clip,height=0.3\columnwidth]{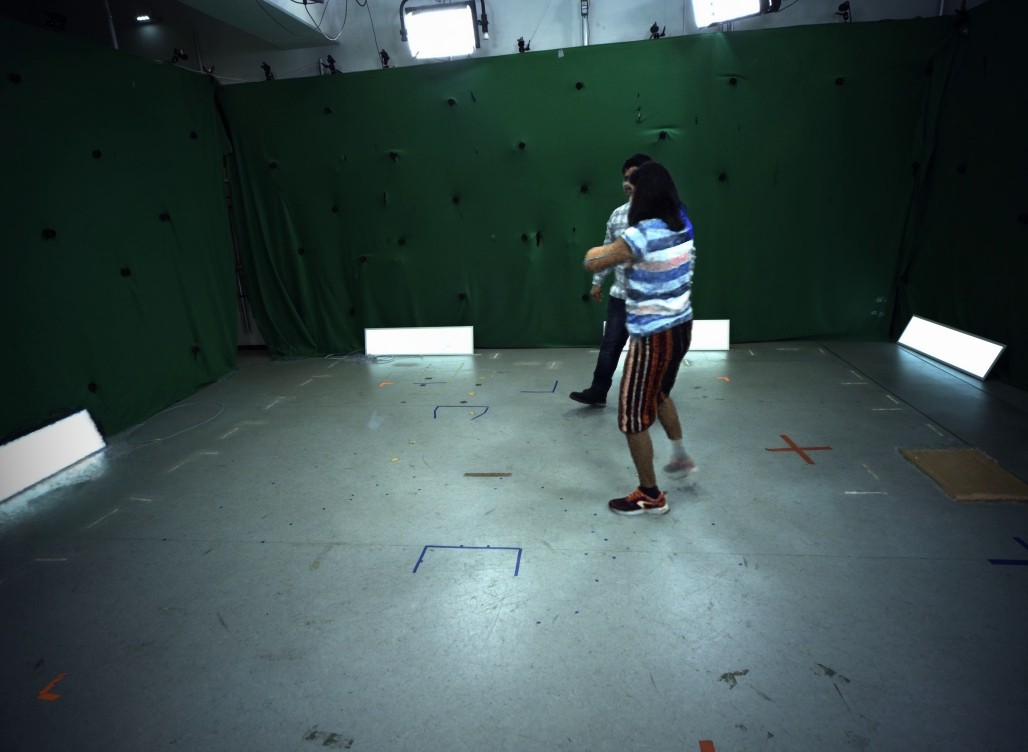}
        &
        \includegraphics[trim={510 220 90 120},clip,height=0.3\columnwidth]{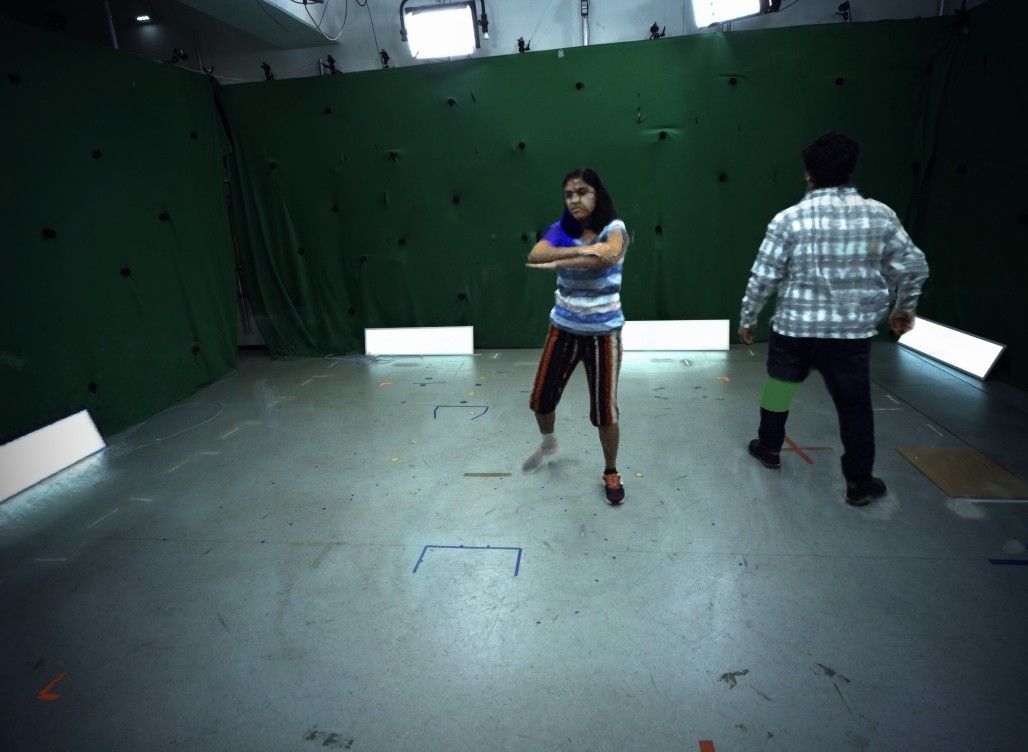}
    \end{tabular}
    
    \caption{\textbf{Scene Editing.} (1) Coloring the left knee of person $1$ green, (2) blending the right shoulder of person $2$ with blue, and (3) making the right foot of person $2$ transparent.}
    \label{fig:editing}
\end{figure}

The time consistency 
enables simple editing of the geometry and appearance. 
As a proof of concept, Fig.~\ref{fig:editing} shows that we can re-color scene parts (\textit{e.g.,} of Seq.~3) 
or make them transparent in a straightforward manner---by modifying the static canonical model. 
The deformations then consistently propagate these changes to all timestamps.

\subsection{Limitations and Future Work}\label{sec:limitations}

While our method design makes a few assumptions, we exploit them as little as possible, \emph{e.g.} we do not use accurate segmentation masks, depth estimates, or 3D estimates. 
This makes it easier to extend our work to other input settings. 

Adjacent work offers promising remedies for our assumptions. 
Most dynamic-NeRF methods are monocular (but not time-consistent for large motion) and some static-NeRF works \cite{yu2020pixelnerf,tewari2021advances} use only few cameras. %
Adapting these ideas might reduce the number of cameras required by our method. 
Since 
 our core problem lies in the dynamic foreground, 
we exclude the background in a soft manner. 
The background could be included via a static NeRF as some existing dynamic-NeRF approaches do. 
Na\"ively modeling time-varying appearance loosens correspondences. 
Sophisticated regularization from Shape-from-Shading \cite{zhang1999shape, barron2014shape,zhang2021nerfactor} might help. 
To preserve time consistency, we do not update the canonical model with newly visible geometry from $t{>}1$. 
However, multi-view input mitigates the effect of occlusions at $t{=}1$. 
Finally, reconstructing topology changes \emph{in a time-consistent manner} remains an open challenge in the field and we do not currently aim to address it.

\section{Conclusion}\label{sec:conclusion}

SceNeRFlow demonstrates the great potential of modern neural scene representations for time-consistent reconstruction of general dynamic objects. 
In particular, we showed how backwards deformation models can be adapted to tracking large motion. 
As a proof of concept, we demonstrated how SceNeRFlow enables simple edits that are consistent over time. 
We also showed how conditioning the canonical model on time trades off novel-view and correspondence quality. 
As discussed in Sec.~\ref{sec:limitations}, we make some simplifying assumptions %
but exploit them as little as possible, paving the way for future extensions to weaker inputs.  

{
\small
\noindent\textbf{Acknowledgements.} 
All data capture and evaluation was done at MPII. 
Research conducted by Vladislav Golyanik and Christian Theobalt at MPII was supported in part by the ERC Consolidator Grant 4DReply (770784). 
This work was also supported by a Meta Reality Labs research grant.
}

{\small
\bibliographystyle{ieeenat_fullname}
\bibliography{main}
}

\newpage
\appendix

\section*{Supplementary Material}

We provide details on how we visualize correspondences in Sec.~\ref{sec:correspondence_visualization}, 
a description of each scene in the dataset in Sec.~\ref{sec:data_details}, 
per-scene quantitative novel-view-synthesis results in Sec.~\ref{sec:per_scene}, 
novel-view-synthesis comparisons with D-NeRF and DeVRF in Sec.~\ref{sec:nvs_prior}, 
more qualitative joint-tracking and novel-view results in Sec.~\ref{sec:more_novel_view}, 
quantitative ablation results in Sec.~\ref{sec:quant_ablation}, 
further architecture and training details in Sec.~\ref{sec:further_architecture}, 
details on how we adapt the baselines to joint tracking in Sec.~\ref{sec:joint_evaluation_details}, 
and more details on the foreground masks used for evaluation in Sec.~\ref{sec:foreground_masks}.

\section{Correspondence Visualization Details}\label{sec:correspondence_visualization}

\begin{figure}
    \centering
    \begin{tabular}{cc}
    \includegraphics[width=0.45\columnwidth]{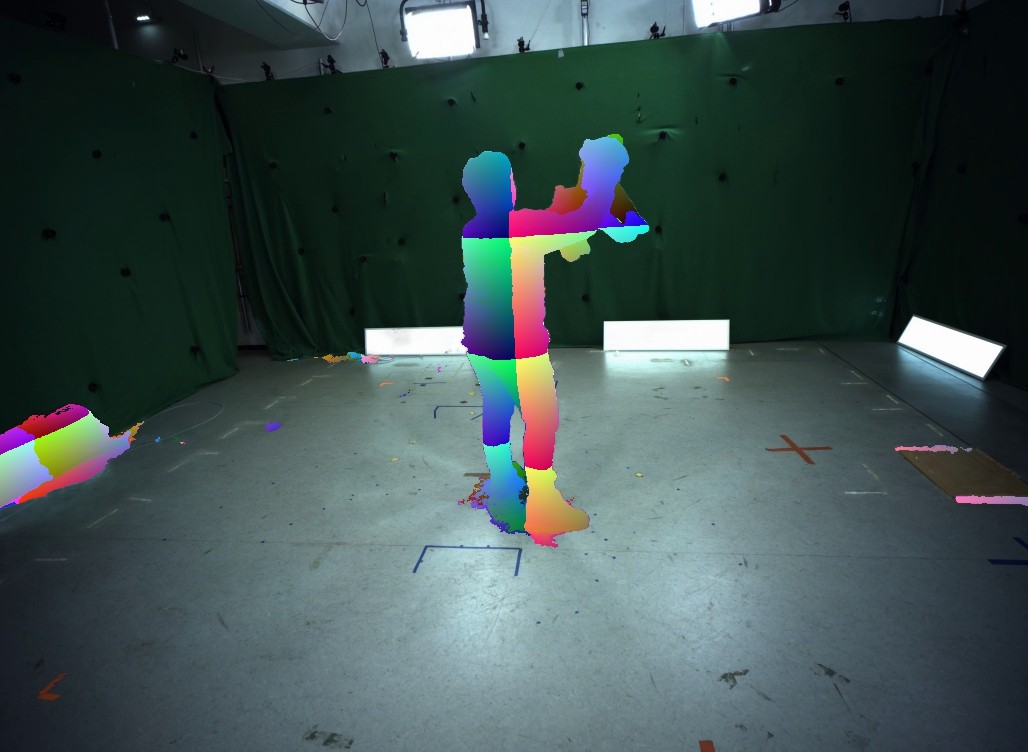}
    &\includegraphics[width=0.45\columnwidth]{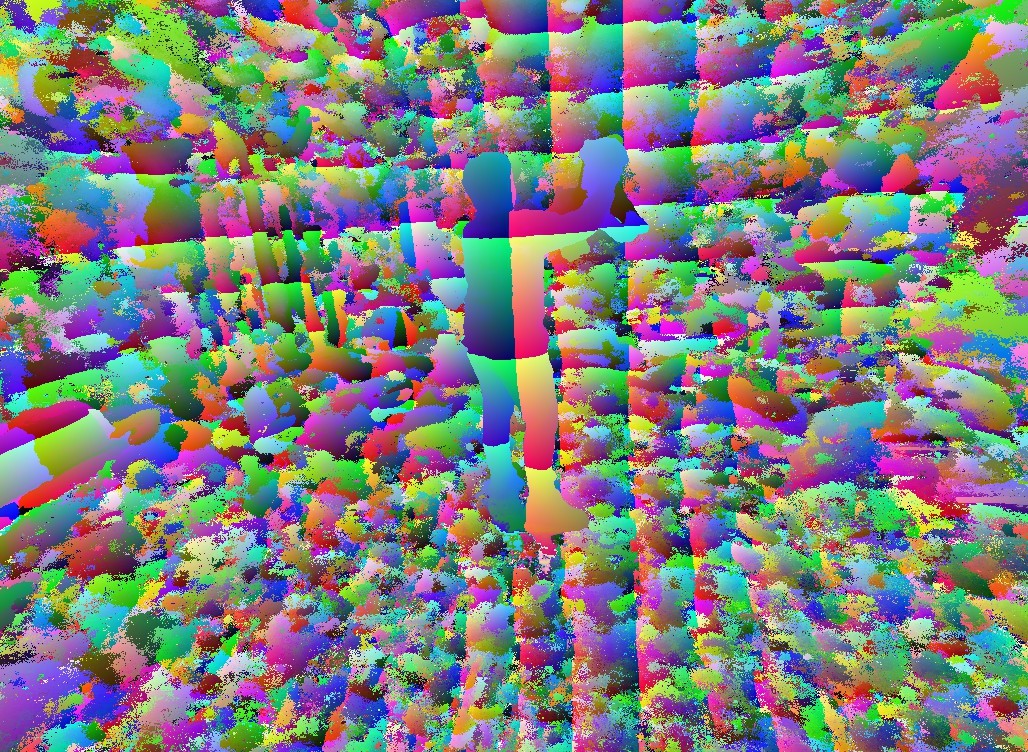}
    \\
    With Filtering &  Without Filtering\\
    \end{tabular}
    
    \caption{\textbf{Filtering for Correspondence Visualization.} For better visualization, we filter out rays with an accumulated transmittance below $0.4$. }
    \label{fig:correspondence_filtering}
\end{figure}

We follow NR-NeRF's~\cite{tretschk2021non} visualization and replace the appearance in the canonical model with a voxel grid of RGB cubes. 
Like prior work~\cite{park2021nerfies,tretschk2021non}, we pick that sample $i'$ of the ray as the surface point $\mathbf{r}(s_{i'})$ that is closest to an accumulated transmittance $\sum_{i=1}^{i'}w_i$ of $0.5$. 
For better visualization, we filter rays with a total accumulated transmittance below $0.4$ (\emph{i.e.} those hitting the background) and visualize them as transparent. 
Fig.~\ref{fig:correspondence_filtering} shows an example.

\section{Data Details}\label{sec:data_details}

Tab.~\ref{tab:scene_descriptions} contains a description and the length in frames of each scene. 
We record at $25$fps.

\section{Per-Scene Quantitative Results}\label{sec:per_scene}

We collect the quantitative results per scene for novel-view synthesis in Tab.~\ref{tab:novel_view_per_scene}.

\section{Novel-View-Synthesis Comparison with Prior Work}\label{sec:nvs_prior}

Tab.~\ref{tab:more_novel_view} contains further comparisons, without Seq.~8 due to memory constraints. 
SceNeRFlow outperforms D-NeRF and DeVRF, which both fail on large motion.

\section{More Qualitative Joint-Tracking and Novel-View Results}\label{sec:more_novel_view}

\begin{figure}
    \centering
    \setlength{\tabcolsep}{1pt}
    \begin{tabular}{cccc}

    & Ours & PREF & NR-NeRF \\
    
    \parbox[t]{2mm}{\rotatebox[origin=c]{90}{Seq. $1$}}
    &
    \raisebox{-0.5\height}{\includegraphics[trim={50 50 50 30},clip,width=0.3\columnwidth]{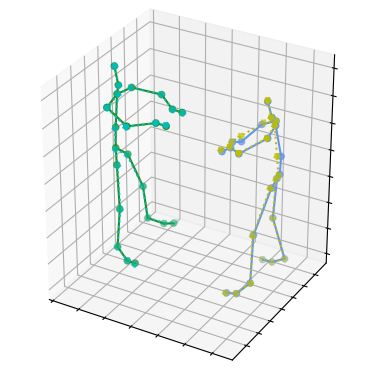}}
    &
    \raisebox{-0.5\height}{\includegraphics[trim={50 50 50 30},clip,width=0.3\columnwidth]{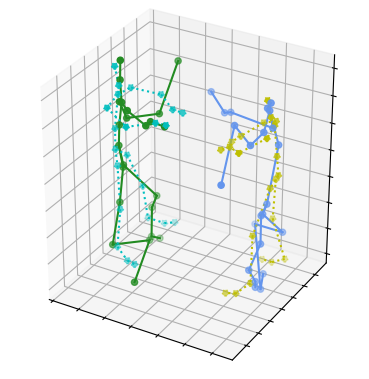}}
    &
    \raisebox{-0.5\height}{\includegraphics[trim={50 50 50 30},clip,width=0.3\columnwidth]{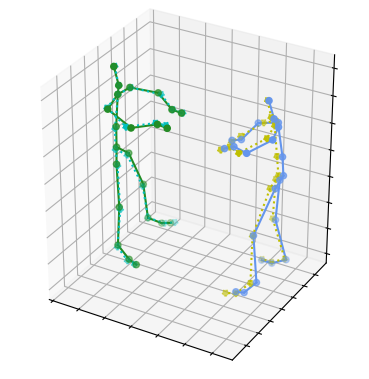}}
    \\

    \parbox[t]{2mm}{\rotatebox[origin=c]{90}{Seq. $2$}}
    &
    \raisebox{-0.5\height}{\includegraphics[trim={50 50 50 30},clip,width=0.3\columnwidth]{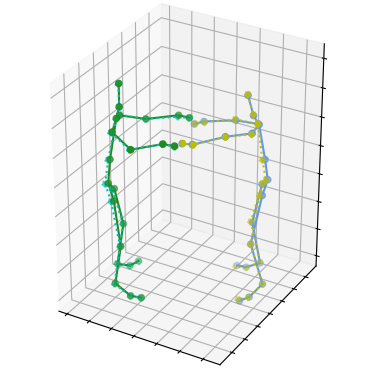}}
    &
    \raisebox{-0.5\height}{\includegraphics[trim={50 50 50 30},clip,width=0.3\columnwidth]{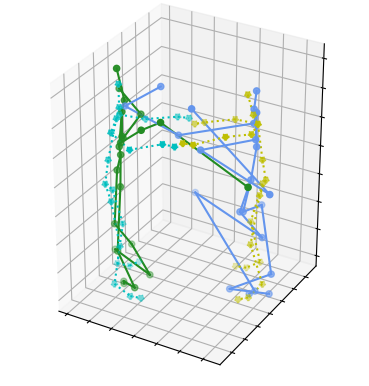}}
    &
    \raisebox{-0.5\height}{\includegraphics[trim={50 50 50 30},clip,width=0.3\columnwidth]{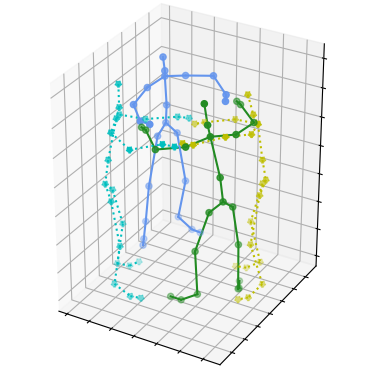}}
    \\

    \parbox[t]{2mm}{\rotatebox[origin=c]{90}{Seq. $3$}}
    &
    \raisebox{-0.5\height}{\includegraphics[trim={50 50 90 30},clip,width=0.3\columnwidth]{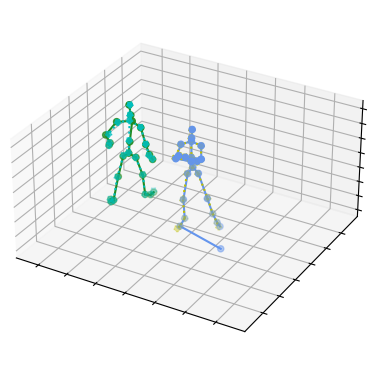}}
    &
    \raisebox{-0.5\height}{\includegraphics[trim={50 50 90 30},clip,width=0.3\columnwidth]{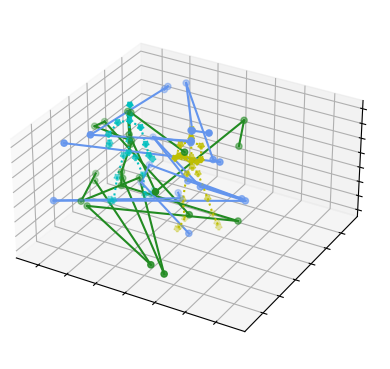}}
    &
    \raisebox{-0.5\height}{\includegraphics[trim={50 50 90 30},clip,width=0.3\columnwidth]{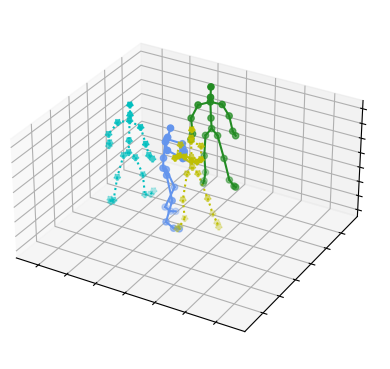}}
    \\

    \parbox[t]{2mm}{\rotatebox[origin=c]{90}{Seq. $5$}}
    &
    \raisebox{-0.5\height}{\includegraphics[trim={60 60 70 40},clip,width=0.3\columnwidth]{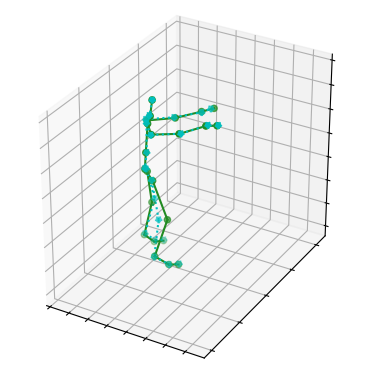}}
    &
    \raisebox{-0.5\height}{\includegraphics[trim={60 60 70 40},clip,width=0.3\columnwidth]{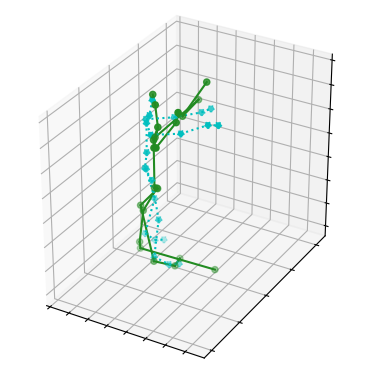}}
    &
    \raisebox{-0.5\height}{\includegraphics[trim={60 60 70 40},clip,width=0.3\columnwidth]{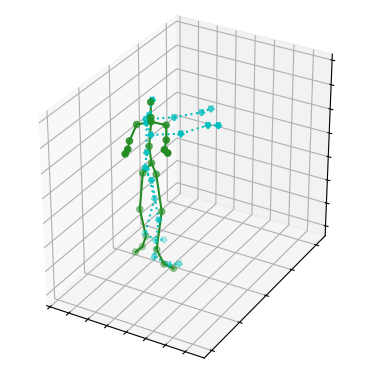}}
    \\

    \parbox[t]{2mm}{\rotatebox[origin=c]{90}{Seq. $7$}}
    &
    \raisebox{-0.5\height}{\includegraphics[trim={60 20 70 0},clip,width=0.3\columnwidth]{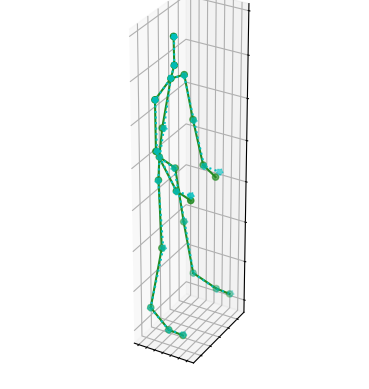}}
    &
    \raisebox{-0.5\height}{\includegraphics[trim={60 20 70 0},clip,width=0.3\columnwidth]{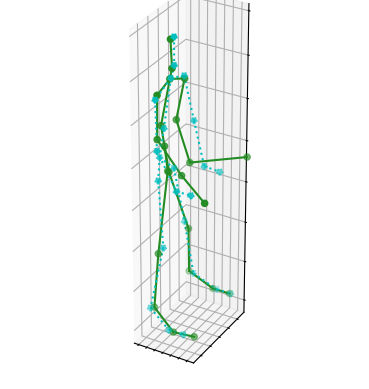}}
    &
    \raisebox{-0.5\height}{\includegraphics[trim={60 20 70 0},clip,width=0.3\columnwidth]{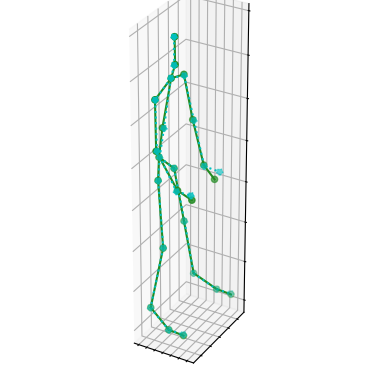}}
    \\
    
    \end{tabular}
    
    \caption{\textbf{Time Consistency.} The solid skeleton is the tracking estimate at $t{=}T$. The dotted skeleton is the pseudo-ground truth at $t{=}T$.}
    \label{fig:more_joints}
\end{figure}

\begin{figure*}
    \centering
    \footnotesize
    \setlength{\tabcolsep}{1pt}
    \begin{tabular}{cccccc}

    & Ground Truth & Ours & NR-NeRF & SNF-A & SNF-AG\\
    
    \parbox[t]{2mm}{\rotatebox[origin=c]{90}{View $1$}}
    &
    \raisebox{-0.5\height}{\includegraphics[trim={250 200 350 200},clip,width=0.19\textwidth]{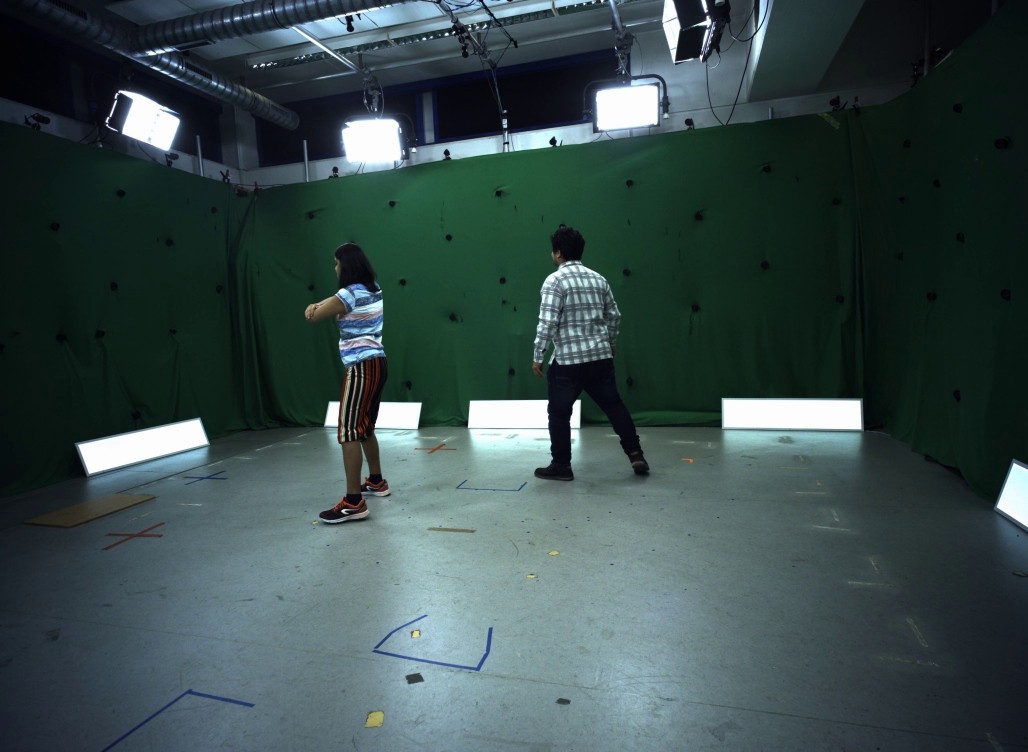}}
    &
    \raisebox{-0.5\height}{\includegraphics[trim={250 200 350 200},clip,width=0.19\textwidth]{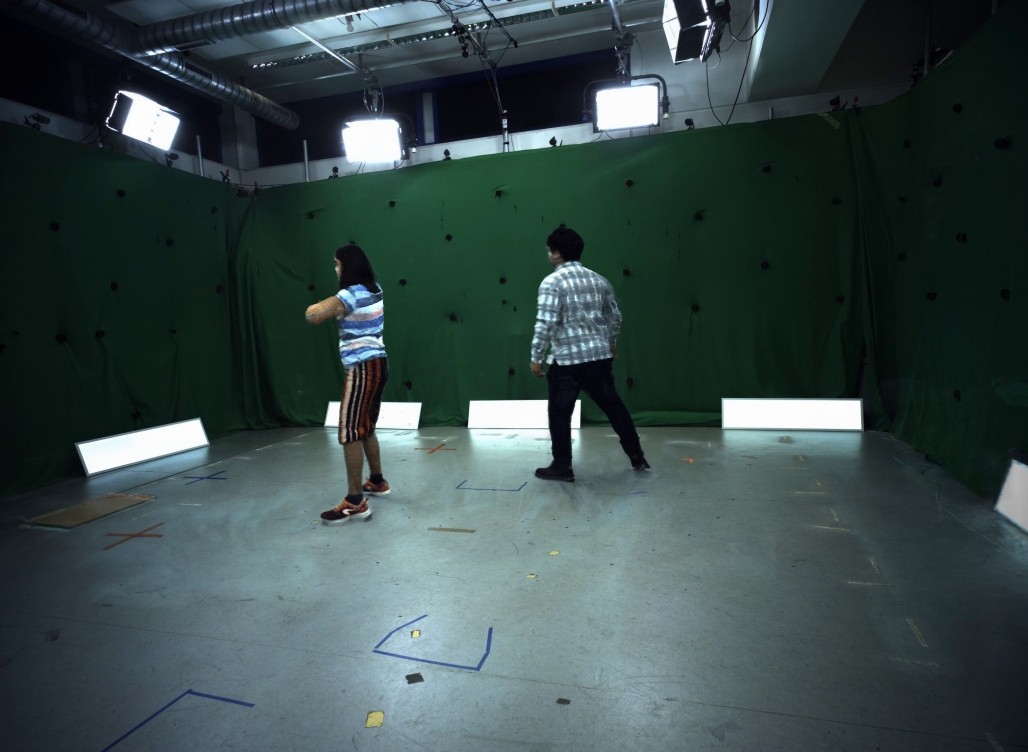}}
    &
    \raisebox{-0.5\height}{\includegraphics[trim={250 200 350 200},clip,width=0.19\textwidth]{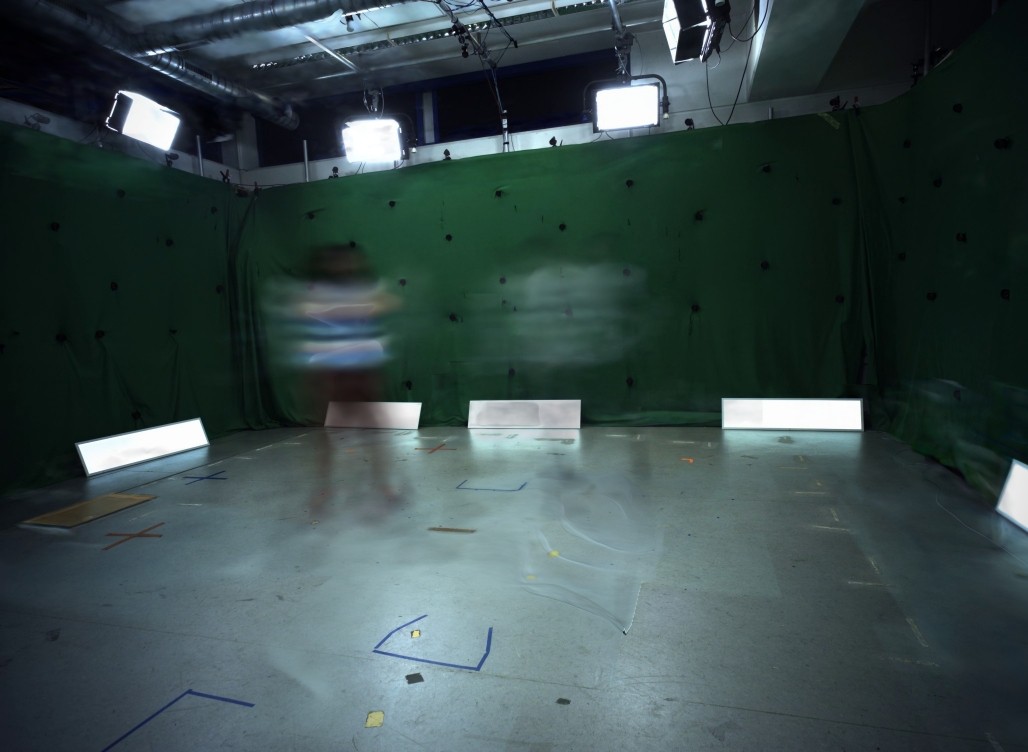}}
    &
    \raisebox{-0.5\height}{\includegraphics[trim={250 200 350 200},clip,width=0.19\textwidth]{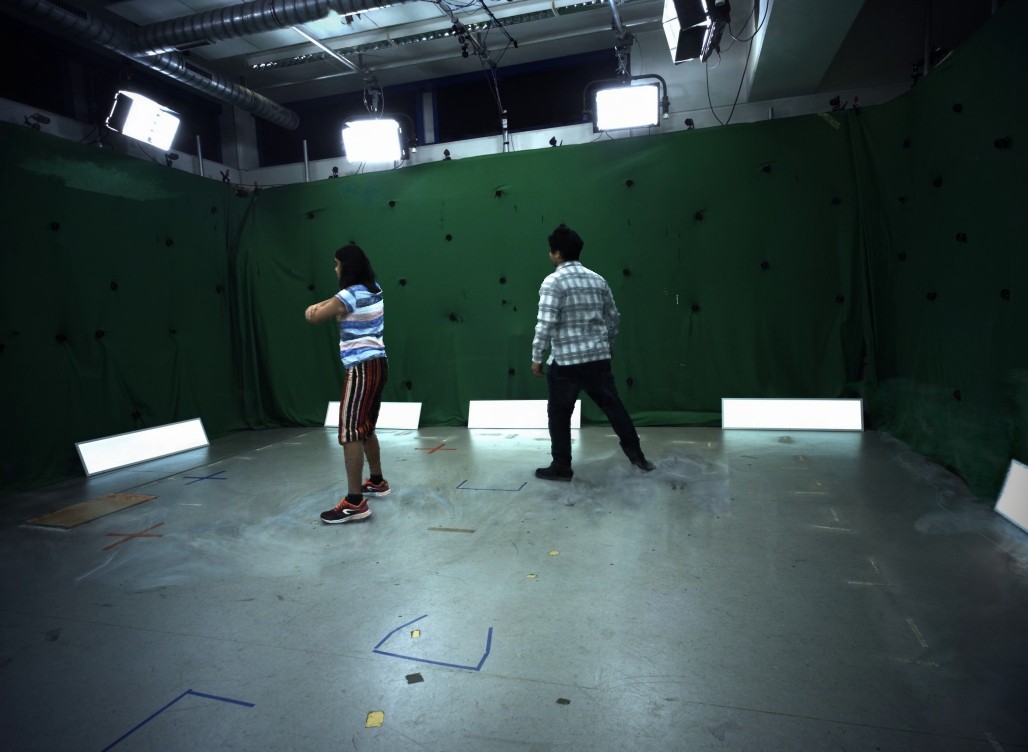}}
    &
    \raisebox{-0.5\height}{\includegraphics[trim={250 200 350 200},clip,width=0.19\textwidth]{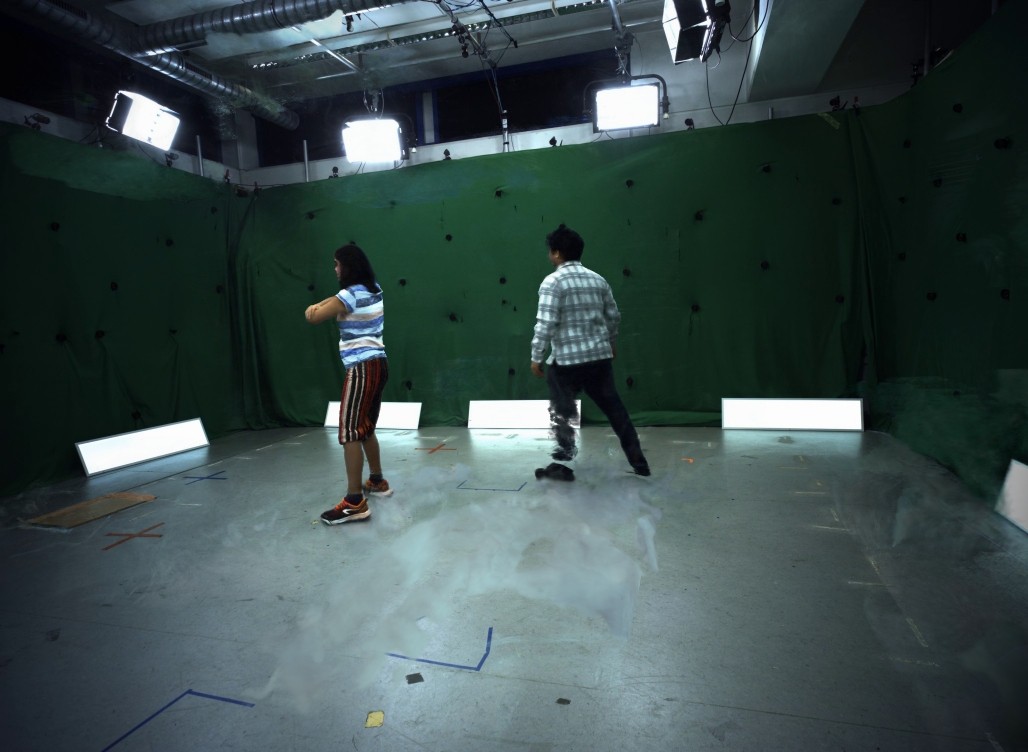}}
    \\
    
    \parbox[t]{2mm}{\rotatebox[origin=c]{90}{View $2$}}
    &
    \raisebox{-0.5\height}{\includegraphics[trim={130 100 250 120},clip,width=0.19\textwidth]{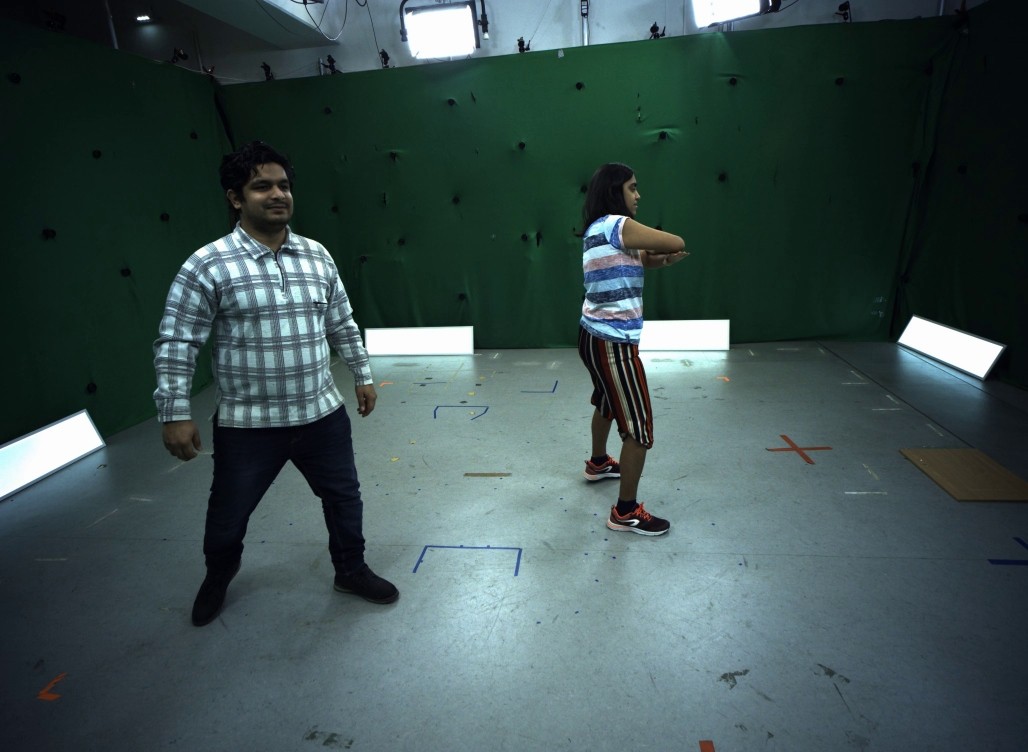}}
    &
    \raisebox{-0.5\height}{\includegraphics[trim={130 100 250 120},clip,width=0.19\textwidth]{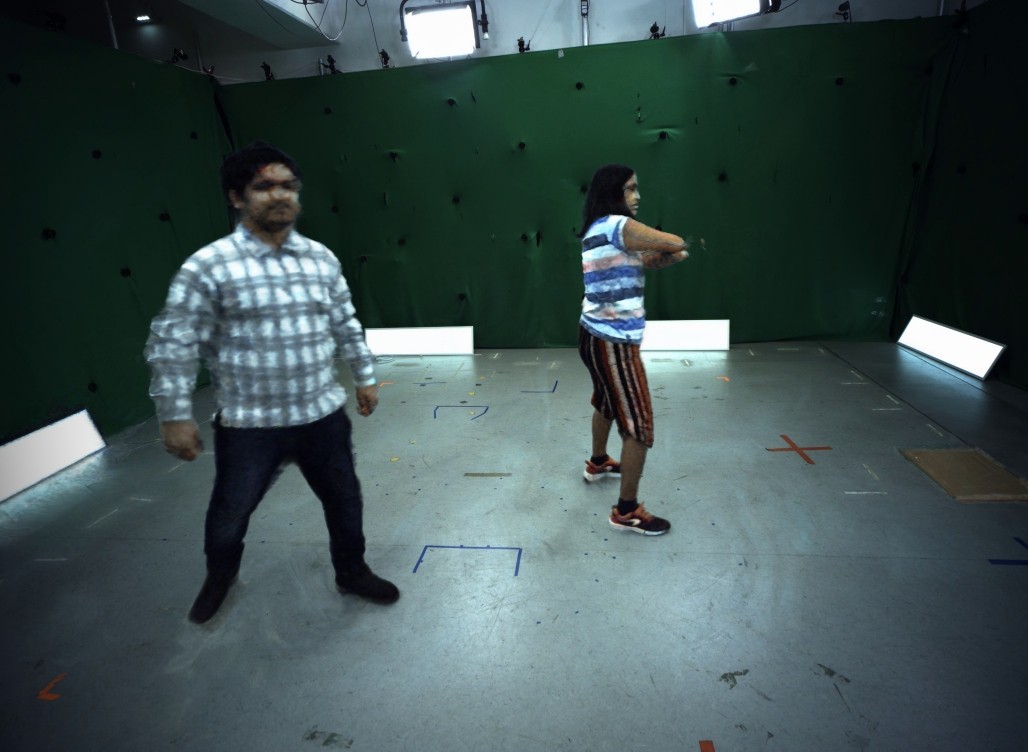}}
    &
    \raisebox{-0.5\height}{\includegraphics[trim={130 100 250 120},clip,width=0.19\textwidth]{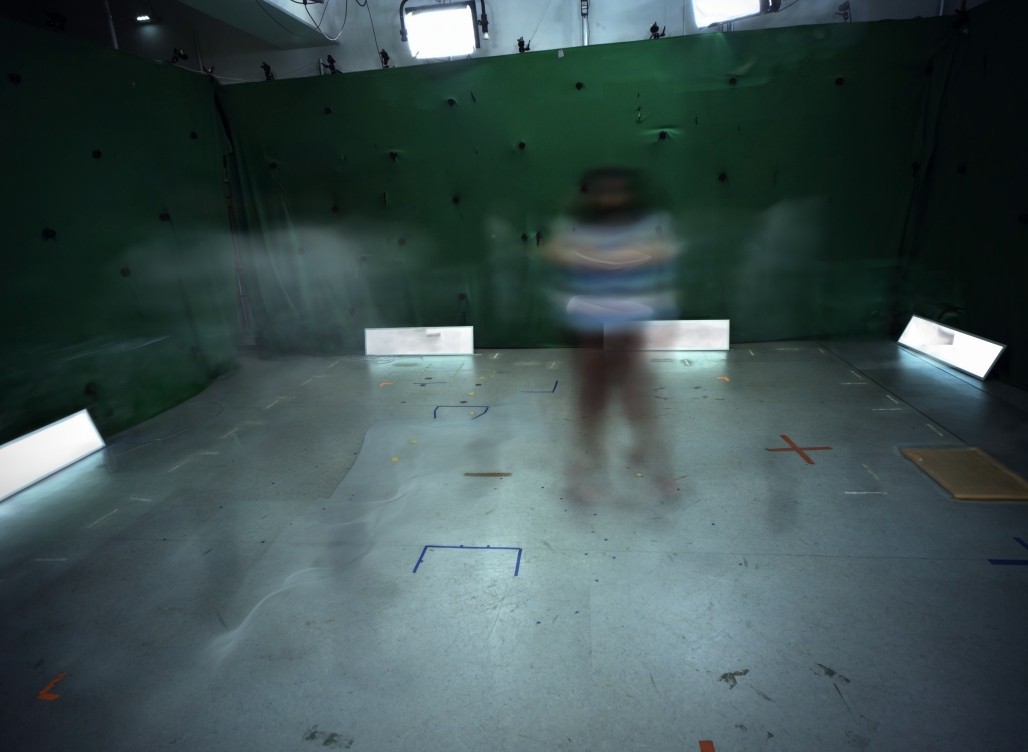}}
    &
    \raisebox{-0.5\height}{\includegraphics[trim={130 100 250 120},clip,width=0.19\textwidth]{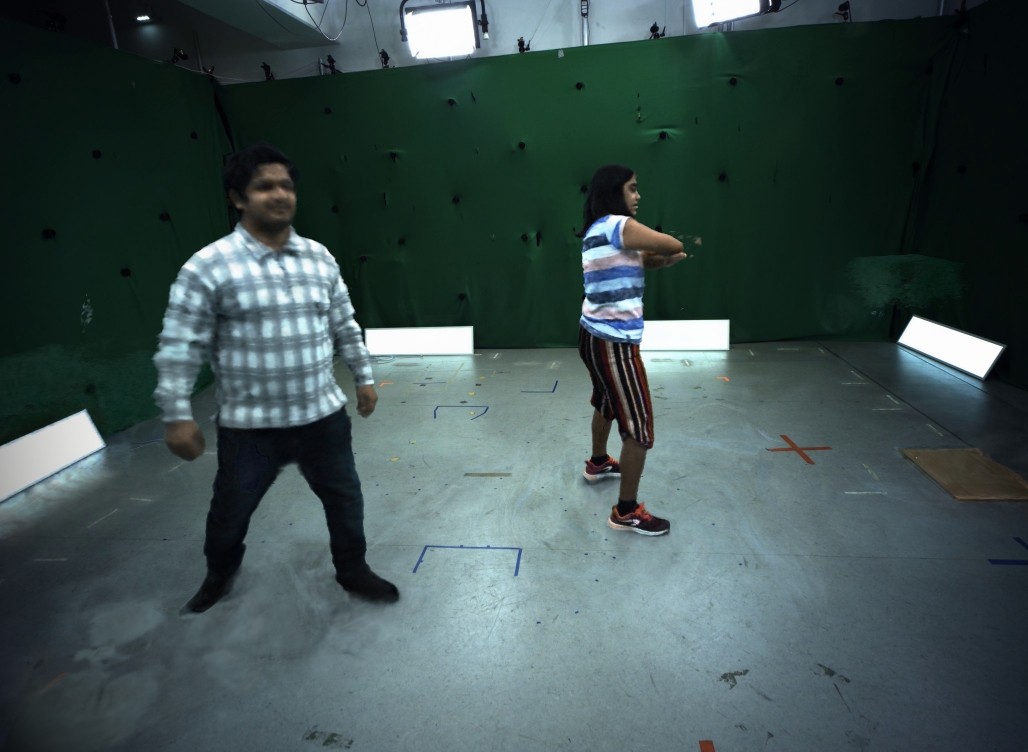}}
    &
    \raisebox{-0.5\height}{\includegraphics[trim={130 100 250 120},clip,width=0.19\textwidth]{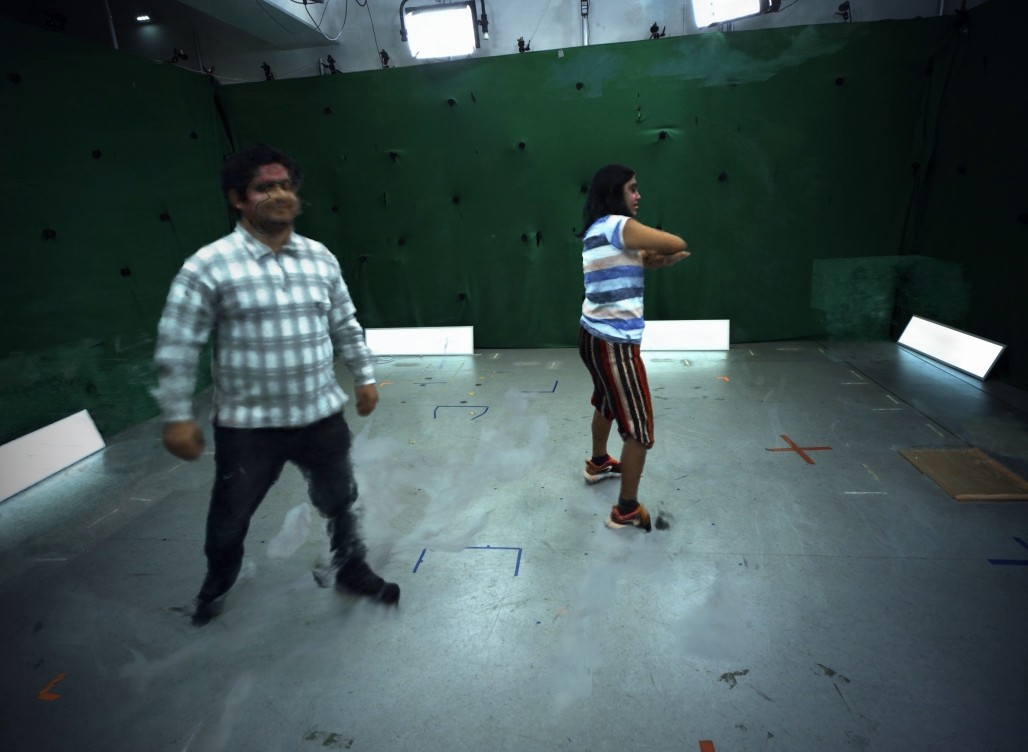}}
    \\
    
    \hline
    \hline
    
    \parbox[t]{2mm}{\rotatebox[origin=c]{90}{View $1$}}
    &
    \raisebox{-0.5\height}{\includegraphics[trim={250 200 500 200},clip,width=0.19\textwidth]{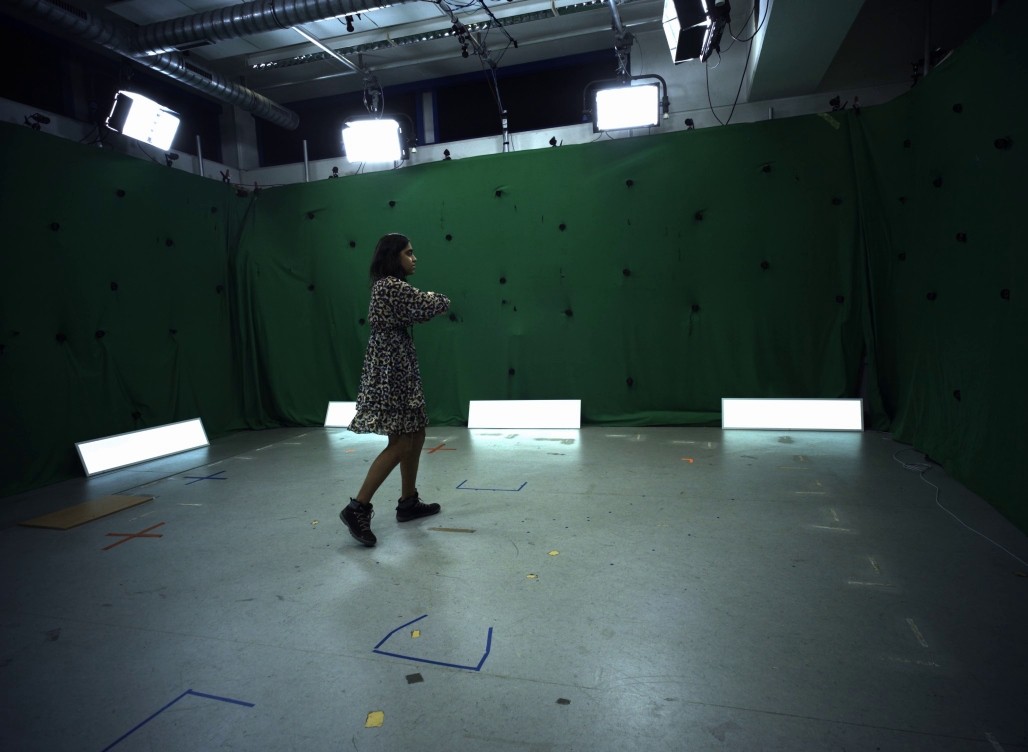}}
    &
    \raisebox{-0.5\height}{\includegraphics[trim={250 200 500 200},clip,width=0.19\textwidth]{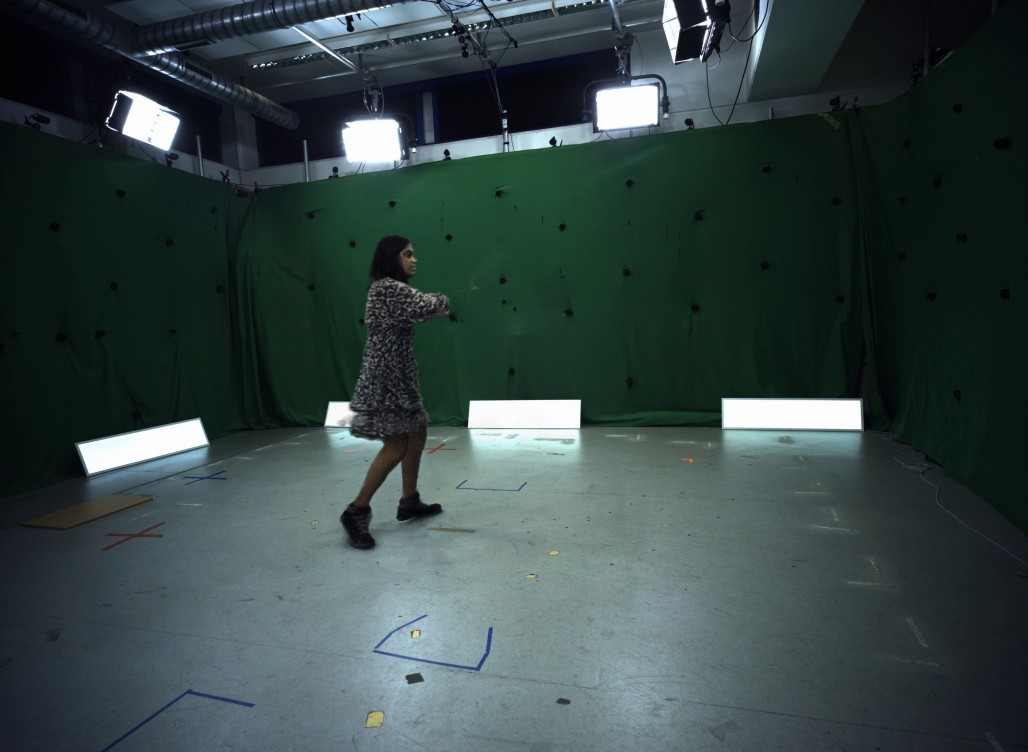}}
    &
    \raisebox{-0.5\height}{\includegraphics[trim={250 200 500 200},clip,width=0.19\textwidth]{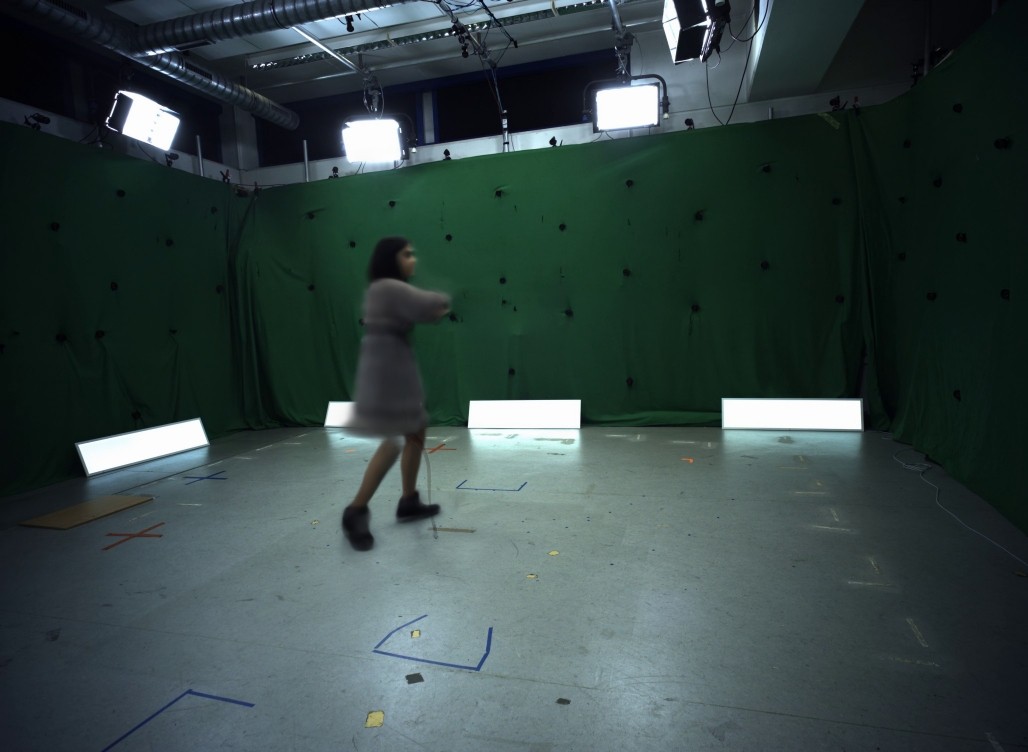}}
    &
    \raisebox{-0.5\height}{\includegraphics[trim={250 200 500 200},clip,width=0.19\textwidth]{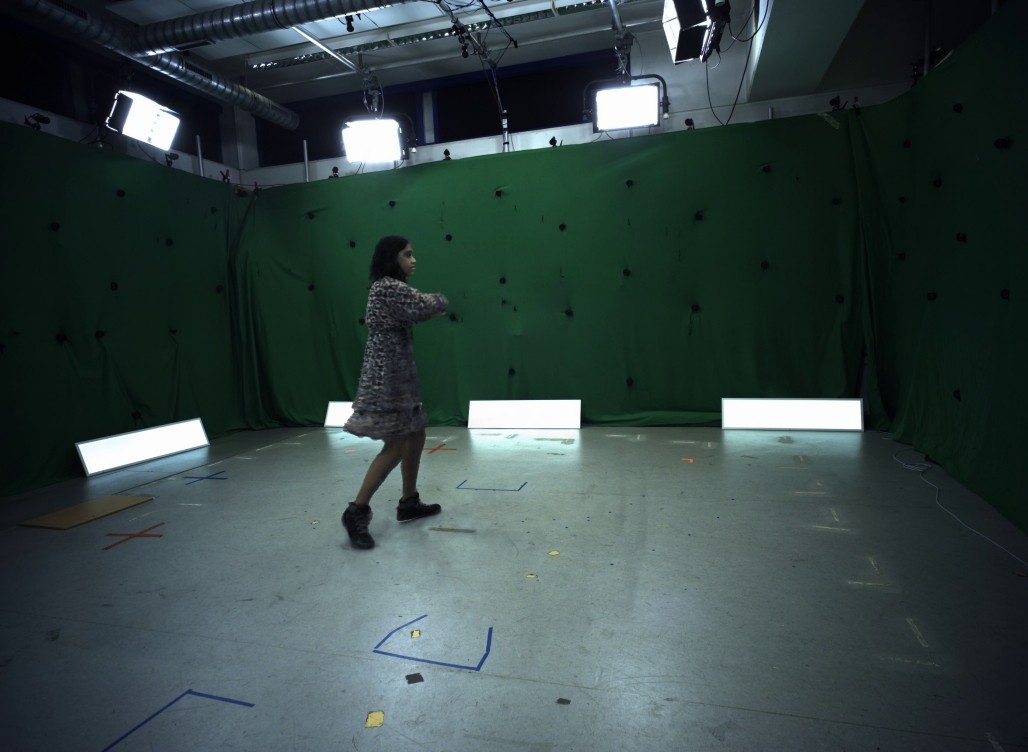}}
    &
    \raisebox{-0.5\height}{\includegraphics[trim={250 200 500 200},clip,width=0.19\textwidth]{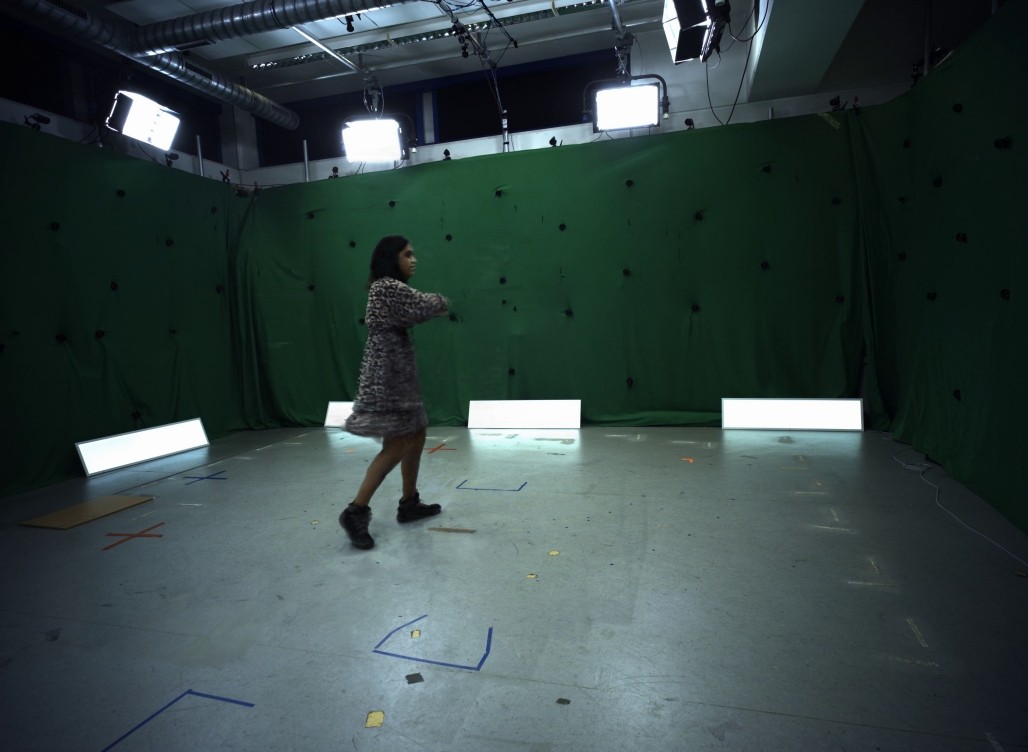}}
    \\
    
    \parbox[t]{2mm}{\rotatebox[origin=c]{90}{View $2$}}
    &
    \raisebox{-0.5\height}{\includegraphics[trim={380 220 340 160},clip,width=0.19\textwidth]{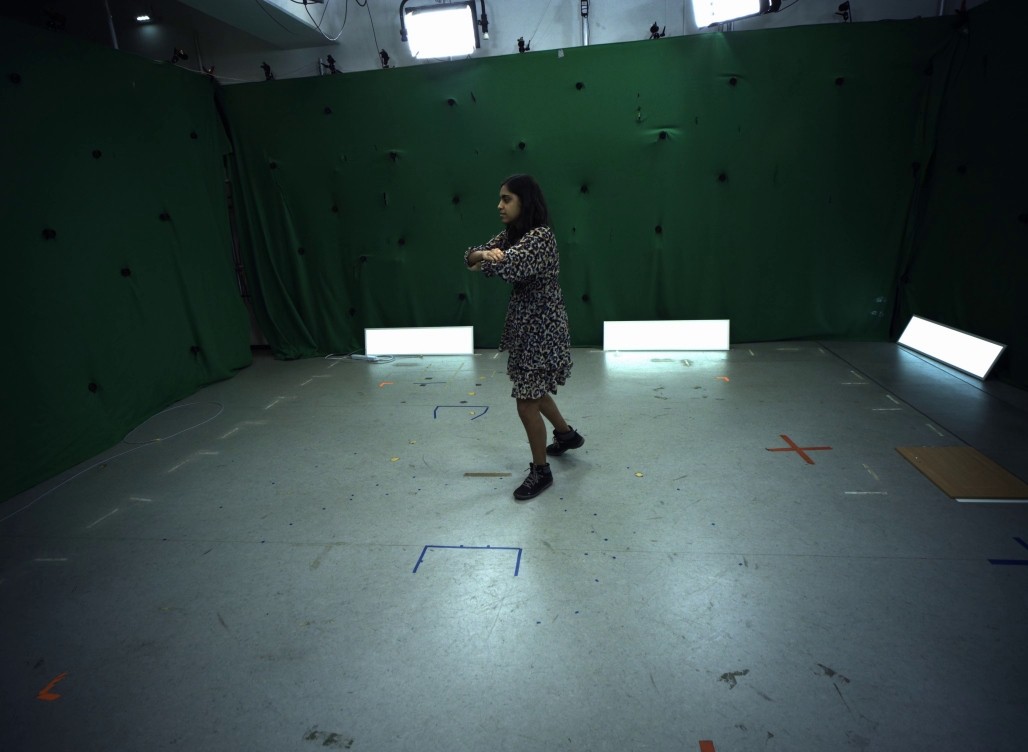}}
    &
    \raisebox{-0.5\height}{\includegraphics[trim={380 220 340 160},clip,width=0.19\textwidth]{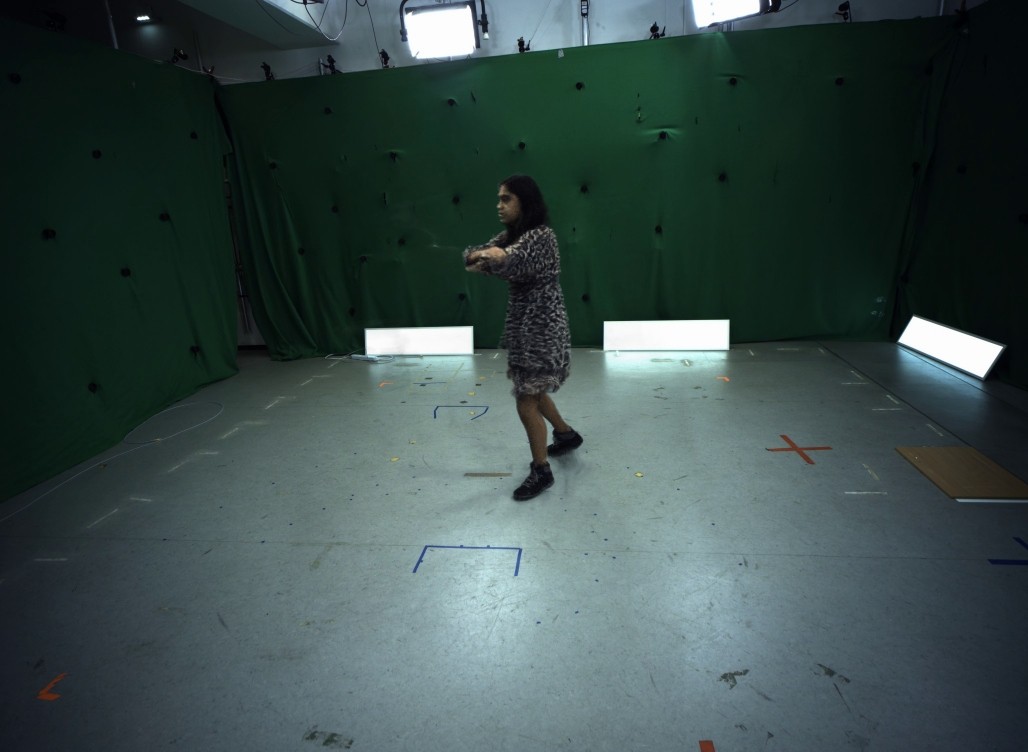}}
    &
    \raisebox{-0.5\height}{\includegraphics[trim={380 220 340 160},clip,width=0.19\textwidth]{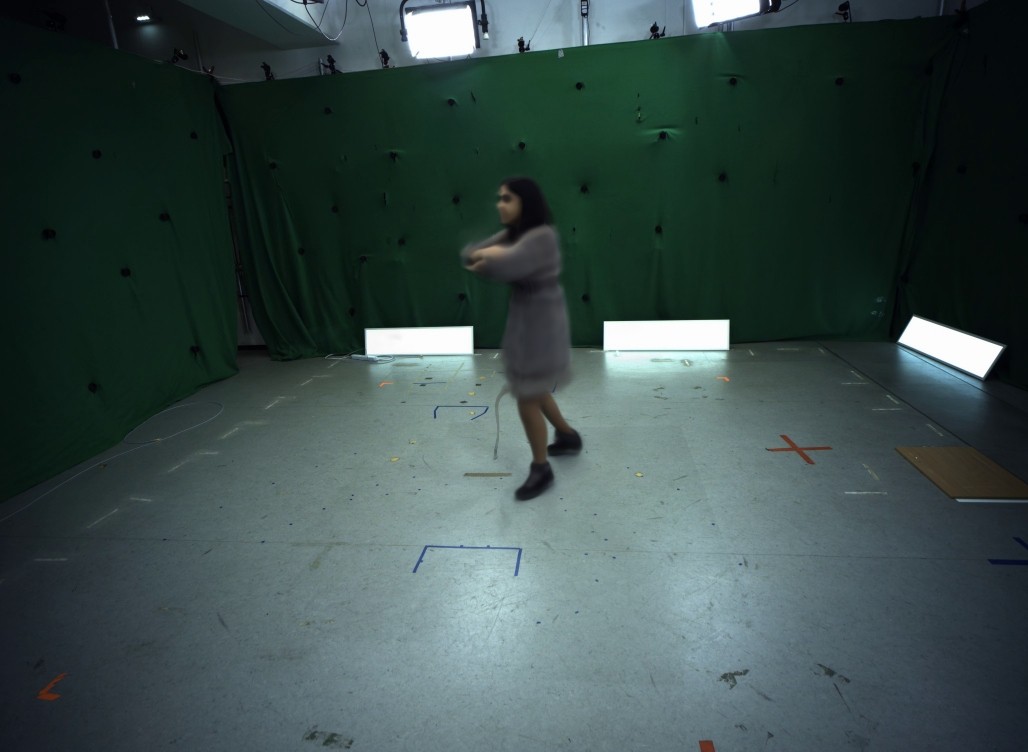}}
    &
    \raisebox{-0.5\height}{\includegraphics[trim={380 220 340 160},clip,width=0.19\textwidth]{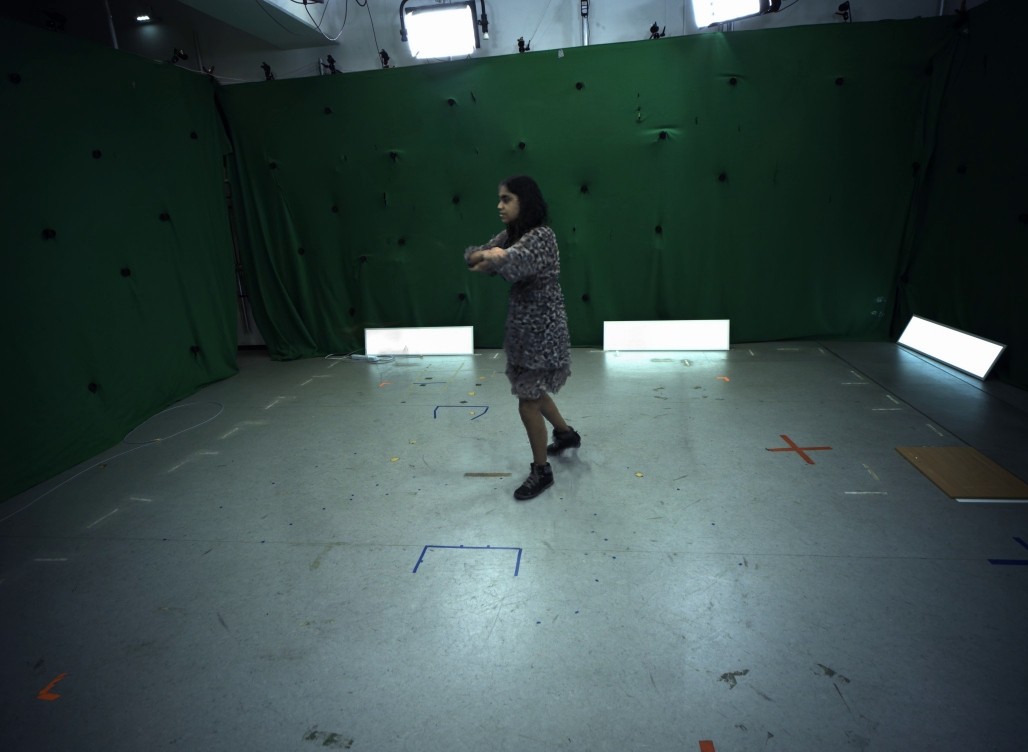}}
    &
    \raisebox{-0.5\height}{\includegraphics[trim={380 220 340 160},clip,width=0.19\textwidth]{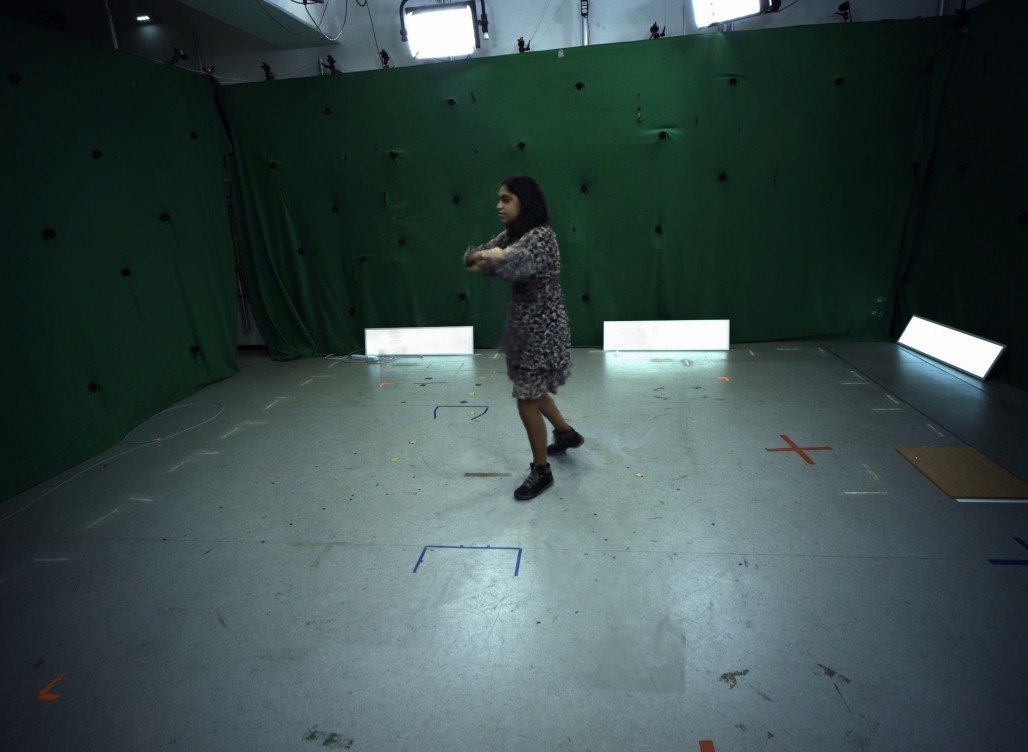}}
    
    \end{tabular}
    
    \caption{\textbf{Novel-View Synthesis.} (First row) Seq. 3 at $t{=}\frac{T}{2}$. (Second row) Seq. 4 at $t{=}\frac{T}{2}$.}
    \label{fig:more_novel_view}
\end{figure*}

\begin{figure*}
    \centering
    \footnotesize
    \setlength{\tabcolsep}{1pt}
    \begin{tabular}{cccccc}

    & Ground Truth & Ours & NR-NeRF & SNF-A & SNF-AG\\
    
    \parbox[t]{2mm}{\rotatebox[origin=c]{90}{View $1$}}
    &
    \raisebox{-0.5\height}{\includegraphics[trim={300 200 450 200},clip,width=0.19\textwidth]{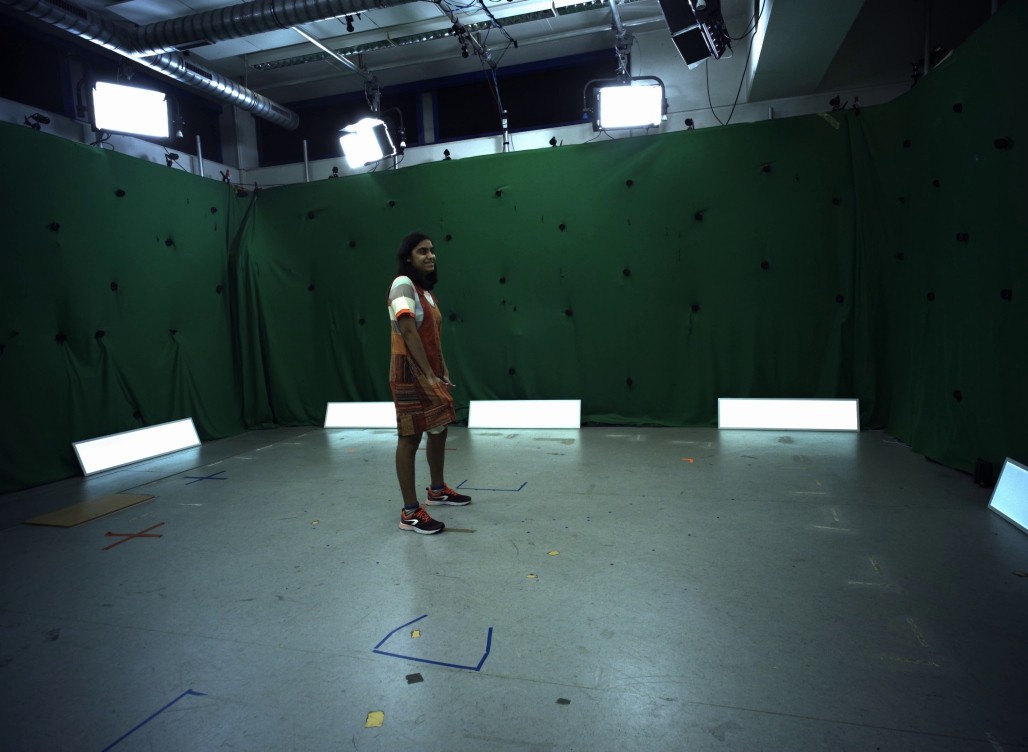}}
    &
    \raisebox{-0.5\height}{\includegraphics[trim={300 200 450 200},clip,width=0.19\textwidth]{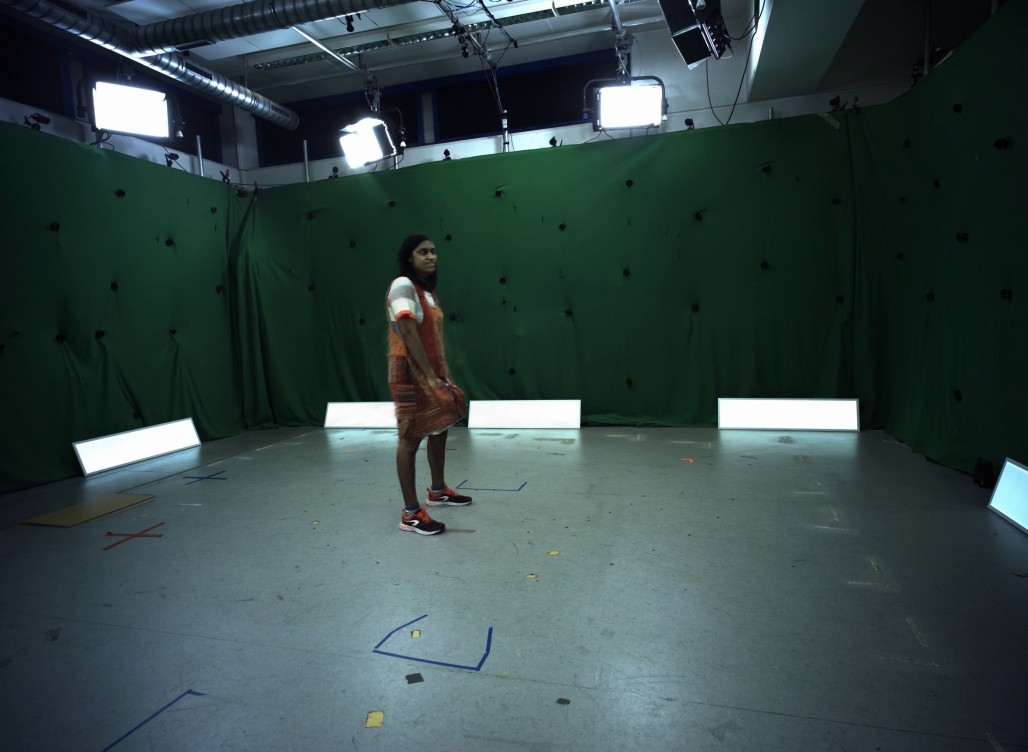}}
    &
    \raisebox{-0.5\height}{\includegraphics[trim={300 200 450 200},clip,width=0.19\textwidth]{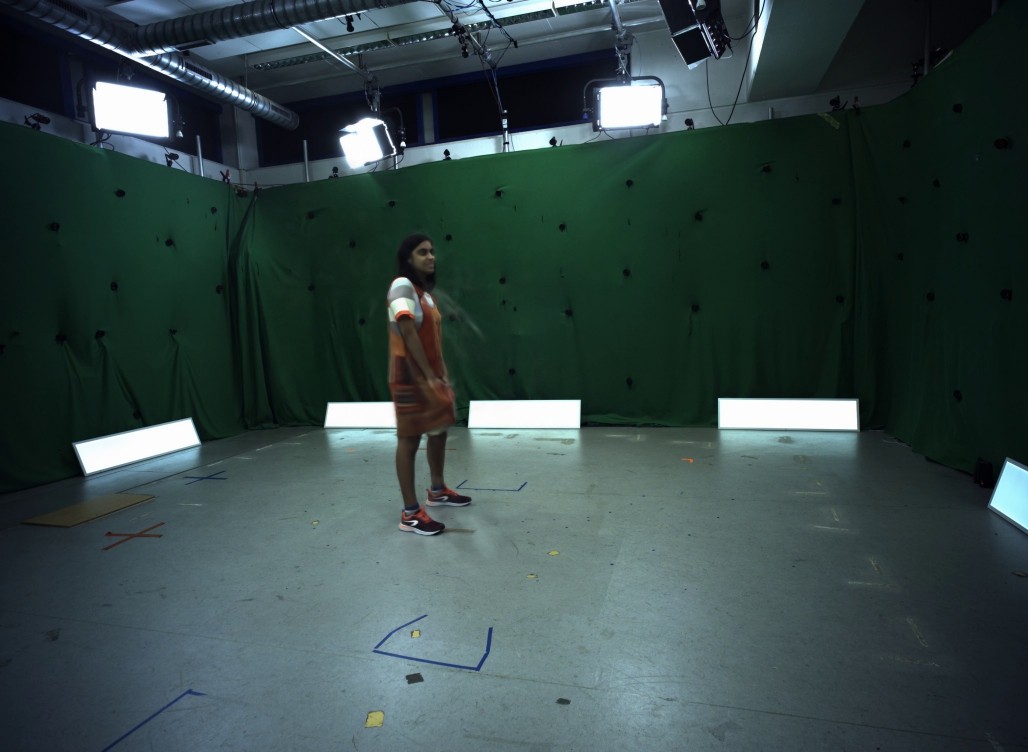}}
    &
    \raisebox{-0.5\height}{\includegraphics[trim={300 200 450 200},clip,width=0.19\textwidth]{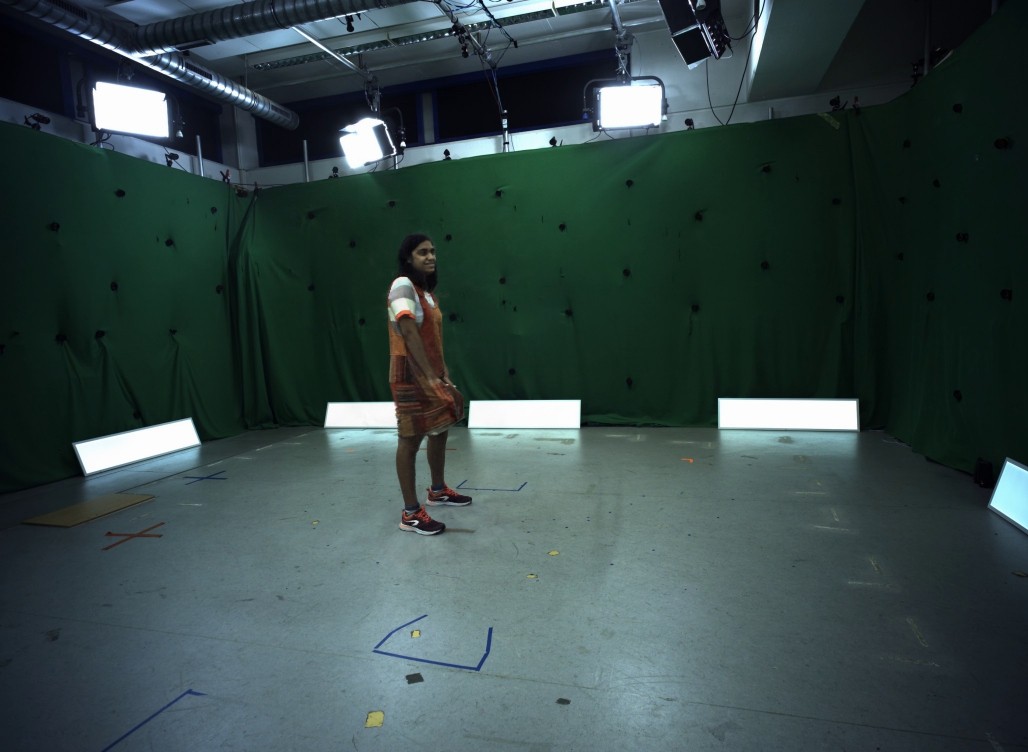}}
    &
    \raisebox{-0.5\height}{\includegraphics[trim={300 200 450 200},clip,width=0.19\textwidth]{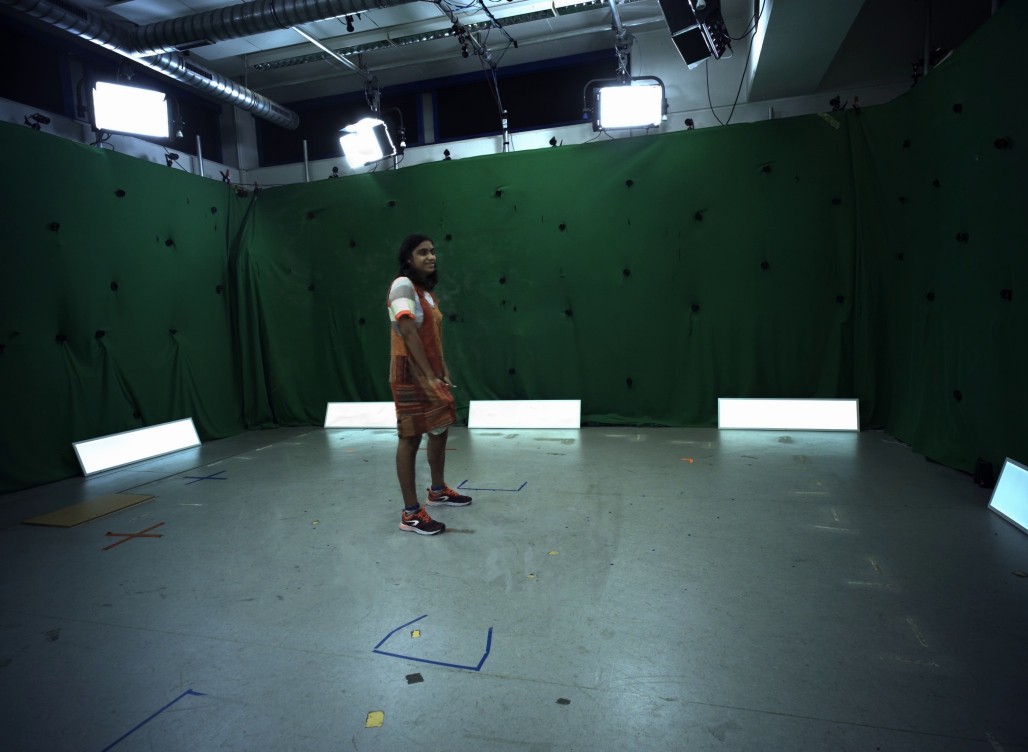}}
    \\
    
    \parbox[t]{2mm}{\rotatebox[origin=c]{90}{View $2$}}
    &
    \raisebox{-0.5\height}{\includegraphics[trim={330 200 350 120},clip,width=0.19\textwidth]{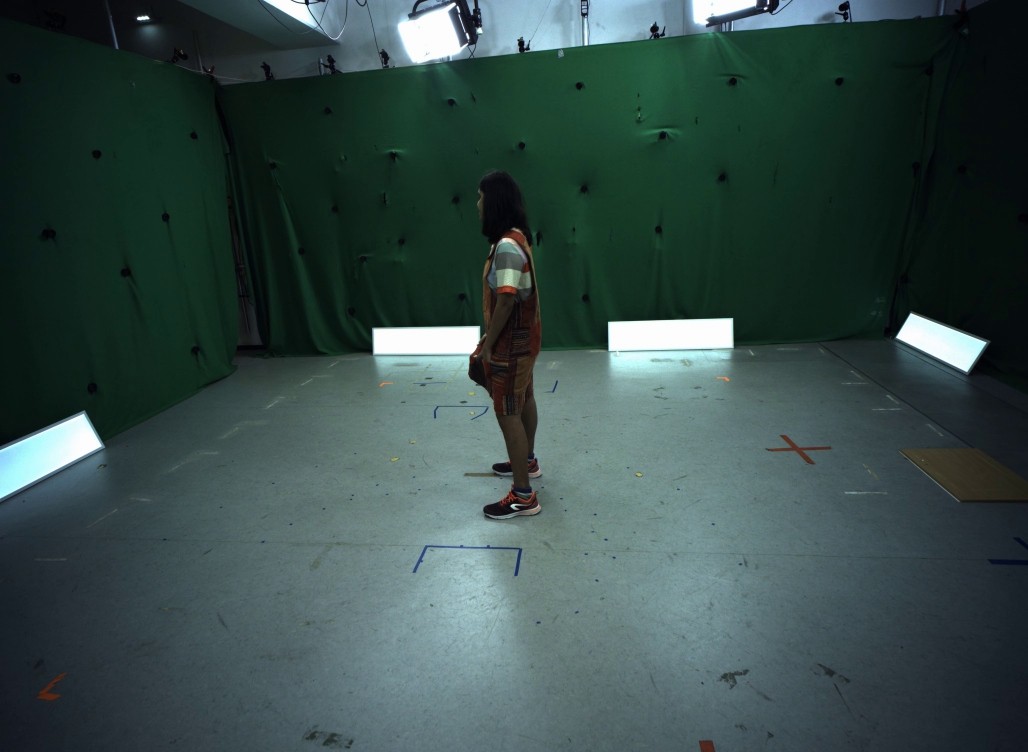}}
    &
    \raisebox{-0.5\height}{\includegraphics[trim={330 200 350 120},clip,width=0.19\textwidth]{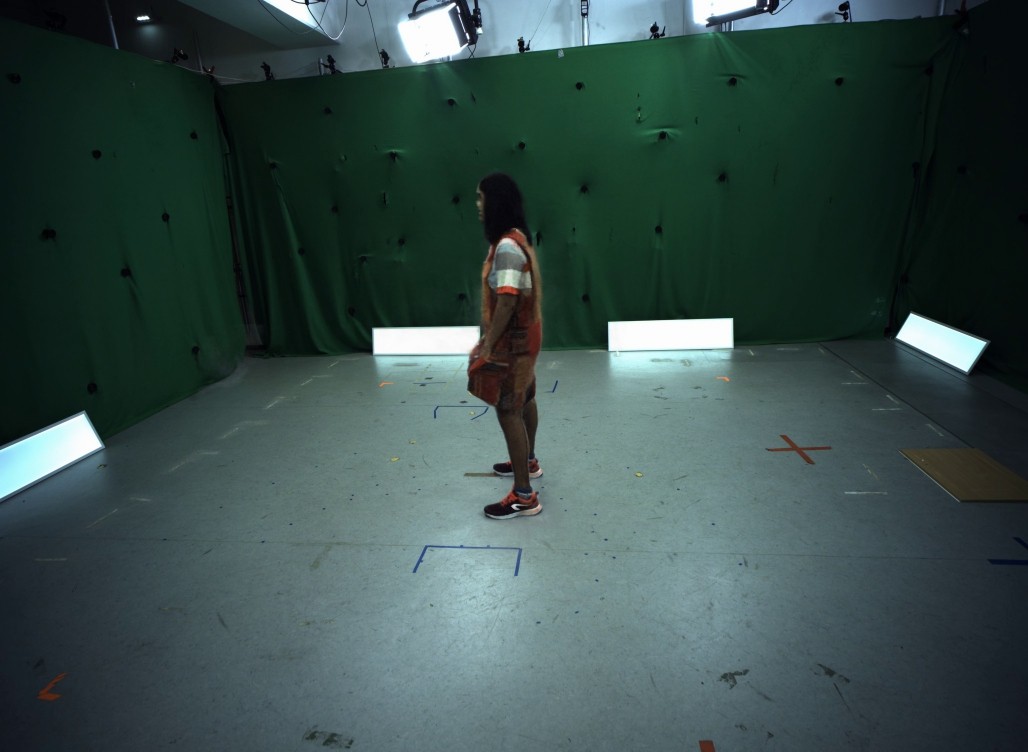}}
    &
    \raisebox{-0.5\height}{\includegraphics[trim={330 200 350 120},clip,width=0.19\textwidth]{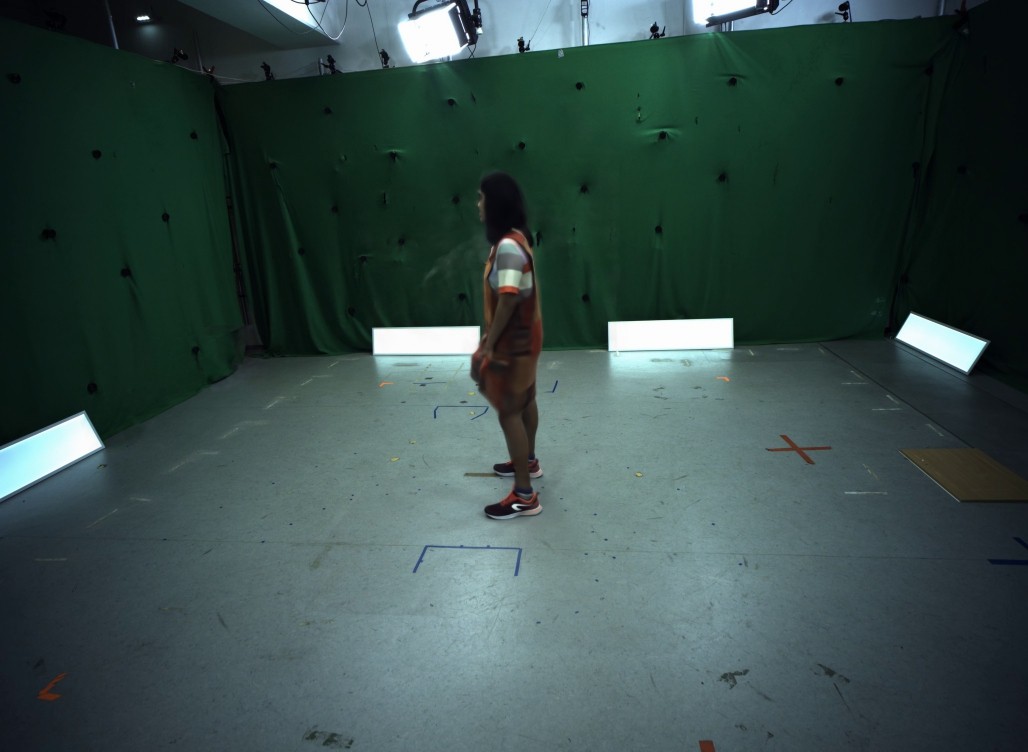}}
    &
    \raisebox{-0.5\height}{\includegraphics[trim={330 200 350 120},clip,width=0.19\textwidth]{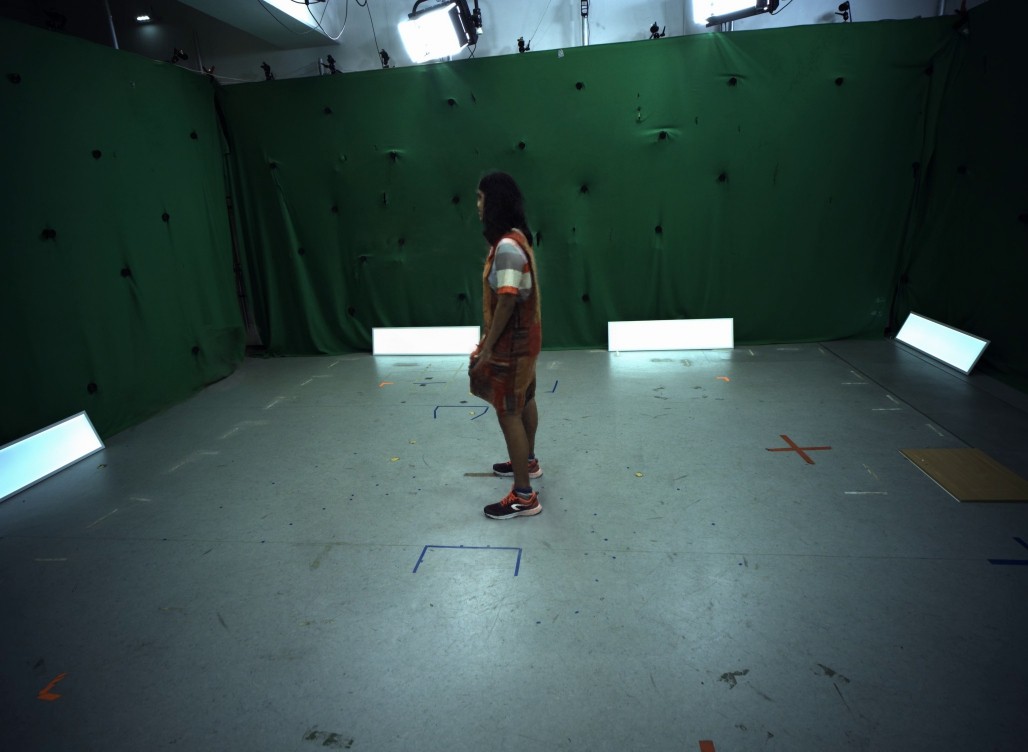}}
    &
    \raisebox{-0.5\height}{\includegraphics[trim={330 200 350 120},clip,width=0.19\textwidth]{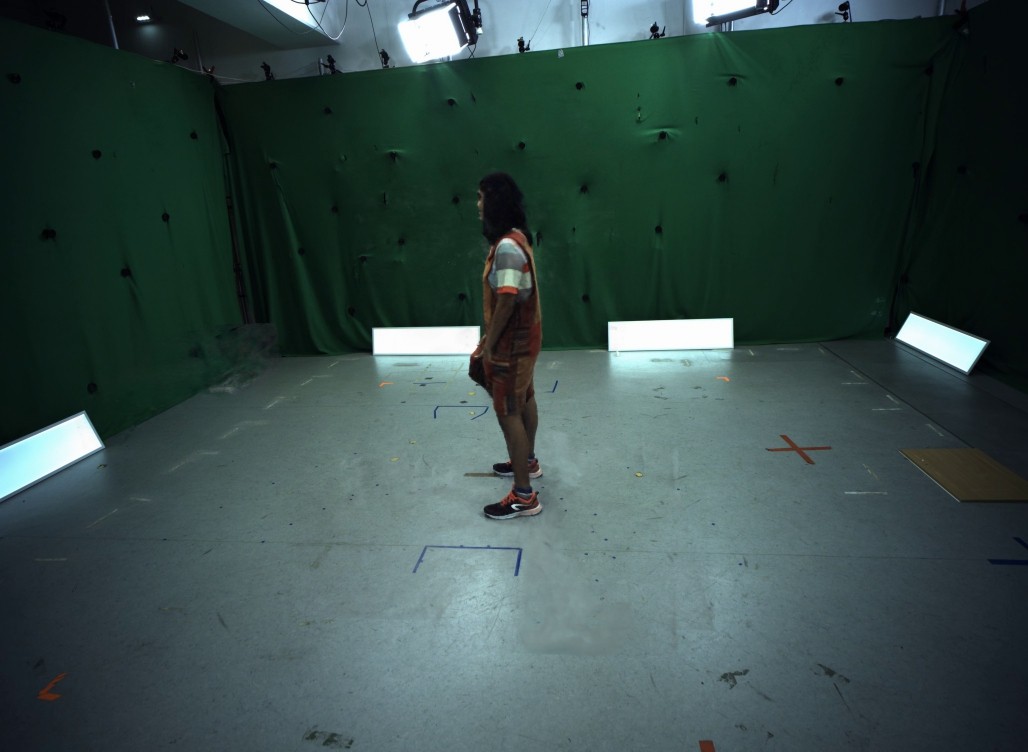}}
    \\
    
    \hline
    \hline
    
    \parbox[t]{2mm}{\rotatebox[origin=c]{90}{View $1$}}
    &
    \raisebox{-0.5\height}{\includegraphics[trim={800 300 300 500},clip,angle=-90,origin=c,width=0.19\textwidth]{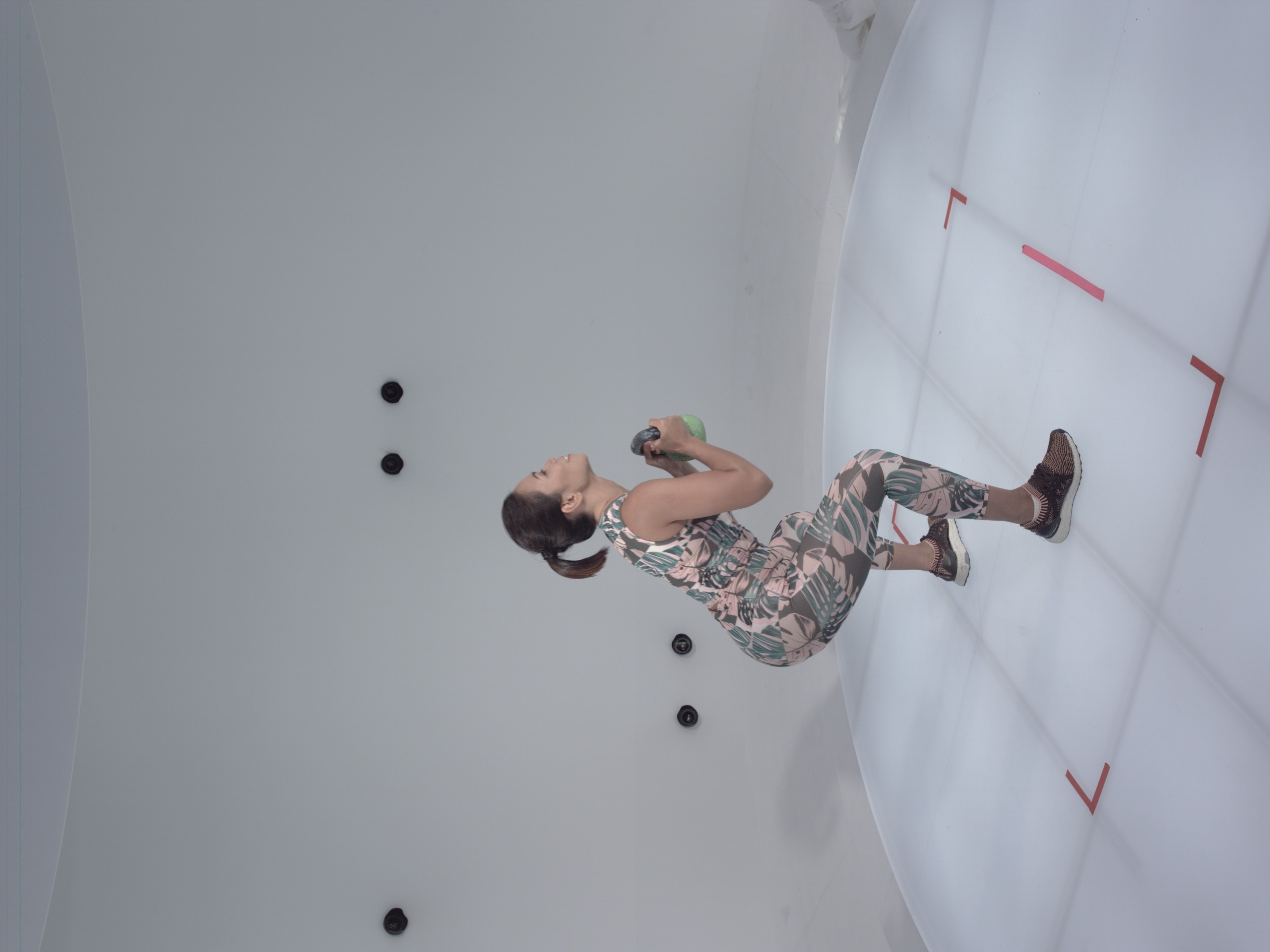}}
    &
    \raisebox{-0.5\height}{\includegraphics[trim={800 300 300 500},clip,angle=-90,origin=c,width=0.19\textwidth]{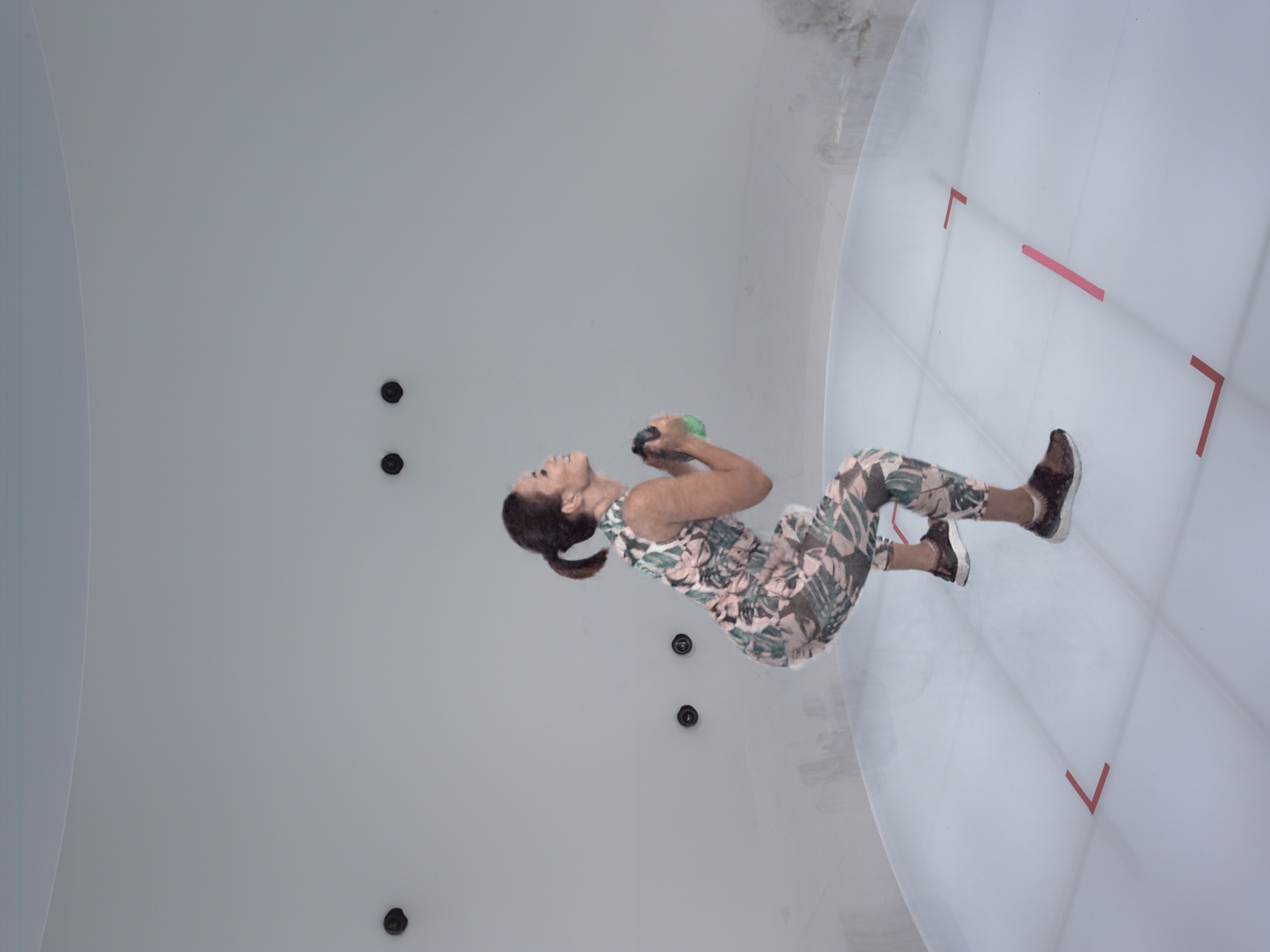}}
    &
    \raisebox{-0.5\height}{\includegraphics[trim={800 300 300 500},clip,angle=-90,origin=c,width=0.19\textwidth]{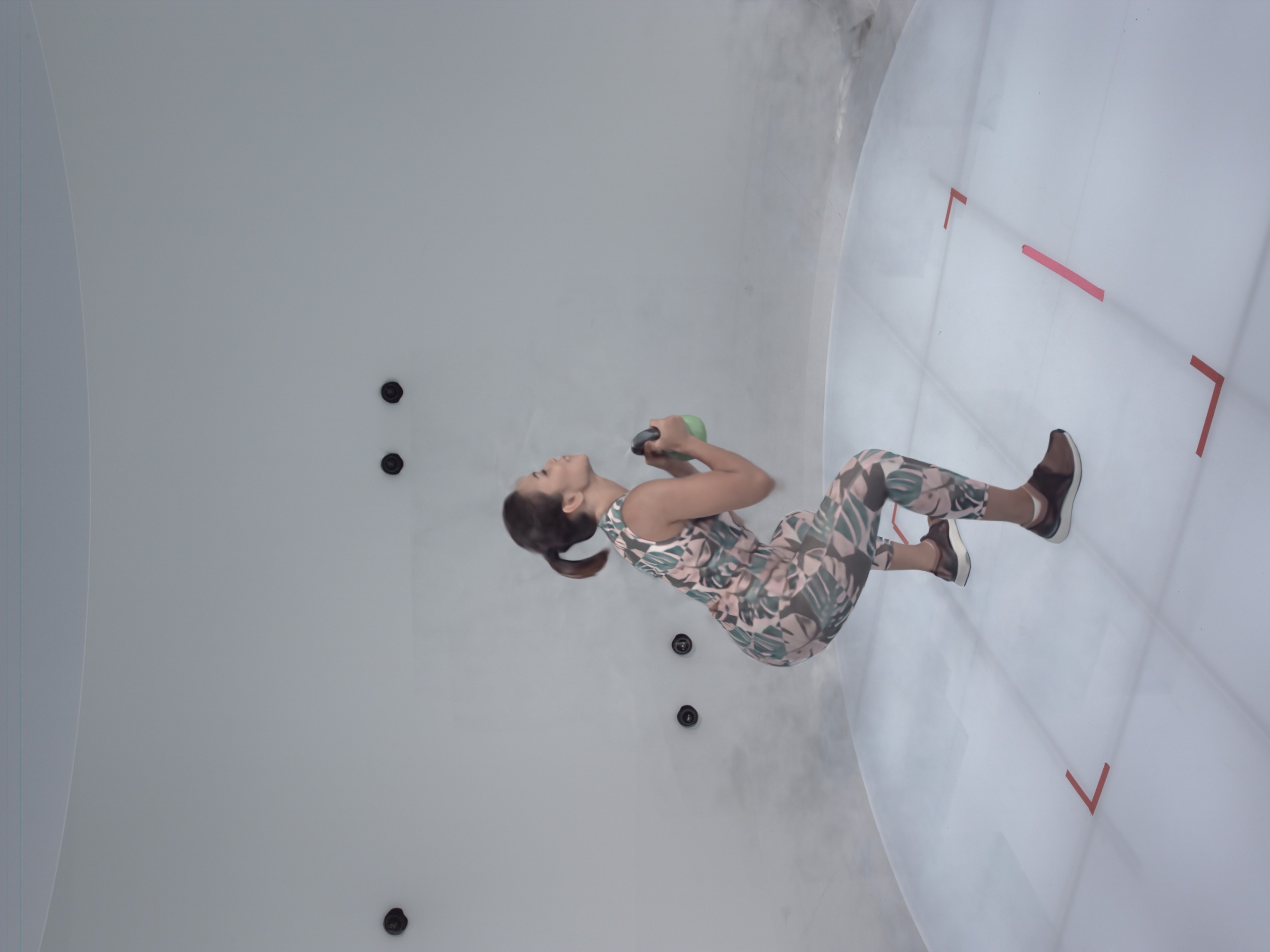}}
    &
    \raisebox{-0.5\height}{\includegraphics[trim={800 300 300 500},clip,angle=-90,origin=c,width=0.19\textwidth]{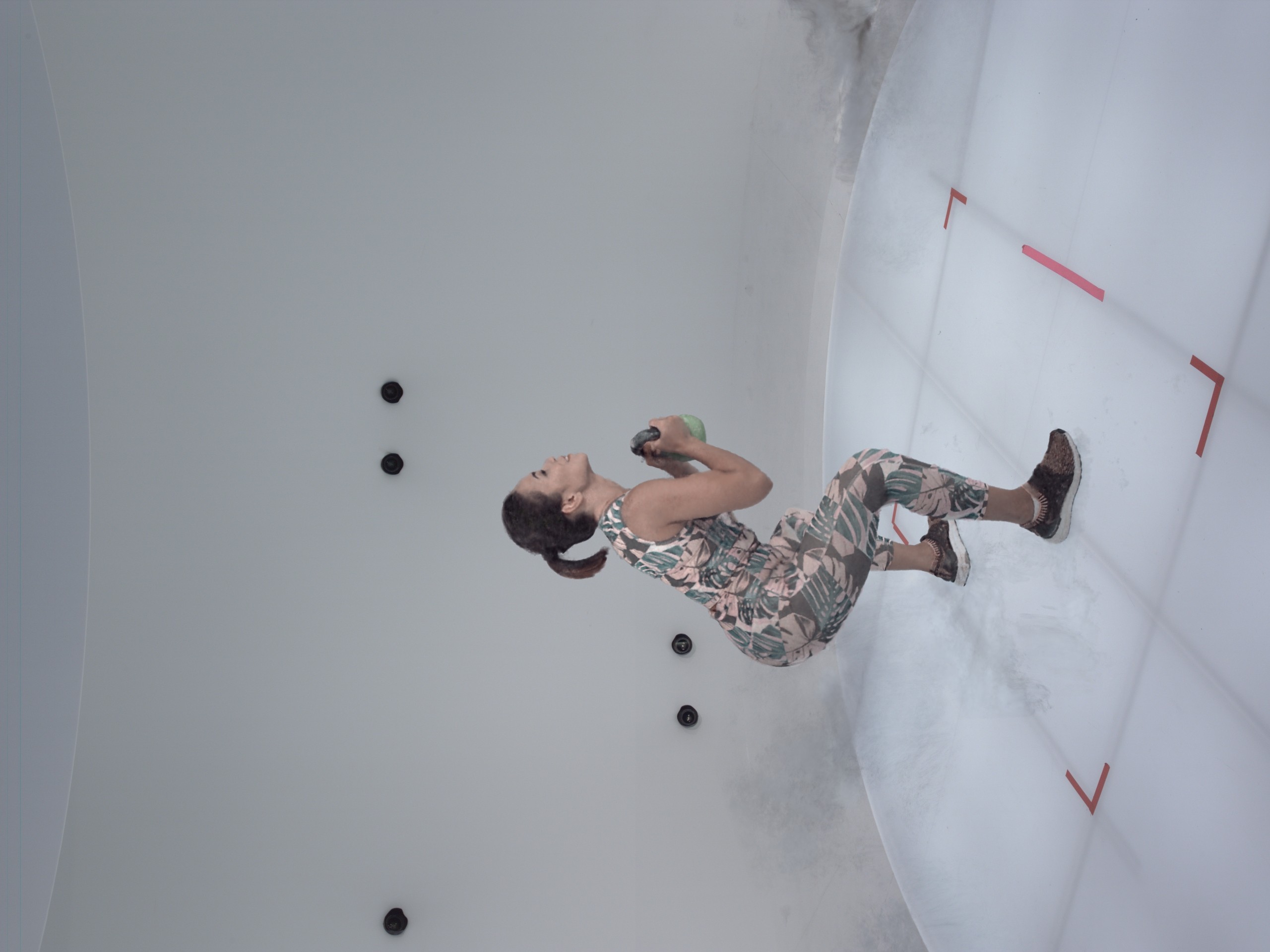}}
    &
    \raisebox{-0.5\height}{\includegraphics[trim={800 300 300 500},clip,angle=-90,origin=c,width=0.19\textwidth]{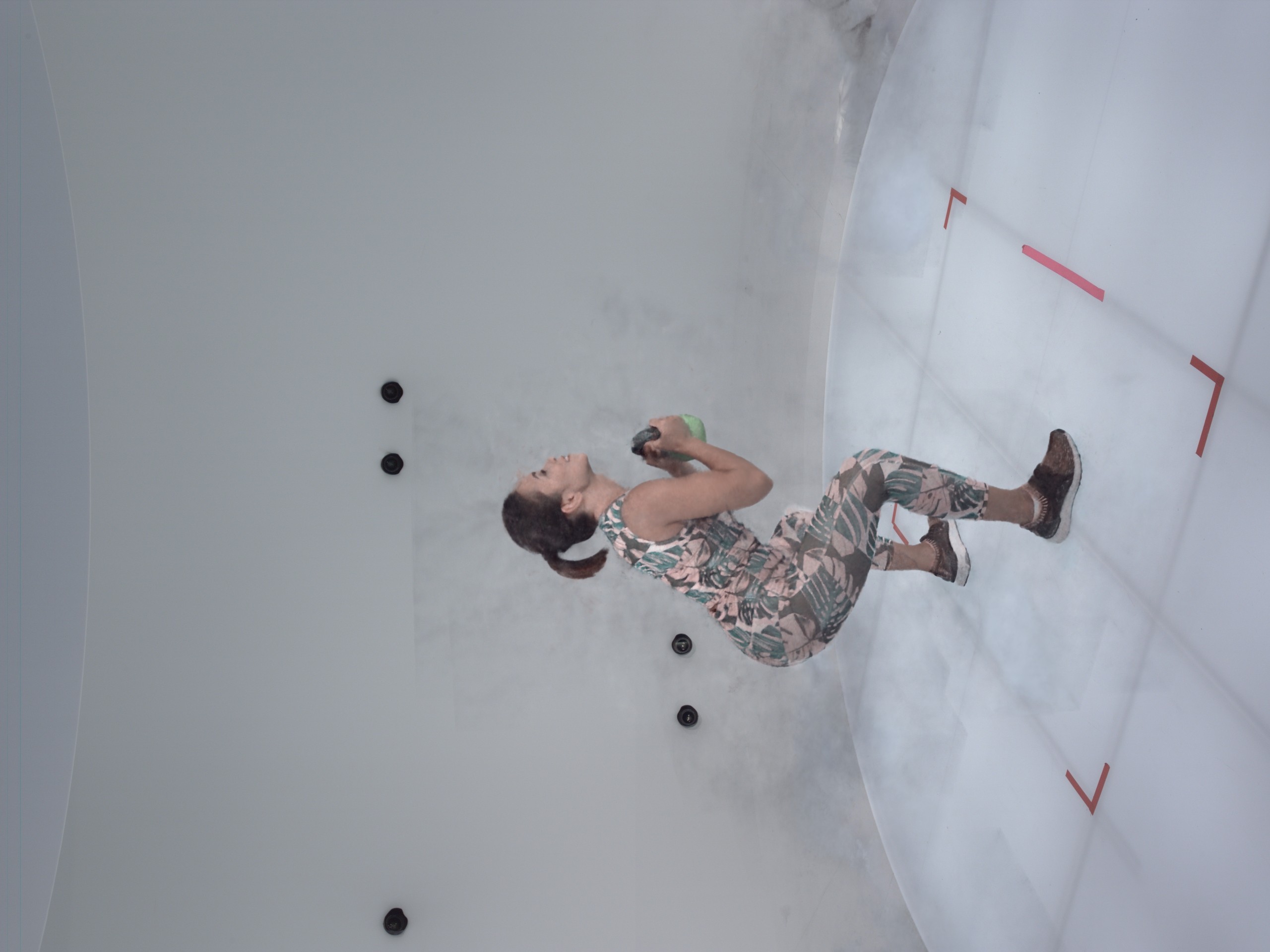}}
    \\
    
    \parbox[t]{2mm}{\rotatebox[origin=c]{90}{View $2$}}
    &
    \raisebox{-0.5\height}{\includegraphics[trim={700 550 340 260},clip,angle=-90,origin=c,width=0.19\textwidth]{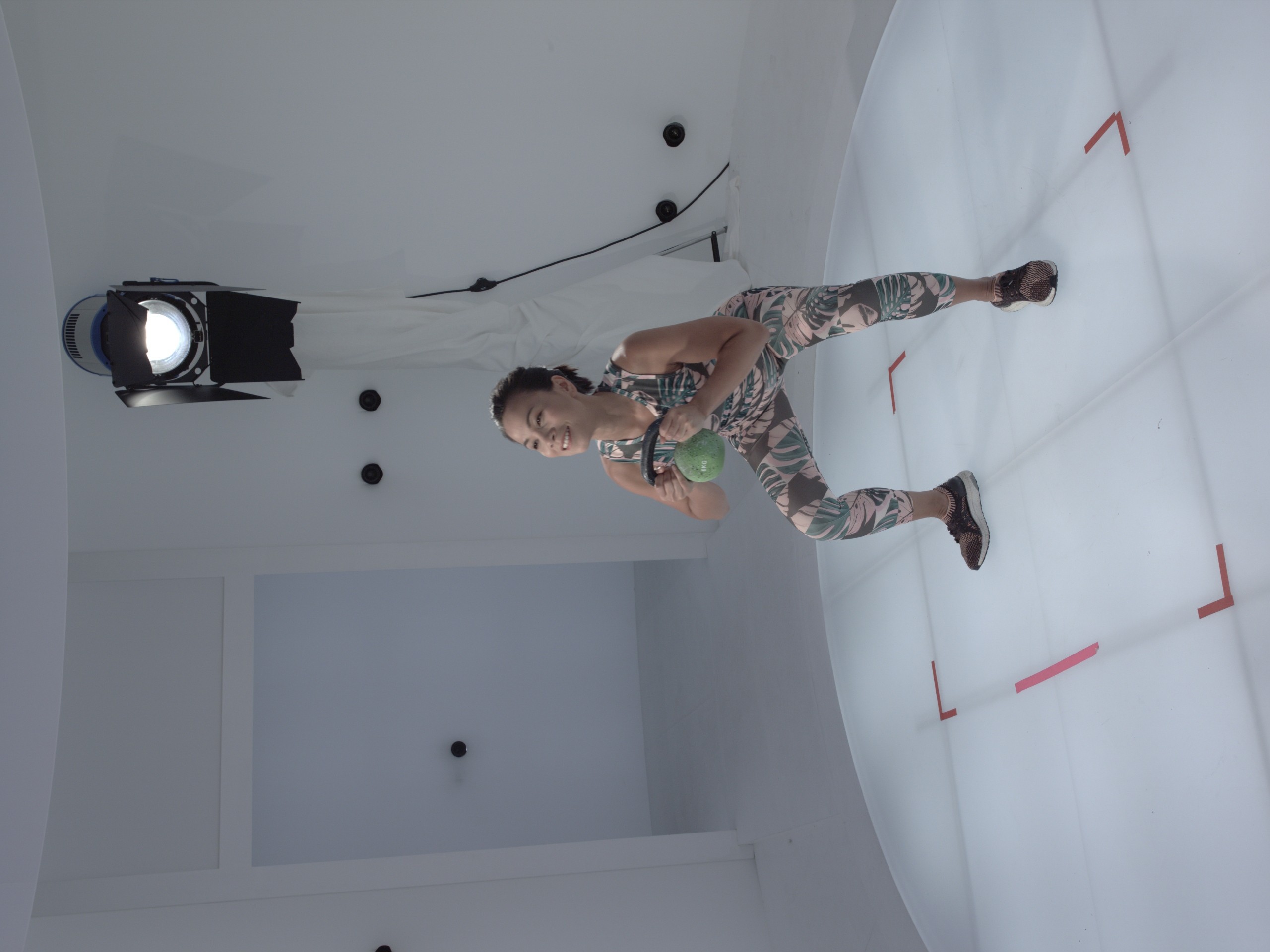}}
    &
    \raisebox{-0.5\height}{\includegraphics[trim={700 550 340 260},clip,angle=-90,origin=c,width=0.19\textwidth]{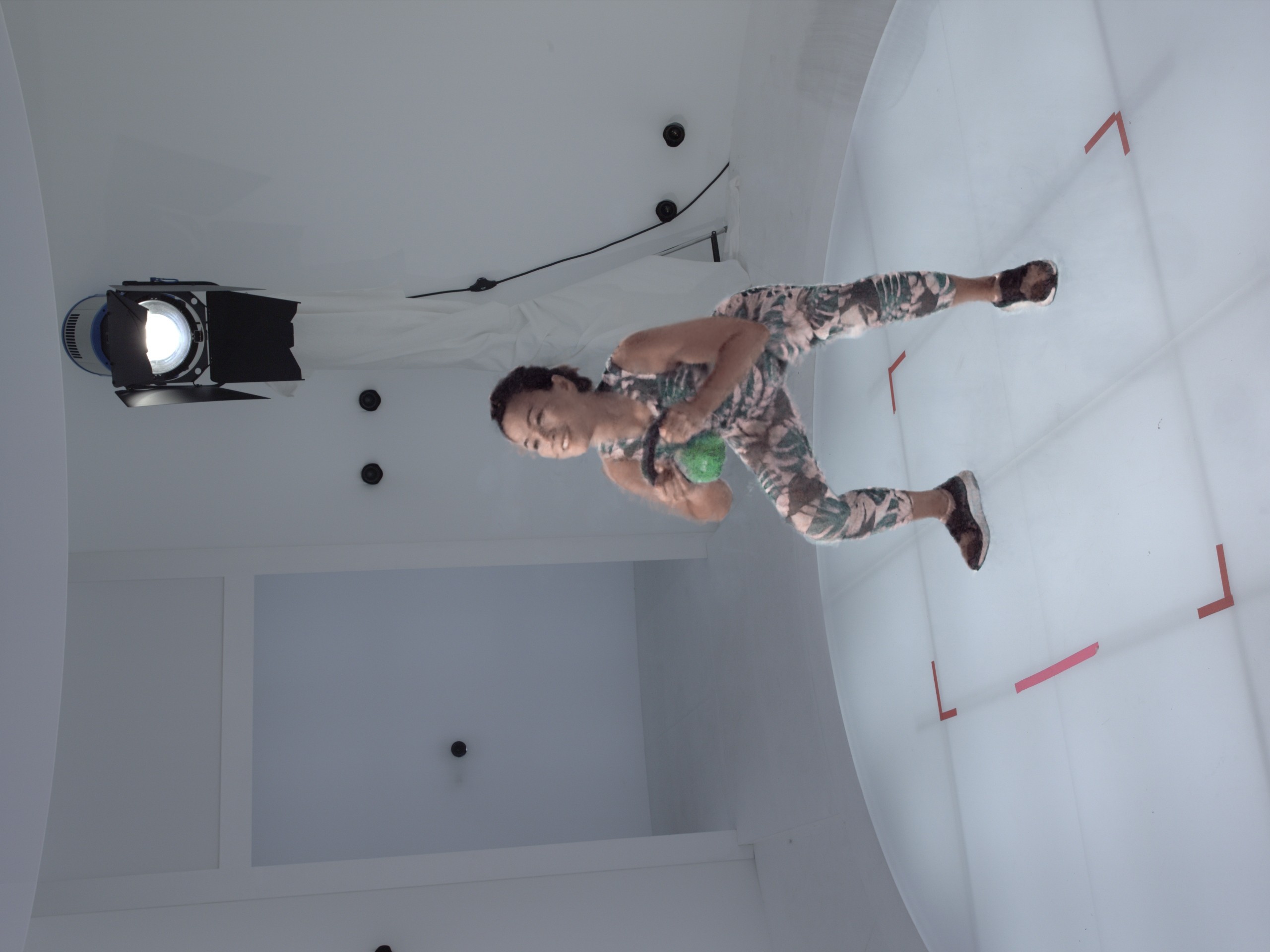}}
    &
    \raisebox{-0.5\height}{\includegraphics[trim={700 550 340 260},clip,angle=-90,origin=c,width=0.19\textwidth]{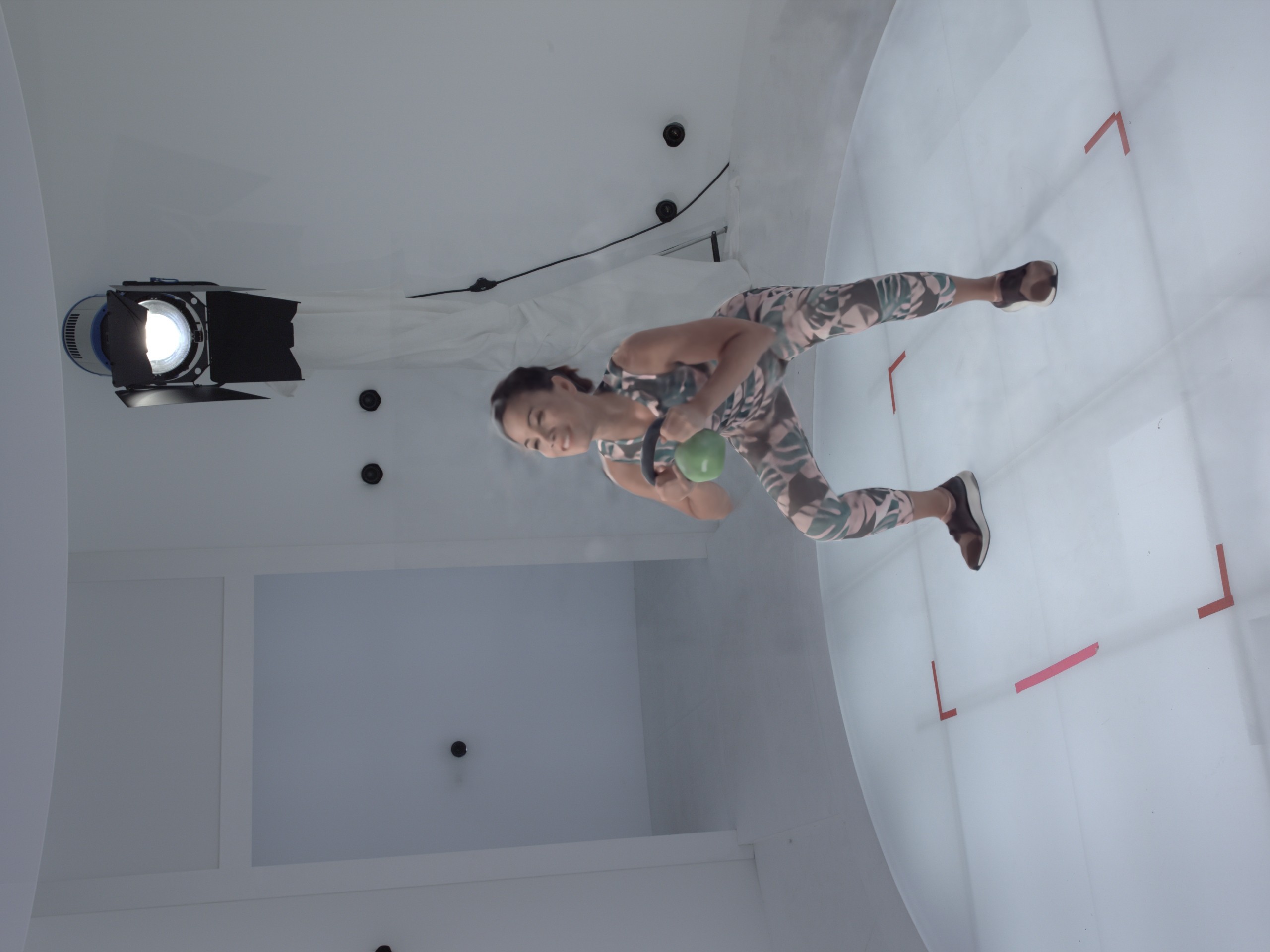}}
    &
    \raisebox{-0.5\height}{\includegraphics[trim={700 550 340 260},clip,angle=-90,origin=c,width=0.19\textwidth]{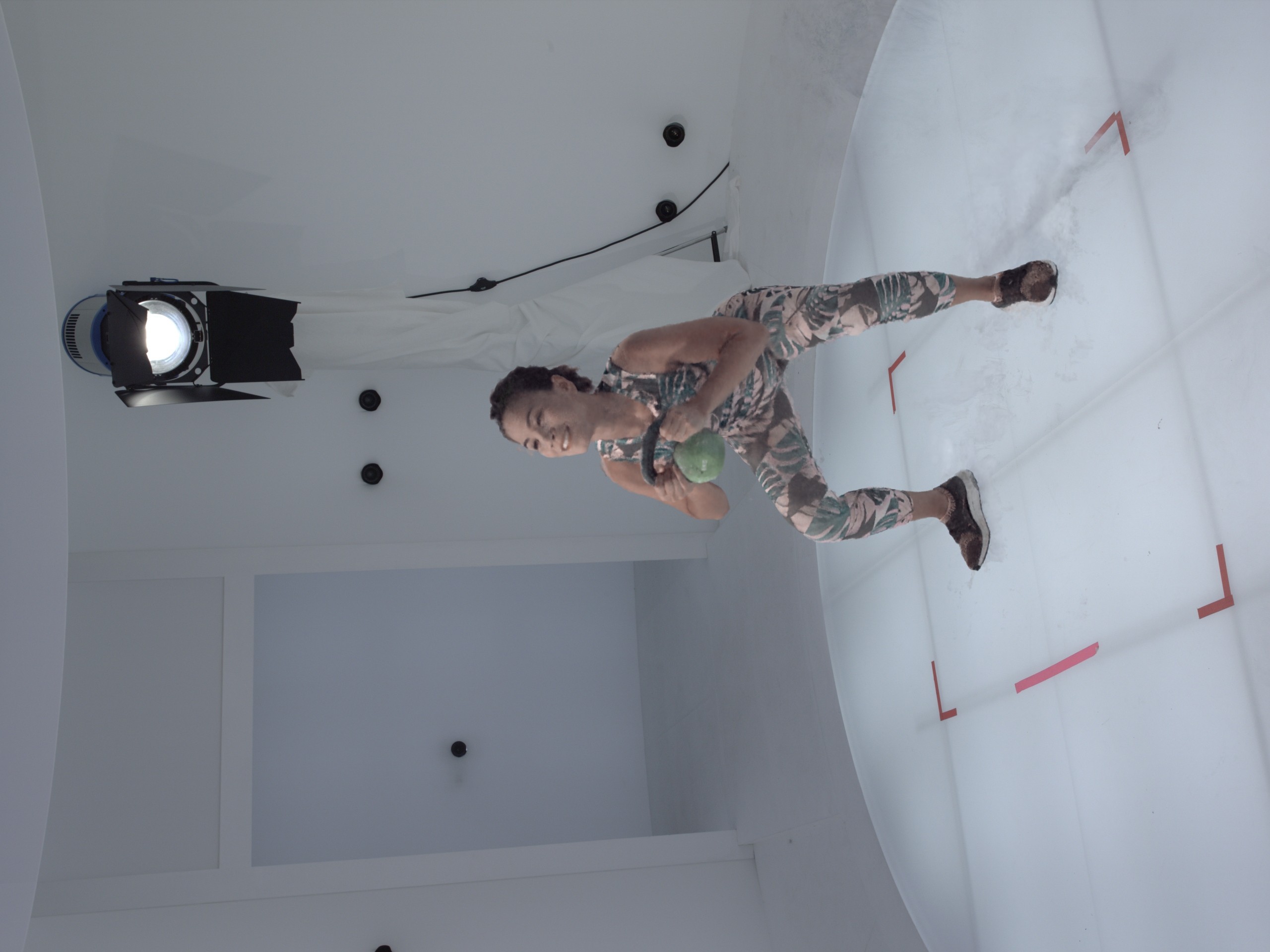}}
    &
    \raisebox{-0.5\height}{\includegraphics[trim={700 550 340 260},clip,angle=-90,origin=c,width=0.19\textwidth]{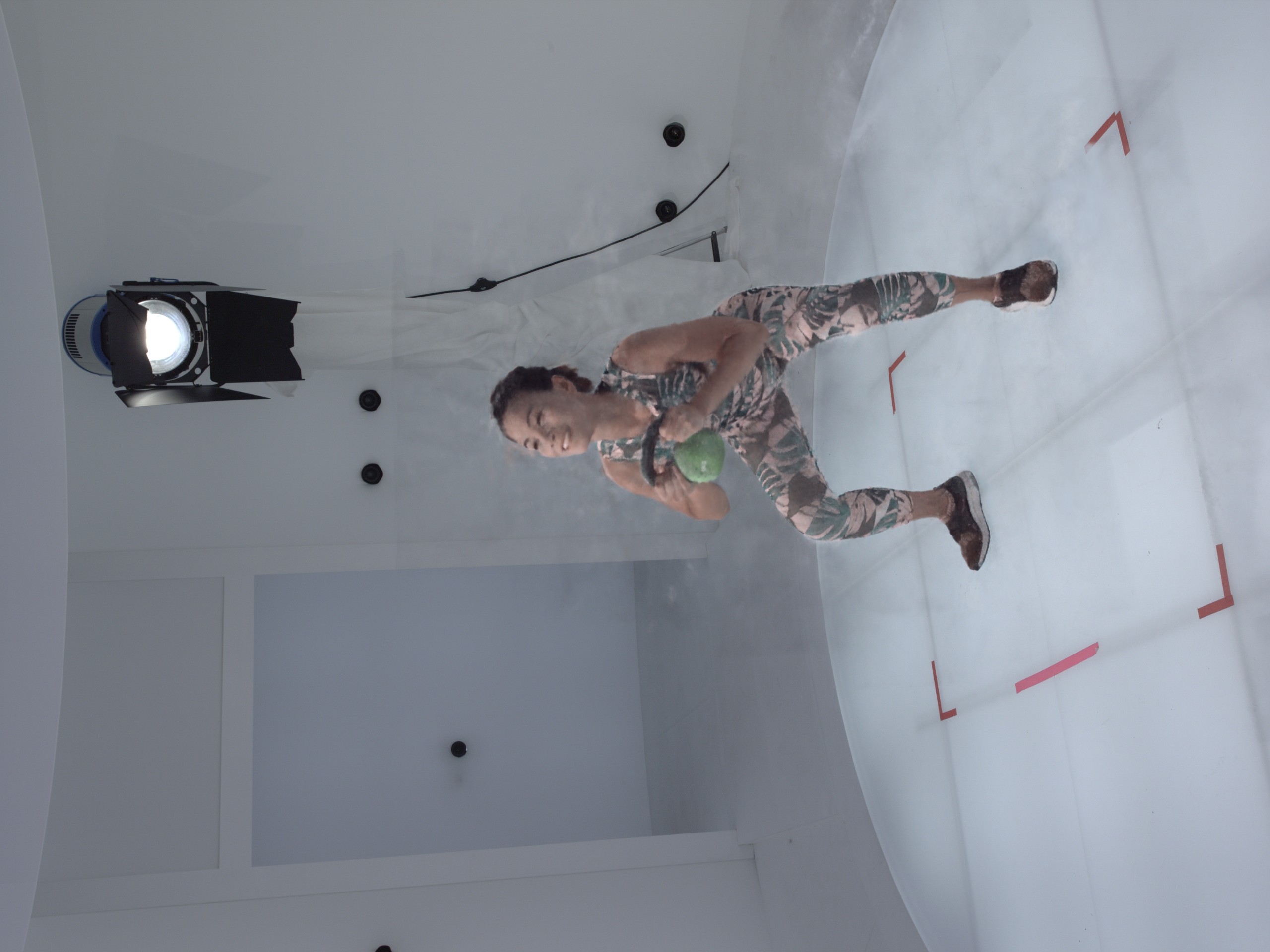}}
    
    \end{tabular}
    
    \caption{\textbf{Novel-View Synthesis.} (First row) Seq. 7 at $t{=}\frac{T}{2}$. (Second row) Seq. 8 at $t{=}T$. }
    \label{fig:more_novel_view_2}
\end{figure*}

Fig.~\ref{fig:more_joints} contains the qualitative joints estimated at $t{=}T$ for all sequences not shown in the main paper. 
Fig.~\ref{fig:more_novel_view} and Fig.~\ref{fig:more_novel_view_2} contain more novel-view results of all methods for four scenes at $t{=}\frac{T}{2}$.

\begin{table}
    \centering
    \resizebox{\columnwidth}{!}{
        \begin{tabular}{l|l|l}
        \hline
        Name & Description & Frames\\
        \hline
        Seq. $1$ & two people playing with a plush dog & $125$\\
        Seq. $2$ & two people holdings hands & $125$\\
        Seq. $3$ & one person rotating, one person walking & $300$\\
        Seq. $4$ & a person dancing while wearing a dark dress & $100$\\
        Seq. $5$ & a person walking like a zombie while wearing a light dress & $125$\\
        Seq. $6$ & a person playing with a plush dog & $125$\\
        Seq. $7$ & a person standing with a brown dress & $125$ \\
        Seq. $8$ & a person doing squads (from NR-NeRF~\cite{tretschk2021non}) & $5$ \\
        \end{tabular}
    }
    \caption{\textbf{Dataset Description.} 
    }
    \label{tab:scene_descriptions}
\end{table}

\section{Quantitative Ablation Results}\label{sec:quant_ablation}

Tab.~\ref{tab:quant_ablations} confirms the qualitative ablation results quantitatively.

\section{Further Architecture and Training Details}\label{sec:further_architecture}

For all methods, we normalize the scene into the unit cube by tightly fitting an axis-aligned bounding box to all near and far plane samples of all images. 

\subsection{Ours} 
\noindent\textbf{Architecture.} 
Both the coarse and fine deformation fields use the same architecture. 
The hash grid consists of $16$ levels, with two feature dimensions per level. 
The coarsest level has a resolution of $32^3$, and each subsequent level has a $1.3819$ times higher resolution. 
The hashmap has a size of $2^{20}$. 
The shallow MLP has one hidden layer with $64$ hidden dimensions. 

The canonical model uses a hash grid with $13$ levels, two feature dimensions per level, a coarsest resolution of $128^3$, and a scaling factor of $1.3819$. 
The shallow MLP that outputs the opacity has one hidden layer with $64$ hidden dimensions. 
A second MLP outputs the appearance and takes as input a $15$-dimensional vector additionally regressed by the first shallow MLP. 
The second MLP has two hidden layers with $64$ hidden dimensions.

\begin{table}
    \centering
    \resizebox{\columnwidth}{!}{
        \begin{tabular}{c|clc|ccccc}
        \hline
            & & & & Ours & NR-NeRF & SNF-A & SNF-AG & Background \\
        \hline
        \hline
        
        \parbox[t]{2mm}{\multirow{6}{*}{\rotatebox[origin=c]{90}{Seq. $1$}}} 
        & \parbox[t]{2mm}{\multirow{3}{*}{\rotatebox[origin=c]{90}{\footnotesize Unmasked}}} 
          & PSNR  &$\uparrow$   & $27.49$ & $26.84$ & $\mathbf{27.84}$ & $26.73$ & $18.64$ \\
        & & SSIM  &$\uparrow$   & $\mathbf{0.928}$ & $0.915$ & $\mathbf{0.928}$ & $0.906$ & $0.894$ \\
        & & LPIPS &$\downarrow$ & $\mathbf{0.074}$ & $0.117$ & $0.076$ & $0.138$ & $0.139$ \\
        \cline{2-9}
        & \parbox[t]{2mm}{\multirow{3}{*}{\rotatebox[origin=c]{90}{Masked}}} 
          & PSNR  &$\uparrow$   & $29.73$ & $29.38$ & $\mathbf{30.02}$ & $29.89$ & --- \\
        & & SSIM  &$\uparrow$   & $0.970$ & $0.966$ & $\mathbf{0.971}$ & $0.970$ & --- \\
        & & LPIPS &$\downarrow$ & $0.021$ & $0.036$ & $\mathbf{0.017}$ & $0.018$ & --- \\
        \hline
        \hline
        
        \parbox[t]{2mm}{\multirow{6}{*}{\rotatebox[origin=c]{90}{Seq. $2$}}} 
        & \parbox[t]{2mm}{\multirow{3}{*}{\rotatebox[origin=c]{90}{\footnotesize Unmasked}}} 
          & PSNR  &$\uparrow$   & $27.93$ & $21.64$ & $\mathbf{28.24}$ & $26.99$ & $18.57$ \\
        & & SSIM  &$\uparrow$   & $\mathbf{0.929}$ & $0.875$ & $\mathbf{0.929}$ & $0.907$ & $0.898$\\
        & & LPIPS &$\downarrow$ & $\mathbf{0.069}$ & $0.183$ & $0.074$ & $0.136$ & $0.130$ \\
        \cline{2-9}
        & \parbox[t]{2mm}{\multirow{3}{*}{\rotatebox[origin=c]{90}{Masked}}} 
          & PSNR  &$\uparrow$   & $29.65$ & $22.75$ & $\mathbf{30.23}$ & $29.96$ & --- \\
        & & SSIM  &$\uparrow$   & $0.970$ & $0.931$ & $\mathbf{0.972}$ & $0.970$ & --- \\
        & & LPIPS &$\downarrow$ & $0.019$ & $0.081$ & $\mathbf{0.017}$ & $0.019$ & --- \\
        \hline
        \hline
        
        \parbox[t]{2mm}{\multirow{6}{*}{\rotatebox[origin=c]{90}{Seq. $3$}}} 
        & \parbox[t]{2mm}{\multirow{3}{*}{\rotatebox[origin=c]{90}{\footnotesize Unmasked}}} 
          & PSNR  &$\uparrow$   & $\mathbf{27.72}$ & $21.48$ & $27.50$ & $26.61$ & $18.94$\\
        & & SSIM  &$\uparrow$   & $\mathbf{0.923}$ & $0.874$ & $0.920$ & $0.904$ & $0.895$\\
        & & LPIPS &$\downarrow$ & $\mathbf{0.075}$ & $0.181$ & $0.089$ & $0.136$ & $0.140$\\
        \cline{2-9}
        & \parbox[t]{2mm}{\multirow{3}{*}{\rotatebox[origin=c]{90}{Masked}}} 
          & PSNR  &$\uparrow$   & $\mathbf{28.95}$ & $22.25$ & $28.90$ & $28.47$ & --- \\
        & & SSIM  &$\uparrow$   & $\mathbf{0.961}$ & $0.925$ & $\mathbf{0.961}$ & $0.958$ & --- \\
        & & LPIPS &$\downarrow$ & $\mathbf{0.029}$ & $0.089$ & $\mathbf{0.029}$ & $0.033$ & --- \\
        \hline
        \hline
        
        \parbox[t]{2mm}{\multirow{6}{*}{\rotatebox[origin=c]{90}{Seq. $4$}}} 
        & \parbox[t]{2mm}{\multirow{3}{*}{\rotatebox[origin=c]{90}{\footnotesize Unmasked}}} 
          & PSNR  &$\uparrow$   & $31.59$ & $31.79$ & $31.92$ & $\mathbf{32.08}$ & $24.68$\\
        & & SSIM  &$\uparrow$   & $0.950$ & $0.948$ & $\mathbf{0.951}$ & $0.949$ & $0.943$\\
        & & LPIPS &$\downarrow$ & $\mathbf{0.034}$ & $0.055$ & $0.036$ & $0.044$ & $0.070$\\
        \cline{2-9}
        & \parbox[t]{2mm}{\multirow{3}{*}{\rotatebox[origin=c]{90}{Masked}}} 
          & PSNR  &$\uparrow$   & $33.68$ & $33.67$ & $34.15$ & $\mathbf{34.41}$ & --- \\
        & & SSIM  &$\uparrow$   & $0.980$ & $0.978$ & $\mathbf{0.981}$ & $\mathbf{0.981}$ & --- \\
        & & LPIPS &$\downarrow$ & $\mathbf{0.011}$ & $0.031$ & $0.012$ & $0.014$ & --- \\
        \hline
        \hline
        
        \parbox[t]{2mm}{\multirow{6}{*}{\rotatebox[origin=c]{90}{Seq. $5$}}} 
        & \parbox[t]{2mm}{\multirow{3}{*}{\rotatebox[origin=c]{90}{\footnotesize Unmasked}}} 
          & PSNR  &$\uparrow$   & $32.25$ & $31.73$ & $32.58$ & $\mathbf{32.77}$ & $23.33$\\
        & & SSIM  &$\uparrow$   & $0.946$ & $0.944$ & $\mathbf{0.947}$ & $0.945$ & $0.938$\\
        & & LPIPS &$\downarrow$ & $\mathbf{0.042}$ & $0.057$ & $0.044$ & $0.049$ & $0.076$\\
        \cline{2-9}
        & \parbox[t]{2mm}{\multirow{3}{*}{\rotatebox[origin=c]{90}{Masked}}} 
          & PSNR  &$\uparrow$   & $34.08$ & $33.29$ & $34.49$ & $\mathbf{34.95}$ & --- \\
        & & SSIM  &$\uparrow$   & $0.976$ & $0.973$ & $\mathbf{0.978}$ & $\mathbf{0.978}$ & --- \\
        & & LPIPS &$\downarrow$ &$\mathbf{0.017}$ & $0.033$ & $\mathbf{0.017}$ & $\mathbf{0.017}$ & --- \\
        \hline
        \hline
        
        \parbox[t]{2mm}{\multirow{6}{*}{\rotatebox[origin=c]{90}{Seq. $6$}}} 
        & \parbox[t]{2mm}{\multirow{3}{*}{\rotatebox[origin=c]{90}{\footnotesize Unmasked}}} 
          & PSNR  &$\uparrow$   & $28.52$ & $26.69$ & $\mathbf{29.05}$ & $27.70$ & $19.76$\\
        & & SSIM  &$\uparrow$   & $\mathbf{0.938}$ & $0.917$ & $0.937$ & $0.918$ & $0.916$\\
        & & LPIPS &$\downarrow$ & $\mathbf{0.060}$ & $0.123$ & $0.069$ & $0.120$ & $0.105$\\
        \cline{2-9}
        & \parbox[t]{2mm}{\multirow{3}{*}{\rotatebox[origin=c]{90}{Masked}}} 
          & PSNR  &$\uparrow$   & $30.24$ & $28.85$ & $\mathbf{31.19}$ & $30.99$ & --- \\
        & & SSIM  &$\uparrow$   & $0.978$ & $0.969$ & $\mathbf{0.980}$ & $0.979$ & --- \\
        & & LPIPS &$\downarrow$ & $0.015$ & $0.035$ & $\mathbf{0.013}$ & $\mathbf{0.013}$ & --- \\
        \hline
        \hline
        
        \parbox[t]{2mm}{\multirow{6}{*}{\rotatebox[origin=c]{90}{Seq. $7$}}} 
        & \parbox[t]{2mm}{\multirow{3}{*}{\rotatebox[origin=c]{90}{\footnotesize Unmasked}}} 
          & PSNR  &$\uparrow$   & $34.38$ & $34.79$ & $35.32$ & $\mathbf{35.43}$ & $26.08$\\
        & & SSIM  &$\uparrow$   & $0.959$ & $0.958$ & $\mathbf{0.960}$ & $0.956$ & $0.950$\\
        & & LPIPS &$\downarrow$ & $\mathbf{0.026}$ & $0.032$ & $\mathbf{0.026}$ & $0.036$ & $0.057$\\        
        \cline{2-9}
        & \parbox[t]{2mm}{\multirow{3}{*}{\rotatebox[origin=c]{90}{Masked}}} 
          & PSNR  &$\uparrow$   & $37.88$ & $38.01$ & $39.33$ & $\mathbf{39.83}$ & --- \\
        & & SSIM  &$\uparrow$   & $0.991$ & $0.990$ & $\mathbf{0.992}$ & $\mathbf{0.992}$ & --- \\
        & & LPIPS &$\downarrow$ & $0.005$ & $0.007$ & $\mathbf{0.003}$ & $0.004$ & --- \\
        \hline
        \hline
        
        \parbox[t]{2mm}{\multirow{6}{*}{\rotatebox[origin=c]{90}{Seq. $8$}}} 
        & \parbox[t]{2mm}{\multirow{3}{*}{\rotatebox[origin=c]{90}{\footnotesize Unmasked}}} 
          & PSNR  &$\uparrow$   & $32.67$ & $\mathbf{35.18}$ & $34.58$ & $34.38$ & $23.53$\\
        & & SSIM  &$\uparrow$   & $0.937$ & $0.942$ & $\mathbf{0.944}$ & $0.941$ & $0.926$\\
        & & LPIPS &$\downarrow$ & $0.046$ & $\mathbf{0.041}$ & $0.042$ & $0.047$ & $0.088$\\
        \cline{2-9}
        & \parbox[t]{2mm}{\multirow{3}{*}{\rotatebox[origin=c]{90}{Masked}}} 
          & PSNR  &$\uparrow$   & $34.82$ & $\mathbf{38.17}$ & $38.13$ & $37.26$ & --- \\
        & & SSIM  &$\uparrow$   & $0.980$ & $0.984$ & $\mathbf{0.986}$ & $0.984$ & --- \\
        & & LPIPS &$\downarrow$ & $0.022$ & $0.017$ & $\mathbf{0.013}$ & $0.018$ & --- \\
        \hline
        
        \end{tabular}
    }
    \caption{\textbf{Novel-View Synthesis.} We report per-scene PSNR, SSIM, and LPIPS.
    }
    \label{tab:novel_view_per_scene}
\end{table}

\begin{table}
    \centering
        \begin{tabular}{clc|ccccc}
        \hline
           &   &              & SceNeRFlow & D-NeRF & DeVRF\\
        \hline
        \hline
        \parbox[t]{2mm}{\multirow{3}{*}{\rotatebox[origin=c]{90}{\footnotesize Unmasked}}} & 
          PSNR  & $\uparrow$   & $\mathbf{29.98}$ & $24.89$ & $25.24$ \\
        & SSIM  & $\uparrow$   & $\mathbf{0.939}$ & $0.913$ & $0.905$ \\
        & LPIPS & $\downarrow$ & $\mathbf{0.054}$ & $0.107$ & $0.125$ \\
        \hline
        \parbox[t]{2mm}{\multirow{3}{*}{\rotatebox[origin=c]{90}{Masked}}} 
        & PSNR  & $\uparrow$   & $\mathbf{32.03}$ & $26.13$ & $27.66$ \\
        & SSIM  & $\uparrow$   & $\mathbf{0.975}$ & $0.951$ & $0.957$ \\
        & LPIPS & $\downarrow$ & $\mathbf{0.017}$ & $0.060$ & $0.041$ \\
        \end{tabular}
    \caption[More quantitative novel-view synthesis results]{\textbf{Novel-View Synthesis.} Mean PSNR, SSIM, and LPIPS across scenes, except for Seq. 8 (due to memory limitations from the high resolution). 
    }
    \label{tab:more_novel_view}
\end{table}

\begin{table}
    \centering
    \resizebox{\columnwidth}{!}{
        \begin{tabular}{clc|ccccc}
        \hline
        &       &              & SceNeRFlow & No Online & No Extend & No Coarse & No Fine \\
        \hline
        \hline
        \parbox[t]{2mm}{\multirow{3}{*}{\rotatebox[origin=c]{90}{\footnotesize Unmasked}}} 
        & PSNR  & $\uparrow$   & $30.32$ & $30.36$ & $28.53$ & $\mathbf{30.40}$ & $28.95$ \\
        & SSIM  & $\uparrow$   & $\mathbf{0.939}$ & $0.929$ & $0.927$ & $0.936$ & $0.933$ \\
        & LPIPS & $\downarrow$ & $\mathbf{0.054}$ & $0.074$ & $0.067$ & $0.056$ & $0.056$ \\
        \hline
        \parbox[t]{2mm}{\multirow{3}{*}{\rotatebox[origin=c]{90}{Masked}}} 
        & PSNR  & $\uparrow$   & $32.38$ & $\mathbf{32.61}$ & $30.20$ & $32.45$ & $30.45$ \\
        & SSIM  & $\uparrow$   & $\mathbf{0.976}$ & $0.974$ & $0.966$ & $0.974$ & $0.970$ \\
        & LPIPS & $\downarrow$ & $\mathbf{0.018}$ & $0.020$ & $0.027$ & $0.020$ & $0.021$ \\
        \hline
        \hline        
        & MPJPE & $\downarrow$ & $\mathbf{1.5}$ & $33.4$ & $15.6$ & $10.4$ & $\mathbf{1.5}$ \\
        \end{tabular}
    }
    \caption[Quantitative ablation results]{\textbf{Ablations.} Mean PSNR, SSIM, LPIPS, and MPJPE across scenes. 
    }
    \label{tab:quant_ablations}
\end{table}

\noindent\textbf{Weighting Scheme for Smoothness Loss.}
We weigh sample $i$ on ray $r$ depending on its closeness to the object. 
Mathematically, we start with $\hat{\sigma}_{r,i}=\exp(-\sigma_{r,i}\delta)$, where $\sigma_{r,i}$ is the opacity of the $i$-th sample on ray $r$. 
We next apply max-pooling with window size $k=\lfloor f \cdot S \rfloor$, where we empirically set $f{=}0.005$: 
\begin{equation}
    \hat{\sigma}'_{r,i} = \max_{i' \in [i-k,\ldots, i+k]} \hat{\sigma}_{r,i'}.
\end{equation}
We then weaken the regularization on empty space by $u{=}10$: 
\begin{equation}
    \hat{\sigma}''_{r,i} = 
        \begin{cases}
            \frac{1}{u}\hat{\sigma}'_{r,i} ~~~~~~~\text{if } \hat{\sigma}'_{r,i} > u \cdot \hat{\sigma}_{r,i}\\
            \hat{\sigma}'_{r,i} ~~~~~~~~~~\text{else.}
        \end{cases}
\end{equation}
Finally, we weaken the regularization on very small offsets $\Delta \in\mathbb{R}^3$ with a soft threshold of $s_t{=}0.001$: 
\begin{equation}
    \hat{\sigma}'''_{r,i}(\Delta) = \operatorname{sig}\Big( \frac{4\lVert \Delta \rVert_2}{s_t} -2 \Big) \hat{\sigma}''_{r,i},
\end{equation}
where $\operatorname{sig}(x){=}\frac{1}{1+\exp(-x)}$ is the sigmoid function. 
We thus have: 
\begin{equation}
    \mathcal{L}_{\text{norm,w}} = \frac{1}{RS} \sum_r \sum_i \hat{\sigma}'''_{r,i}(\Delta) \mathbb{E}_{\mathbf{e}}\Big[\big\lvert \lVert \mathbf{J}_{\mathbf{r}_r(s_i)}^\top \mathbf{e} \rVert_2 - 1 \big\rvert\Big],
\end{equation}
where $\Delta{=}\Delta_c(\mathbf{r}_r(s_i))$ when regularizing the coarse deformations, and $\Delta{=}\Delta_f(d_c(\mathbf{r}_r(s_i)))$ when regularizing the fine deformations.

\noindent\textbf{Learning Rates.}
We use exponential decay for the learning rates when constructing the canonical model and for each timestamp $t{>}1$. 
For $t{=}1$, we decay by a factor of $0.01$ for all parameters. 
For $t{>}1$, we decay the coarse deformations by $0.1$ when they get optimized, and the fine deformations by $0.1$ when they get optimized. 
All parameters of the canonical model have an initial learning rate of $10^{-2}$. 
All deformation parameters have an initial learning rate of $10^{-3}$.
The vignetting parameters have an initial learning rate of $10^{-2}$ and are the only parameters with no weight decay. 

During the first $1000$ iterations when building the canonical model, we warm up the learning rates with an additional factor that exponentially increases from $0.01$ to $1$. 

\subsection{Variants}

We compare against two variants of our method that use time-varying canonical models and thus neglect correspondences. 

The first variant, SNF-A, allows the appearance but not the geometry of the canonical model to change for $t{>}1$. 
We implement this via a separate appearance model that has the same HashMLP architecture as the standard canonical model. 
We can then keep the standard canonical model, which now only predicts the geometry, fixed for $t{>}1$ while allowing the appearance to vary. 

The second variant, SNF-AG, has time-varying appearance and geometry. 
We thus use the standard canonical model. 
We apply the geometry regularization losses $\mathcal{L}_{\text{back}}$ and $\mathcal{L}_{\text{hard}}$ to the canonical model at all timestamps $t{\geq}1$. 

For both variants, we seek to explain as much of the reconstruction as possible via the deformations, such that the canonical model needs to change as little as possible. 
We thus split the $10k$ iterations per timestamp into three equally long phases: only coarse deformations (as before), then only fine deformations (as before), then only the canonical model. 
\emph{I.e.}, the canonical model gets optimized only during the third phase, when the deformations are fixed. 

We found these variants unstable to train, with frequent divergence, and the following remedies helped: 
The canonical hash grid(s) use the settings from Instant NGP~\cite{mueller2022instant} (a coarsest resolution of $16$, with the resolution of each finer level being $1.5$ times finer than the previous level, with a total of $16$ levels). 
During the third phase, we use Instant NGP's Huber loss with a threshold of $0.1$ as reconstruction loss instead of our $\ell_1$ reconstruction loss. 
We use an exponentially decaying learning rate for the third phase that goes from $10^{-2}$ to $10^{-3}$ (except for the $300$-frame Seq.~3, where we use a ten times lower learning rate for long-term training stability). 

\subsection{NR-NeRF} 
We adapt NR-NeRF~\cite{tretschk2021non} to our settings as follows: 
Since we provide background images to NR-NeRF, we do not use its rigidity network. 
Furthermore, due to our large-motion setting, we remove its offsets loss that encourages the deformations to remain small. 
To keep runtime reasonable, we apply pruning. 
Doing so also enables us to always sample densely and we thus do not apply hierarchical sampling. 
We also use foreground-focused batches and vignetting correction. 
Since the recommended~\cite{tretschk2021non} training time of $200k$ iterations is only shown for very short scenes, we instead train NR-NeRF for longer on our scenes. 
Specifically, we extend NR-NeRF's training time to that of our method and thus train NR-NeRF for one third of the number of iterations of our method. 
We train Seq. 8 for $200k$ iterations, as that is longer. 

\subsection{PREF}
Like for NR-NeRF, we also supply background images, apply pruning, use foreground-focused batches, and learn the vignetting correction. 
Since we follow the authors of PREF and split our long scenes into chunks of $25$ frames (see Sec.~\ref{sec:joint_evaluation_details}), we keep the training time the same.

\section{Joint Evaluation Details}\label{sec:joint_evaluation_details}

We provide details on how we adapt PREF and NR-NeRF to long-term 3D joint tracking. 

\noindent\textbf{PREF.}
To train on a longer sequence, PREF~\cite{uii_eccv22_pref} splits the sequence into chunks of $25$ frames and trains on each chunk independently. 
We do the same with our scenes. 
To obtain long-term correspondences across chunks, we overlap the chunks for three frames. 

\noindent\textbf{NR-NeRF.} 
Unlike SceNeRFlow, NR-NeRF's canonical model does not coincide with the world space at $t{=}1$. 
We therefore cannot use $\{\hat{\mathbf{p}}^1_j\}_{j}$ directly as the target canonical positions. 
Instead, we apply the backward deformation model at $t{=}1$ to $\{\hat{\mathbf{p}}^1_j\}_{j}$ to obtain their positions in canonical space. 
We then have the joint positions in canonical space and the world-space positions at $t{=}1$, which allows us to apply the same tracking procedure as for SceNeRFlow.

\section{Foreground Masks for Evaluation}\label{sec:foreground_masks}

Since we did not tune the variants beyond making their training stable, they exhibit some significant undesired artifacts in empty space that we did not try to remove. 
In addition to scores on the full images, we therefore also compute scores that are focused on the actual dynamic object of interest. 
To this end, we use foreground masks. 

\begin{figure*}
    \centering
    \begin{tabular}{ccc|cc}
    & \multicolumn{2}{c|}{View $1$, $t{=}1$} & \multicolumn{2}{c}{View $2$, $t{=}T$} 
    \\
    & Ground Truth & Mask & Ground Truth  & Mask  
    \\
    
    \parbox[t]{2mm}{\rotatebox[origin=c]{90}{Seq. $1$}}
    & 
    \raisebox{-0.5\height}{\includegraphics[width=0.2\textwidth]{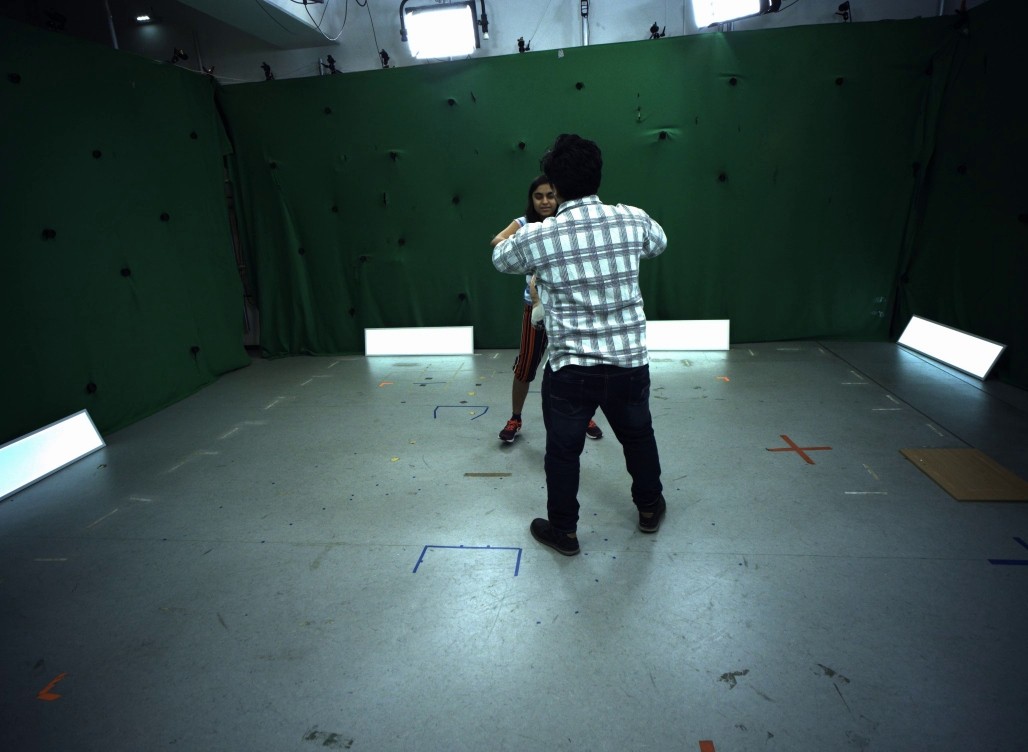}}
    &
    \raisebox{-0.5\height}{\includegraphics[width=0.2\textwidth]{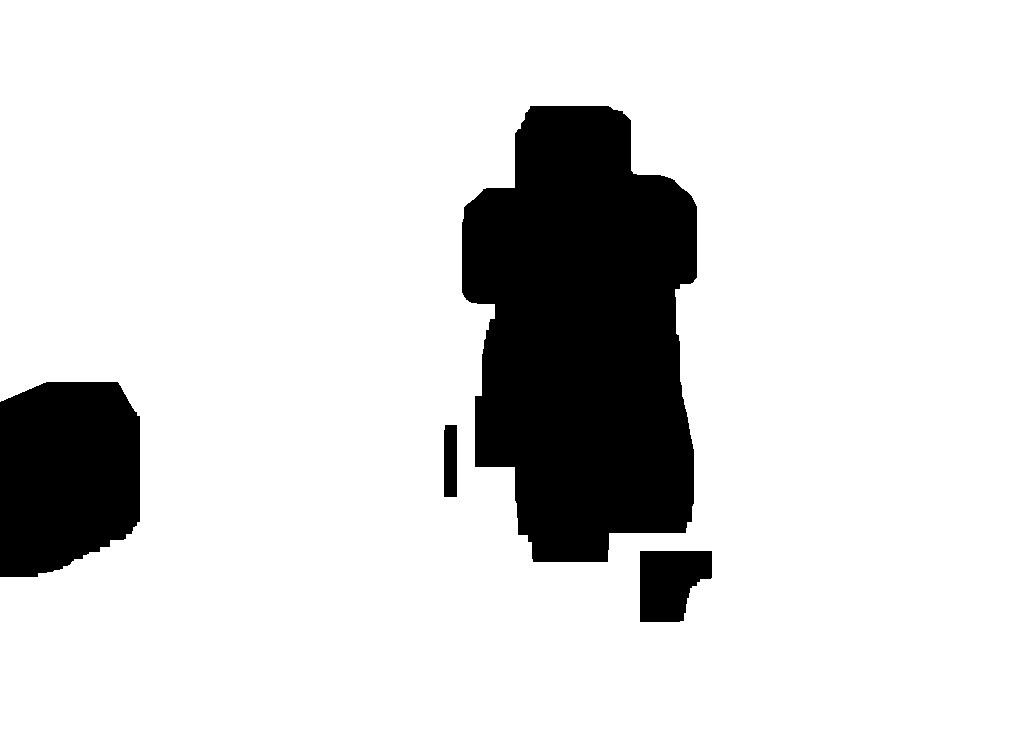}}
    &
    \raisebox{-0.5\height}{\includegraphics[width=0.2\textwidth]{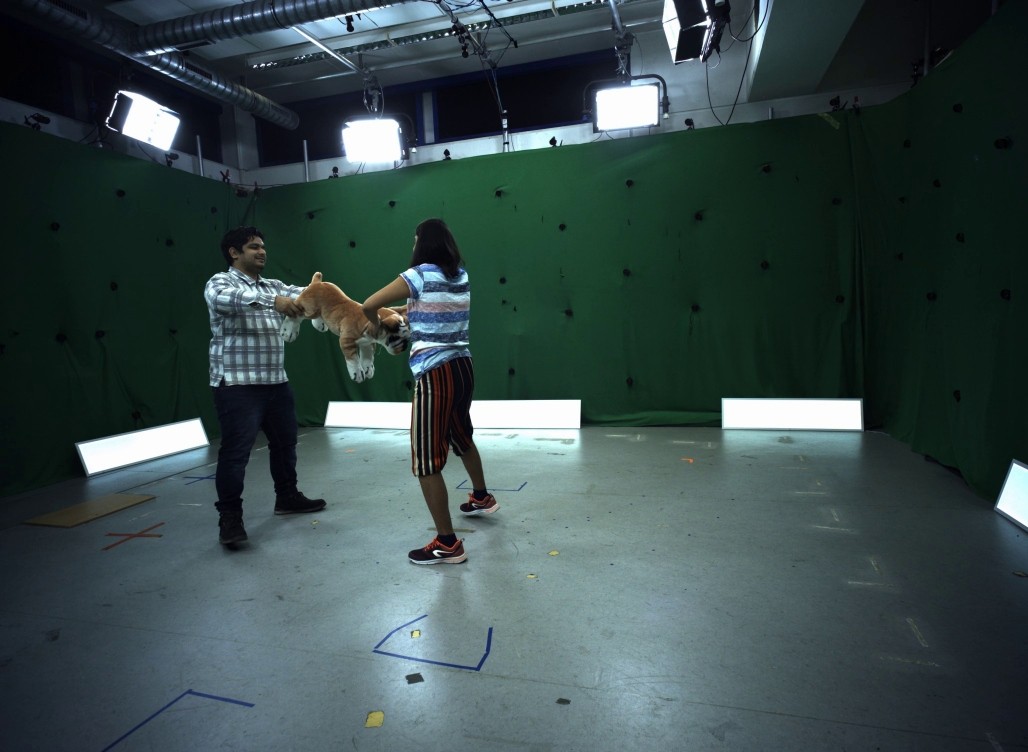}}
    &
    \raisebox{-0.5\height}{\includegraphics[width=0.2\textwidth]{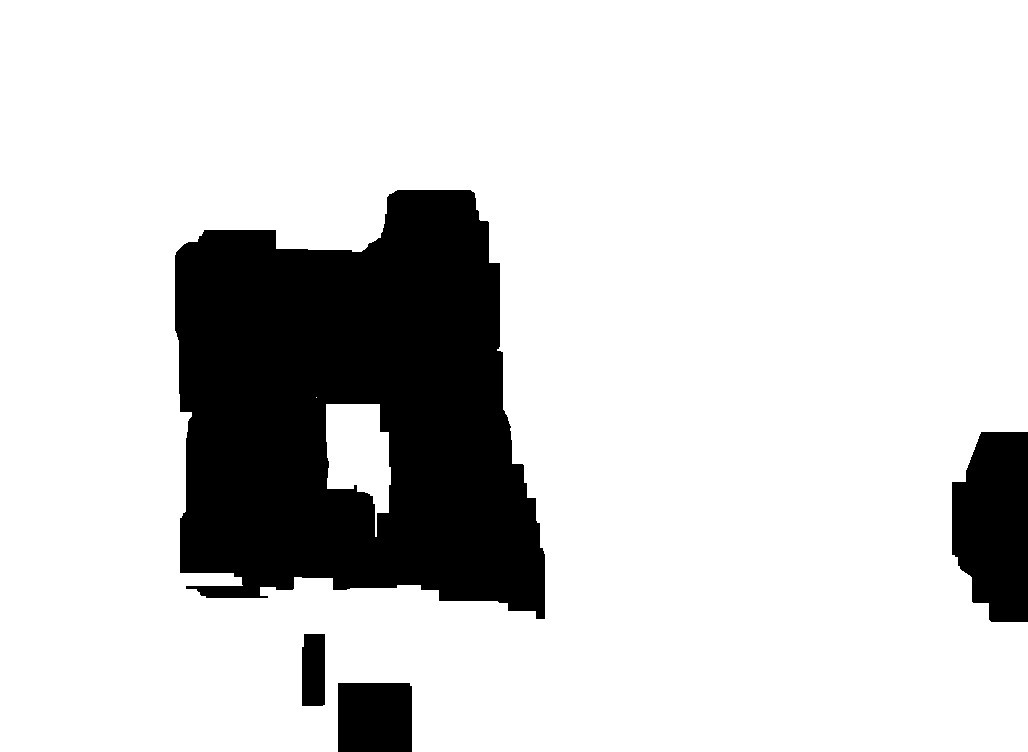}}
    \\
    
    \parbox[t]{2mm}{\rotatebox[origin=c]{90}{Seq. $2$}}
    &
    \raisebox{-0.5\height}{\includegraphics[width=0.2\textwidth]{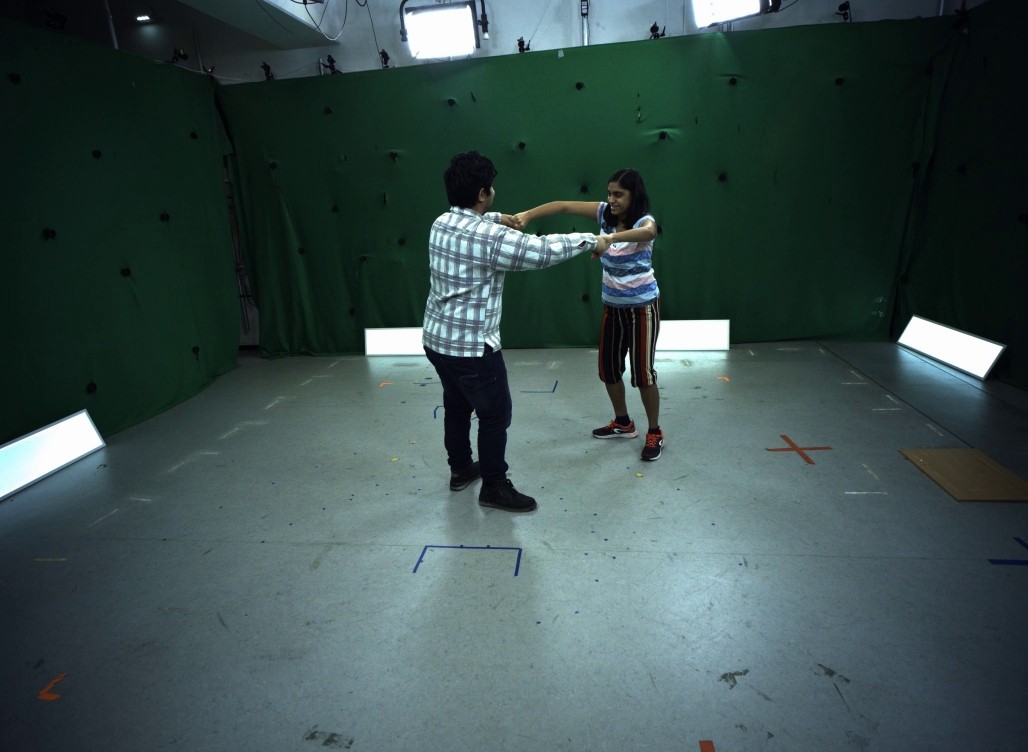}}
    &
    \raisebox{-0.5\height}{\includegraphics[width=0.2\textwidth]{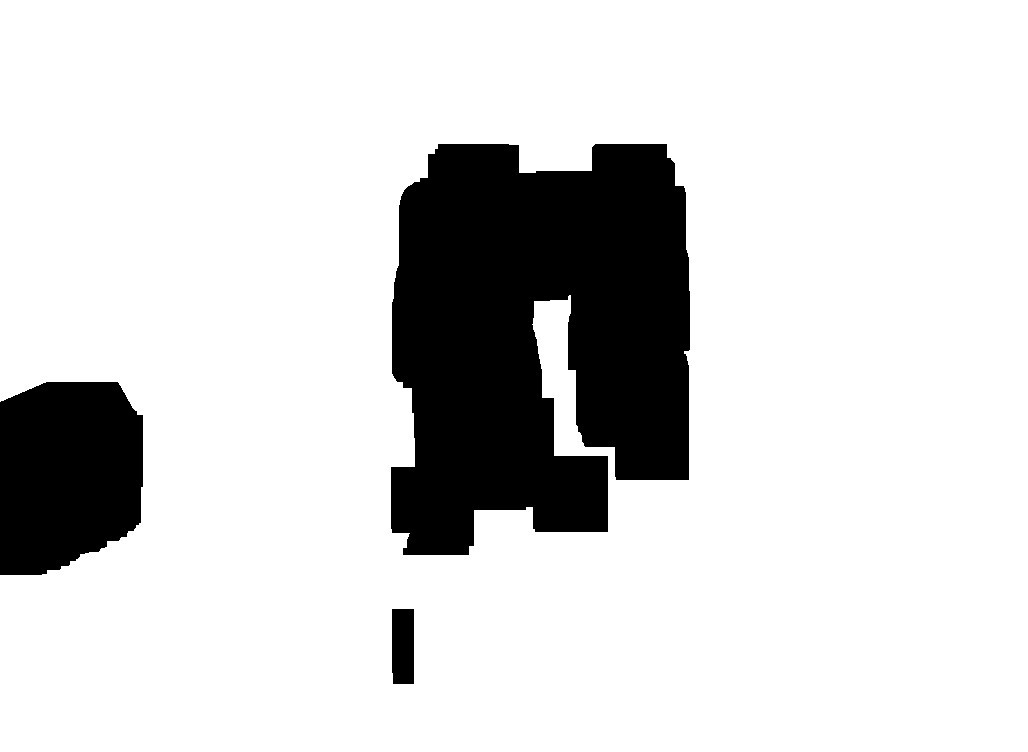}}
    &
    \raisebox{-0.5\height}{\includegraphics[width=0.2\textwidth]{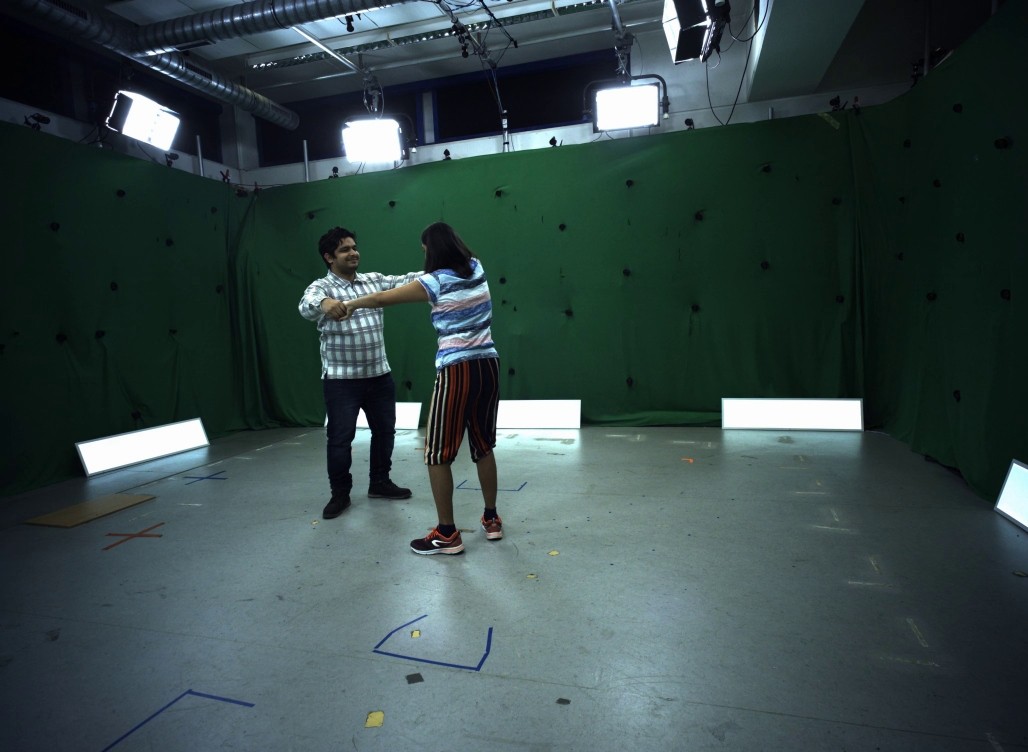}}
    &
    \raisebox{-0.5\height}{\includegraphics[width=0.2\textwidth]{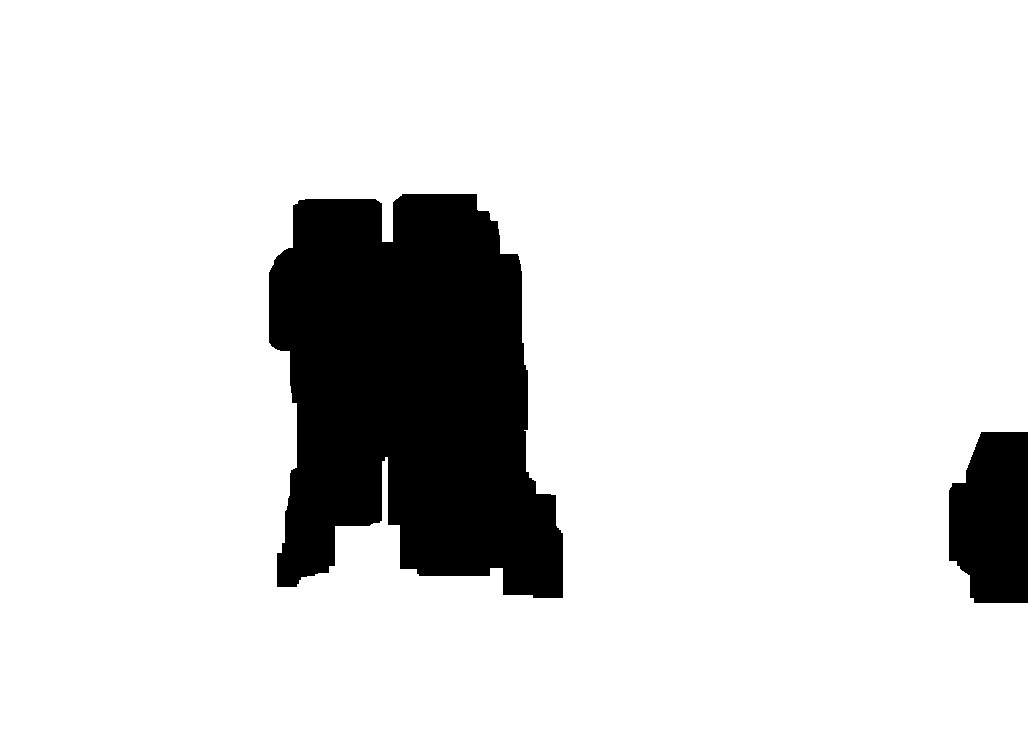}}
    \\
    
    \parbox[t]{2mm}{\rotatebox[origin=c]{90}{Seq. $3$}}
    &
    \raisebox{-0.5\height}{\includegraphics[width=0.2\textwidth]{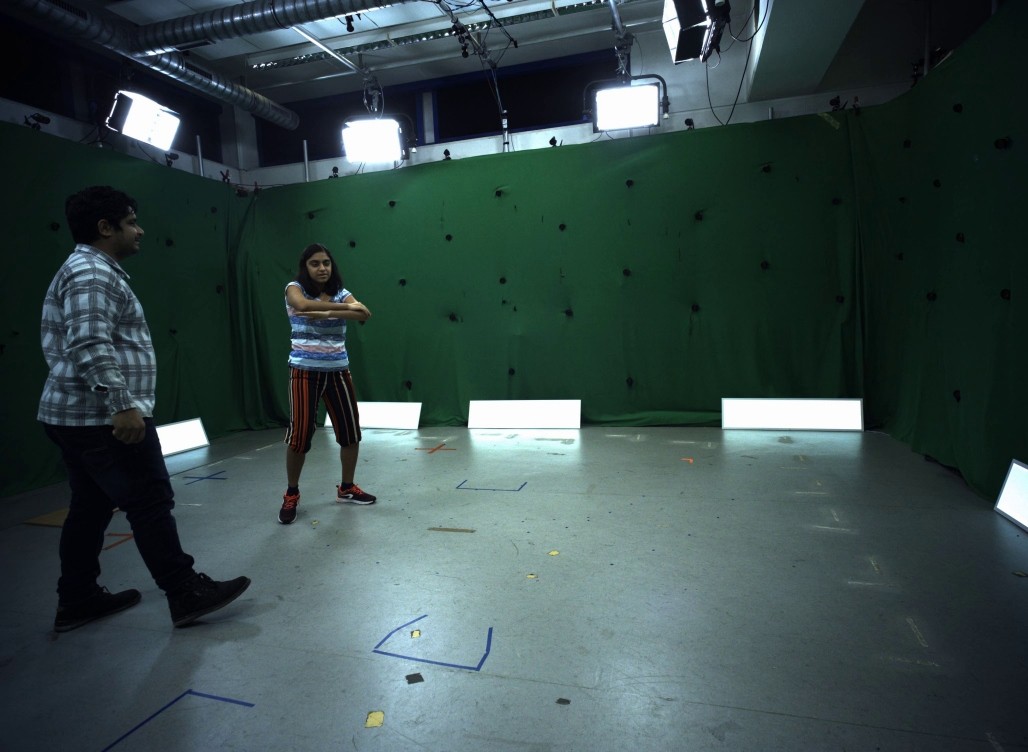}}
    &
    \raisebox{-0.5\height}{\includegraphics[width=0.2\textwidth]{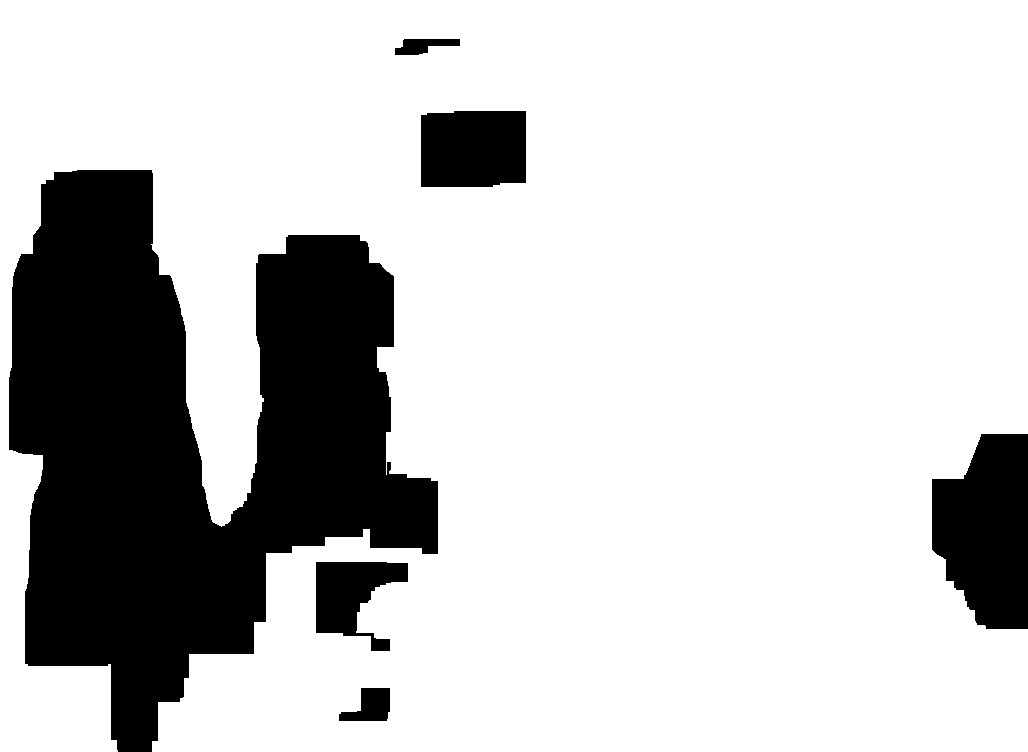}}
    &
    \raisebox{-0.5\height}{\includegraphics[width=0.2\textwidth]{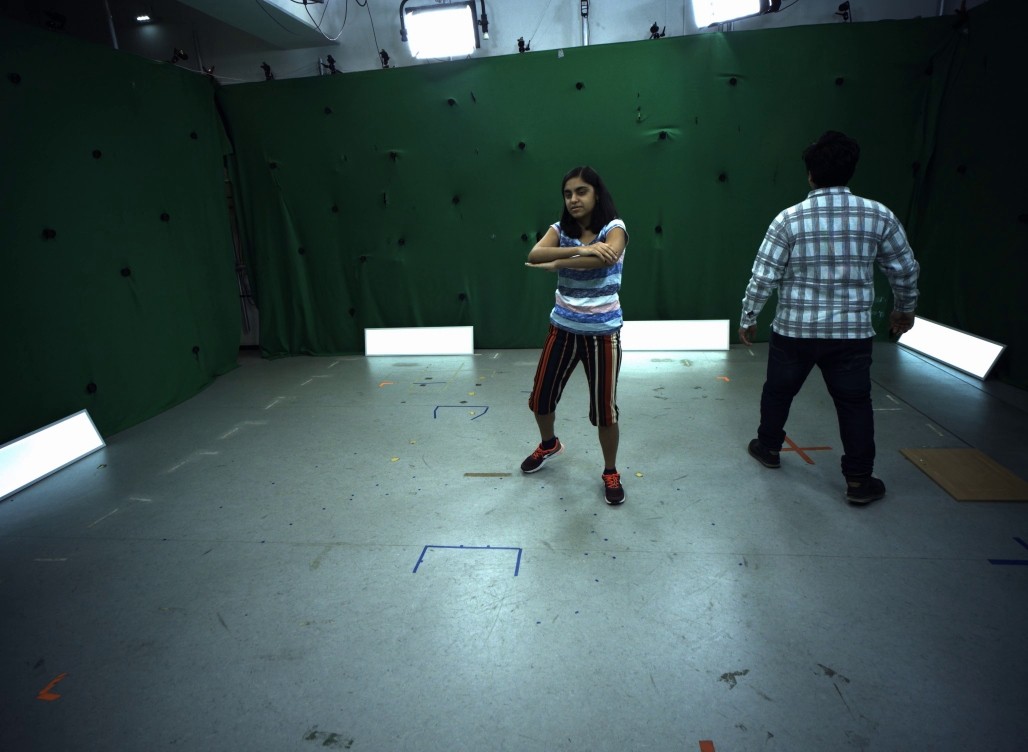}}
    &
    \raisebox{-0.5\height}{\includegraphics[width=0.2\textwidth]{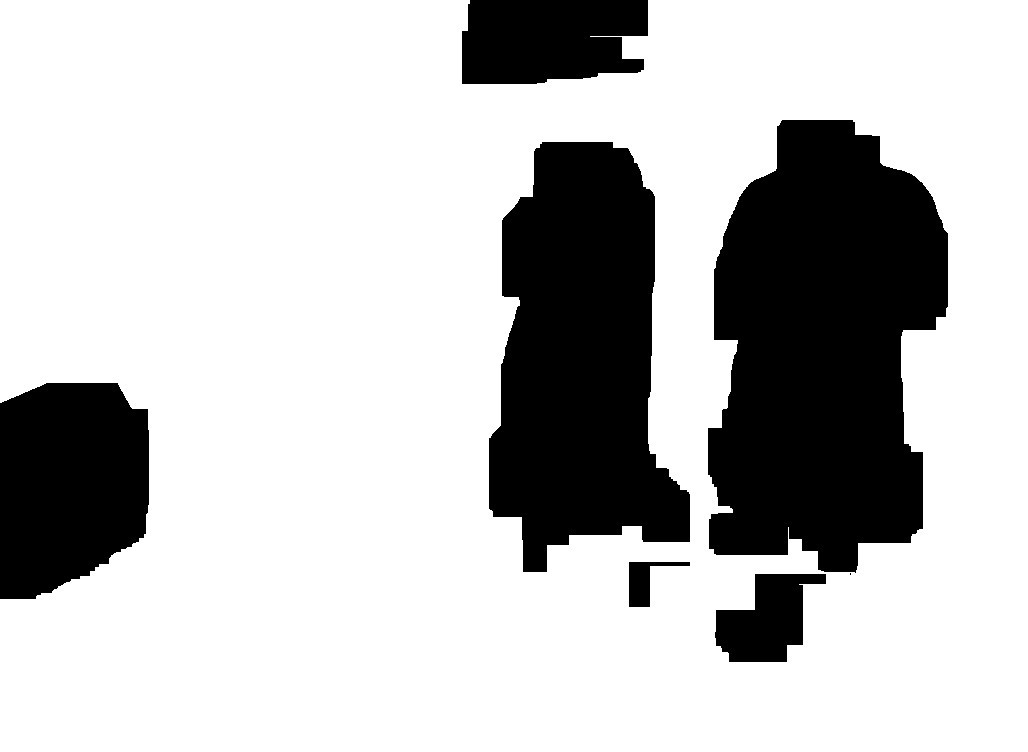}}
    \\

    \parbox[t]{2mm}{\rotatebox[origin=c]{90}{Seq. $4$}}
    &
    \raisebox{-0.5\height}{\includegraphics[width=0.2\textwidth]{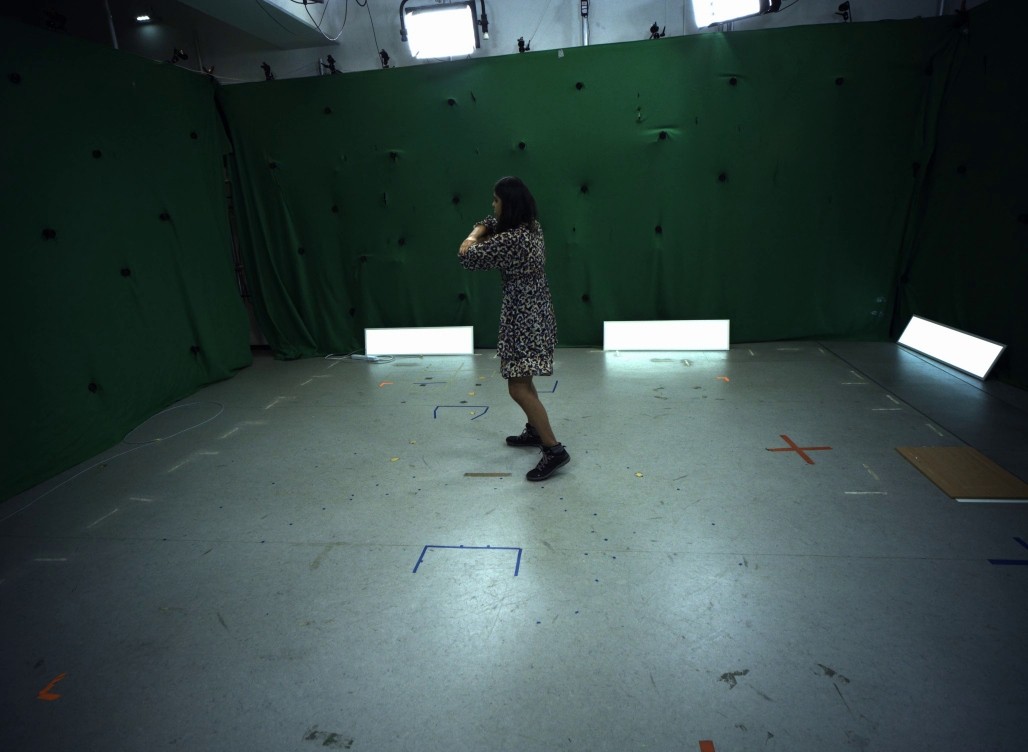}}
    &
    \raisebox{-0.5\height}{\includegraphics[width=0.2\textwidth]{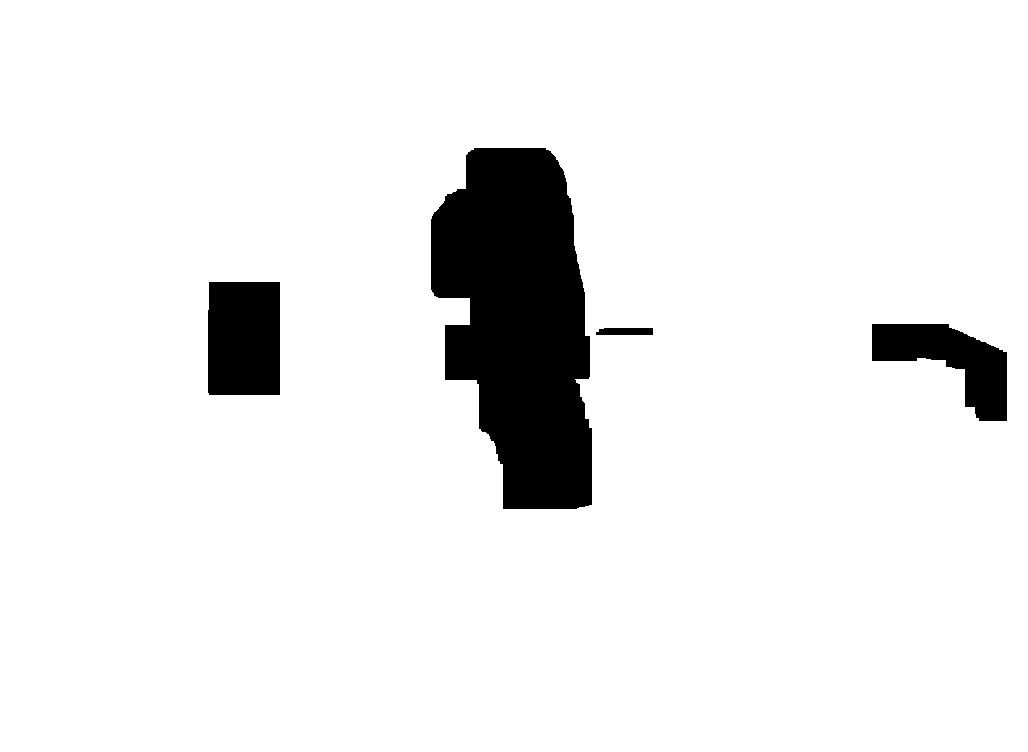}}
    &
    \raisebox{-0.5\height}{\includegraphics[width=0.2\textwidth]{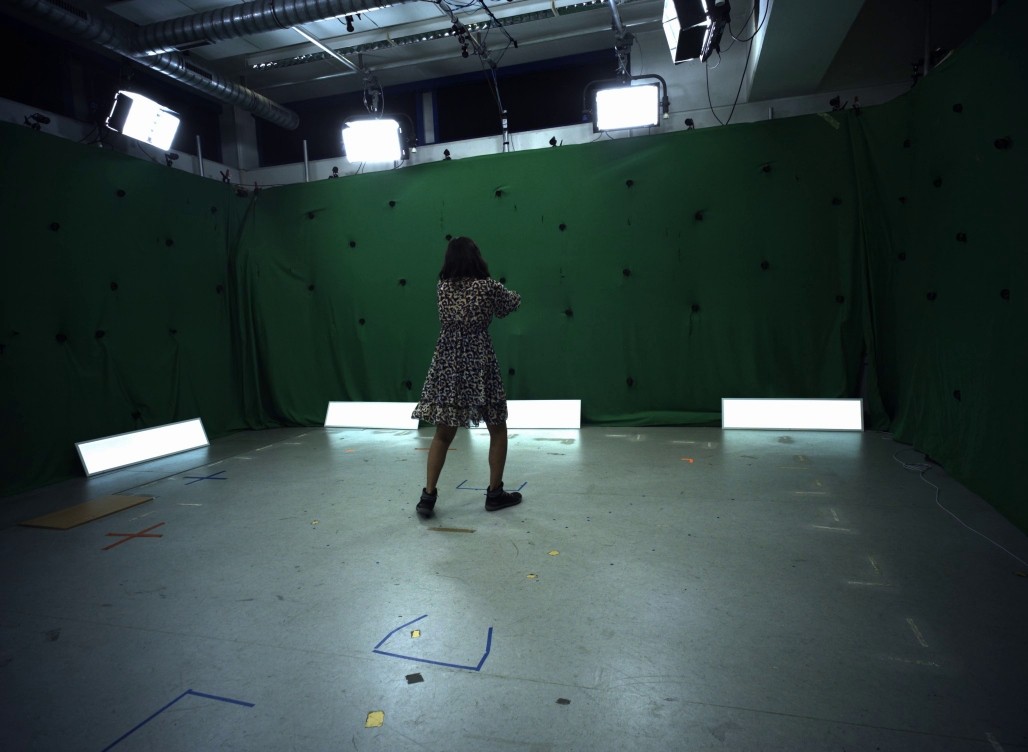}}
    &
    \raisebox{-0.5\height}{\includegraphics[width=0.2\textwidth]{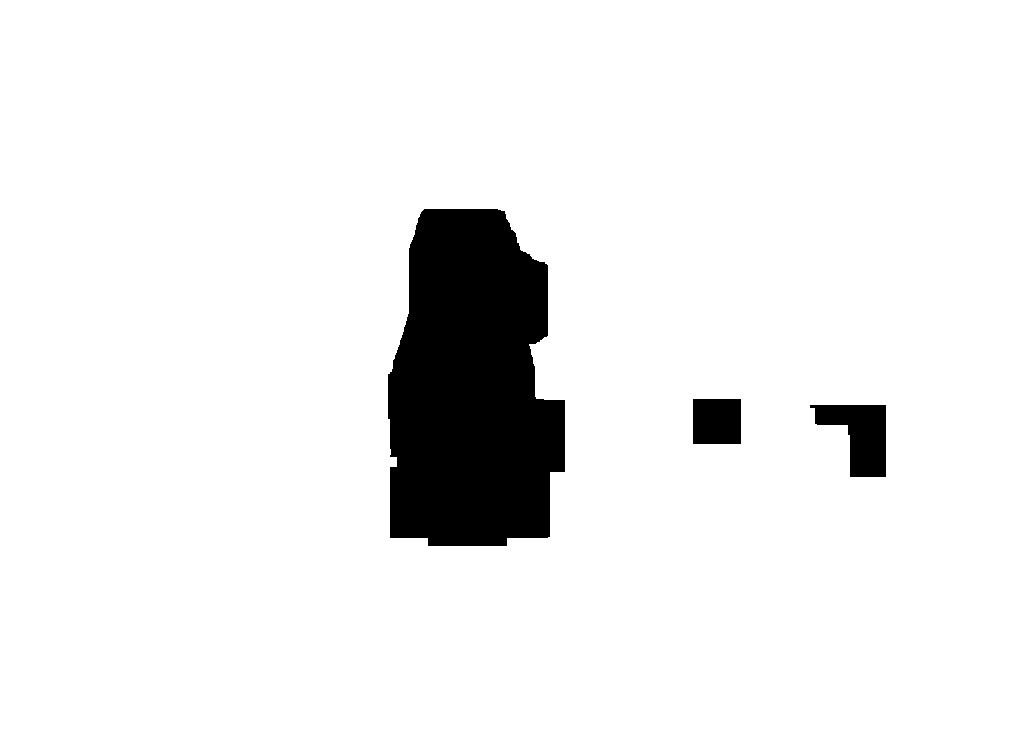}}
    \\
    
    \parbox[t]{2mm}{\rotatebox[origin=c]{90}{Seq. $5$}}
    &
    \raisebox{-0.5\height}{\includegraphics[width=0.2\textwidth]{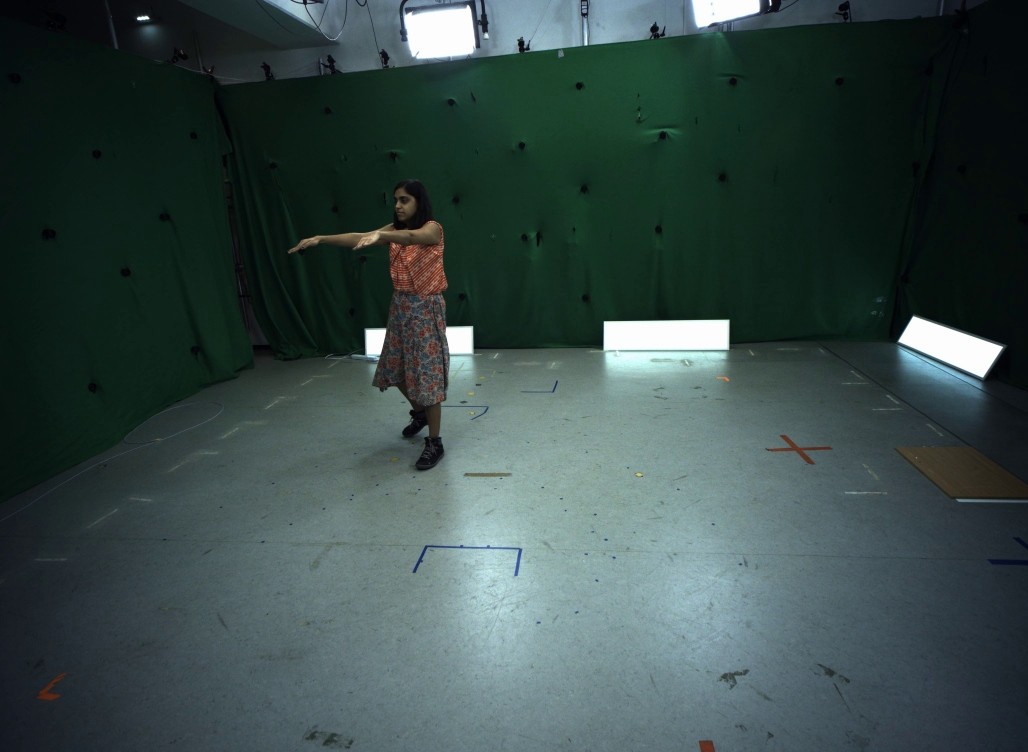}}
    &
    \raisebox{-0.5\height}{\includegraphics[width=0.2\textwidth]{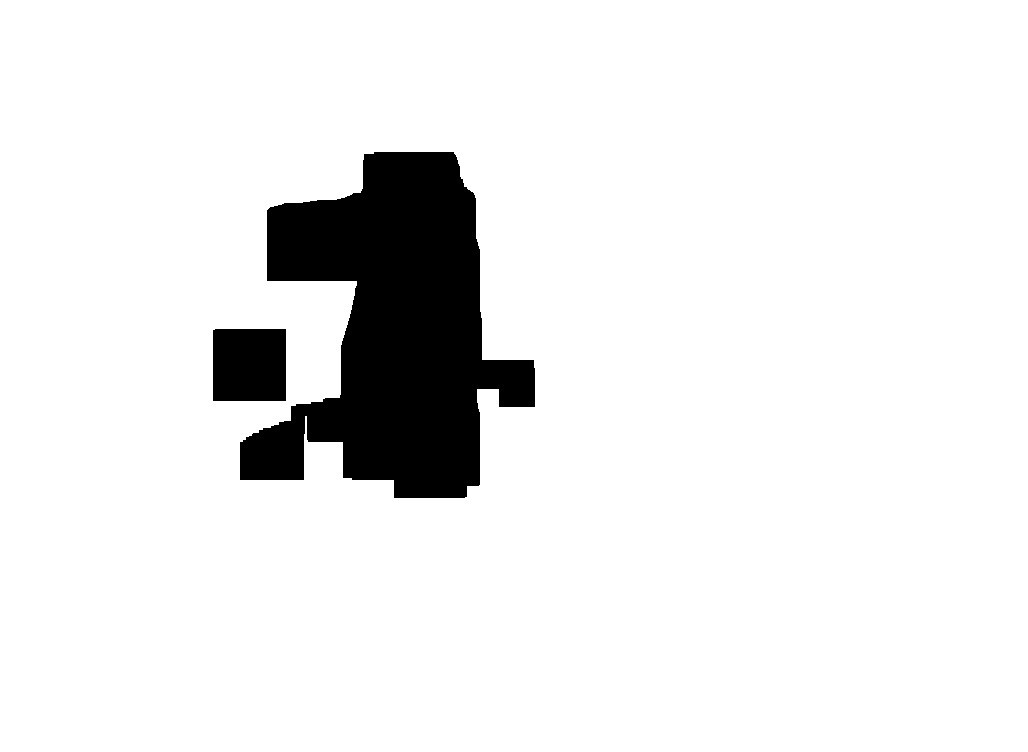}}
    &
    \raisebox{-0.5\height}{\includegraphics[width=0.2\textwidth]{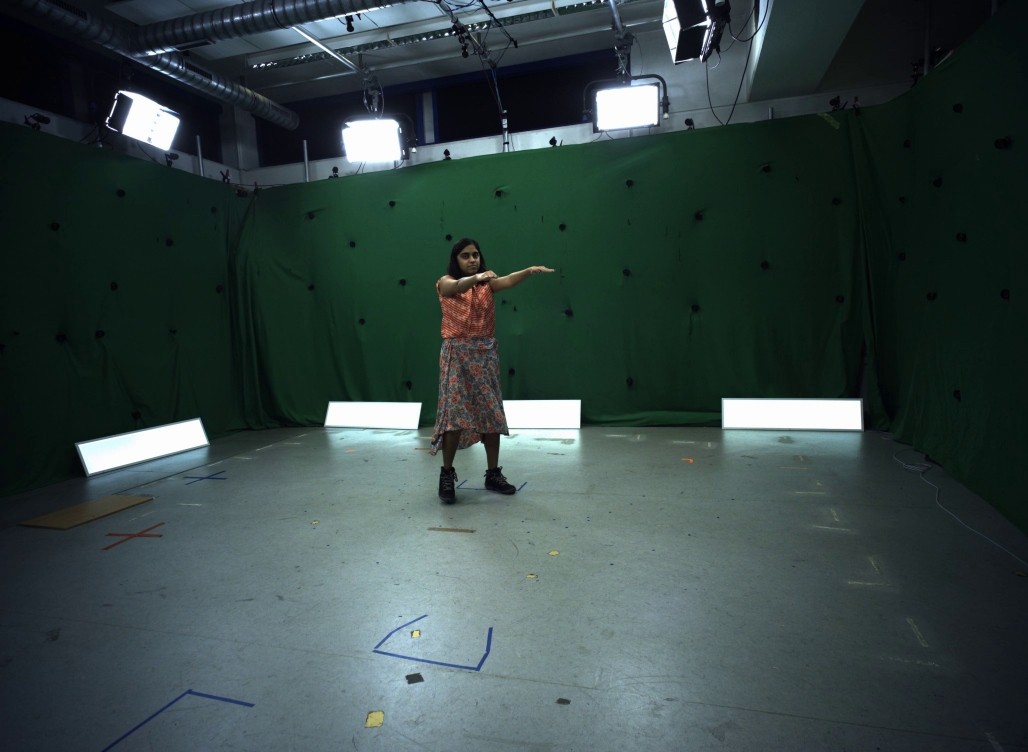}}
    &
    \raisebox{-0.5\height}{\includegraphics[width=0.2\textwidth]{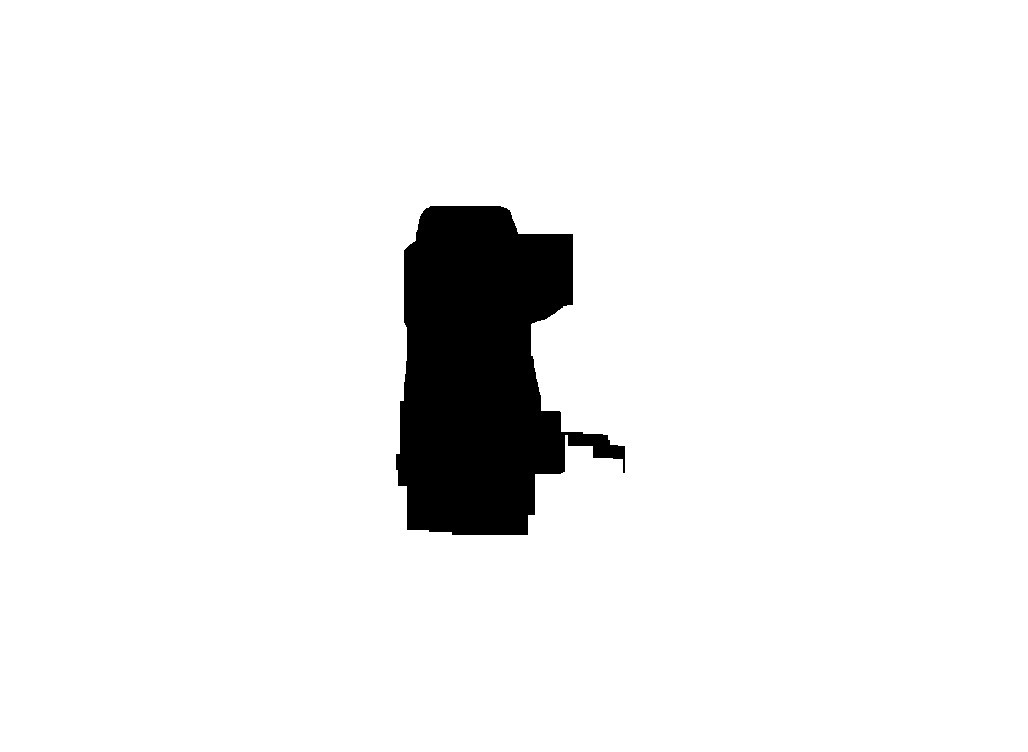}}
    \\

    \parbox[t]{2mm}{\rotatebox[origin=c]{90}{Seq. $6$}}
    &
    \raisebox{-0.5\height}{\includegraphics[width=0.2\textwidth]{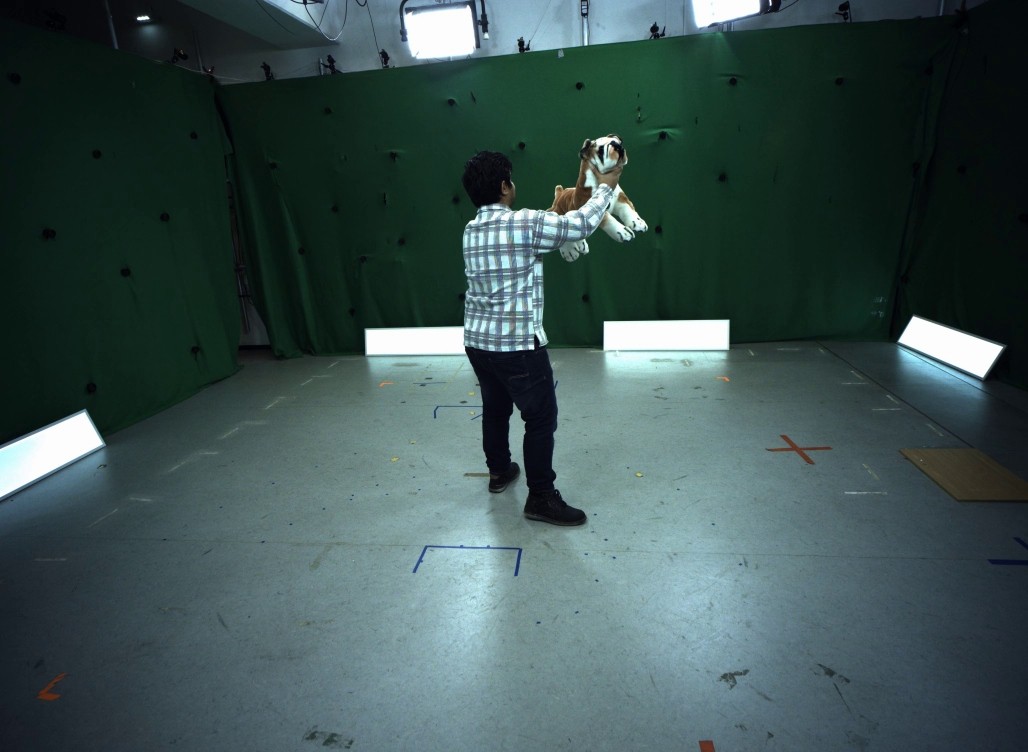}}
    &
    \raisebox{-0.5\height}{\includegraphics[width=0.2\textwidth]{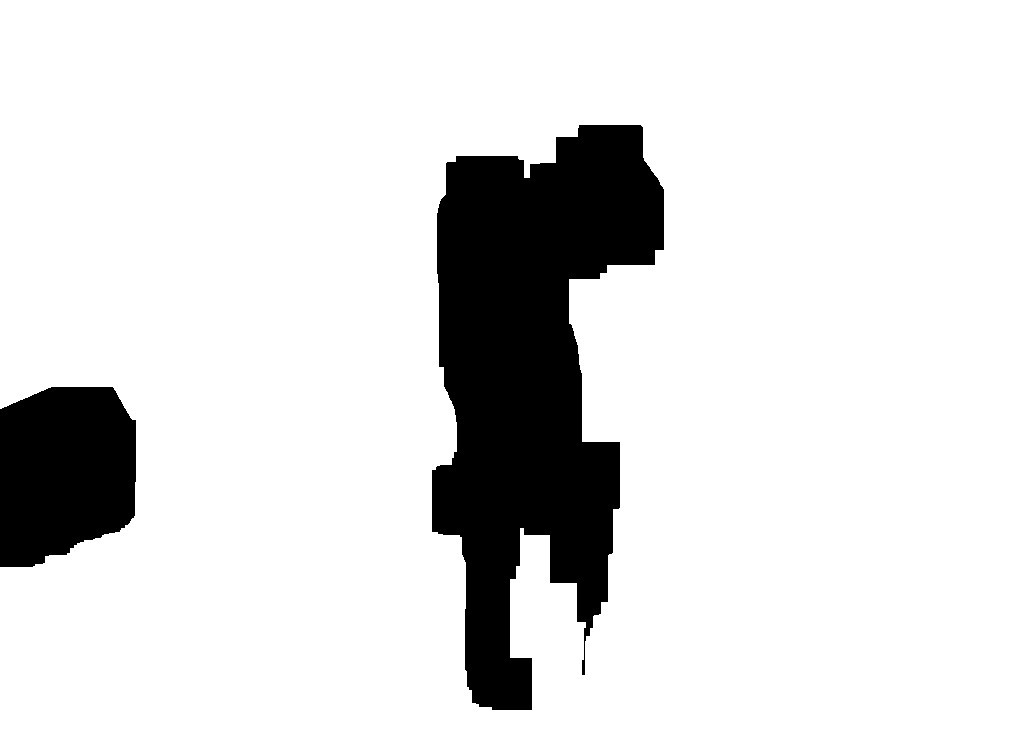}}
    &
    \raisebox{-0.5\height}{\includegraphics[width=0.2\textwidth]{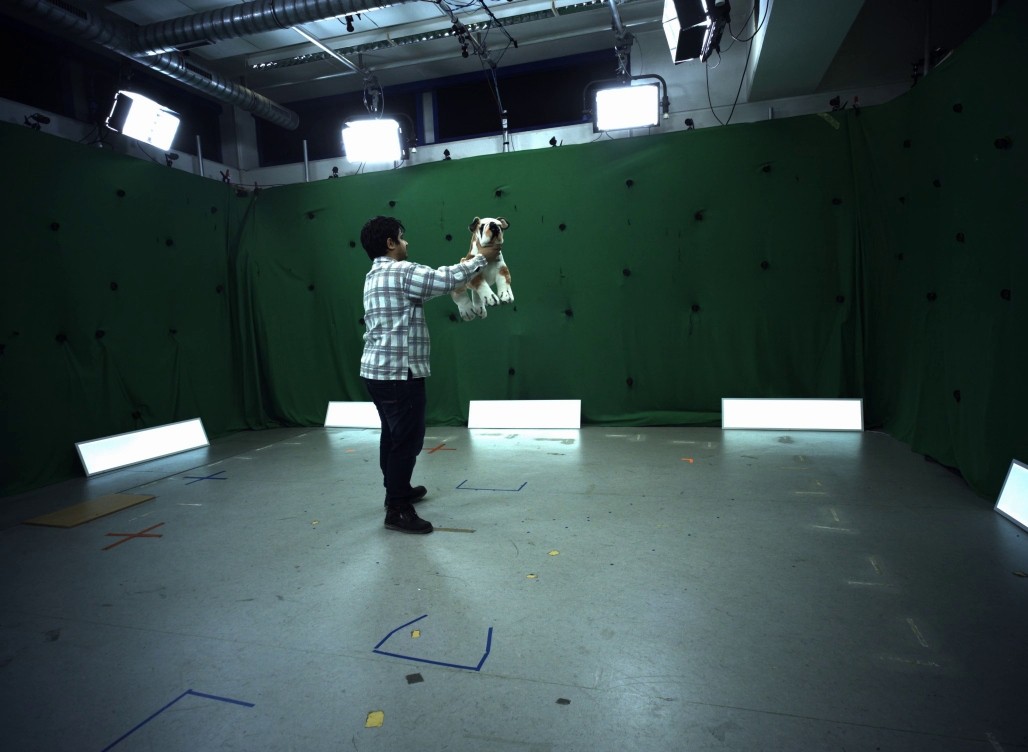}}
    &
    \raisebox{-0.5\height}{\includegraphics[width=0.2\textwidth]{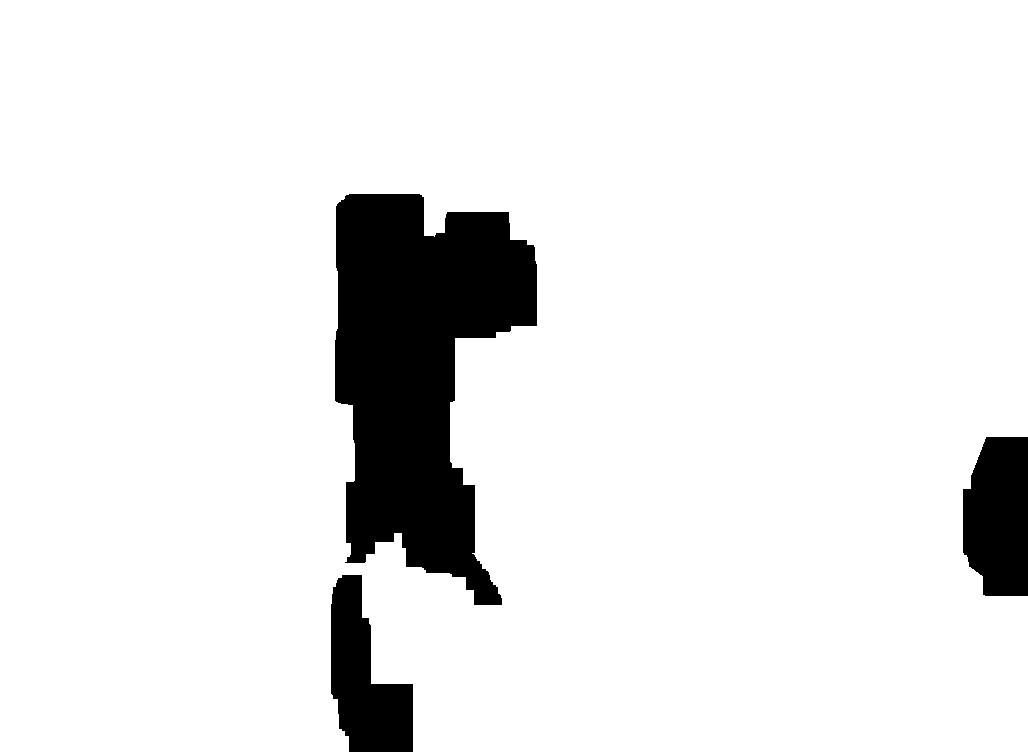}}
    \\

    \parbox[t]{2mm}{\rotatebox[origin=c]{90}{Seq. $7$}}
    &
    \raisebox{-0.5\height}{\includegraphics[width=0.2\textwidth]{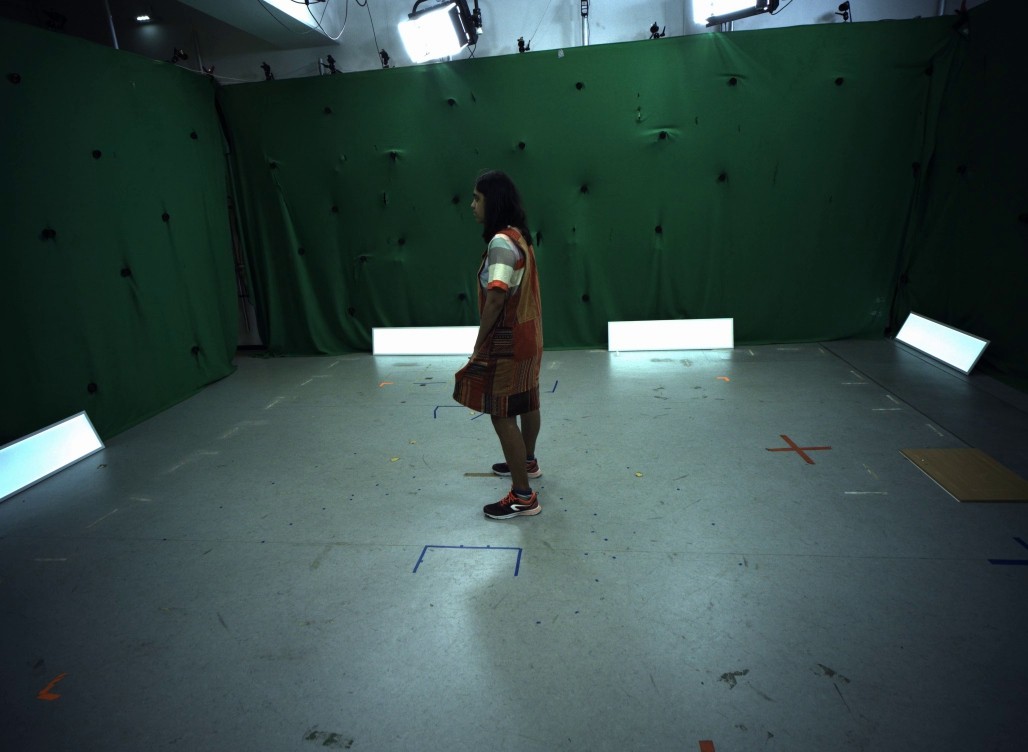}}
    &
    \raisebox{-0.5\height}{\includegraphics[width=0.2\textwidth]{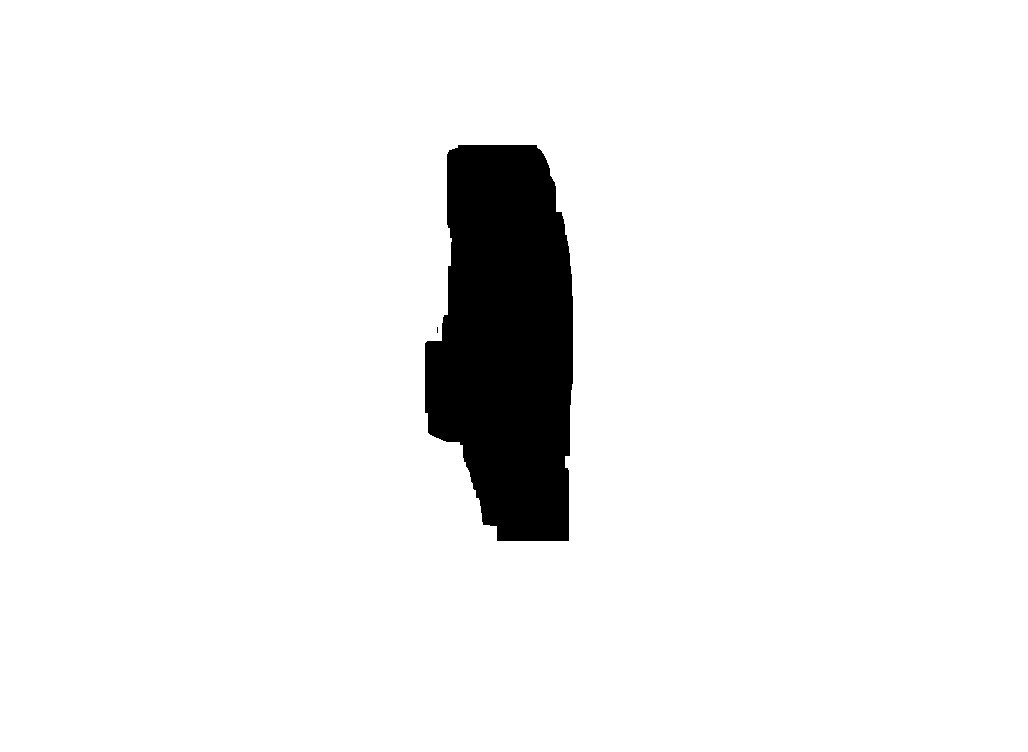}}
    &
    \raisebox{-0.5\height}{\includegraphics[width=0.2\textwidth]{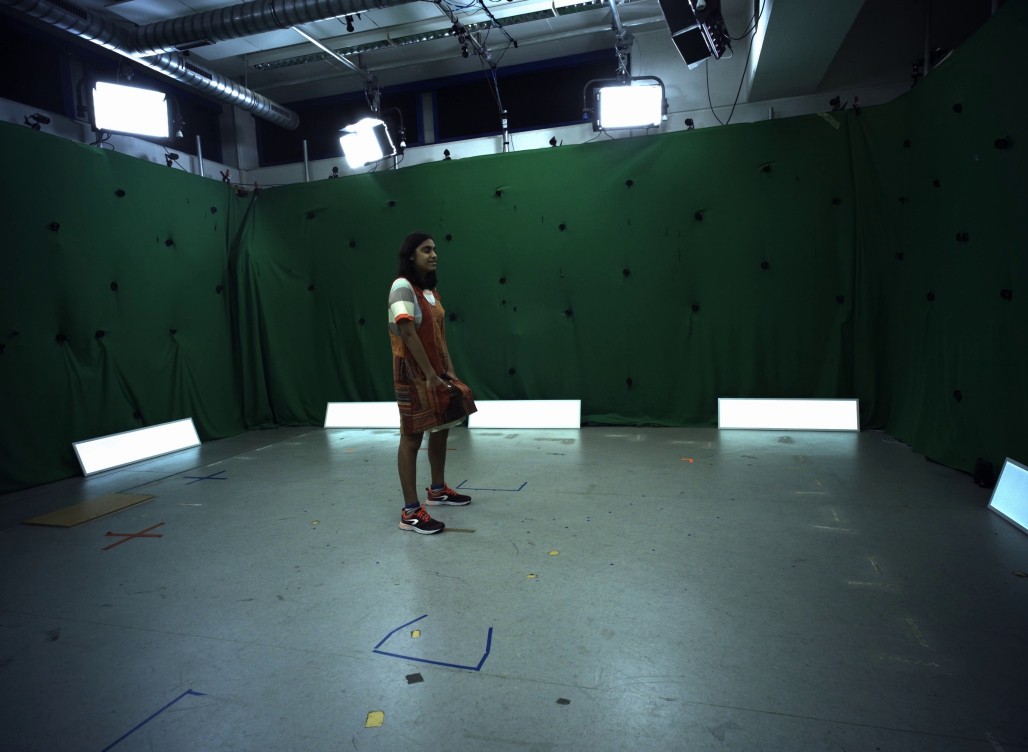}}
    &
    \raisebox{-0.5\height}{\includegraphics[width=0.2\textwidth]{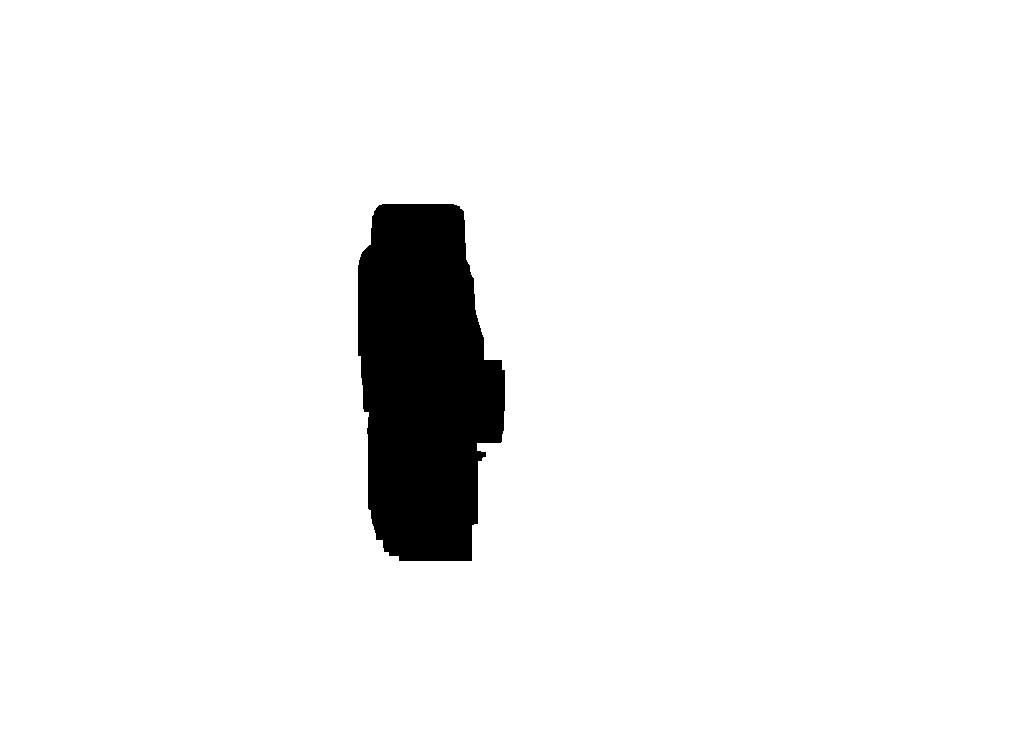}}
    \\

    \parbox[t]{2mm}{\rotatebox[origin=c]{90}{Seq. $8$}}
    &
    \raisebox{-0.5\height}{\includegraphics[angle=-90,origin=c,width=0.1\textwidth]{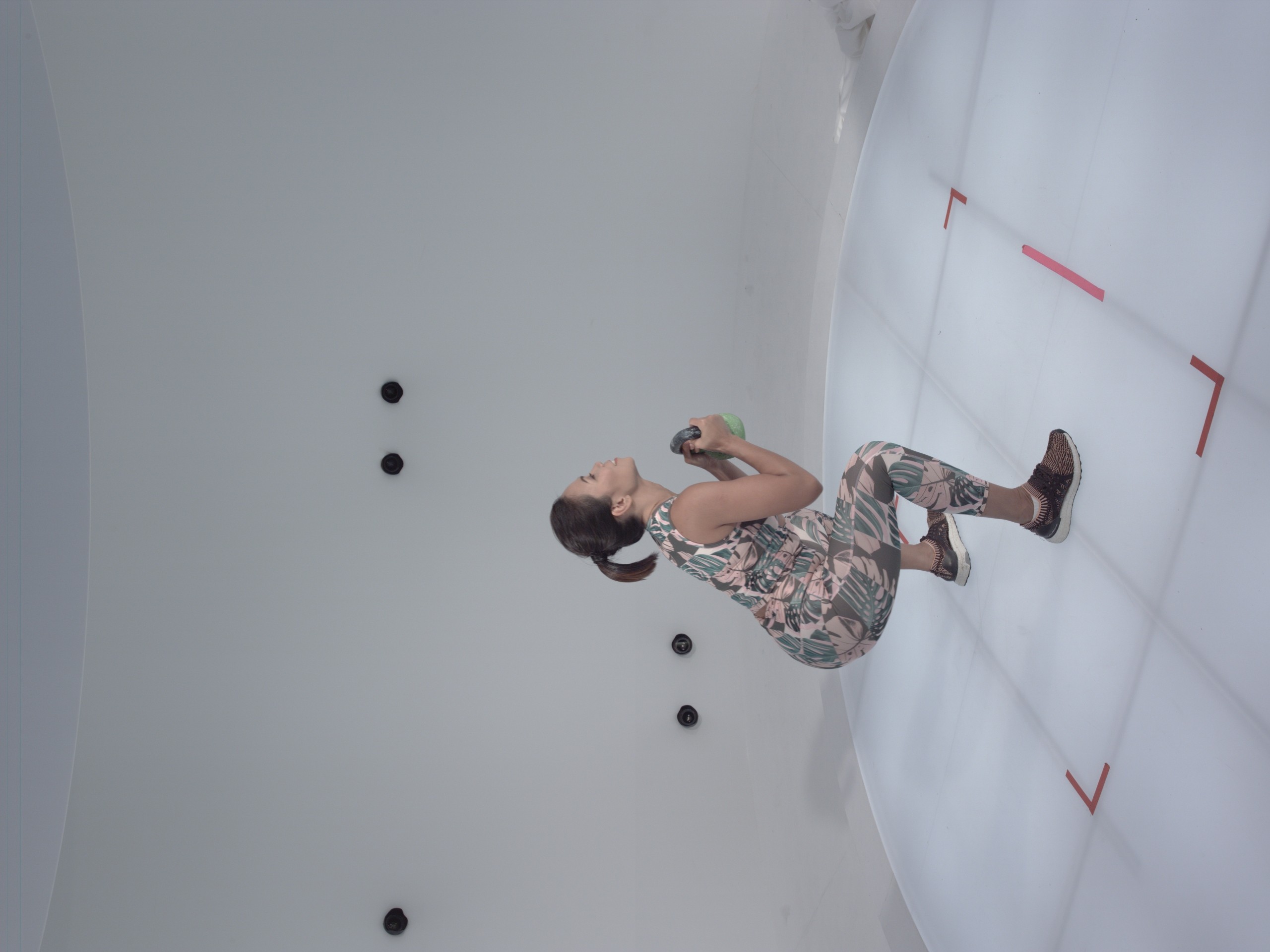}}
    &
    \raisebox{-0.5\height}{\includegraphics[angle=-90,origin=c,width=0.1\textwidth]{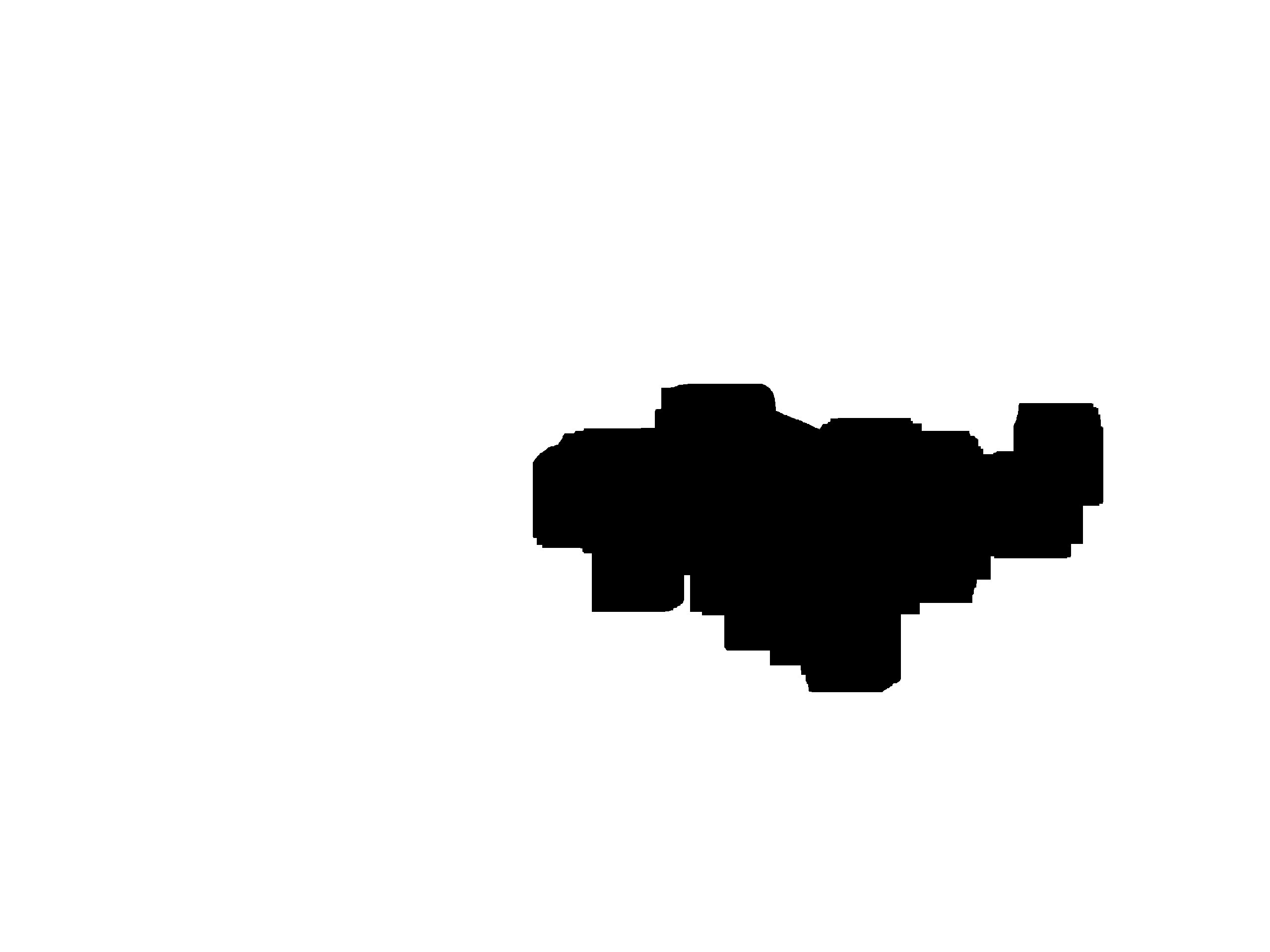}}
    &
    \raisebox{-0.5\height}{\includegraphics[angle=-90,origin=c,width=0.1\textwidth]{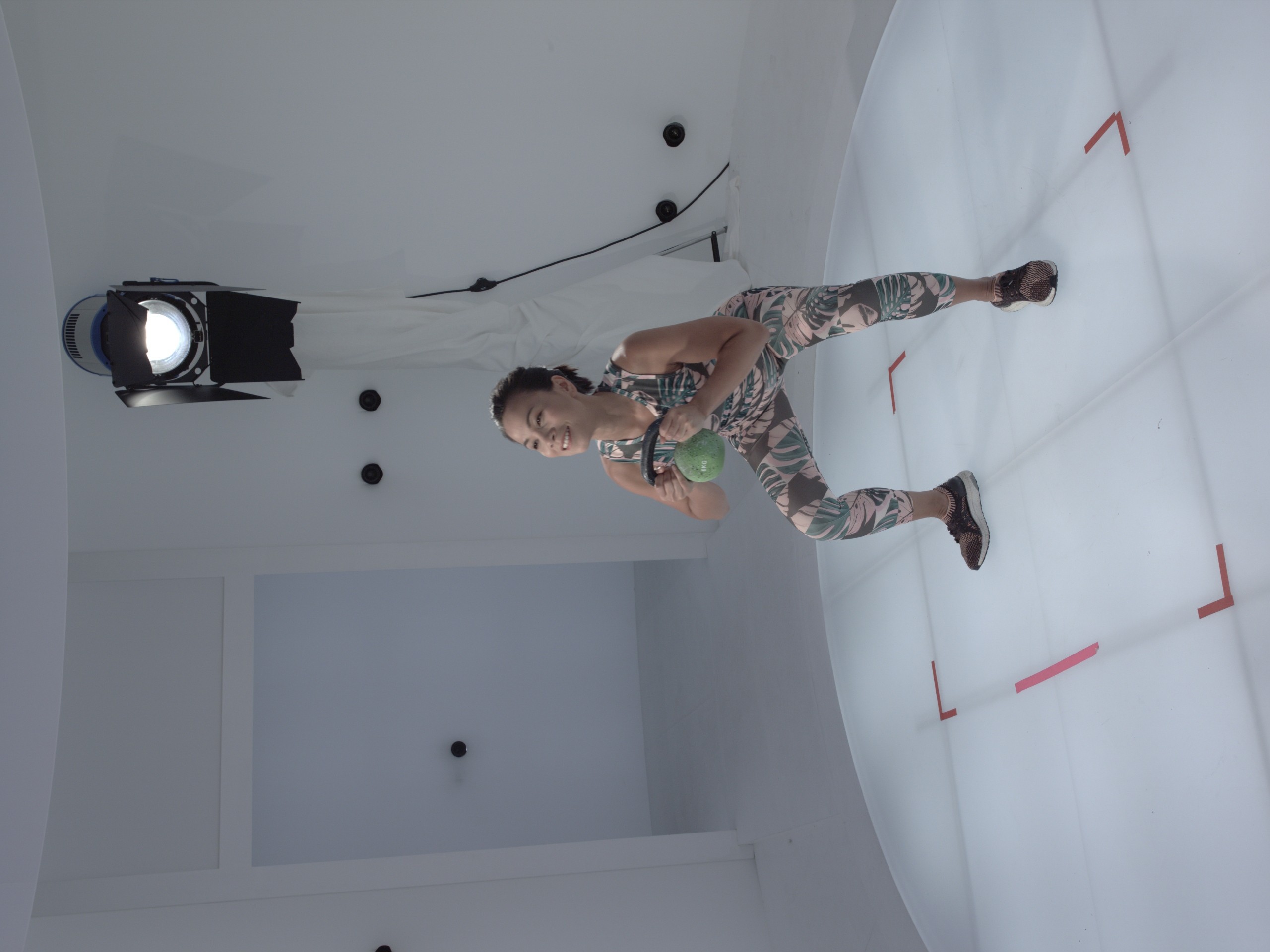}}
    &
    \raisebox{-0.5\height}{\includegraphics[angle=-90,origin=c,width=0.1\textwidth]{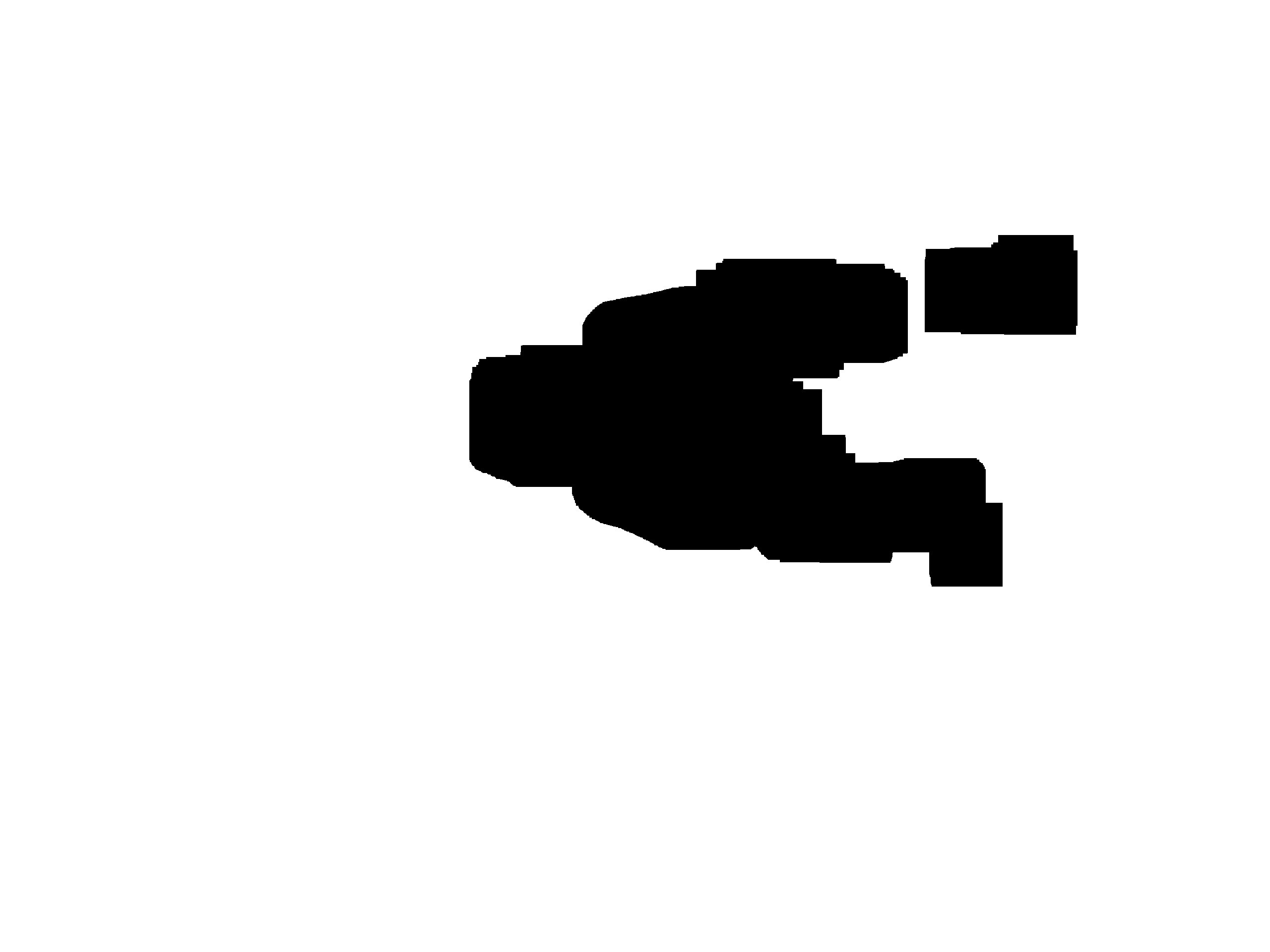}}
    \\
    
    \end{tabular}
    
    \caption{\textbf{Foreground Masks for Evaluation.}}
    \label{fig:foreground_masks_for_evaluation}
\end{figure*}

During training, we use very coarse and inaccurate foreground masks. 
However, we use more accurate foreground masks when computing masked scores during evaluation. 
To also consider reconstructions that are slightly off, these masks are dilated to include the areas surrounding the dynamic foreground in pixel space.
Fig.~\ref{fig:foreground_masks_for_evaluation} shows example masks. 

We use the following procedure to obtain these more accurate foreground masks. 
We first compute two foreground masks separately: (1) $m_b$ using background subtraction with a threshold $\Delta_t$ followed by a morphological opening (\emph{i.e.}, first erosion, then dilation) of the foreground, and (2) $m_\sigma$. 
$m_\sigma$ very roughly detects shadows by determining whether the pixel in $\mathbf{I}^{c,t}$ is a scaled version of the background pixel in $\mathbf{B}^c$ (\emph{i.e.}, whether a na\"ive brightness change of the background has occurred). 
To this end, we first divide each of the three channels of the pixel in $\mathbf{I}^{c,t}$ by the respective channel of the background pixel. 
We then take the standard deviation of the resulting three factors and threshold it at $\sigma_t$. 
We finally apply an opening to obtain the final $m_\sigma$. 
Since we use a very generous opening for both masks, we combine them into a single foreground mask by considering those pixels foreground that are foreground in both masks. 
We manually tune $\Delta_t$, $\sigma_t$, and the opening sizes for each sequence. 

\end{document}